\documentclass[a4paper,12pt]{article}
\usepackage{authblk}
\usepackage{amsfonts}
\usepackage{amssymb}
\usepackage{amsthm}
\usepackage{amsmath}
\usepackage{array}
\usepackage{float} 
\usepackage{booktabs}
\usepackage{cases}
\usepackage{cite}
\usepackage{color}
\usepackage{dsfont}
\usepackage{enumerate}
\usepackage{epsfig}
\usepackage{multirow}
\usepackage{graphicx}
\usepackage{geometry}%
\usepackage[colorlinks,linkcolor=red]{hyperref}
\usepackage{pdflscape}
\usepackage{longtable}
\usepackage{latexsym}
\usepackage{longtable}
\usepackage{lscape}
\usepackage{multirow}
\usepackage{mathrsfs}
\usepackage[numbers]{natbib}
\usepackage{rotating}
\usepackage{setspace}
\usepackage{pst-node}
\usepackage{pst-grad}
\usepackage{pst-xkey}
\usepackage{pstricks}
\usepackage{pst-gantt}
\usepackage[section]{placeins}
\usepackage{tipa}
\usepackage{titlesec}
\usepackage{titletoc}
\usepackage{times}
\usepackage{tikz}
\usepackage[T1]{fontenc}
\usepackage[utf8]{inputenc}
\usepackage{url}
\usepackage{subfigure}
\usepackage{diagbox}
\usepackage{color}
\usepackage{supertabular}
\usepackage{tabularx}
\usepackage{lscape}
\usepackage{amssymb}
\usepackage{bbding}
\usepackage{hyperref}
\usepackage{bm}
\usepackage{verbatim}

\title{Heterogeneous Feature Representation for Digital Twin-Oriented Complex Networked Systems}

\author[1]{Jiaqi Wen, Bogdan Gabrys, Katarzyna Musial \\ Complex Adaptive Systems Lab, Data Science Institute, University of Technology Sydney \\ jiaqi.wen@student.uts.edu.au, bogdan.gabrys@uts.edu.au, katarzyna.musial-gabrys@uts.edu.au}

\begin{document}
\setlength{\baselineskip}{18pt}%

\date{~}
\maketitle
\vspace{-1cm}

\begin{abstract}
Building models of Complex Networked Systems (CNS) that can accurately represent reality forms an important research area. To be able to reflect real world systems, the modelling needs to consider not only the intensity of interactions between the entities but also features of all the elements of the system. This study aims to improve the expressive power of node features in Digital Twin-Oriented Complex Networked Systems (DT-CNSs) with heterogeneous feature representation principles. This involves representing features with crisp feature values and fuzzy sets, each describing the objective and the subjective inductions of the nodes' features and feature differences. Our empirical analysis builds DT-CNSs to recreate realistic physical contact networks in different countries from real node feature distributions based on various representation principles and an optimised feature preference. We also investigate their respective disaster resilience to an epidemic outbreak starting from the most popular node. The results suggest that the increasing flexibility of feature representation with fuzzy sets improves the expressive power and enables more accurate modelling. In addition, the heterogeneous features influence the network structure and the speed of the epidemic outbreak, requiring various mitigation policies targeted at different people.



\vspace{1ex}
\vspace{1ex}


\vspace{0.2cm}

\end{abstract}
\newpage
\tableofcontents
\newpage
\newpage

\section{Introduction}

Research community's efforts have resulted in significant progress in terms of Complex Networked Systems (CNSs) modelling. The accuracy of the approaches consistently increases with the ultimate goal to build a Digital Twin of a real-world Complex Networked System that would faithfully represent real systems. The recent advancements are related to the increasing modelling complexity that includes introducing heterogeneous features and realistic rules of network growth and dynamic processes on networks \citep{wen2023dtcns}. The features include both topological and ascribed characteristics connected with the topology and nodes attributes while the rules aim at capturing such phenomena as the preferential attachment and homophily. 

Digital Twin-Oriented Complex Networked Systems (DT-CNSs) is an approach proposed to deal with different levels of complexity when it comes to dynamics of and on the network with the aim of faithfully representing and modelling real complex networked systems~\citep{wen2022towards}. In the space of social networks, the aim of the DT-CNS is to model interactions between people and dynamics of any processes propagating over the resulting from the interactions social network. The information that is utilised to build DT-CNS includes people's characteristics (e.g. age, gender) and preferences when it comes to who they want to interact with (i.e. preference of people towards specific age of others).

In the network dimension, many studies developed network models on preferential attachment to topological features \citep{barabasi1999emergence,tsiotas2020preferential,abbasi2012betweenness} (e.g. degree centrality and clustering coefficient, etc.). In contrast, some other studies drive the network growth based on the homophily of nodes considering their similarity in terms of the ascribed features \citep{kossinets2009origins,boucher2015structural,ertug2022does} (e.g. age, sex, etc.) and the latent representation space \citep{zhou2018dynamic,liao2018attributed}. In the process dimension, the dynamic processes on the networks generally select a seed for the first contagious nodes based on topological features \citep{comin2011identifying,karczmarczyk2018influencing} (e.g. centrality measures, etc.) and model their transmissibility conditional on specific features about interactions (e.g. multiple/single exposure \citep{min2018competing,jovanovski2021modeling,brodka2020interacting}, close/air contact \citep{liu2020using,krol2015propagation}, etc.). 

CNSs in the current studies get increasingly complex, considering more realistic information about the network and process dimensions. Historically, CNSs first focused on the least complex scenario: static networks without introducing network attributes or dynamic processes \citep{barabasi1999emergence,doye2002network,watts1998collective,fortunato2006scale,lu2009similarity,seaton2004stations,solomonoff1951connectivity,hunter2008goodness,lancichinetti2009benchmarks}. Only recently, the researchers incorporated realistic information for dynamic processes to investigate the impact of different static network topologies on the processes \citep{nekovee2007theory,li2014analysis,stegehuis2016epidemic,zhang2021vulnerability,zhang2018influence,wang2019simulation,wang2021multi}. For example, \cite{li2014analysis} validated the impact of network community structure on the epidemic dynamics by modelling the epidemic spreading process on the clustered scale-free networks. Meanwhile, researchers also included more realistic network features in CNSs \citep{ashraf2019simulation,sun2014epidemic,musial2012triad,jia2021directed,jia2021measuring}. \cite{ashraf2019simulation} proposed a social network simulator based on node features and nodes' preferences for connecting with others. \cite{sun2014epidemic} considered the multi-relation information attributed to the edges and studied their impact on the epidemic spreading process. Our previous study, \cite{wen2023dtcns}, investigates the impact of node features on the networks and the epidemic spreading processes on these networks.

The complexity of CNSs increases even more as the network and the process dimensions evolve with the interrelated changes in the network topology and transmissibility. Current studies generally ignore the interrelated dynamics in CNSs and most of them only focus on the dynamic networks without considering dynamic processes 
\citep{gunecs2016link,petri2018simplicial,ashraf2019simulation,block2020social,shi2020evaluating,musial2013creation,budka2013molecular}. There are just few studies that investigate the impact of dynamic networks on the dynamic processes \citep{liu2020using,kim2020location}. In addition, some studies include nodes' features in the CNS evolution process. Only \cite{ashraf2019simulation} simulates dynamic networks based on features and the changeable connection principles. Overall, the heterogeneous node features and the changes in CNSs remain a research challenge.

Our previous study proposed a DT-CNS modelling framework on the above-mentioned heterogeneous features and feature preference principles, including preferential attachment and homophily \cite{wen2023dtcns}. However, all the features studied in the CNS space are specific qualitative features. The uncertain, subjective and vague inductions of qualitative and quantitative features, such as different views of the linguistic descriptions for height (short, tall, very tall, etc.), emerge as another source of complexity for DT-CNSs and a research gap to be addressed in this study. Therefore, we aim to improve the expressive power of the proposed DT-CNS modelling framework by faithfully representing the uncertain features and the related preferences and validating their influences on approaching reality during the modelling process.  

Current studies generally use fuzzy sets to express vague qualitative or quantitative data based on various fuzzy membership functions, such as Gaussian and triangular membership functions \citep{ajofoyinbo2011development}. They investigate the influence of membership functions on the expressive power as part of sensitivity checks \citep{thiem2014membership,ali2015comparison,lee2001comparison}, and optimise their parameters to improve expressiveness \citep{kaya2006utilizing,lagunes2018optimization}. In contrast, current studies on CNSs generally employ fuzzy sets to represent fuzzy communities \citep{havens2013soft} or fuzzy trust relations \citep{wu2014social,xu2020trust2privacy}. None of these studies uses fuzzy sets to describe the CNS features of nodes. 

This study improves the expressive power of the current DT-CNSs by introducing heterogeneous feature representation principles, including both the crisp and fuzzy value representations. We first extend the proposed in \citep{wen2023dtcns} DT-CNSs by introducing both crisp and fuzzy representations of features with the goal of recreating physical contact networks of different countries from real features, including Belgium, Finland, Germany, Luxembourg, Italy and Poland (extracted from the website \footnote{\url{http://www.socialcontactdata.org/socrates/}}. Next, we validate the impact of various features and feature representation principles on the DT-CNSs' performance. Finally, we identify each country's most realistic physical contact networks and investigate their respective disaster resilience given an epidemic spread from the most popular node. 

The main contributions of this study are:
\begin{itemize}
 \item Enabling heterogeneous node feature representation from the perspectives of entities in their decisions on evaluating and connecting with others
 by introducing fuzzy representation principles;
 \item Recreating real-world social contact patterns based on real features and principles of feature representation and preference;
 \item Validating the impact of heterogeneous feature representation on network growth and epidemic spread;
 \item Suggesting mitigation policies targeted at heterogeneous populations for disaster resilience.
\end{itemize}

The rest of this study is structured as follows. Section \ref{Rep1-2section2} presents the methodology of DT-CNSs driven by heterogeneous features and rules. It also presents the feature representation principles, including the crisp value and the fuzzy set representations. 
Following this, Section \ref{Rep1-2section3} presents the empirical analysis, discussion and validation of DT-CNSs that recreate social contact networks from real features. Finally, we conclude with Section \ref{Rep1-2section4}.

\section{Digital Twin-Oriented Complex Networked Systems built on fuzzy representation of features}
\label{Rep1-2section2}
In this section, we propose a DT-CNS modelling approach for real-world social networked systems based on heterogeneous features, feature representation principles and feature preference principles. 

To facilitate an understanding of the concepts used the following example is used. The nodes (entities) in the social networks can have features such as sex or age. These nodes observe and evaluate the features of others to determine their preferred connections. The specific and uncertain features can be represented (feature representation) and preferred (feature preference) differently in this process. In the used datasets, the gender feature take specific binary values between females and males in the nodes' mutual evaluation processes, where they can have specific preferences for females or males (crisp values). In contrast, age, though it can be described with specific values, can only be perceived and preferred by other nodes with uncertainty in the evaluation process. The middle-aged nodes may prefer to contact a young-age node in the social interaction process due to parenthood. In this case, the respective age range can be uncertain and represented with fuzzy sets and the nodes have heterogeneous preferences for each fuzzy set.



We represent this real system as $\mathbf{GP}_t = (\mathbf{G}_t,\mathbf{P}_t)$, composed of a social network $\mathbf{G}_t =(\mathbf{V}_t,\mathbf{E}_t)$ and an epidemic spreading process $\mathbf{P}_t = (\mathbf{S}_t,\mathbf{R}_t, \gamma_t, \eta_t)$ that takes place on this social network at the time $t$.

The social entities (nodes) are represented as $\mathbf{V}_t = \{\cdots, v_{i,t}, \cdots\}$, with their undirected pairwise interactions denoted in the edge set $\mathbf{E}_t = \{\cdots, \mathbf{e}(v_{i,t},v_{j,t}), \cdots|v_{i,t},v_{j,t}\in V_t, i\neq j\}$. Among these entities, the epidemic spread starts from specific social entities $\mathbf{S}_t=\{\cdots,v_{i,t},\cdots|v_{i,t}\in V_t\}$ with a transmissibility at the level of $\gamma_t$ and recovery rate of $\eta_t$ and results in changes of the others' infection status $\mathbf{R}_t = \{\mathbf{r}(v_{i,t})|v_{i,t}\in V_t\}$. The infection status is represented by a binary value, $0$ or $1$, each representing the node's healthy and infected status.

The social contact patterns and the epidemic propagation, as represented with $\mathbf{G}_t =(\mathbf{V}_t,\mathbf{E}_t)$ and $\mathbf{P}_t = (\mathbf{S}_t,\mathbf{R}_t)$, are each driven by heterogeneous features $\mathbf{f}(v_{i,t})$ related to each social entity $v_{i,t}$ and the corresponding principles of feature preference $\mathbf{fp}(v_{i,t})$ and feature representation $\mathbf{fr}(v_{i,t})$ as explained in the sections below. The features, preferences related to features and the respective representation principles vary for the entity $v_{i,t}$ and can change with the time $t$. 

\subsection{Feature representation}
\label{Fr}


If we assume there are $k$ features considered in society, each social entity $v_{i,t}$ has a $k$ dimensional feature vector:
\begin{equation}
    \mathbf{f}(v_{i,t}) = [f_1(v_{i,t}),f_2(v_{i,t}),\cdots,f_k(v_{i,t})]
\end{equation}
which, correspondingly, is followed with a $k$ dimensional vector $\mathbf{fr}(v_{i,t})$ to determine the feature representation principles. The representation of the $k$-th feature is determined by $fr_k(v_{i,t}) = (cf_k^p(v_{i,t}),\mathbf{m}_k^p(v_{i,t}),cf_k^h(v_{i,t}),\mathbf{m}_k^h(v_{i,t}))$, 
including a binary indicator $cf_{k}^p = 0 \quad\textrm{or}\quad 1$ for the usage of crisp feature values or fuzzy sets respectively, and, if using fuzzy sets, the $l_k^p$--length vectors $\mathbf{m}_k^p(v_{i,t})$ that define the respective membership function of each fuzzy set. Similarly, the binary indicator $cf_{k}^h = 0 \quad\textrm{or}\quad 1$ decides the usage of crisp feature differences or fuzzy feature differences, and is followed with the $l_k^h$--length vectors $\mathbf{m}_k^p(v_{i,t})$ that determines the corresponding Membership functions.

Specifically, for any entity $v_{i,t}$, his perception of the feature $f_k(v_{j,t})$ of any other node such as $v_{j,t}$ is unfolded as: 
\begin{equation}
\small
\begin{split}
f_k^u(v_{j,t}|v_{i,t}) =\left \{
\begin{aligned}
&f_k(v_{j,t})  &\textrm{if $cf_k^p(v_{i,t})=0$.} \\
& [f_{k,1}(v_{j,t}|v_{i,t}),f_{k,2}(v_{j,t}|v_{i,t}),\cdots,f_{k,l_k^p}(v_{j,t}|v_{i,t})] &\textrm{if $cf_k^p(v_{i,t})=1$.} \\
\end{aligned}
\right.
\end{split}
\end{equation}
where we unfold an uncertain perception of feature $f_k(v_{j,t})$  into a series of membership degrees according to the Membership functions of fuzzy sets. 

We can calculate membership degree to the $y_{th}$ fuzzy set based on a Gaussian Membership function as: 
\begin{equation}
    f_{k,y}(v_{j,t}|v_{i,t}) = exp\{-\frac{1}{2}(\frac{f_k(v_{j,t})-\mu_{y}^p(v_{i,t})}{\sigma_{y}^p(v_{i,t})})^2\} 
\end{equation}
where $\mu_y^p(v_{i,t})$ and $\sigma_y^p(v_{i,t})$, included in the vector $\mathbf{m}_k^p(v_{i,t})$, respectively represents the location and the spread of the involved Gaussian distribution for the specific focal node $v_{i,t}$.

As a result, the $k$-dimensional feature vector $\mathbf{f}(v_{j,t})$ of each node is unfolded into a $z$-dimensional vector $\mathbf{f}^u(v_{j,t}|v_{i,t})$, dependent on the feature representation principle of the other node. Imagining a $k$-length feature vector $\mathbf{f}(v_{j,t})$ which only has one uncertain feature $f_{2}(v_{j,t})$, it is unfolded in to the following $z$-length vector $\mathbf{f}^u(v_{j,t}|v_{i,t})$:
\begin{equation}
\begin{split}
\begin{aligned}
\mathbf{f}^u(v_{j,t}|v_{i,t}) = [f_1(v_{j,t}),f_{2,1}(v_{j,t}|v_{i,t}),f_{2,2}(v_{j,t}|v_{i,t}),\\
\cdots, 
    f_{2,l_2}(v_{j,t}|v_{i,t}),f_{3}(v_{j,t}),\cdots,f_k(v_{j,t})] \\
\end{aligned}
\end{split}
\end{equation}

Similarly, the $k$-dimensional vector $\Delta \mathbf{f} (v_{i,t},v_{j,t})$ represents the difference between entities in society: 
\begin{equation}
    \Delta \mathbf{f} (v_{i,t},v_{j,t}) = [\Delta f_1 (v_{i,t},v_{j,t}),\Delta f_2 (v_{i,t},v_{j,t}),\cdots,\Delta f_k (v_{i,t},v_{j,t})]
\end{equation}

The difference $\Delta f_k (v_{i,t},v_{j,t}) = |f_k(v_{i,t})- f_k(v_{j,t})|$ between the entities in terms of feature $f_k$ considering the fuzzy set denoted with $l_k^h$, from the subjective view of $v_{i,t}$, is unfolded as:
\begin{equation}
\small
\begin{split}
\Delta f^u_k (v_{i,t},v_{j,t}|v_{i,t}) =\left \{
\begin{aligned}
&\Delta f_k (v_{i,t},v_{j,t})  &\textrm{if $cf_k(v_{i,t})=0$.} \\
& [\Delta f_{k,1}(v_{j,t}|v_{i,t}),\Delta f_{k,2}(v_{j,t}|v_{i,t}),\cdots,\Delta f_{k,l_k^h}(v_{j,t}|v_{i,t})] &\textrm{if $cf_k(v_{i,t})=1$.} \\
\end{aligned}
\right.
\end{split}
\end{equation}
where an uncertain perception of feature difference $\Delta f_k(v_{j,t},v_{i,t})$ can also be unfolded from the perspective of entity $v_{i,t}$ into a series of membership degrees. 

Similarly, we can also calculate the membership degree to the $y_{th}$ fuzzy set based on a Gaussian Membership function as:
\begin{equation}
    \Delta f^u_k (v_{i,t},v_{j,t}|v_{i,t}) = exp\{-\frac{1}{2}(\frac{\Delta f_k(v_{i,t},v_{j,t})-\mu_{y}^h(v_{i,t})}{\sigma_{y}^h(v_{i,t})})^2\} 
\end{equation}
where $\mu_y^h(v_{i,t})$ and $\sigma_y^h(v_{i,t})$ each represents the location and the spread of the involved Gaussian distribution for the specific focal node $v_{i,t}$ and are included in the parameter vector $\mathbf{m}_k^h(v_{i,t})$.

Correspondingly, the difference between entities $v_{i,t}$ and $v_{j,t}$, from the subjective view of $v_{i,t}$ can be unfolded into the vector $\Delta\mathbf{f}^u(v_{i,t},v_{j,t}|v_{i,t})$:
\begin{equation}
\begin{split}
\begin{aligned}
\Delta\mathbf{f}^u(v_{i,t},v_{j,t}|v_{i,t}) = [\Delta f_1(v_{i,t},v_{j,t}),\Delta f_{2,1}(v_{i,t},v_{j,t}|v_{i,t}),\Delta f_{2,2}(v_{i,t},v_{j,t}|v_{i,t}),\\
\cdots, 
    \Delta f_{2,l_2}(v_{i,t},v_{j,t}|v_{i,t}),\Delta f_{3}(v_{i,t},v_{j,t}),\cdots,f_k(v_{i,t},v_{j,t})] \\
\end{aligned}
\end{split}
\end{equation}

To help illustrate, we again take the age and sex features of nodes as an example. According to the crisp representation principle, the sex features can be represented with specific binary values such as $0$ and $1$, each denoting the females and the males, respectively. Similarly, the age feature takes specific values and can be represented with the respective crisp value. However, rather than considering the specific age values and determining whether to prefer a specific age or age difference, the nodes can prefer middle-aged nodes around $35$ and the nodes with an age difference of around $30$. In this case, a single age can be represented with three membership values, each denoting the respective membership to the young-aged (age around $15$), middle-aged (age around $35$) and the old-aged (age around $65$). The age difference between a pair of nodes can also be represented by the degree of membership to the fuzzy set (concerning the age difference around $30$).


\subsection{Feature preference}
\label{Fp}
The nodes determine their connection preferences based on the others' features and their feature differences in comparison to others. These preferences are respectively captured by the preferential attachment and homophily principles, each respectively concerning the preferences for features and feature differences.

Referring to our previous study on DT-CNS \citep{wen2023dtcns}, the preference for the unfolded features $\mathbf{f}^u(v_{i,t})$ and the unfolded feature differences $\mathbf{f}^u(v_{i,t},v_{j,t}|v_{i,t})$ is defined with four same length vectors:
\begin{equation}
    \mathbf{fp} = \{\mathbf{p},\mathbf{w}^p,\mathbf{h},\mathbf{w}^h\}
\end{equation}
As an example, the uncertain deductions on a single feature of age can be unfolded with three fuzzy sets, each describing the respective membership to the young-age, middle-age and old-age group.

The trinity value vectors $\mathbf{p}$ and $\mathbf{h}$ indicate the negative/zero/positive preference for features and feature differences. 
$\mathbf{w}^p$ and $\mathbf{w}^h$ each represents the weight of preference, valuing within $(0,1]$.

Following the toy example in section \ref{Fr}, the nodes have the corresponding preferences for the respective features and feature differences. The preferential attachment principles can e.g. describe the nodes' preferences for connecting with female and middle-aged nodes. The homophily effect can describe the nodes' preferences for example for the same gender and an age difference of around $30$, which involves comparing the features between any node pair.

\subsection{Network formation}
\label{Nf}
In our proposed DT-CNSs, social networks grow based on the preference for the unfolded features. The pairwise interactions between entities are each scored as: 
\begin{equation}
 \pi(v_{i,t},v_{j,t}) = (\frac{1}{2}\pi^p(v_{i,t},v_{j,t}) + \frac{1}{2}\pi^h(v_{i,t},v_{j,t})+\epsilon_{ij,t})*I(\zeta) \quad v_i,v_j\in V, i\neq j
\end{equation}
where the encounter factor $I(\zeta)$ determines whether two nodes encounter each other based on a binary value between $0$ and $1$, following a Bernoulli distribution at the probability of $\zeta$. The node pair can encounter, denoted as $I(\zeta)=1$, by the chance of $\zeta$ and then evaluate each other for a connection. Tn contrast, the node pair, without an encounter (as denoted by $I(\zeta)=0$), cannot conduct mutual evaluations and stays unconnected. The random interference $\epsilon_{ij,t} \sim\mathcal{N}(0,\sigma^2)$ follows a random normal distribution and causes variations of scores of node pairs when they are similarly scored through mutual evaluation.

Constrained by the encounters, the network growth is driven by ranking dynamics based on the score $\pi(v_{i,t},v_{j,t})$ of any node pair through mutual evaluation concerned about their preferences for features and feature differences.

The preferential attachment score $\pi^p(v_{i,t},v_{j,t})$ incorporates the preference for the other nodes with larger/smaller crisp/fuzzy features:
\begin{equation}
\small
\begin{aligned}
 \pi^p(v_{i,t},v_{j,t}) = \mathbf{f}^u(v_{j,t}|v_{i,t})^\tau (\mathbf{p}(v_{i,t}) \odot \mathbf{w}^p(v_{i,t}))+\mathbf{f}(v_{i,t}|v_{j,t})^\tau (\mathbf{p}(v_{j,t}) \odot \mathbf{w}^p(v_{j,t})) 
\end{aligned}
\end{equation}

The homophily score $\pi^h(v_{i,t},v_{j,t})$, concerned with feature differences, incorporates the preference for the other nodes with similar/dissimilar crisp/fuzzy features:
\begin{equation}
\small
\begin{aligned}
 \pi^h(v_{i,t},v_{j,t}) &=& \Delta\mathbf{f}(v_{i,t},v_{j,t}|v_{i,t})^\tau (\mathbf{h}(v_{i,t}) \odot \mathbf{w}^h(v_{i,t}))
  + \Delta\mathbf{f}(v_{i,t},v_{j,t}|v_{j,t})^\tau (\mathbf{h}(v_{j,t}) \odot \mathbf{w}^h(v_{j,t}))
\end{aligned}
\end{equation}

The score list $\Pi_t = \{\pi(v_{i,t},v_(j,t))|v_{i,t},v_{i,t}\in V, i \neq j\}$ keeps frozen with fixed features and feature preference and representation principles. Given the required edge number $\lambda_{e,t}$, we create a static network to represent the first $\lambda_{e,t}$ discrete interactions at this fixed time point.

Based on the respective set-ups in section \ref{Fr} and section \ref{Fp}, the nodes in the toy example evaluate and select preferred connections considering the sex and age features. The female and middle-aged nodes around $35$ are preferred and, thus, likely to have more connections than the other nodes. Meanwhile, the nodes with the same gender and an age difference of around $30$ tend to cluster in the network formation process. However, it is hard to determine the most popular nodes with the superposition of preferential attachment and homophily principles. We must create more realistic networks based on real features to explore feature-driven network patterns.

\subsection{Epidemic transmission}
The epidemic spreads from the seed nodes $\mathbf{S}_t=\{\cdots,v_{i,t},\cdots|v_{i,t}\in V_t\}$. A healthy node, given a connection with an infected node, gets infected at the probability of $\gamma_t$. Correspondingly, the infection probability $\mathcal{P}^s$ of a healthy node $v_{i,t}$ increases with the increased connections with the infected nodes.
\begin{equation}
    \mathcal{P}^s(v_{i,t}) = 1-\prod\limits_{v_{j,t}\in \mathbf{V}_t}(1-\gamma_t)^{e(v_{i,t},v_{j,t})}
\end{equation}
where $\prod\limits_{v_{j,t}\in \mathbf{V}_t}(1-\gamma_t)^{e(v_{i,t},v_{j,t})}$ denotes the probability of keeping healthy given the connections $e(v_{i,t},v_{j,t})$ with any other node. In contrast, the recovery rate $\eta_{t}$ denotes the probability of the recovery of any infected node despite of the connections with infected others.

The features and the corresponding preferences determine the network patterns, which impact the epidemic spreading process and the well-being of any node in a social network. Based on the toy example presented in section \ref{Fr}, section \ref{Fp}, and section \ref{Nf}, the popular nodes, with preferred features such as females at the age of around $30$, tend to have more epidemic risks due to higher number of connections with others. Meanwhile, nodes with similar genders and an age difference of around $30$ can also have higher infection risks due to the clustering network patterns. To create realistic social networks and to investigate the influence of features on the epidemic spreading process, we conduct an empirical analysis and present its results and discussion in section \ref{rep1-2section32}.

\section{Empirical Analysis}
\label{Rep1-2section3}

In this section, we use the proposed DT--CNS model to generate the physical contact networks for six countries (data available for Belgium, Finland, Germany, Luxembourg, Poland and Italy from 2005) and investigate their respective disaster resilience to an epidemic spread. Given partial observable information about this real scenario (as described in section \ref{Rep1-2data}), we build hybrid DT-CNSs based on both real and simulated data, including (i) a hybrid network built on real nodes' features and simulated edges and (ii) a simulation-based epidemic spreading process.

We build hybrid network models with available age \& sex features and feature representation principles (section \ref{rep1-2section31}) and evaluate their performance in simulating social networks and simulating realistic social contact matrices in section \ref{rep1-2section32}. The selected model is validated using sensitivity analysis considering the increasingly flexible feature representation principles in section \ref{rep1-2section33}. In section \ref{rep1-2section34}, we simulate the social networks of each country and simulate the epidemic spreading process as a reference for policy-making in disaster resilience.

\subsection{Data Description}
\label{Rep1-2data}

We have partial observable information about the social contact patterns in different countries, including the social contact matrices between different age groups, as disclosed on the website \footnotemark[1], and the age \& sex distributions of the population gathered for year 2005 (see Fig.~\ref{DNfeat2} in the appendix \ref{featapp}).

We obtain the age and sex distributions of the population in each country from the website \footnotemark[2] and, referring to these distributions, generate age features for each country. As shown in Fig.~\ref{DNfeat2} in the appendix \ref{featapp}, the age \&sex distributions of the five countries have a similar shape. There are more people in the middle age ranges (25-60) than in other age groups, and more females in the older groups (60-90+) than males. The entire population's age is between 0 and 90.

\footnotetext[1]{\url{http://www.socialcontactdata.org/socrates/}}
\footnotetext[2]{\url{https://www.populationpyramid.net/}}

We obtain social contact matrices for countries, including Belgium, Finland, Germany, Luxembourg, Italy and Poland, from the social contact data website \footnotemark[1]. The social contact matrix $\{m_{i,j}\}$ describes the average number of physical contacts between each pair of age classes reported weekly between May 2005 and September 2006 in a specific country's survey \citep{mossong2008social}. Considering the reciprocal nature of social connections, the social contact matrices $\{m_{i,j}\}$ are adjusted based on the population size in the respective age groups \citep{willem2020socrates}:
\begin{equation}
 m_{ij}^{reciprocal} = \frac{m_{ij}N_{i}+m_{ji}N_{j}}{2N_i}
 \label{eqq}
\end{equation}
where $N_i$ and $N_j$ each represents the population size in age group $i$ and $j$. $m_{ij}^{reciprocal}N_i$ represents the average number of social contacts between age group $i$ and $j$. Given a specific number of nodes and the corresponding age features, we can create their social contact matrix by adjusting the country's social contact matrix $m_{i,j}$ according to equation \ref{eqq} based on the information related to the population size $N_i$ and $N_j$ in the network simulations.

In this study, we assume $90$ nodes for the simulated social networks in each country. Given the social contact matrix, we derive a number of edges for each network. As shown in Tab.~\ref{DNedge},  the social networks generally have around $700$ edges. However, the social network of Finland only has $339$ edges, and in contrast, the social network of Poland has $800$ edges. 


\begin{table}[h]
\centering
\small
\caption{Number (No.) of nodes and edges.}
\label{DNedge}
\setlength{\tabcolsep}{3pt}
\renewcommand{\arraystretch}{1.5}
\begin{tabular}{|c|c|c|c|c|c|c|}
\hline
Country& Belgium & Finland & Germany & Italy& Luxembourg & Poland \\
\hline
No.Nodes& 90& 90& 90& 90& 90& 90\\
\hline
No.Edges&614&336&462&770&687&793\\
\hline
\end{tabular}
\end{table}

\subsection{Network Models Initialisation}
\label{rep1-2section31}
This section provides a description of the process to build network models to uncover the unobservable social preferences behind complex social interactions and unfold the social contact matrices into social networks (see Tab.~\ref{DNedge}). We also introduce different features and their representation principles to investigate their respective impact on the faithfulness of network representation (see Table.~\ref{HNmodel}). This involves the random network model $RN$ built on the random encounters of each node pair and the hybrid network models $HN$ that introduce various features (age and sex) and feature representation principles (crisp/fuzzy) and preferences to those features (a mixture of preferential attachment and homophily).

\begin{table}[htp]
\centering
\small
\caption{Model description.}
\label{HNmodel}
\setlength{\tabcolsep}{3pt}
\renewcommand{\arraystretch}{1.5}
\begin{tabular}{|c|c|c|c|}
\hline
\multirow{2}{*}{Model Name}& \multirow{2}{*}{Features} &\multicolumn{2}{|c|}{Rules} \\
\cline{3-4}
& &Representation &Preference \\
\hline
\multirow{1}{*}{$RN$}&---- & ---- & ---- \\
\hline
\multirow{1}{*}{$HN-A_{c}$}& Age & Crisp & Preferential attachment \& Homophily \\
\hline
\multirow{1}{*}{$HN-A_{f}$}& Age &Fuzzy & Preferential attachment \& Homophily \\
\hline
\multirow{2}{*}{$HN-A_c-S_c$}&Age & Crisp & Preferential attachment \& Homophily \\
\cline{2-4}
&Sex & Crisp & Preferential attachment \& Homophily \\
\hline
\multirow{2}{*}{$HN-A_f-S_c$}&Age & Fuzzy &Preferential attachment \& Homophily \\
\cline{2-4}
&Sex & Crisp & Preferential attachment \& Homophily \\
\hline
\end{tabular}
\end{table}


\subsubsection{Features}

\begin{figure}[H] 
	\centering
	\subfigure[Belgium]{
		\begin{minipage}[b]{0.32\linewidth}
			\includegraphics[width=1\linewidth]{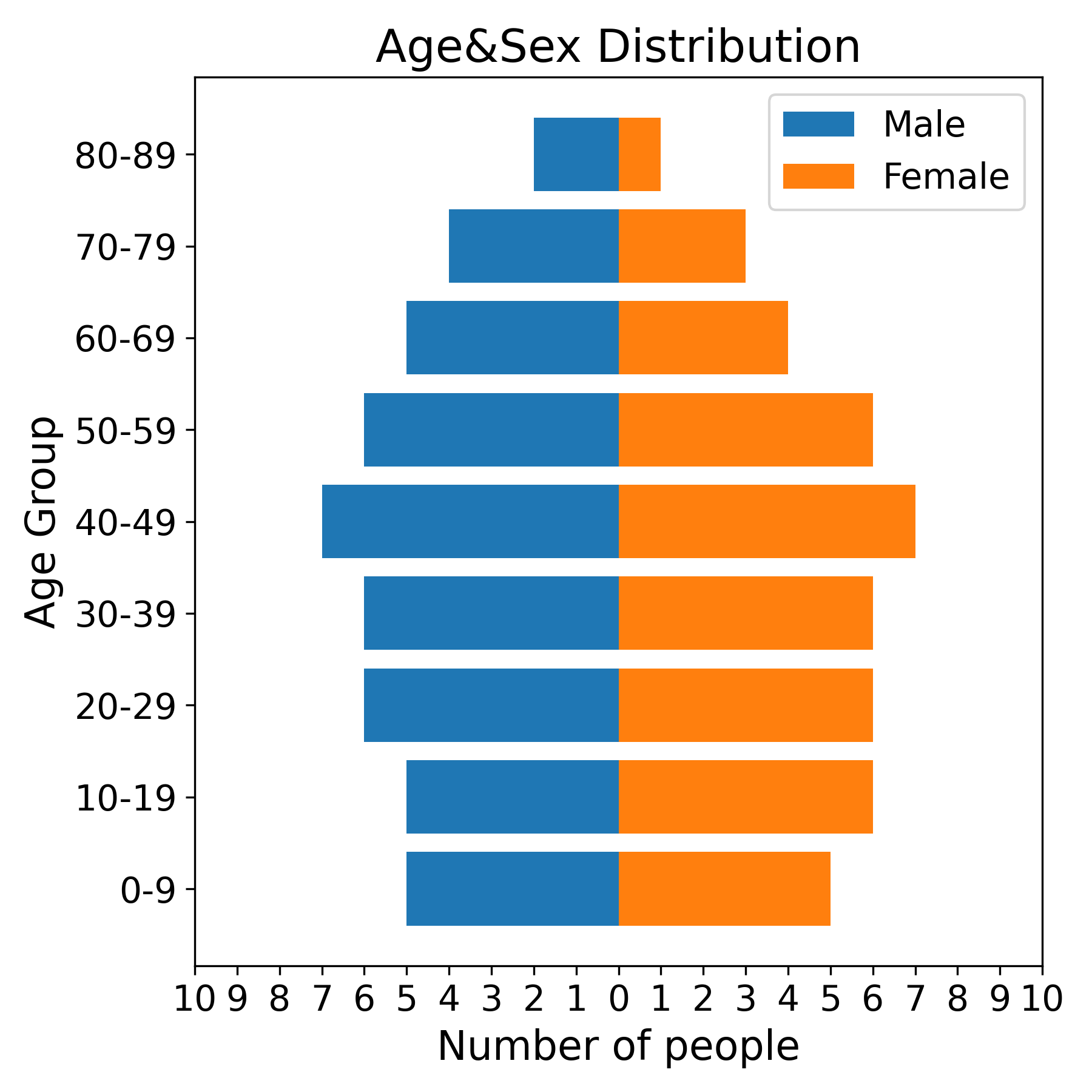}
	\end{minipage}}
	\subfigure[Finland]{
		\begin{minipage}[b]{0.32\linewidth}
			\includegraphics[width=1\linewidth]{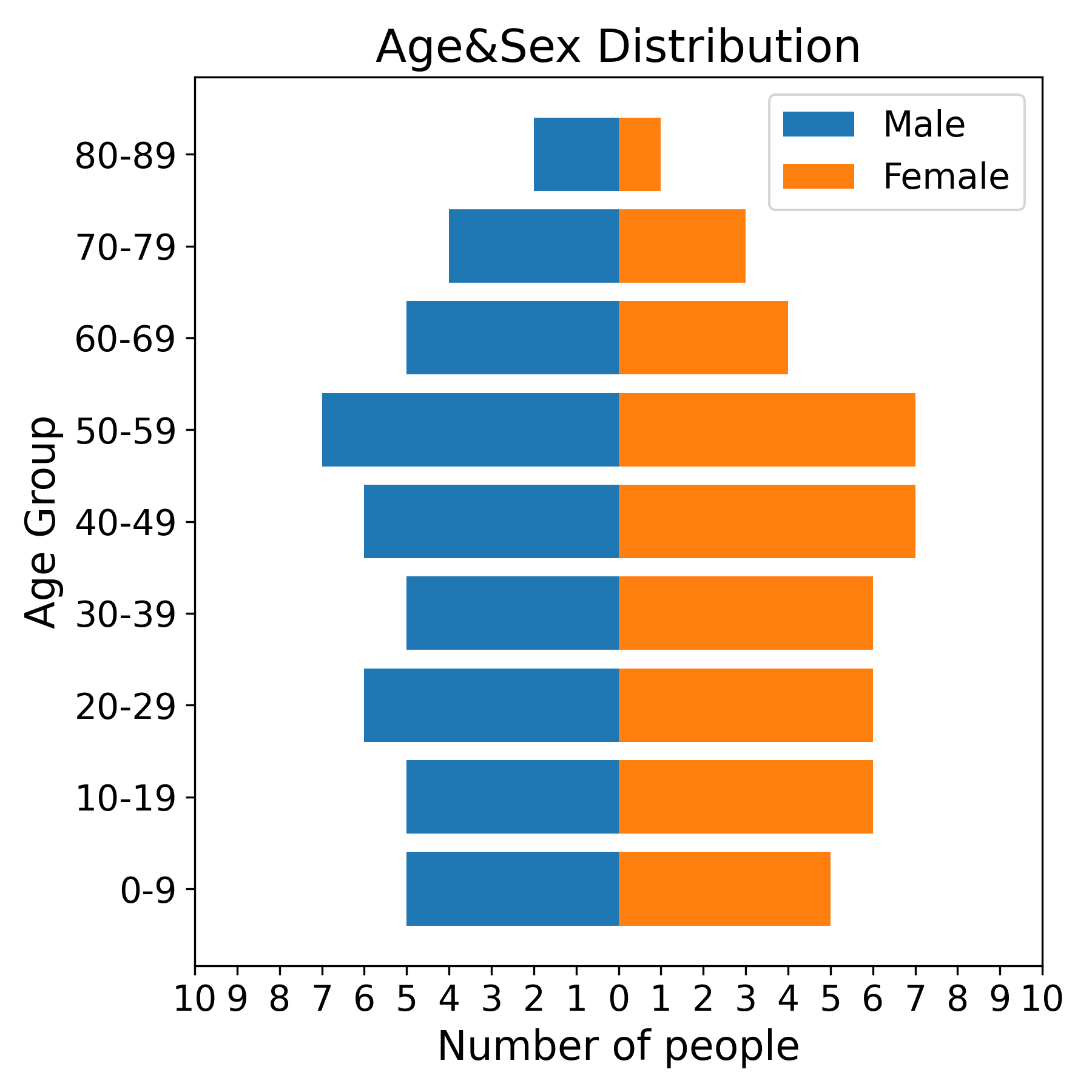}
	\end{minipage}}
	\subfigure[Germany]{
		\begin{minipage}[b]{0.32\linewidth}
			\includegraphics[width=1\linewidth]{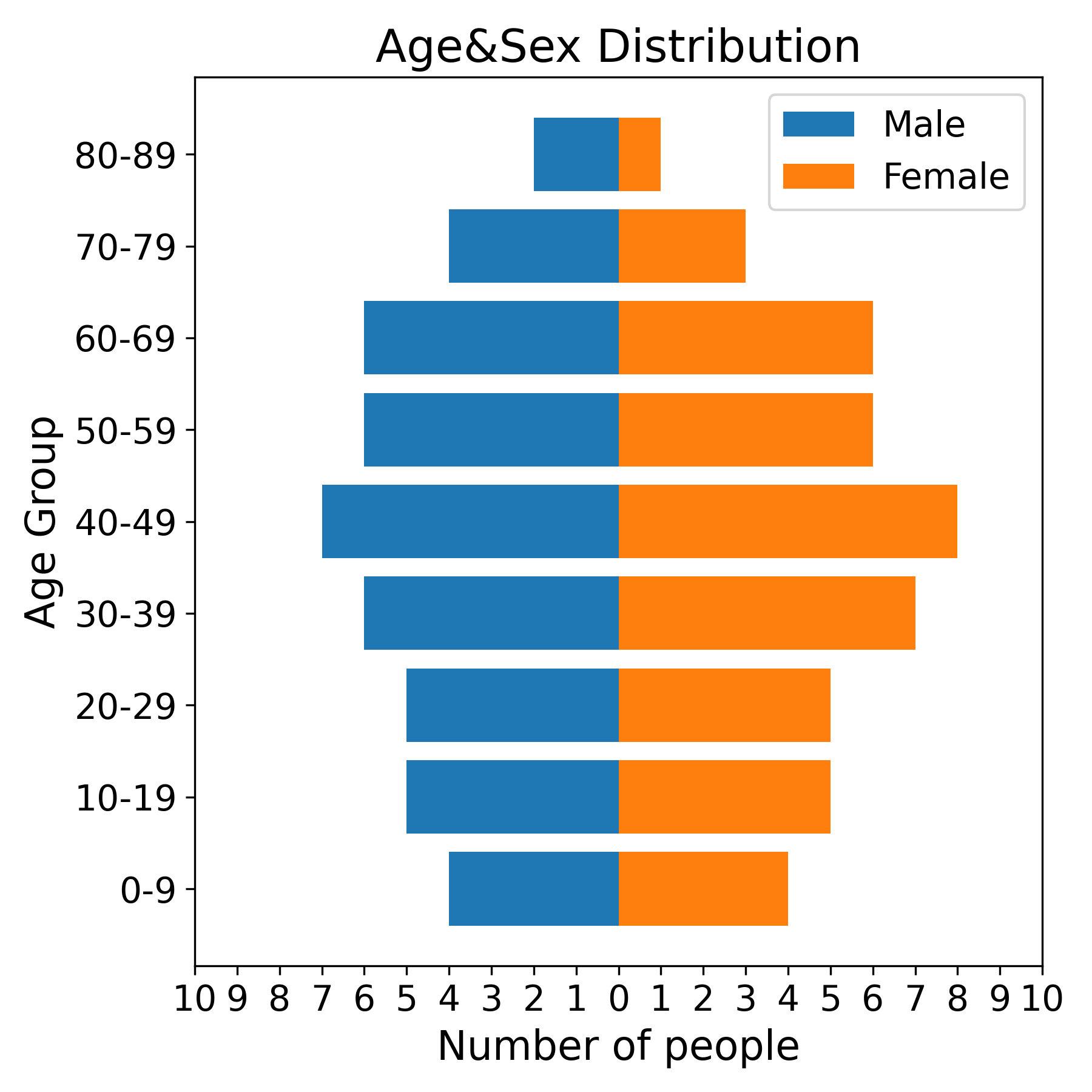}
	\end{minipage}}\\
		\subfigure[Italy]{
		\begin{minipage}[b]{0.32\linewidth}
			\includegraphics[width=1\linewidth]{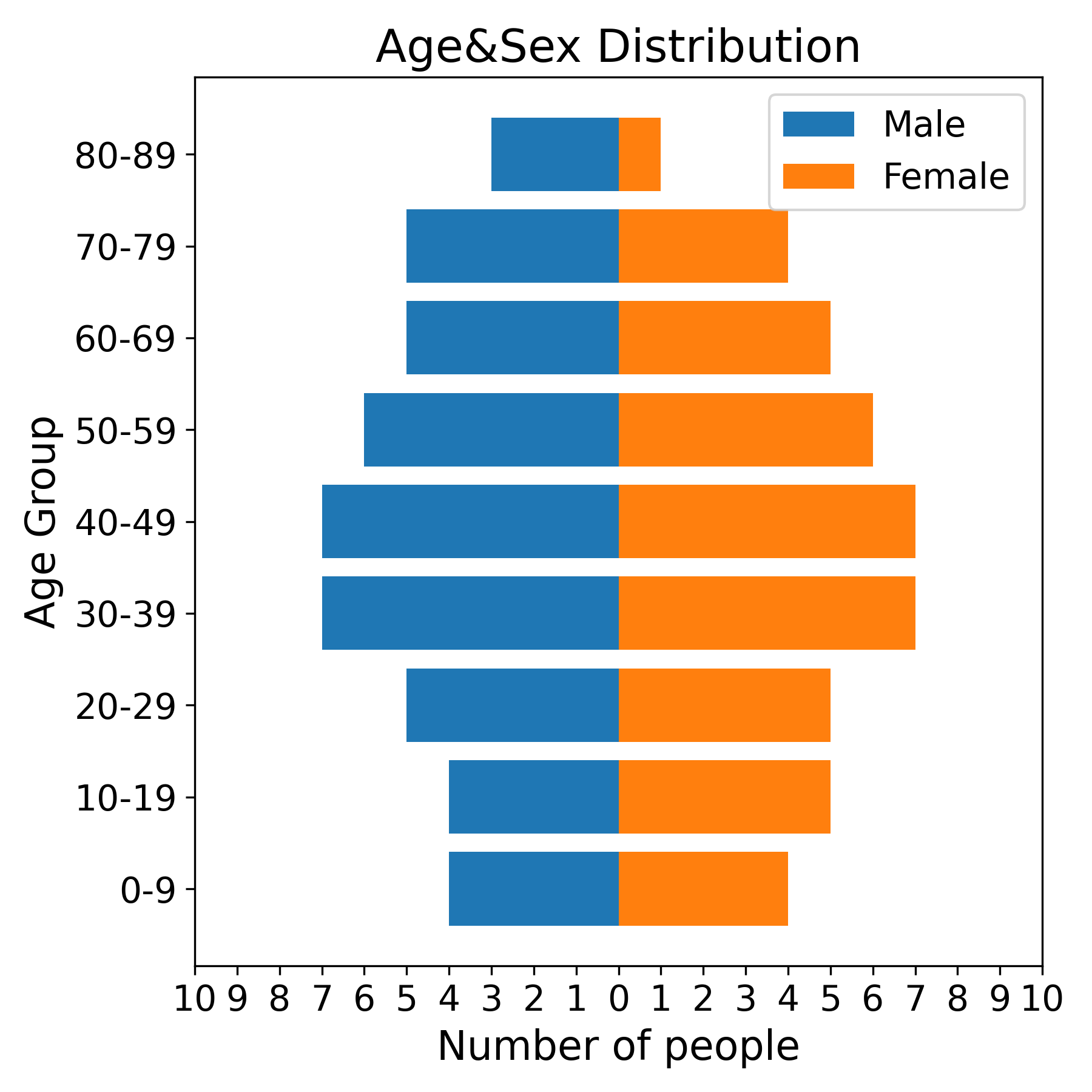}
	\end{minipage}}
	\subfigure[Luxembourg]{
		\begin{minipage}[b]{0.32\linewidth}
			\includegraphics[width=1\linewidth]{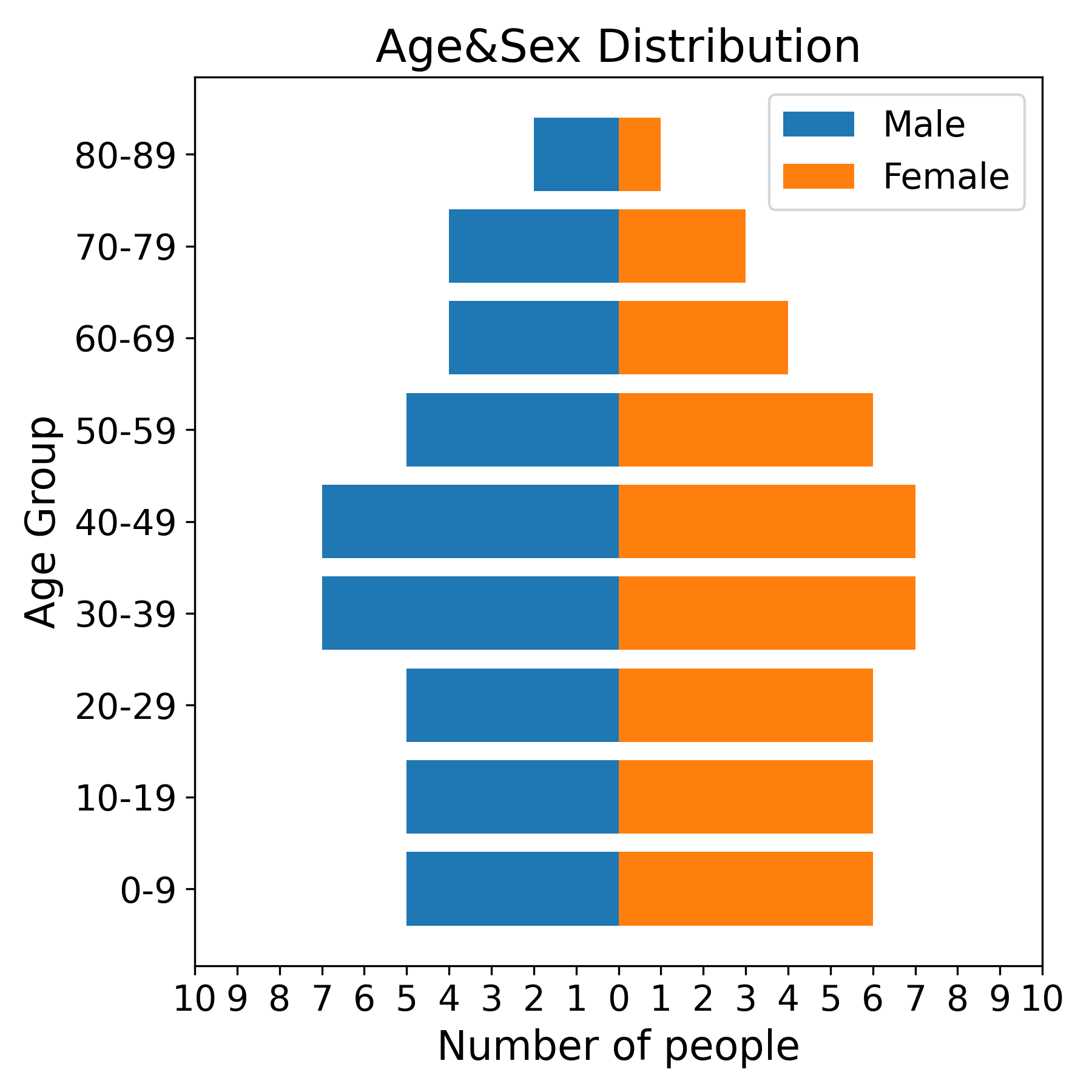}
	\end{minipage}}
		\subfigure[Poland]{
		\begin{minipage}[b]{0.32\linewidth}
			\includegraphics[width=1\linewidth]{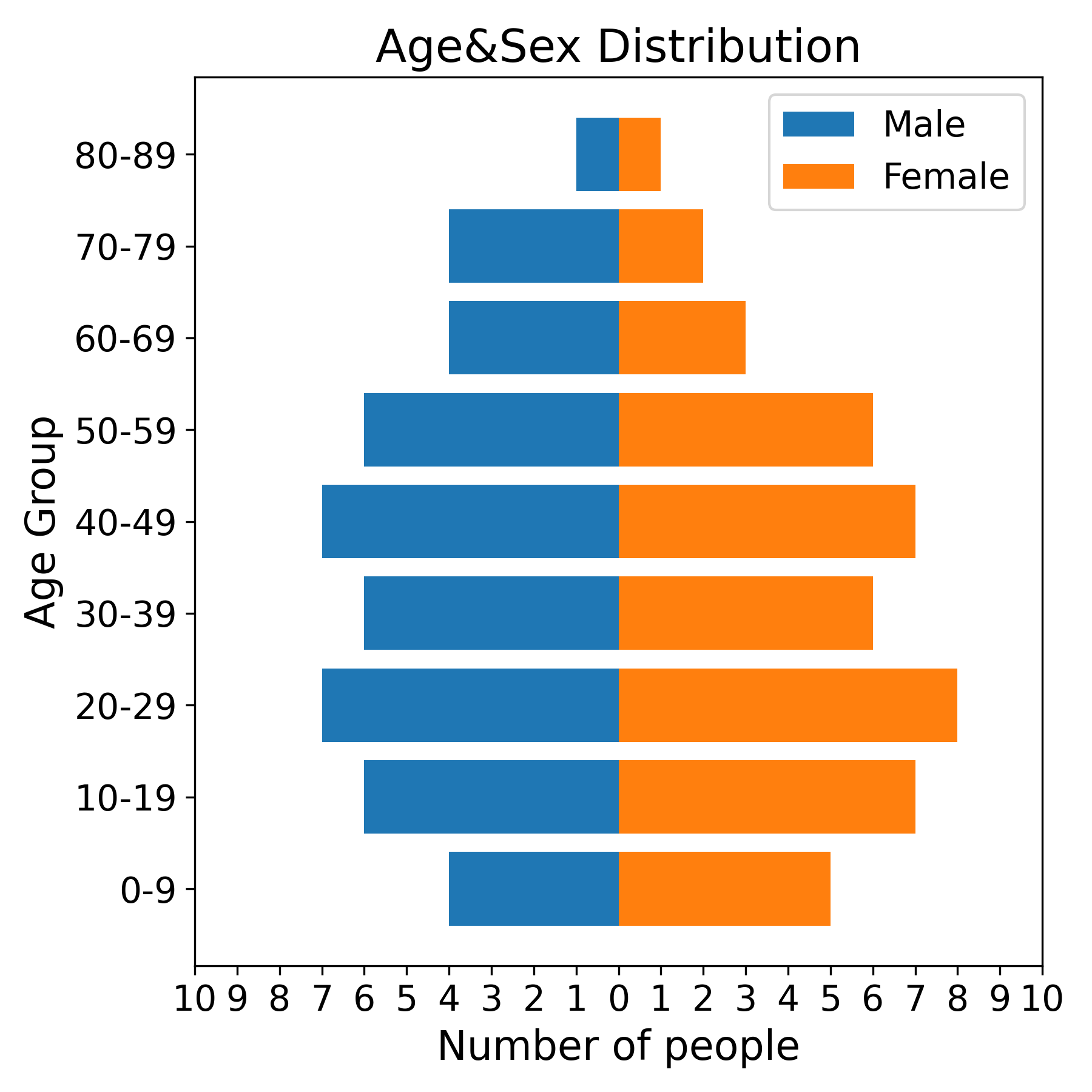}
	\end{minipage}}
	\caption{The simulated age and sex-distributions in 2005 for countries including (a) Belgium, (b) Finland, (c) Germany, (d) Italy, (e) Luxembourg and (f) Poland.}
\label{DNfeat22}
\end{figure}

Features, including the age and the sex of nodes, are generated according to the age and sex-distributions of each country in 2005 (see Fig.~\ref{DNfeat2} for real data and Fig.~\ref{DNfeat22} for simulated data). As the population age is between $0$ and $90$, we assume the age range in the simulations as $[0,89]$ and simulate the age and sex features accordingly. We classify the $90$ nodes into eighteen age and sex groups, considering the age classes separated every ten years and the differences between males and females. Within each age and sex class, we randomly simulate the age features for each of the nodes.    

\begin{figure}[H] 
	\centering
	\subfigure[Age Group]{
		\begin{minipage}[b]{0.32\linewidth}
			\includegraphics[width=1\linewidth]{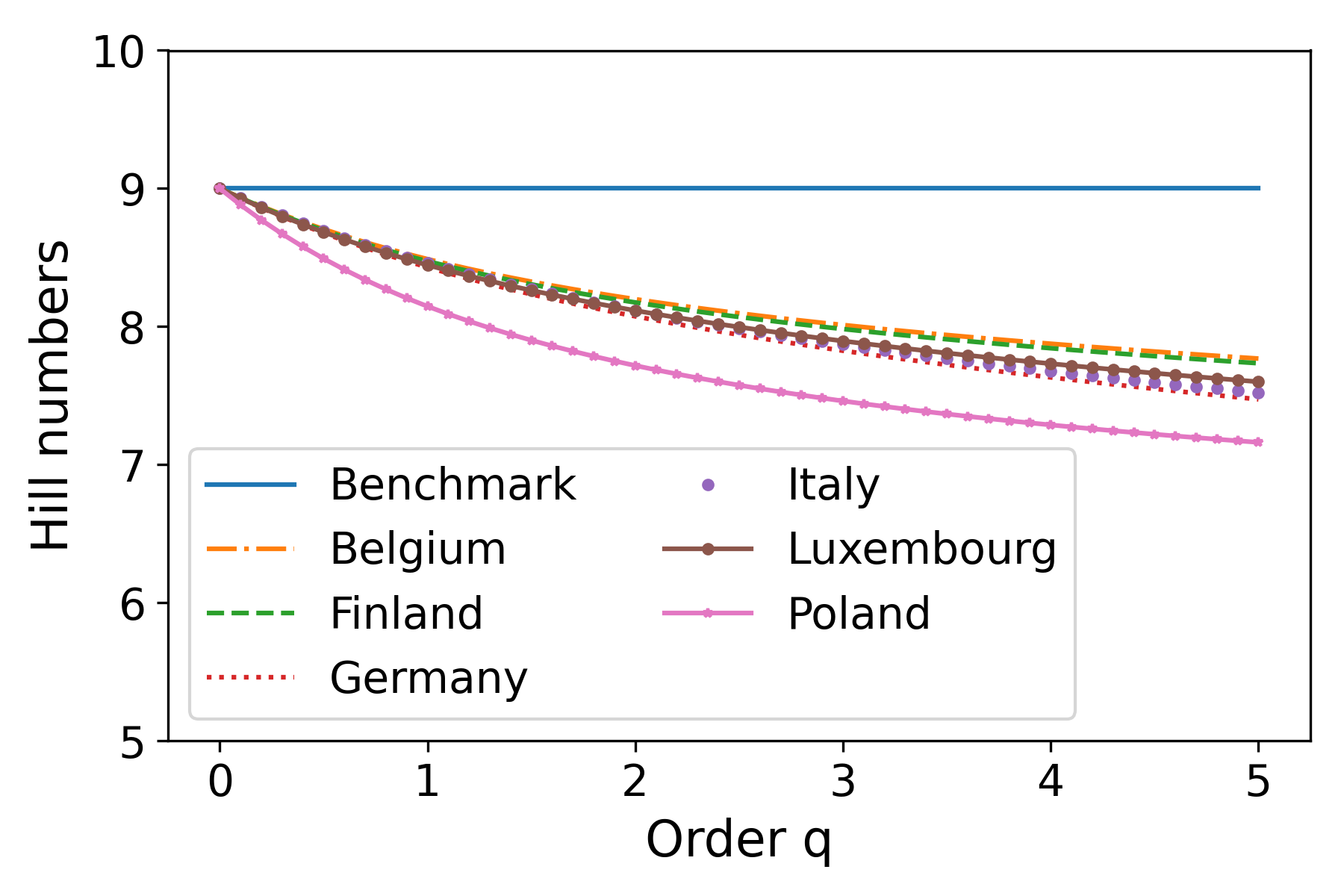}
	\end{minipage}}
	\subfigure[Sex Group]{
		\begin{minipage}[b]{0.32\linewidth}
			\includegraphics[width=1\linewidth]{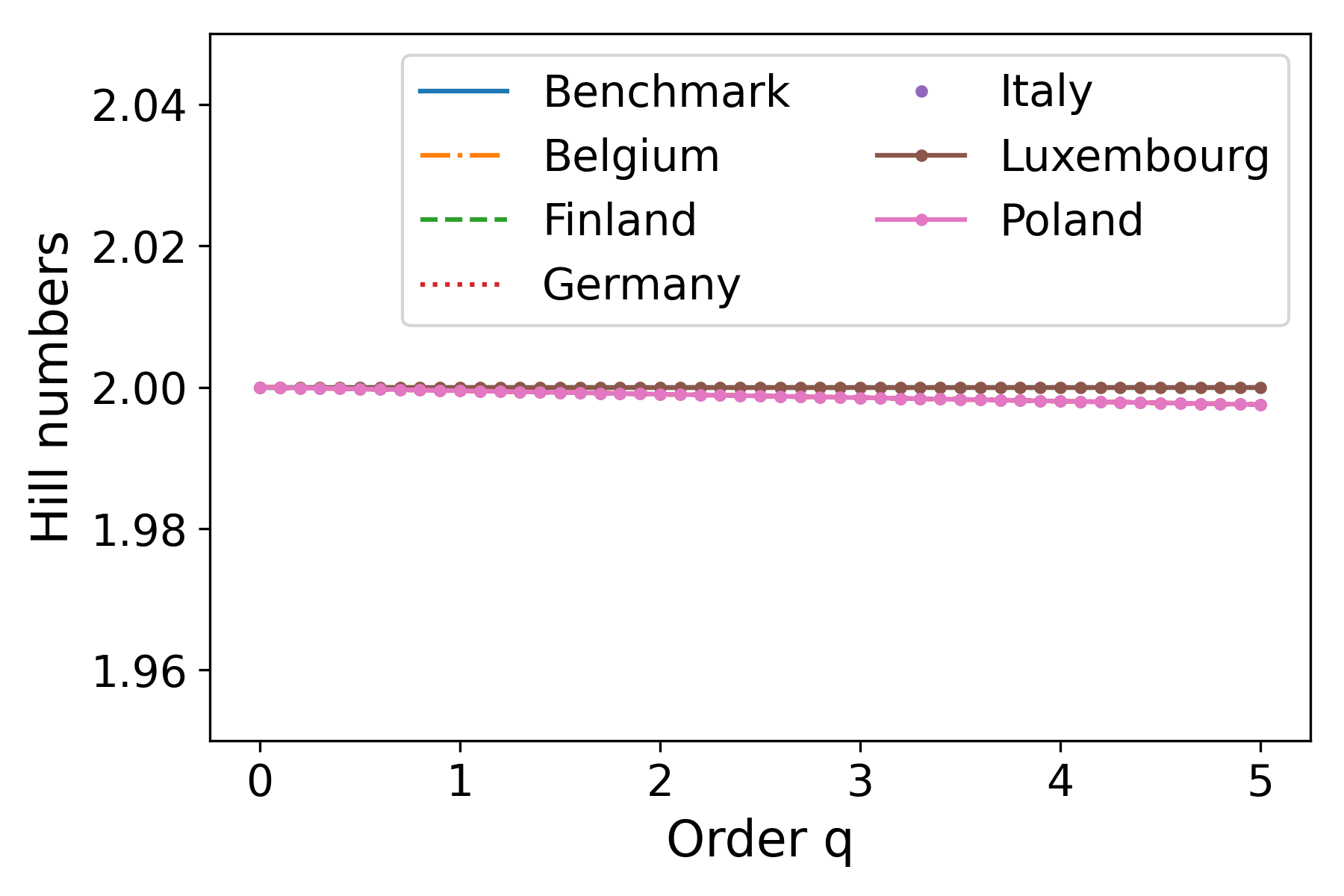}
	\end{minipage}}
	\subfigure[Age\&Sex Group]{
		\begin{minipage}[b]{0.32\linewidth}
			\includegraphics[width=1\linewidth]{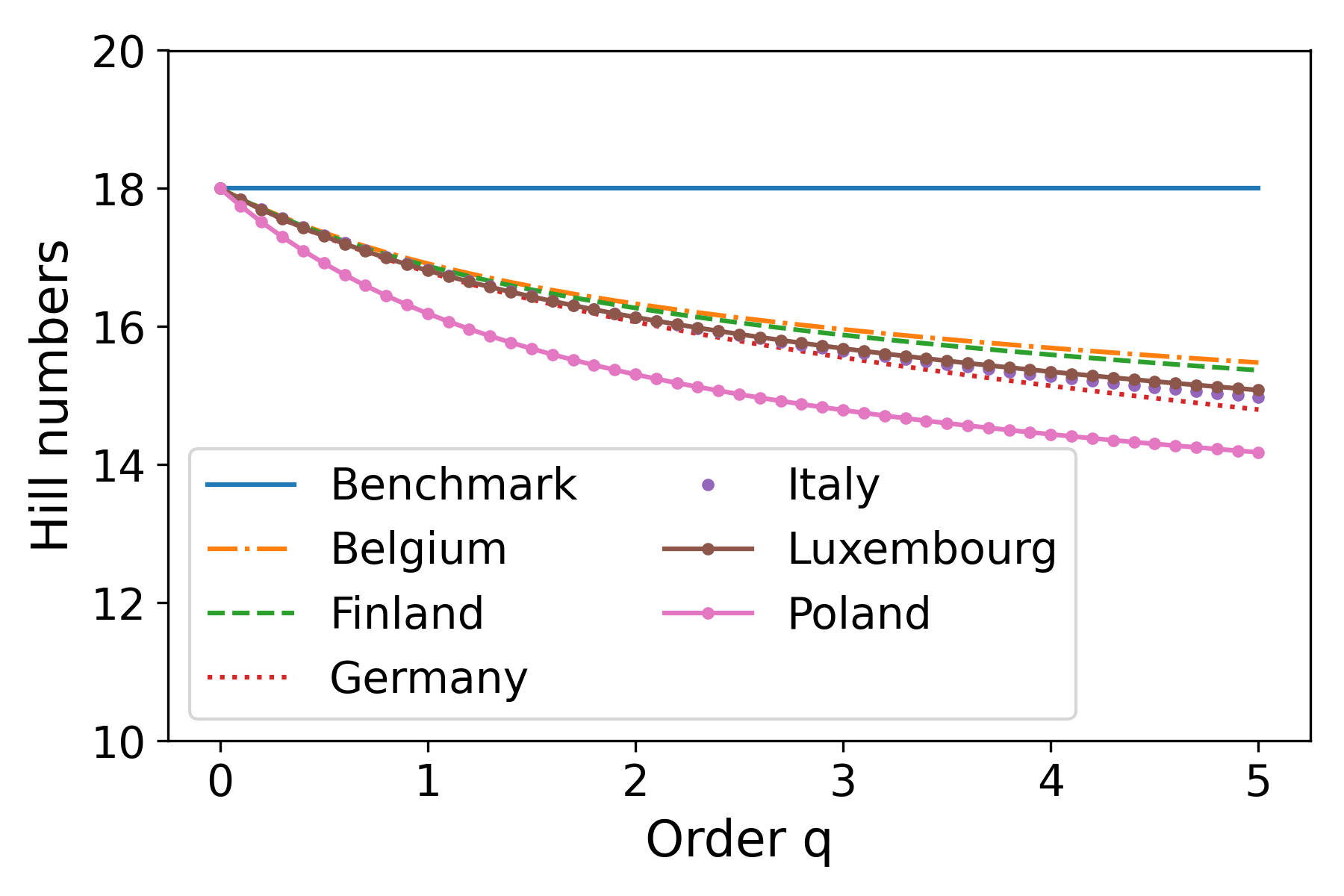}
	\end{minipage}}\\
	\caption{The diversity of the population in each country, as measured by the Hill numbers based on age, sex and age\&sex groups.}
\label{DNdiversity}
\end{figure}

Based on the differences in the simulated age distribution, sex distribution and age and sex distribution, we analyse the diversity of each country's population by calculating the Hill numbers (see Fig.~\ref{DNdiversity}). The Hill numbers are calculated with an order value $q$. A higher $q$ value indicates a higher sensitivity level to the relative frequencies of the species (different groups considering the features of age and sex). In addition, higher Hill numbers represent a higher diversity level, characterised with higher abundance of the respective species and more even distribution of nodes in the respective species \citep{qiu2016effects}. Therefore, the sex diversity of the countries is similar, which approaches the value of 2, which is the most diverse case (see Fig.~\ref{DNdiversity}(b)). The age diversity of the countries varies depending on the country, which deviates from the most diverse case possible at the value of 9 (see Fig.~\ref{DNdiversity}(a)). The diversity of the population increases a lot when considering the differences in age and sex. The differences in each country's population diversity have also increased (see Fig.~\ref{DNdiversity}(c)). Generally speaking, Belgium has the highest level of diversity given any diversity references (age, sex, or age\&sex) considering the calculated Hill numbers.

\subsubsection{Feature representation and preference} 

\label{principleselection}
Principles involved in the hybrid modelling of social networks cover two aspects: (i) feature representation principles and (ii) feature preference principles. 

\begin{figure}
 \centering
 \includegraphics[width=0.8\linewidth]{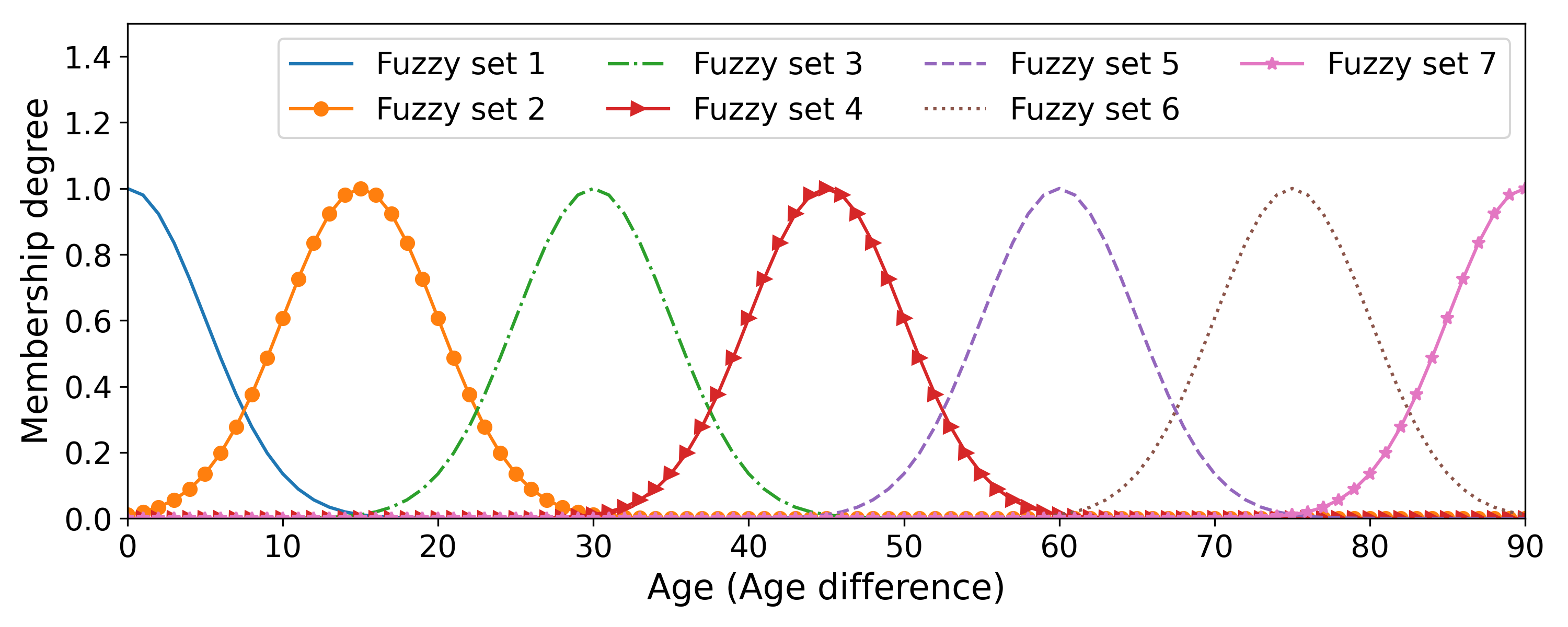}
 \caption{The fuzzy representation of age (age difference).}
 \label{fuzzyage}
\end{figure}

In this study, we represent the sex features with binary values, where males and females are each represented respectively as ones and zeros with a crisp boundary. The representation of age features follows either a crisp or fuzzy principle. Given a crisp principle, we represent the age and the age differences with their corresponding crisp values. Given a fuzzy principle, we initialise seven fuzzy sets to represent the uncertain inductions of the ages and the age differences. Each fuzzy set describes the respective membership of age/age difference to each abstract age/age difference group (see Fig.\ref{fuzzyage}).

As shown in Fig.~\ref{fuzzyage}, we represent age (age difference) value with a vector incorporating seven membership values for the corresponding age ranges. We employ Gaussian membership functions to describe the membership of a single age (age difference) value to the age ranges that centre around the list of $\mu$ values, each initialised as $0$, $15$, $30$, $45$, $60$, $75$ and $90$. In addition, we initialised the standard deviations of each Gaussian membership function with an identical value at $5$.

We employ preferential attachment and homophily principles to describe the preference for features and generate social networks. With a crisp representation principle, the homophily principle describes the preference for similar/dissimilar features by preferring small/large feature differences. In contrast, the preferential attachment principle describes the preference for small/large features values. In contrast, with a fuzzy representation principle, the homophily principle transforms to the preference for fuzzy feature difference ranges. The preferential attachment principle describes the preference for fuzzy feature value ranges.

\subsection{Model Selection considering Features and Rules}
\label{rep1-2section32}

Given the ground-truth information about social contact matrices and age and sex distributions, we optimise the 
preferences and weights of preferences for the age and sex to generate social networks. Based on an assumed encounter rate at $0.8$, we can also recreate the corresponding social contact matrices based on the social networks simulated by the network models. 
Referring to our previous report \cite{wen2023dtcns}, we measure the faithfulness of the network representation by local-level similarity, reproducible network topology and modelling efficiency.

In our empirical analysis, we ensure the \textbf{reproducibility} of the same networks by setting a random seed for the Bernoulli distribution that generates binary values to represent encounters. We can recreate the same networks with the same sDNA and encounter rate. In addition, we measure the \textbf{efficiency} of generating hybrid networks with the runtime of a simulation. As shown in Tab.~\ref{HNfaithful}, the runtime of simulations does not vary much with the feature values changes, given a crisp representation principle. In contrast, the runtime increases when a fuzzy representation principle is employed, caused by the fuzzification process. In addition, the runtime of simulation differs depending on the country, which increases with the required number of edges (see Tab.~\ref{DNedge}).

\begin{table}[h]
\centering
\small
\caption{The similarity and the efficiency of network representation with the network models, as measured by the Euclidean distance ($EU$ distance) between social contact matrices and the Runtime (second).}
\label{HNfaithful}
\setlength{\tabcolsep}{3pt}
\renewcommand{\arraystretch}{1.5}
\begin{tabular}{|c|c|c|c|c|c|c|c|}
\hline
&Country& Belgium & Finland & Germany & Italy& Luxembourg & Poland \\
\hline
\multirow{2}{*}{$RN$} &$EU$ distance & 5.51 & 4.80 & 4.60 & 11.62 & 8.48 & 9.37\\
\cline{2-8}
&Runtime & 0.31&  0.23&  0.22&  0.24&  0.23&  0.23\\
\hline
\multirow{2}{*}{$HN-A_{c}$}&$EU$ distance & 5.76 & 4.16 & 4.68 & 10.39 & 7.36 & 8.06\\
\cline{2-8}
&Runtime & 0.24&  0.22&  0.43&  0.24&  0.23&  0.23\\
\hline
\multirow{2}{*}{$HN-A_{f}$}&$EU$ distance &  6.30&  4.08&  4.15 & 10.40 &7.19 & 7.78\\ 
\cline{2-8}
&Runtime &    0.62&  0.62&  0.63& 0.62&  0.66&  0.63\\
\hline
\multirow{2}{*}{$HN-A_c-S_c$}&$EU$ distance &  4.51 & 3.65 &3.82& 9.19&5.84 &  6.63\\
\cline{2-8}
&Runtime &  0.33&  0.31&  0.31&  0.32&  0.32& 0.32\\
\hline
\multirow{2}{*}{$HN-A_f-S_c$}&$EU$ distance & 4.02  & 2.55&  3.32 & 9.05 &5.40& 5.62\\
\cline{2-8}
&Runtime &    0.71&  0.71&  0.76&  0.73&  0.71& 0.74\\
\hline
\end{tabular}
\end{table}

For the \textbf{faithfulness}, we measure the local similarity between real and hybrid networks based on the Euclidean distance ($EU$ distance) between the social contact matrices. As shown in Tab.~\ref{HNfaithful}, the $EU$ distance generally decreases with the increasing number of features and the introduction of fuzzy age (age difference). This implies the positive impact of increasing structural complexity and dynamics complexity on social network simulations, each concerning the introduction of heterogeneous features and flexible fuzzy representation principles. Generally, $HN-A_f-S_c$ models enable the creation of the most similar hybrid social networks to those in real world for all the countries with the highest complexity level in both structural and dynamics dimensions.

To investigate the influence of structural and dynamics complexity on network generation, we simulate the social networks for each country with the optimised $HN$ models and recreate the social contact matrices. While we use Finland for the illustration purposes (see Fig.~\ref{FinlandMat}) the plots for the other countries can be found in the appendix (see Fig.~\ref{BelgiumMat}, Fig.~\ref{GermanyMat}, Fig.~\ref{ItalyMat}, Fig.~\ref{LuxembourgMat} and Fig.~\ref{PolandMat} in the appendix \ref{appMS}).. 

\begin{figure}[htp] 
	\centering
	\subfigure[Target]{
		\begin{minipage}[b]{0.42\linewidth}
			\includegraphics[width=1\linewidth]{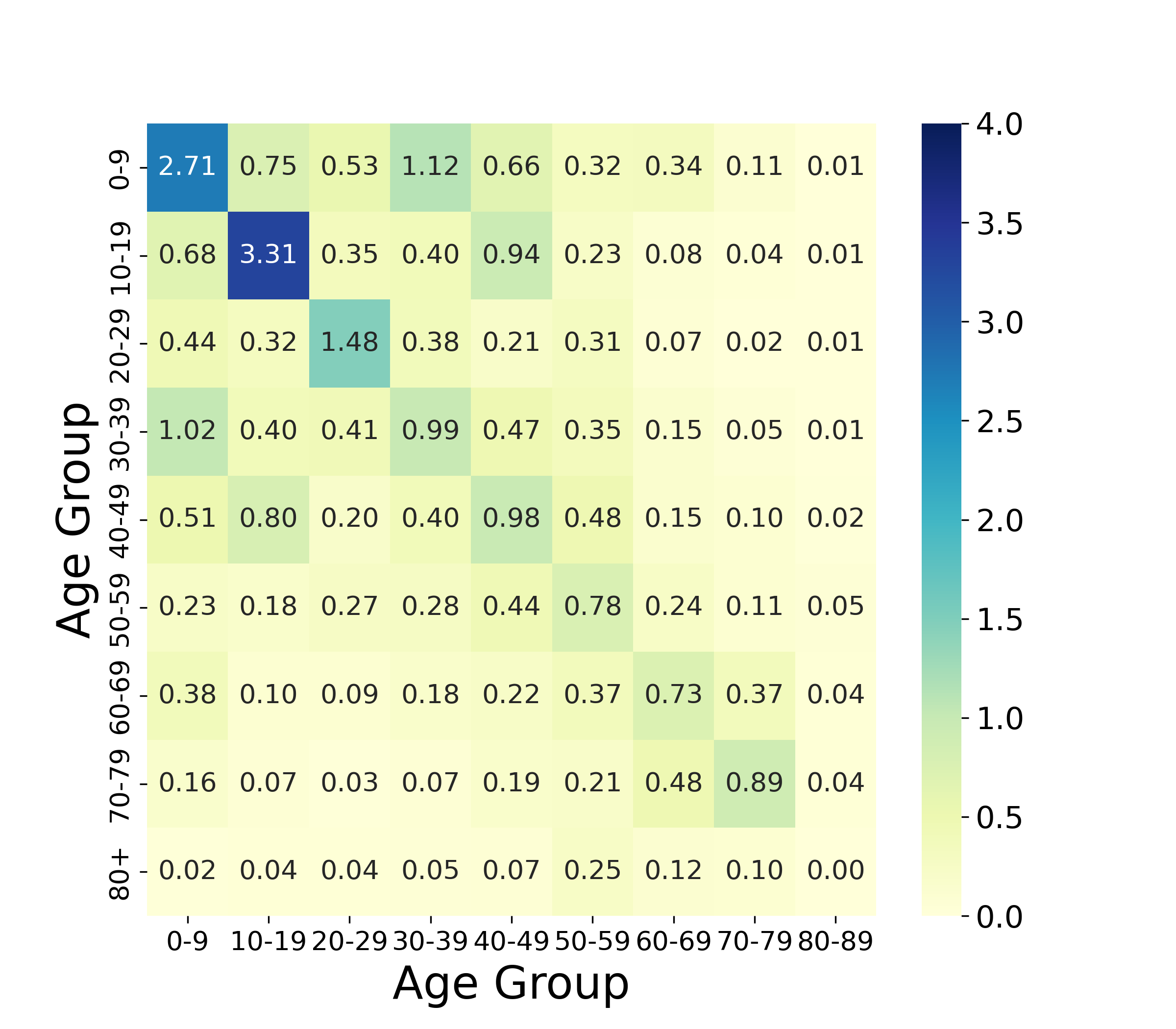}
	\end{minipage}}
  \hspace{-5mm}
	\subfigure[$RN$]{
		\begin{minipage}[b]{0.42\linewidth}
			\includegraphics[width=1\linewidth]{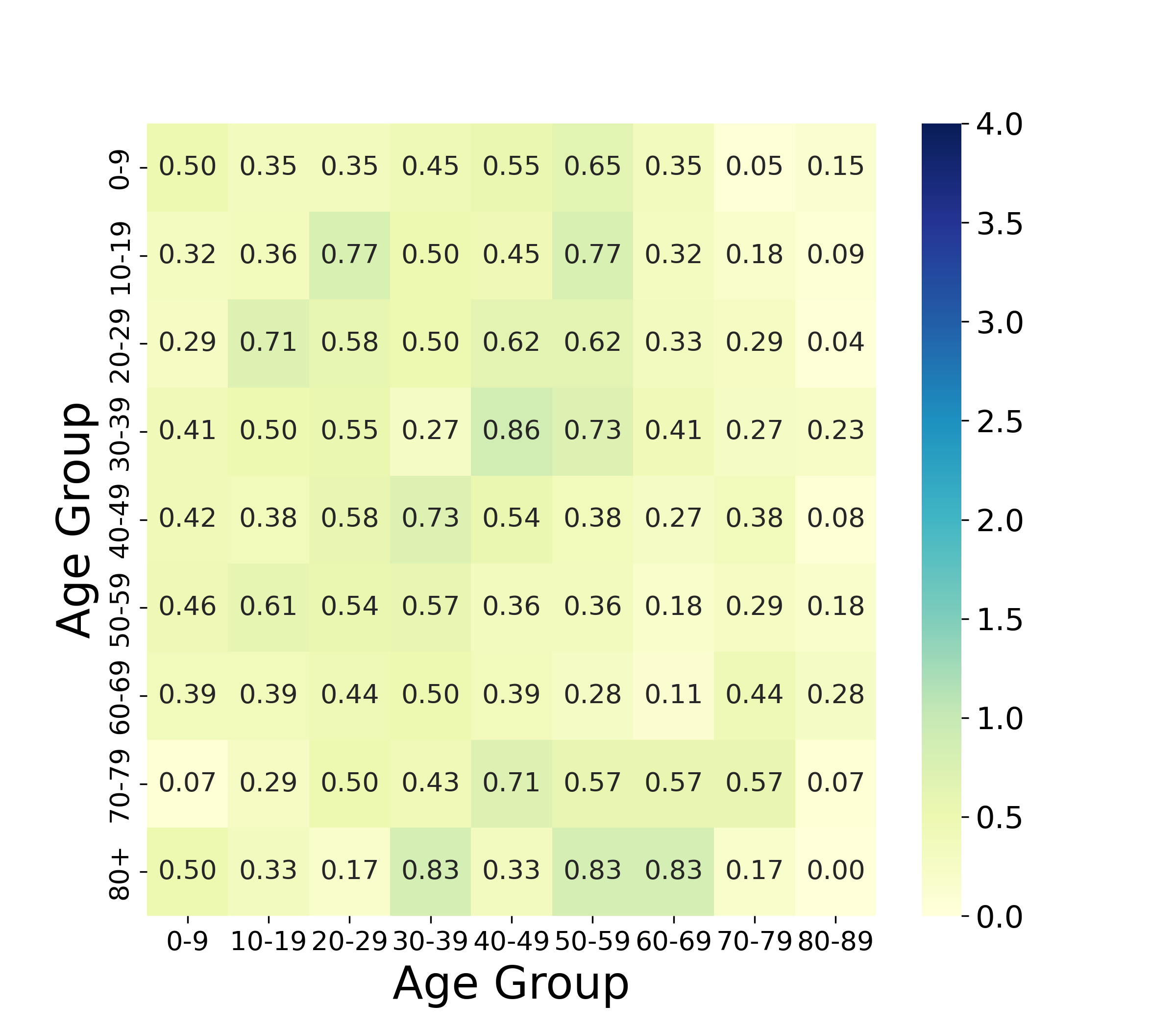}
	\end{minipage}}\\
  \vspace{-3mm}
	\subfigure[$HN-A_c$]{
		\begin{minipage}[b]{0.42\linewidth}
			\includegraphics[width=1\linewidth]{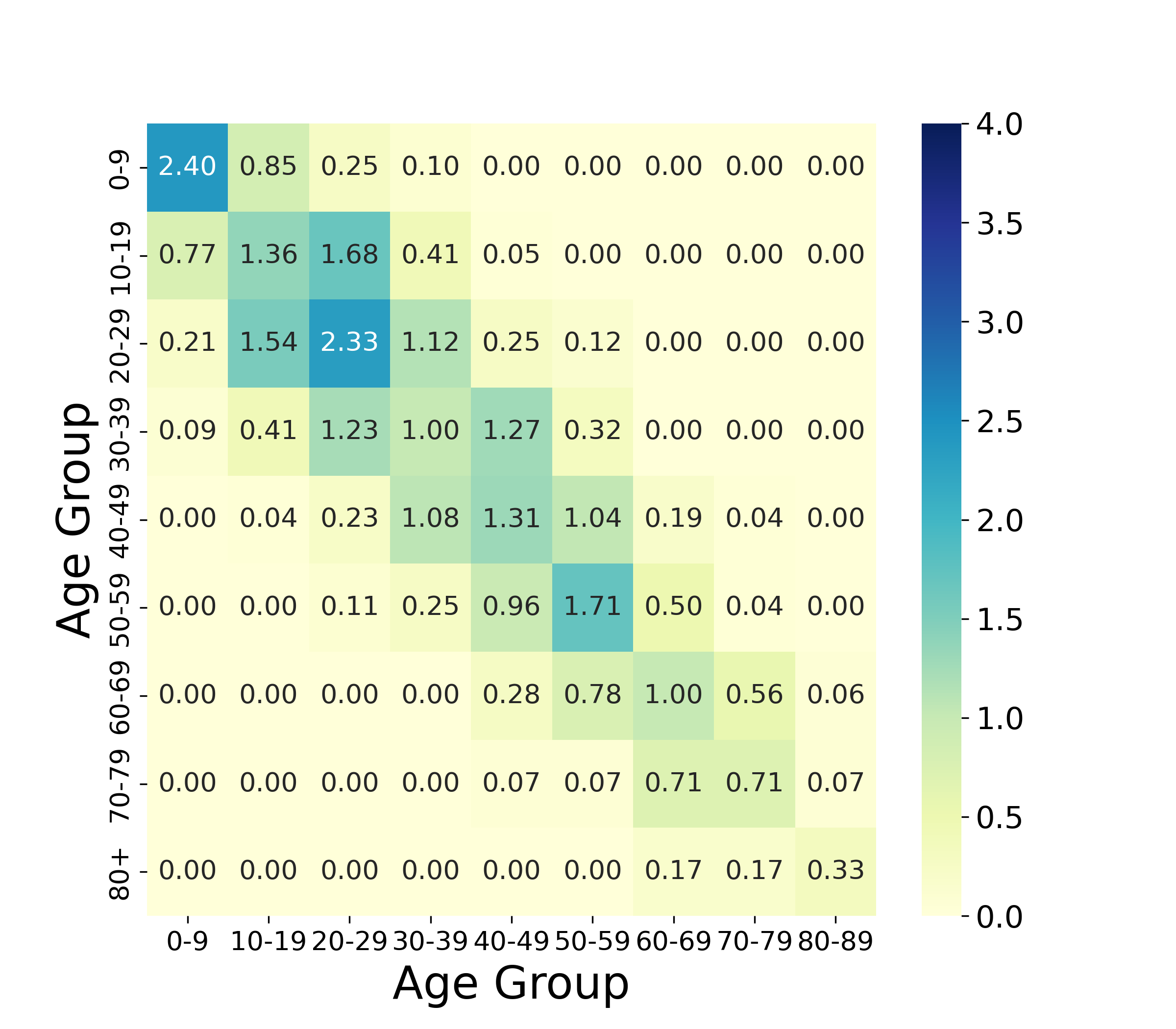}
	\end{minipage}}
  \hspace{-5mm}
		\subfigure[$HN-A_f$]{
		\begin{minipage}[b]{0.42\linewidth}
			\includegraphics[width=1\linewidth]{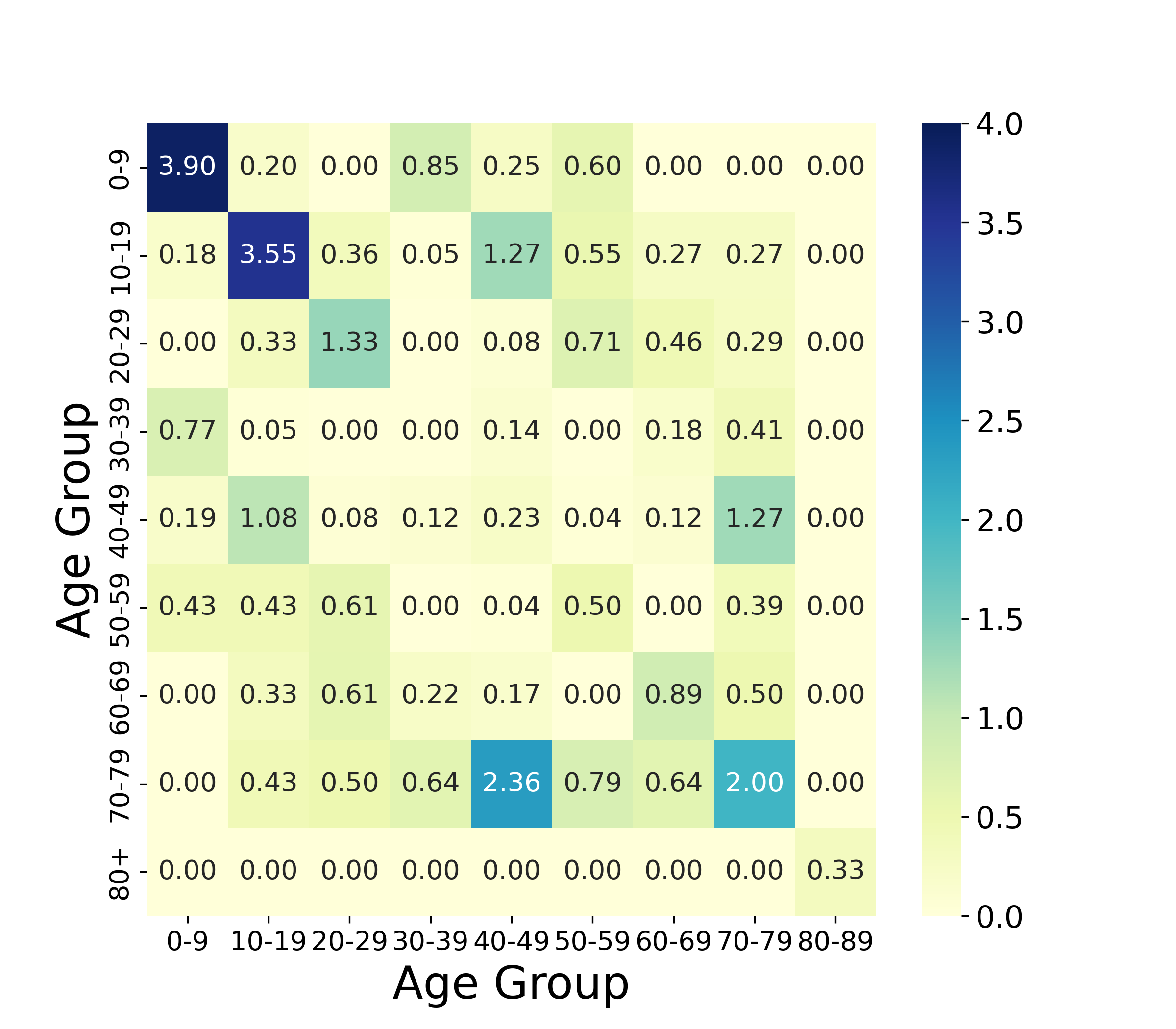}
	\end{minipage}}\\
  \vspace{-3mm}
		\subfigure[$HN-A_c-S$]{
		\begin{minipage}[b]{0.42\linewidth}
	\includegraphics[width=1\linewidth]{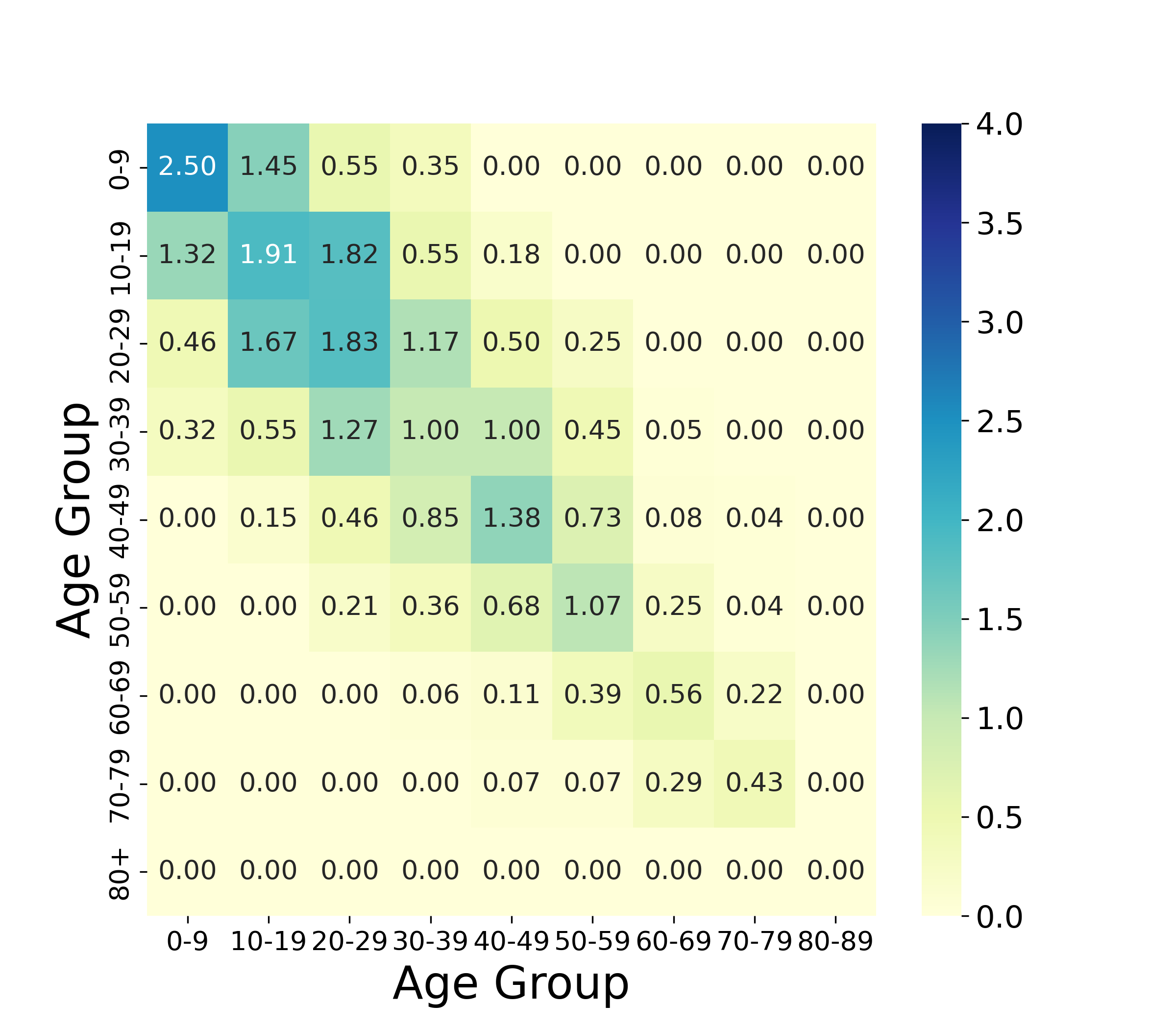}
	\end{minipage}}
  \hspace{-5mm}
		\subfigure[$HN-A_f-S$]{
		\begin{minipage}[b]{0.42\linewidth}
			\includegraphics[width=1\linewidth]{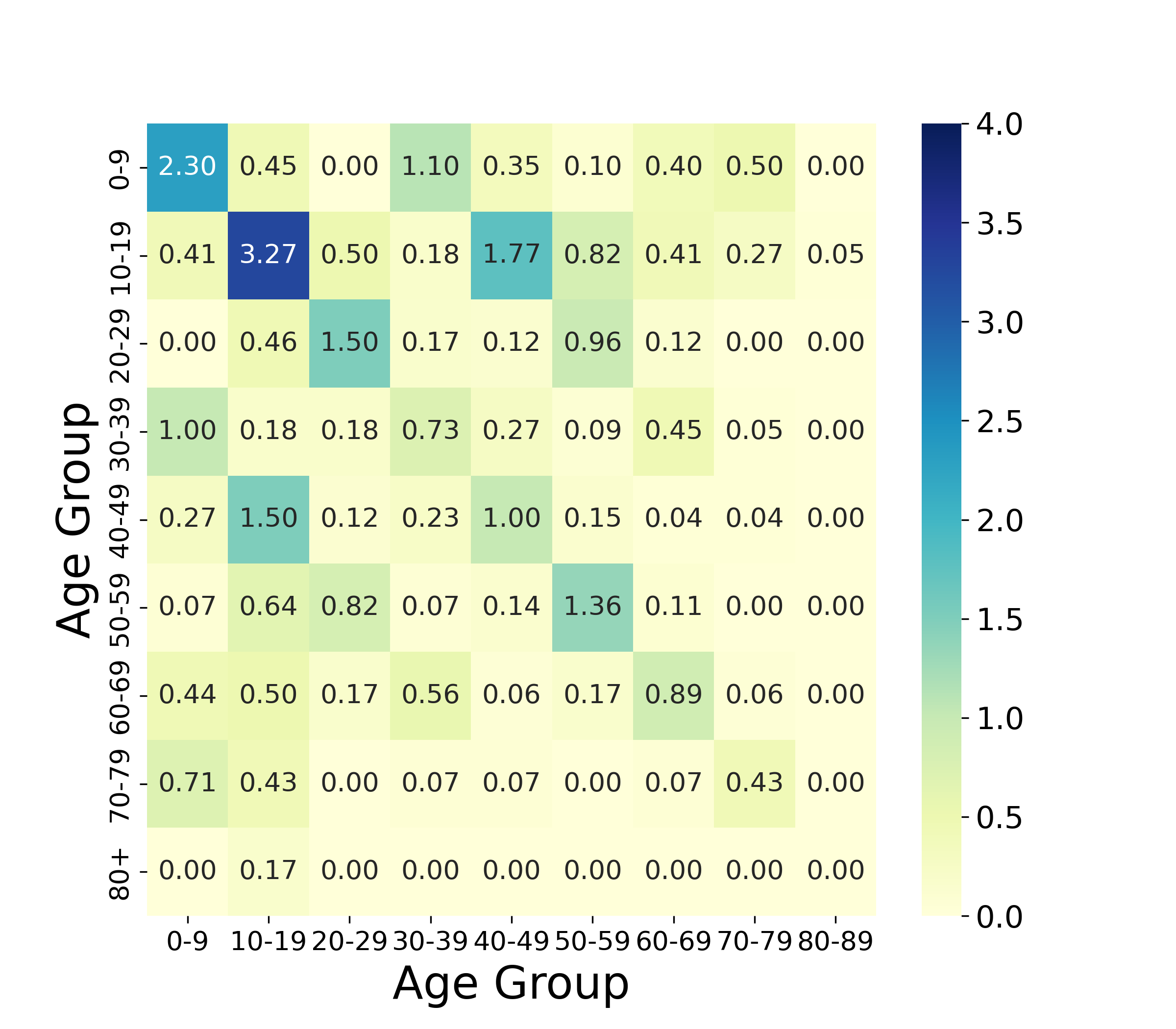}
	\end{minipage}}\\
	\caption{The target social contact matrix ((a) target) and the recreated social contact matrices of Finland, which are respectively generated by the (b) $RN$, (c) $HN-A_c$, (d) $HN-A_f$, (e) $HN-A_c-S$ and (f) $HN-A_f-S$ models.} 
\label{FinlandMat}
\end{figure}

Overall, the social contact matrix of real social network (the target) in each country shows people's preference for similar ages, younger ages and age differences at around $30$. With an optimised encounter rate and sex preference, The $RN$ model manages to fill in the cells of contact matrices by random encounters and recreate a low social contact rate for large age groups due to the small number of people. With an optimised age preference, the $HN-A_c$ model only captures the homophily to age but leaves two blank corners for the recreated social contact matrix. In contrast, the $HN-A_f$ model, with the introduction of fuzzy age representation, fills in more cells and can better capture the preference for similar ages and age differences around $30$. Compared with the $HN-A_c$ and $HN-S_c$ models, the $HN-A_c-S_c$ model combines both advantages by incorporating the preference for age values and similar ages. However, there are still blank cells in the two corners. This problem is solved with the $HN-A_f-S_c$ model by the fuzzy representation of age and the corresponding preference for a fuzzy feature (feature difference). The $HN-A_f-S_c$ model generates the most similar social networks, as suggested by the local-level similarity when recreating the social contact matrices. The increased diversity of the population and the fuzzy representation of features lead to an increased similarity level. This conclusion holds for all the networks involved in the experiments -- the DT-CNSs built on more realistic features can create more realistic networks. However, the selection of appropriate fuzzy sets to recreate social networks remains a research challenge to address. Therefore, we select the $HN-A_f-S_c$ model based on the model performance presented in this section and the next section extends this model with flexible fuzzy sets to investigate the impact of flexible fuzzy representation principles on the model performance in approaching realistic networks.

\subsection{Model Extension With Flexible Fuzzy Representation Principles}
\label{rep1-2section33}
In this section, we investigate the impact of flexible fuzzy representation principles on the model performance of DT-CNSs. We increase the flexibility of fuzzy representation principles by introducing different numbers of fuzzy sets and tuning the parameters of the fuzzy sets. This increases the complexity of modelling DT-CNSs and influences the accuracy of creating realistic social networks. As an example of extending DT-CNS with higher flexibility and complexity, we extend the $HN-A_f-S_c$ model (network model built on the fuzzy representation of age features and the crisp representation of sex features) selected in Section \ref{rep1-2section32} towards better model performance in recreating social contact matrices. 

As presented in the model initialisation process (see section \ref{rep1-2section31}, the fuzzy representation of the ages and age differences follow the assumptions of (i) seven fuzzy sets, which are determined by the same (ii) Gaussian Membership functions, parameterised with the $\mu$ list $[0,15,30,45,60,75,90]$ and the $\sigma$ list $[5,5,5,5,5,5,5]$. This section investigates whether the expressiveness of the age features can be improved when the fuzzy sets become more flexible. We select the best-performing fuzzy sets based on this enlarged searching space in section \ref{setselection}. We also conduct a global sensitivity analysis of the selected fuzzy sets and select parameters to be 
optimised for model performance improvement in section \ref{paramselectsection}.


\subsubsection{Selection of Fuzzy Sets}
\label{setselection}

\begin{figure}[htp] 
	\centering
	\subfigure[One Fuzzy Set]{
		\begin{minipage}[b]{0.46\linewidth}
			\includegraphics[width=1\linewidth]{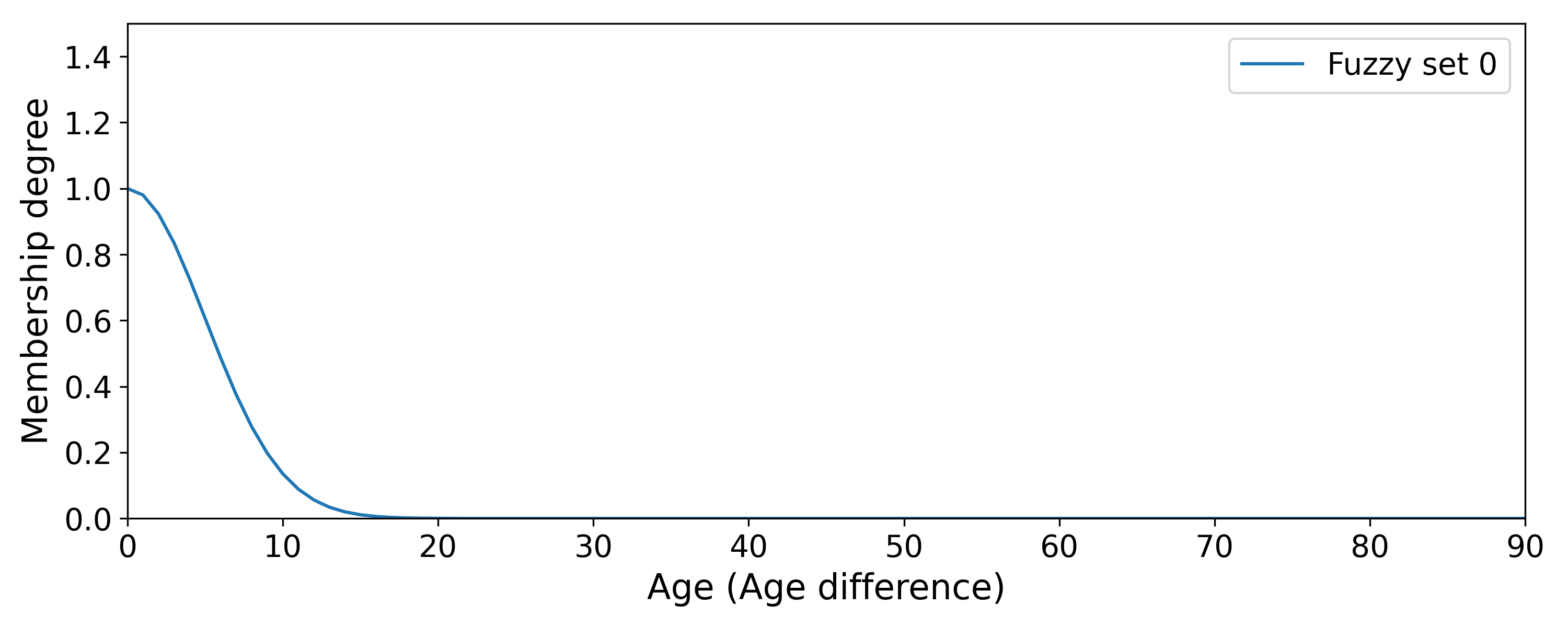}
	\end{minipage}}
	\subfigure[Two Fuzzy Sets]{
		\begin{minipage}[b]{0.46\linewidth}
			\includegraphics[width=1\linewidth]{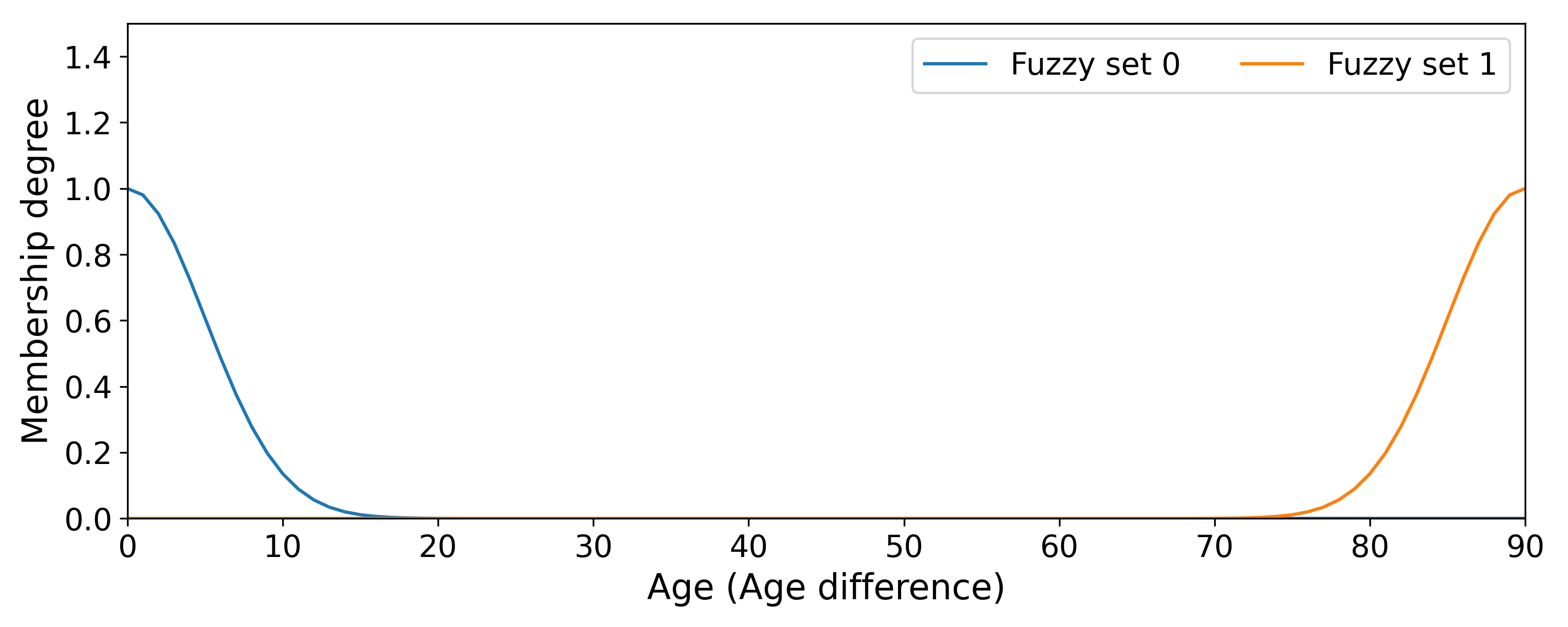}
	\end{minipage}}\\
	\subfigure[Three Fuzzy Sets]{
		\begin{minipage}[b]{0.46\linewidth}
			\includegraphics[width=1\linewidth]{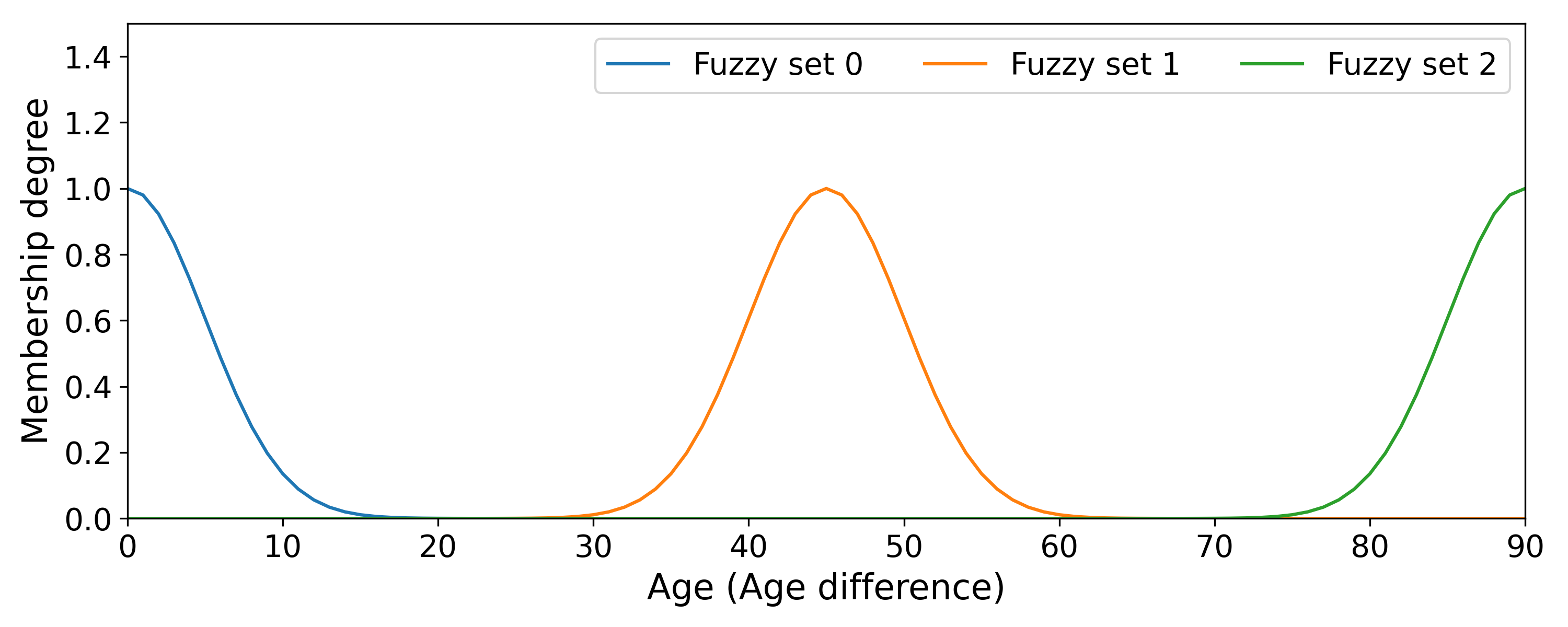}
	\end{minipage}}
		\subfigure[Four Fuzzy Sets]{
		\begin{minipage}[b]{0.46\linewidth}
			\includegraphics[width=1\linewidth]{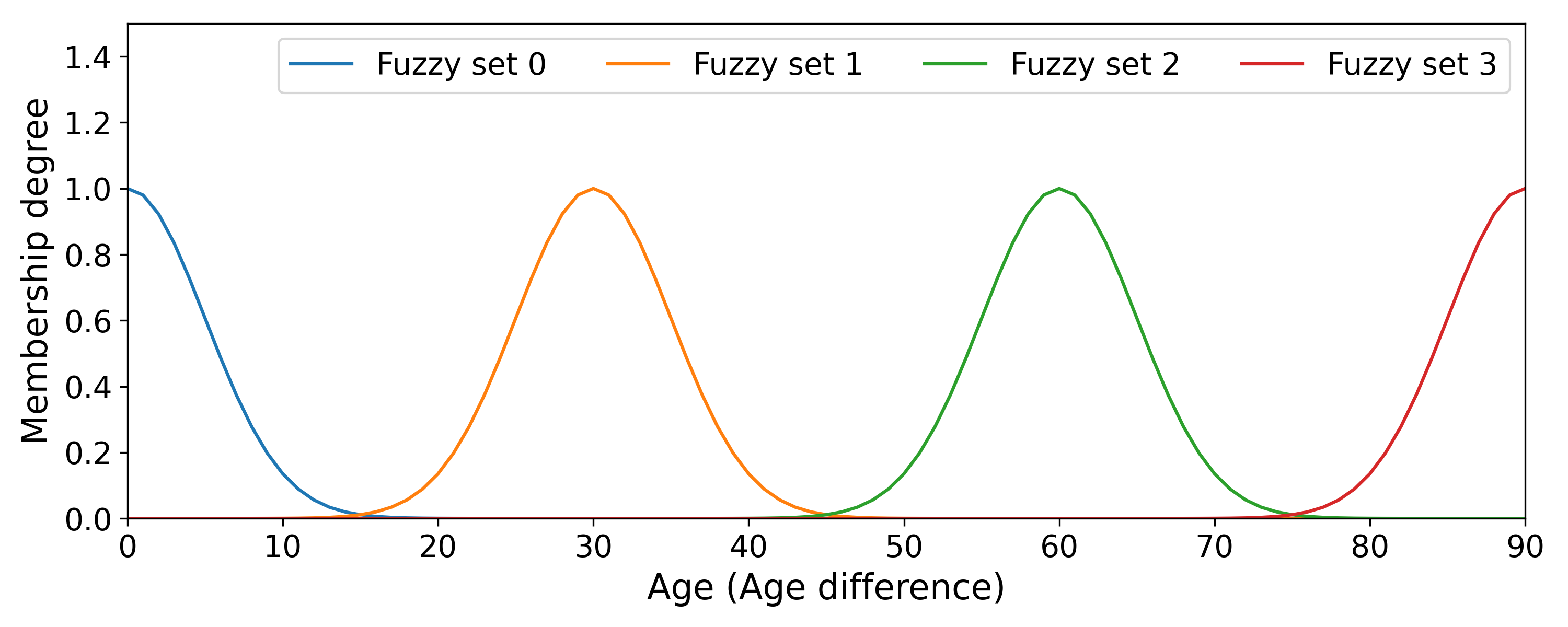}
	\end{minipage}}\\
		\subfigure[Five Fuzzy Sets]{
		\begin{minipage}[b]{0.46\linewidth}
	\includegraphics[width=1\linewidth]{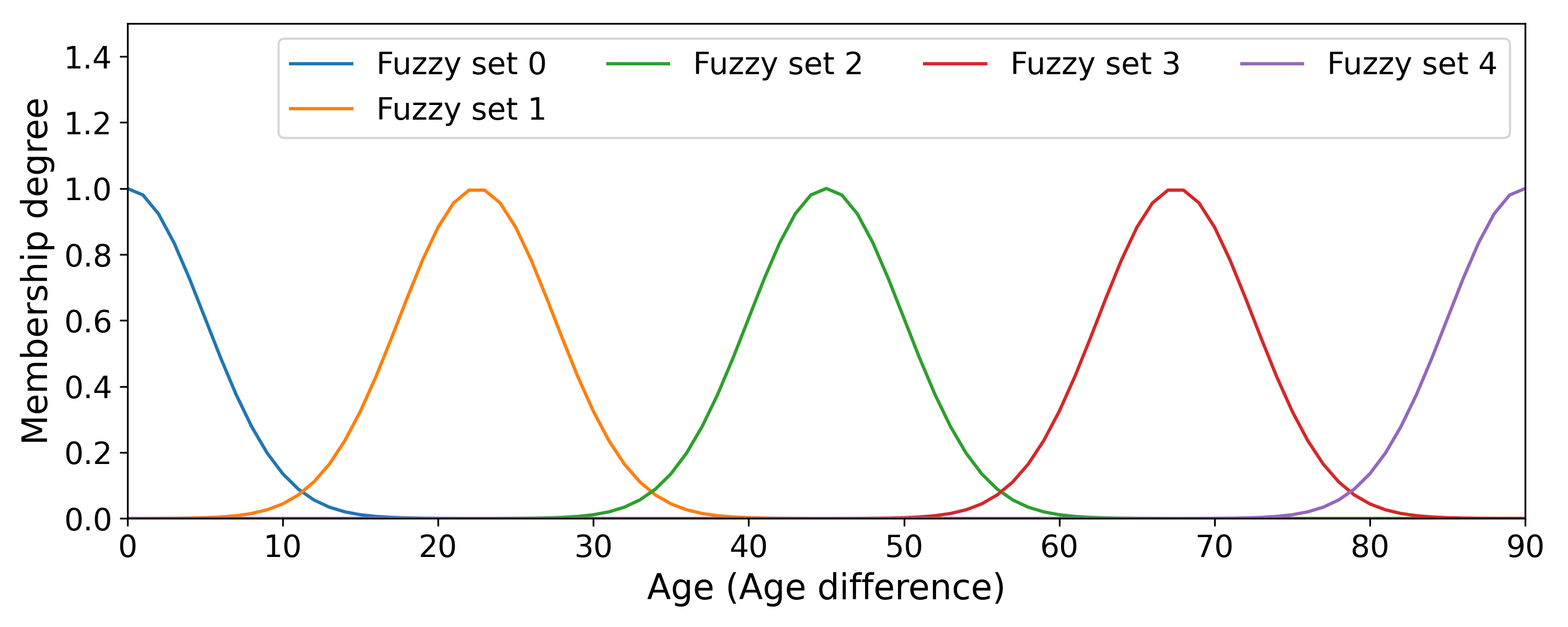}
	\end{minipage}}
		\subfigure[Six Fuzzy Sets]{
		\begin{minipage}[b]{0.46\linewidth}
			\includegraphics[width=1\linewidth]{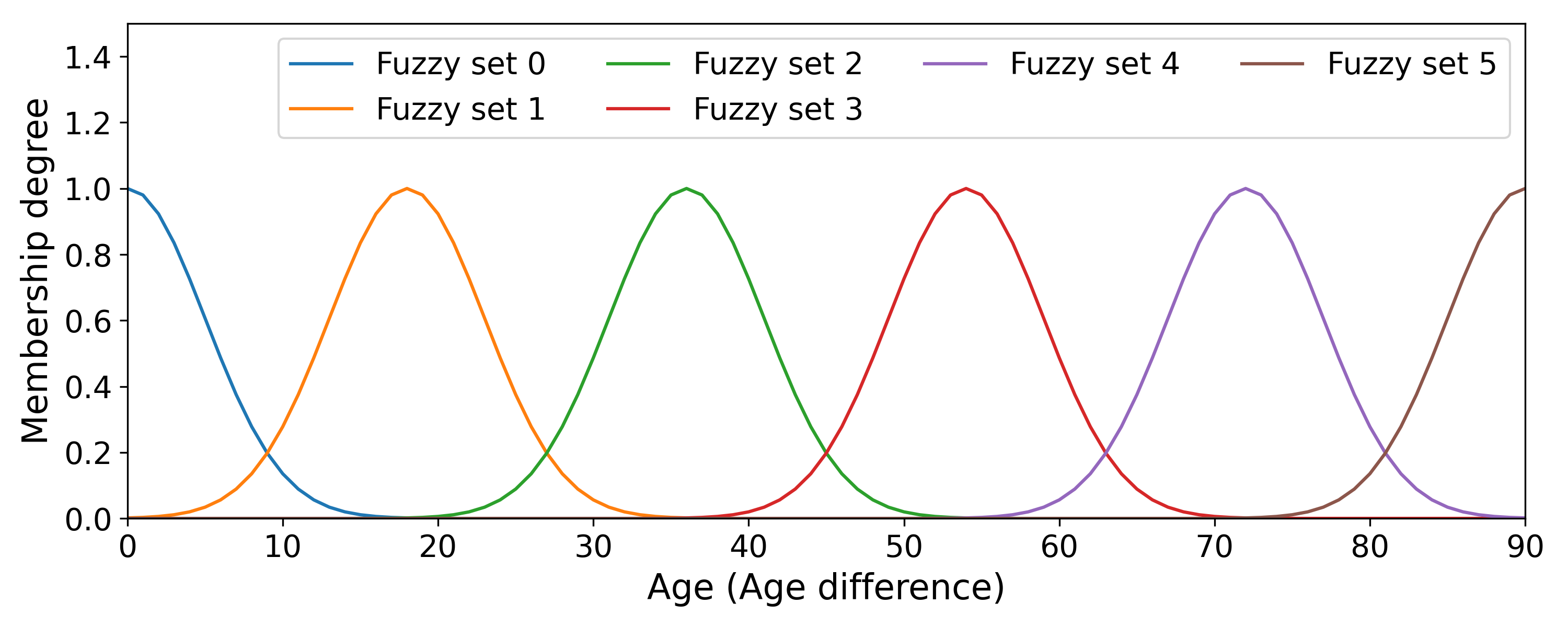}
	\end{minipage}}\\
			\subfigure[Seven Fuzzy Sets]{
		\begin{minipage}[b]{0.46\linewidth}
	\includegraphics[width=1\linewidth]{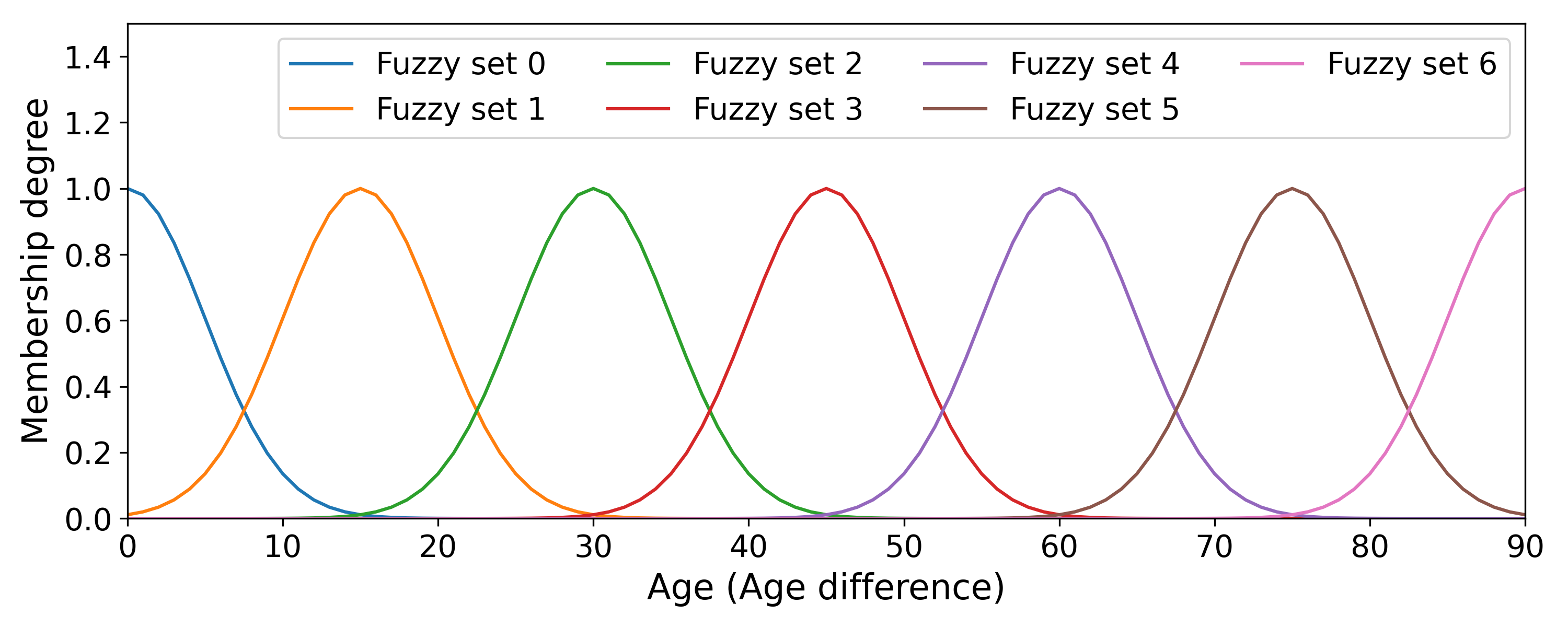}
	\end{minipage}}
		\subfigure[Eight Fuzzy Sets]{
		\begin{minipage}[b]{0.46\linewidth}
			\includegraphics[width=1\linewidth]{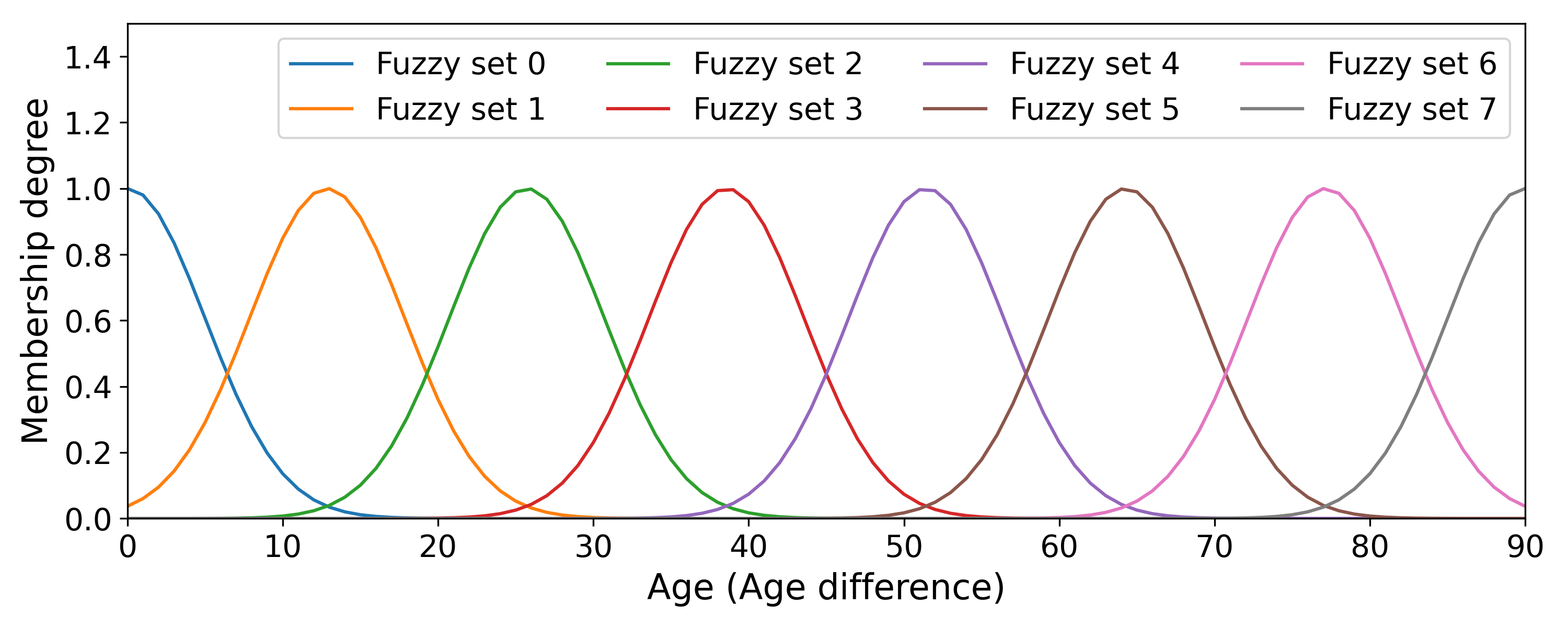}
	\end{minipage}}\\
	\subfigure[Nine Fuzzy Sets]{
		\begin{minipage}[b]{0.46\linewidth}
	\includegraphics[width=1\linewidth]{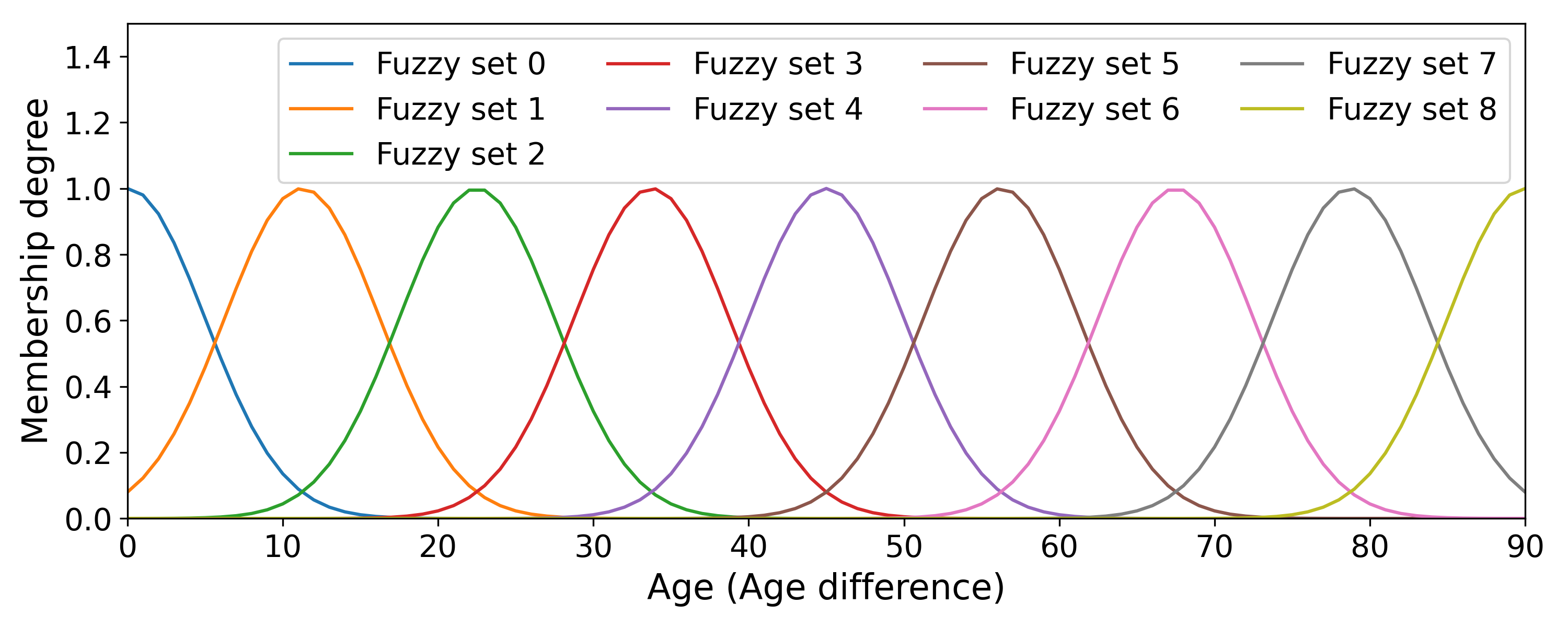}
	\end{minipage}}
		\subfigure[Ten Fuzzy Sets]{
		\begin{minipage}[b]{0.46\linewidth}
			\includegraphics[width=1\linewidth]{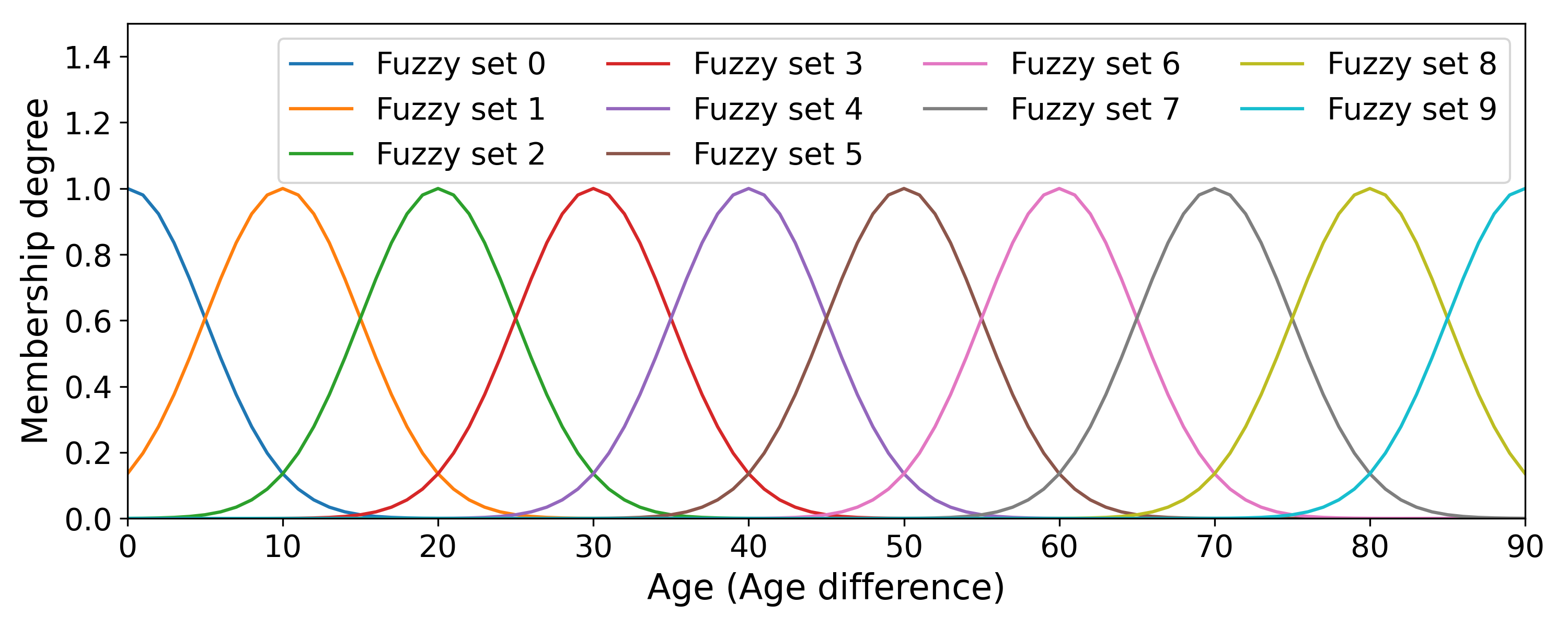}
	\end{minipage}}
	\caption{The searching space of fuzzy sets for age and age difference. In the experiments, both the age and age difference initialise the fuzzy sets based on the same parameter set-ups. Therefore, this figure } 
\label{samefuzzysets}
\end{figure}

We first introduce a flexible number of fuzzy sets to the respective representation of age and age difference. In this space, we re-initialise the $HN-A_f-S_c$ models with different fuzzy set number, ranging from $1$ to $10$. In addition, we initialise the $\mu$ list by equally dividing the age (age difference) value range $[0,90]$ and set up a $\sigma$ list of the same values at $5$ (See Fig.~\ref{samefuzzysets}). The initialised $\sigma$ value at $5$ enables each fuzzy set to cover an age range of around $20$ years that centres around the respective $\mu$ value, corresponding to a generation that nearly excludes parenthood relationships. 

As shown in Fig.~\ref{samefuzzysets}, the combination of fuzzy sets covers more value range with more overlapping areas as the fuzzy set number increases. The representation of features gets more detailed as we unfold a single feature value into a longer vector concerned with more specific age groups. In a real-world scenario, the increasing number of fuzzy sets for age indicates that people pay more attention to age. In a network composed of kids, adults and kinship relations, people focus on the age difference of around $30$. In contrast, people attend to more fuzzy age (age difference) ranges when they develop friendships. However, there is a challenge in optimising the corresponding same-length preference principles and the parameters of membership functions. This study aims to find the optimal fuzzy sets for age and age differences from these initialised fuzzy representation principles. 

\begin{figure}[H]
	\centering
	\subfigure[Belgium]{
		\begin{minipage}[b]{0.46\linewidth}
			\includegraphics[width=1\linewidth]{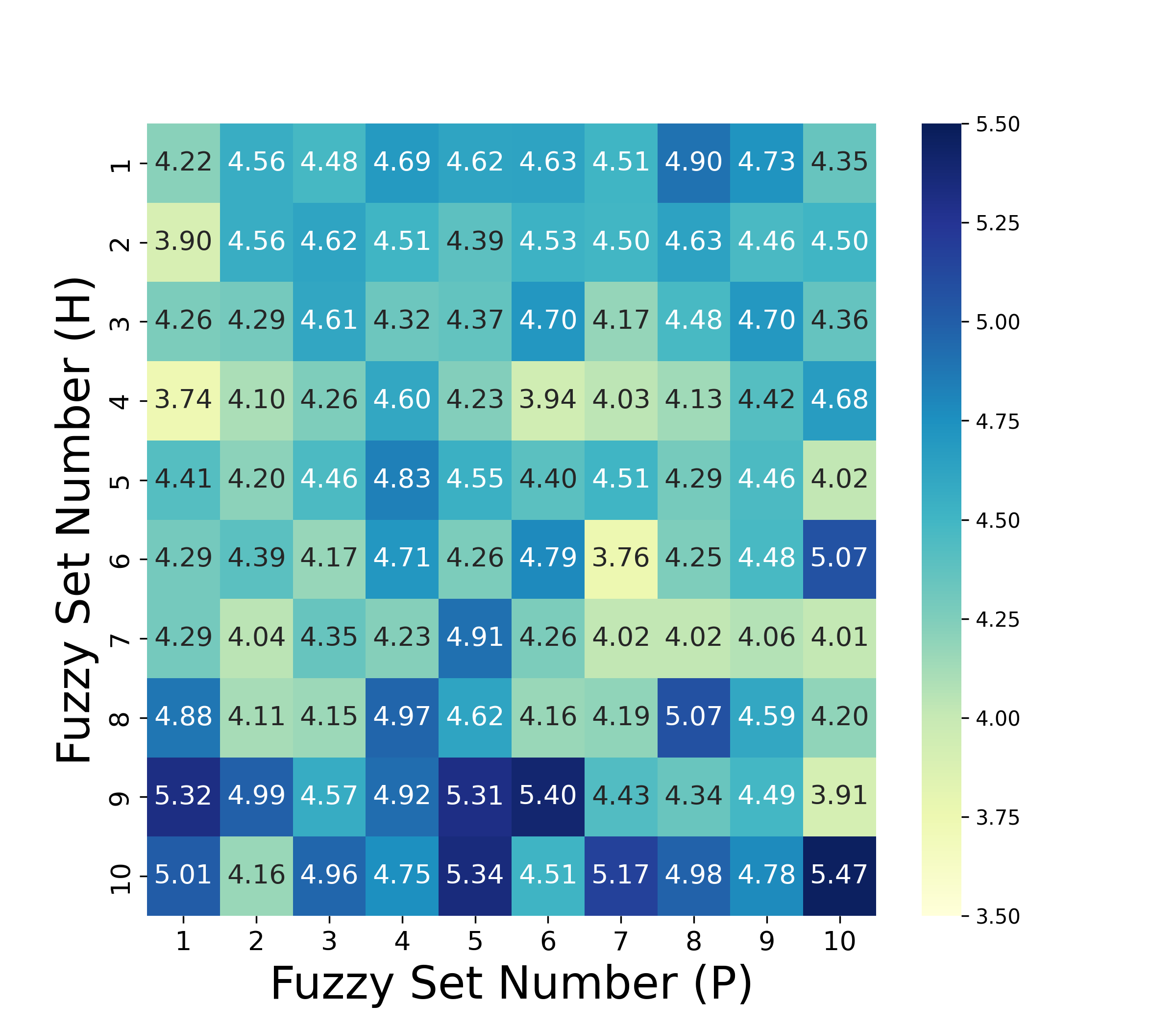}
	\end{minipage}}
 \hspace{-5mm}
	\subfigure[Finland]{
		\begin{minipage}[b]{0.46\linewidth}
			\includegraphics[width=1\linewidth]{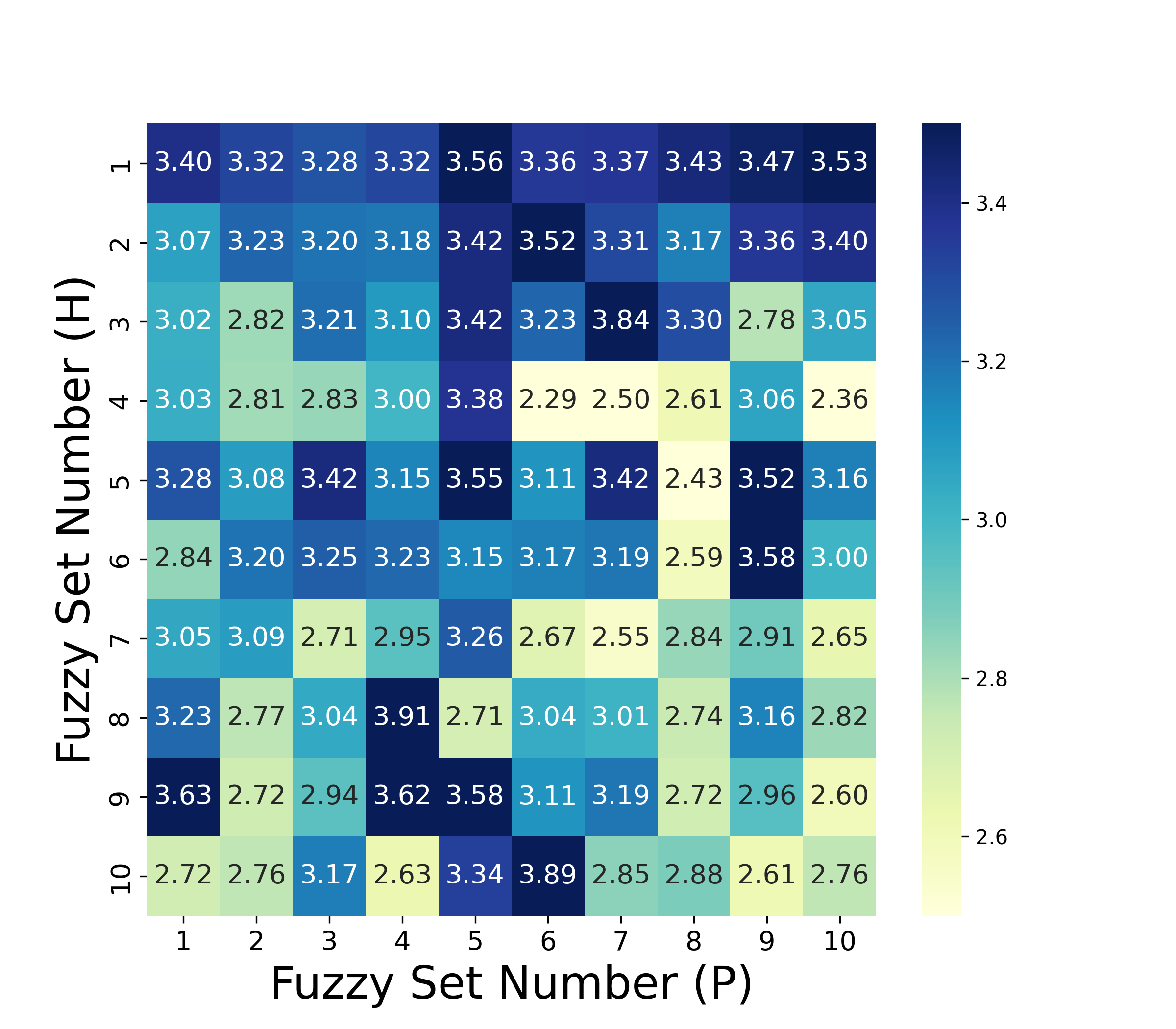}
	\end{minipage}}\\
  \vspace{-3mm}
	\subfigure[Germany]{
		\begin{minipage}[b]{0.46\linewidth}
			\includegraphics[width=1\linewidth]{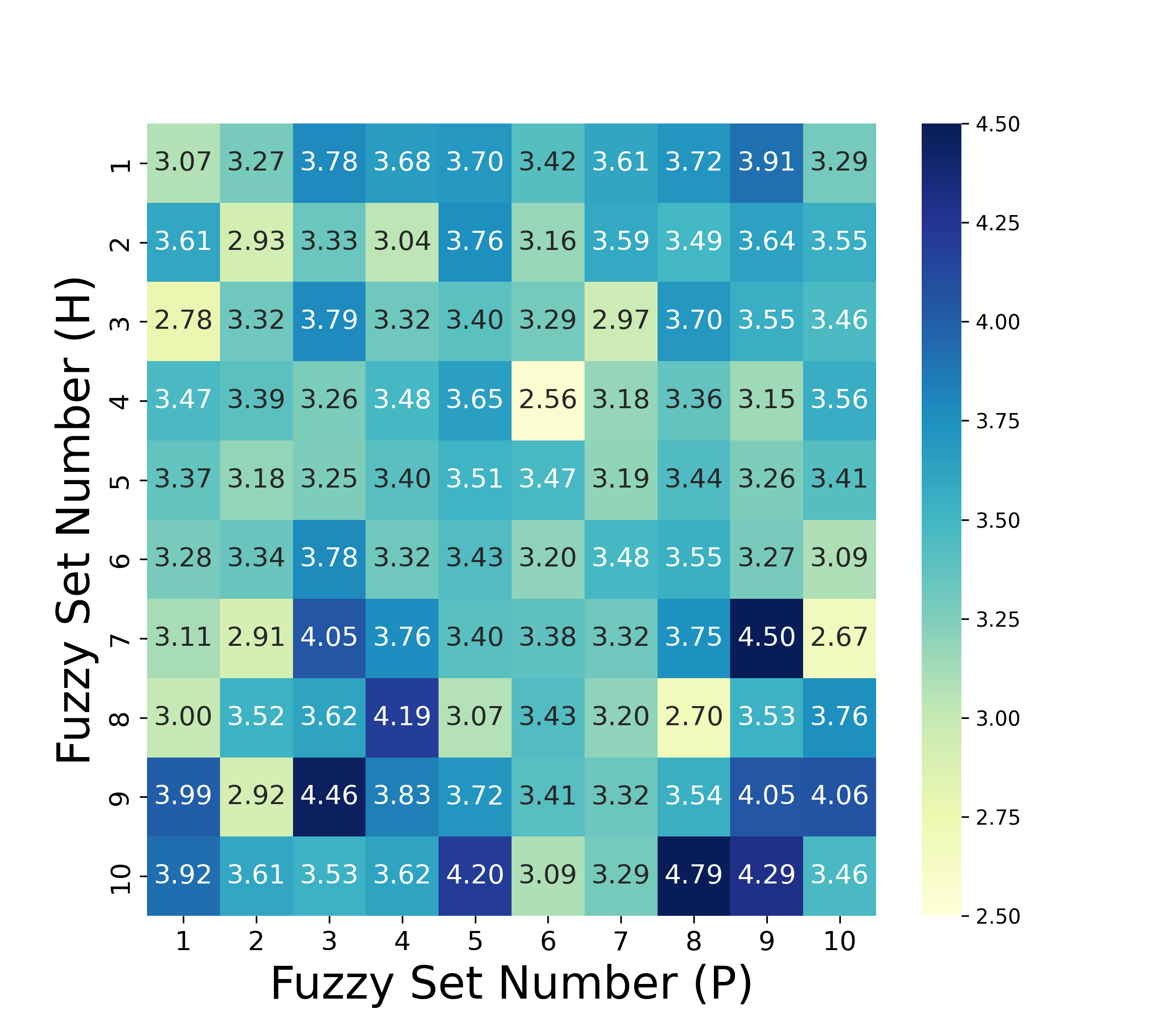}
	\end{minipage}}
  \hspace{-5mm}
		\subfigure[Italy]{
		\begin{minipage}[b]{0.46\linewidth}
			\includegraphics[width=1\linewidth]{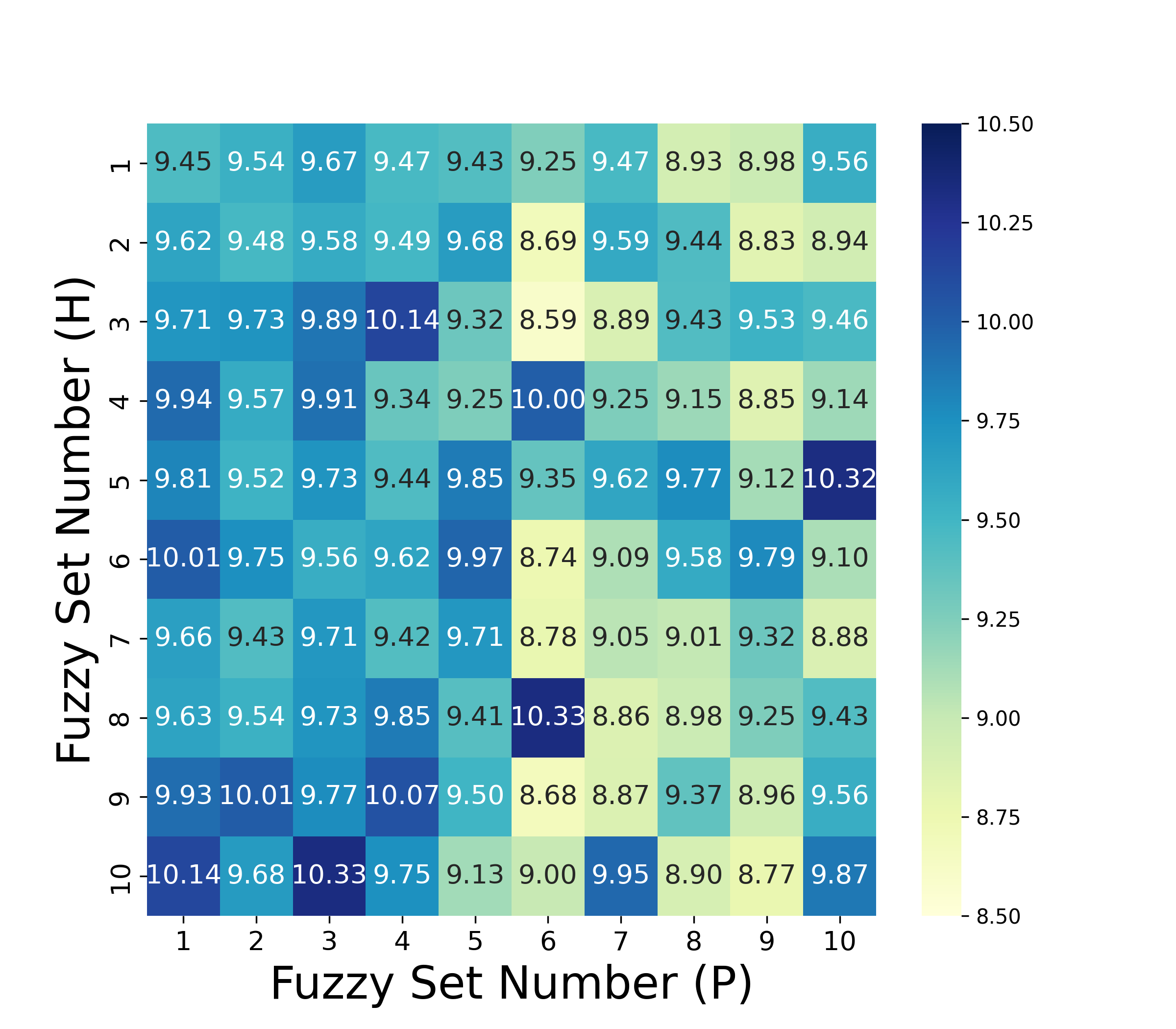}
	\end{minipage}}\\
  \vspace{-3mm}
		\subfigure[Luxembourg]{
		\begin{minipage}[b]{0.46\linewidth}
	\includegraphics[width=1\linewidth]{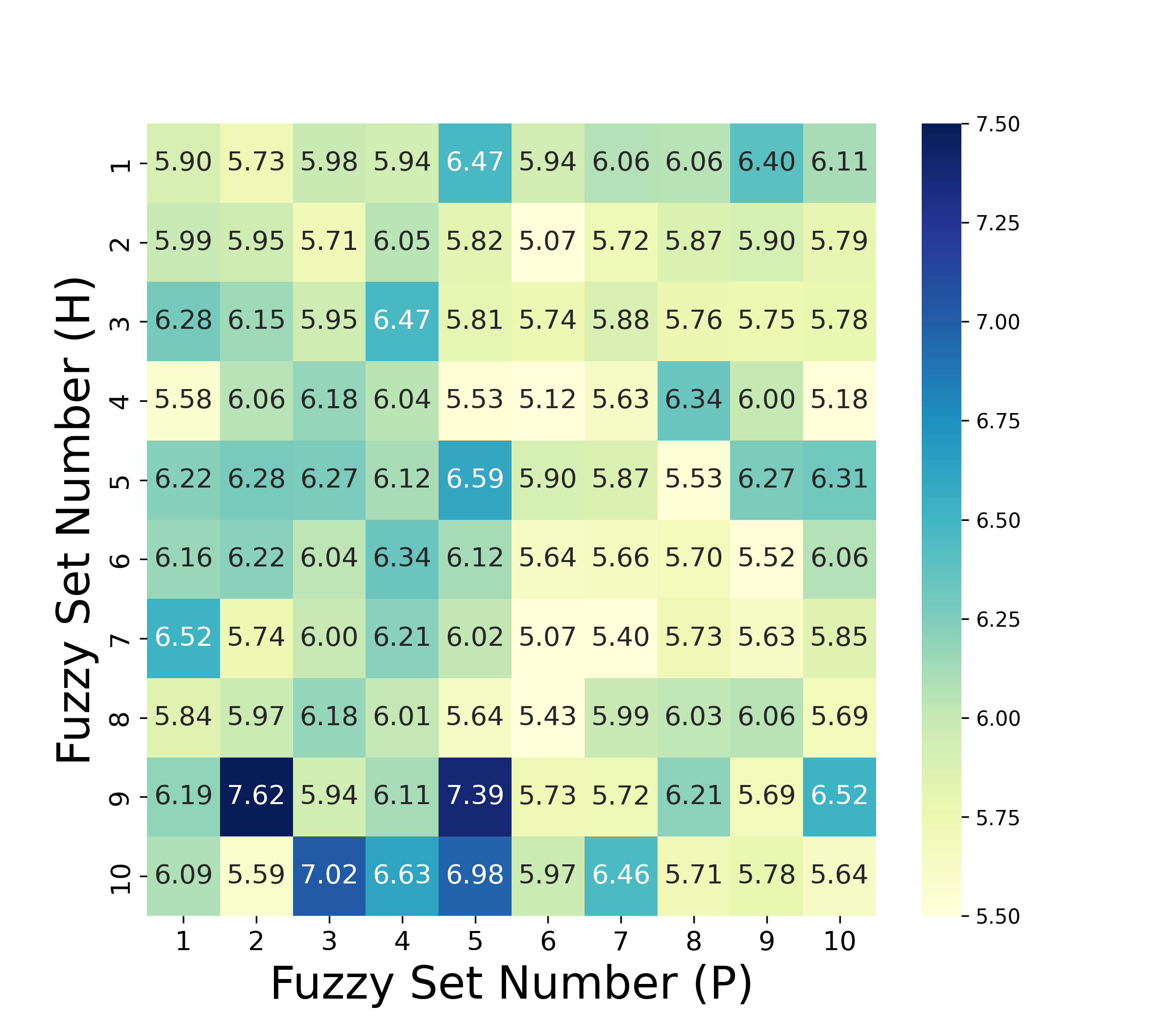}
	\end{minipage}}
  \hspace{-5mm}
		\subfigure[Poland]{
		\begin{minipage}[b]{0.46\linewidth}
			\includegraphics[width=1\linewidth]{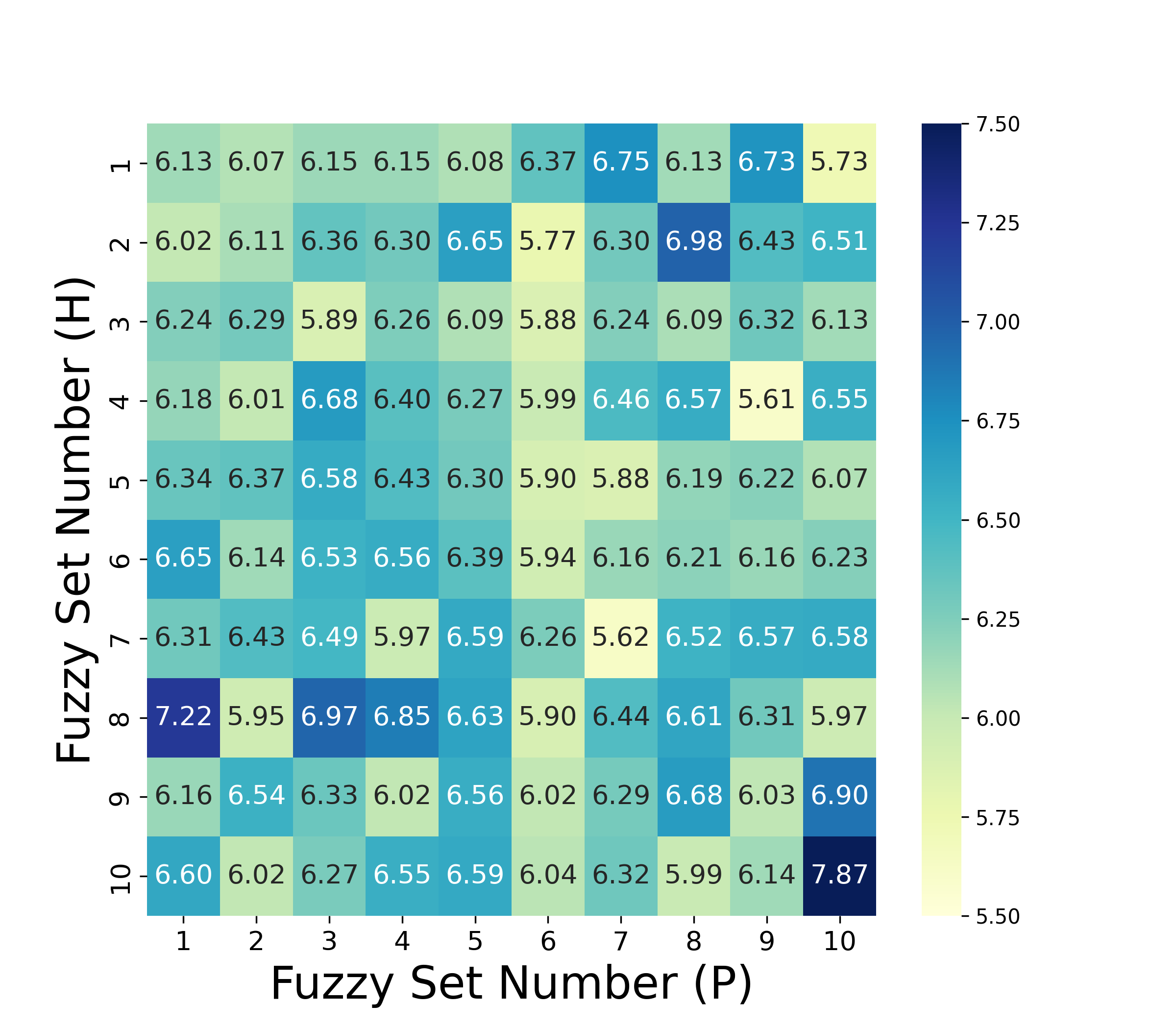}
	\end{minipage}}\\
	\caption{The $EU$ distance achieved with different settings of fuzzy sets (Fuzzy Set Number (H) for age difference on the y-axis and Fuzzy Set Number (P) for age on the x-axis) in experiments on (a) Belgium, (b) Finland, (c) Germany, (d) Italy, (e) Luxembourg and (f) Poland.} 
\label{fuzzysetselection}
\end{figure}

We visualise the influence of the respective fuzzy set number for age and age difference on the $EU$ distance between the target and the recreated social contact matrices (see Fig.\ref{fuzzysetselection}). For each country, the $EU$ distance varies with the fuzzy sets without a clear trend, where an increasing number of fuzzy sets does not necessarily improve the model performance. This indicates that the number of fuzzy sets impacts the expressive power of DT-CNSs but does not directly correlates with the accuracy of recreating realistic social networks. In addition, selecting appropriate fuzzy sets for realistic feature representation is a research challenge to address in DT-CNS modelling.

\begin{table}[h!]
\centering
\scriptsize
\caption{The $EU$ distance achieved with the respective fuzzy sets selected for each country.}
\label{setrank}
\setlength{\tabcolsep}{3pt}
\renewcommand{\arraystretch}{1.5}
\begin{tabular}{|c|c|c|c|}
\hline
\multirow{2}{*}{Country} & \multicolumn{2}{|c|}{Number of fuzzy sets} &
\multirow{2}{*}{$EU$ distance} \\
\cline{2-3}
& Age & Age difference & \\
\hline
\multirow{2}{*}{Belgium} &7& 7& 4.02 \\
\cline{2-4}
& 1 & 4 & 3.74  \\
\hline
\multirow{2}{*}{Finland} &7& 7& 2.55 \\
\cline{2-4}
& 6 &4 & 2.29\\
\hline
\multirow{2}{*}{Germany} &7& 7&3.32\\
\cline{2-4}
&6&4&2.56\\
\hline
\multirow{2}{*}{Italy} &7& 7&9.05 \\
\cline{2-4}
& 6&3 & 8.59\\
\hline
\multirow{2}{*}{Luxembourg} &7& 7&5.40 \\
\cline{2-4}
&6&2& 5.07 \\
\hline
\multirow{2}{*}{Poland} &7& 7&5.62 \\
\cline{2-4}
&9&4 & 5.61\\
\hline
\end{tabular}
\end{table}
 
Compared with the $EU$ distance achieved with seven fuzzy sets (see section \ref{rep1-2section32}), we improve the model performance by introducing more flexible fuzzy sets. Based on Tab.~\ref{setrank}, we select the best-performing fuzzy sets with the smallest $EU$ distance for each country. Generally, smaller number of fuzzy sets is needed to represent age differences. Particularly, the "four fuzzy set" covers the uncertain perception around age differences around $0$, $30$ and $60$, which reveals people's preferences for kinship relations in Belgium, Finland, Germany and Poland. In addition, we select the "six fuzzy set" to represent the age values around $0$, $18$, $36$, $54$, $72$ and $90$. It captures people's preferences for age values in Finland, Germany and Italy. 
In addition, we select a more significant number of fuzzy sets to represent age values than age differences, except for the case of Belgium. This indicates that people in each country put more importance to age differences rather than age. People unfold a single age value into much longer vectors in this condition by considering more fuzzy sets. In contrast, they have limited but focused consideration of age differences, employing fewer fuzzy sets to unveil their consideration for age differences. 


\subsubsection{Selection of Parameters}
\label{paramselectsection}

We conduct a global sensitivity analysis of the Membership Functions' parameters for the selected fuzzy sets of each country in section \ref{setselection}. We use the Variogram Analysis of Response Surfaces (VARS) to estimate parameters' sensitivities. This method uses directional variograms to estimate Integrated Variogram Across a Range of Scales (IVARS), by integrating the directional variogram up to a given perturbation scale of interest (e.g, 10\%, 30\% or 50\%). In this space, these directional variograms represent the sensitivity of the model response to each parameter across the full range of perturbation scales \citep{haghnegahdar2017insights,razavi2016new}. In this study, we measure the parameter sensitivity with $IVARS_{10}$ (scale range = $0$-$10\%$), $IVARS_{30}$ (scale range = $0$-$30\%$) and $IVARS_{50}$ (scale range = $0$-$50\%$). We also measure the variance-based total-order effect (denoted as $Sobol$) for each parameter, which quantifies the respective effects of parameters on the variance of the response of a mathematical model \citep{sudret2008global,saltelli2010variance}. In our experiment, the response of the mathematical model in the sensitivity analysis refers to the changes in similarity level (measured by the $EU$ distance between the target and recreated social contact matrices). The larger values of $IVARS$ and $Sobol$ metrics indicate a more noteworthy impact of the corresponding parameter on the similarity of network generation. 

Each metric generates a set of parameter importance rankings. The parameter rankings are presented in Fig.~\ref{fuzzyBselectparam}(a), Fig.~\ref{fuzzyFselectparam}(a), Fig.~\ref{fuzzyGselectparam}(a), Fig.~\ref{fuzzyIselectparam}(a), Fig.~\ref{fuzzyLselectparam}(a) and Fig.~\ref{fuzzyPselectparam}(a) in the appendix \ref{paramselect}. Based on the aggregated ranks of these sensitivity metrics, we can select the most important parameters in model calibration. As there are no general and definitive criteria in the parameter selection, we select the $6$ most important parameters to ensure the flexibility of parameter changes while preserving the computation efficiency of optimising multiple parameters. In this part, we use the Bayesian optimisation algorithm \citep{bayesOpt} to probe parameter values based on the initial parameter set-ups. Tab.~\ref{boundary} in appendix \ref{app3} shows the bounds of parameters in the sensitivity analysis and the optimisation process. Tab.~\ref{summary} presents the aggregated ranks of parameter importance, the selected parameters in the model calibration and the enhanced model performance.

\begin{table}[h!]
\centering
\scriptsize
\caption{The parameter ranking and selection for the selected fuzzy sets and the resulted model performance evaluated with $EU$ distance. The parameter ranking is determined based on the aggregated ranks of sensitivity indices including $IVARS_{10}$, $IVARS_{30}$, $IVARS_{50}$ and $Sobol$.}
\label{summary}
\setlength{\tabcolsep}{3pt}
\renewcommand{\arraystretch}{1.5}
\begin{tabular}{|l|l|l|l|p{80pt}|l|l|}
\hline
\multirow{2}{*}{Country} & \multicolumn{2}{|c|}{Number of fuzzy sets} &
\multirow{2}{*}{Initial $EU$} &
\multirow{2}{*}{Parameter rankings}  &
\multirow{2}{*}{Parameter selection} & 
\multirow{2}{*}{Improved $EU$} \\
\cline{2-3}
& Age & Age difference & &&&\\
\hline
Belgium & 1 & 4 & 3.74 & $\mu_h^1$> $\sigma_h^1$> $\mu_h^2$> $\sigma_h^2$> $\mu_h^3$> $\sigma_h^3$> $\mu_p^1$> $\mu_h^4$>
       $\sigma_p^1$> $\sigma_h^4$ &[$\mu_h^1$,$\sigma_h^1$,$\mu_h^2$,$\sigma_h^2$,$\mu_h^3$,$\sigma_h^3$] & 3.47\\
\hline
Finland& 6 &4 & 2.29 & $\mu_h^1$> $\sigma_h^1$> $\mu_h^2$> $\sigma_h^2$> $\mu_p^4$> $\sigma_p^4$> $\mu_p^5$>
       $\sigma_p^6$> $\mu_p^6$> $\sigma_p^5$> $\mu_p^3$> $\mu_h^3$> $\mu_p^2$> $\sigma_p^3$>
       $\sigma_h^3$> $\sigma_p^2$> $\mu_p^1$> $\sigma_p^1$> $\mu_h^4$> $\sigma_h^4$& [$\mu_h^1$,$\sigma_h^1$,$\mu_h^2$,$\sigma_h^2$,$\mu_p^4$,$\sigma_p^4$] & 2.10\\
\hline
Germany&6&4&2.56& $\mu_h^1$> $\mu_p^2$> $\sigma_h^1$> $\mu_p^3$> $\sigma_p^2$> $\sigma_p^3$> $\mu_h^2$> $\mu_p^1$>
       $\mu_p^4$> $\sigma_p^4$> $\sigma_p^1$> $\mu_h^3$> $\sigma_h^2$> $\mu_p^5$> $\sigma_h^3$>
       $\sigma_p^5$> $\mu_p^6$> $\sigma_p^6$> $\mu_h^4$> $\sigma_h^4$ &[$\mu_h^1$,$\mu_p^2$,$\sigma_h^1$,$\mu_p^3$,$\sigma_p^2$,$\sigma_p^3$]& 2.28\\ 
\hline
Italy& 6&3&8.59& $\mu_h^1$> $\sigma_h^1$> $\mu_p^2$> $\sigma_p^2$> $\mu_p^1$> $\sigma_p^1$> $\mu_p^3$>
       $\sigma_p^5$> $\sigma_p^3$> $\mu_h^3$> $\mu_h^2$> $\mu_p^5$> $\sigma_h^2$> $\sigma_p^4$>
       $\mu_p^4$> $\sigma_h^3$> $\mu_p^6$> $\sigma_p^6$ & [$\mu_h^1$,$\sigma_h^1$,$\mu_p^2$,$\sigma_p^2$,$\mu_p^1$,$\sigma_p^1$]& 8.34 \\
\hline
Luxembourg& 6&2&5.07&$\mu_h^1$> $\sigma_h^1$> $\mu_p^2$> $\mu_p^1$> $\sigma_p^2$> $\mu_p^5$> $\mu_p^6$> $\sigma_p^1$>
       $\sigma_p^5$> $\sigma_p^6$> $\mu_h^2$> $\mu_p^3$> $\mu_p^4$> $\sigma_p^3$> $\sigma_p^4$>
       $\sigma_h^2$  &[$\mu_h^1$,$\sigma_h^1$,$\mu_p^2$,$\mu_p^1$,$\sigma_p^2$,$\mu_p^5$]&4.77 \\
\hline
Poland &9&4&5.61&
$\mu_h^1$> $\sigma_h^1$> $\mu_h^2$> $\mu_p^2$> $\sigma_p^2$> $\sigma_p^5$> $\sigma_h^2$>
       $\mu_p^5$> $\mu_p^3$> $\sigma_p^7$> $\mu_h^4$> $\mu_h^3$> $\sigma_h^3$> $\sigma_p^3$>
       $\sigma_p^4$> $\sigma_p^9$> $\mu_p^7$> $\mu_p^4$> $\mu_p^1$> $\sigma_p^1$> $\mu_p^9$>
       $\sigma_h^4$> $\mu_p^8$> $\sigma_p^8$> $\sigma_p^6$> $\mu_p^6$&[ $\sigma_h^1$> $\mu_h^2$,$\mu_p^2$,$\sigma_p^2$,$\sigma_p^5$,$\sigma_h^2$]&5.04 \\ 
\hline
\end{tabular}
\end{table}

The outcomes of the optimisation process and the detailed analysis for the six countries are included in the appendix \ref{paramselect} while below we provide the summary and main conclusions. Tab.~\ref{summary} shows the parameter rankings for each country based on the aggregated ranks of sensitivity metrics obtained from the global sensitivity analysis. The selected parameters for each country all incorporate $\mu_1^h$ and $\sigma_1^h$. These two parameters initially describe people's interest in age difference at around $0$. This indicates that the homophily effect, related to the age difference around $0$, can have a greater impact on the social contact matrices than the other characteristics. In addition, the selected parameters for Germany, Italy and Luxembourg include $\mu_p2$ and $\sigma_p^2$ based on the "six fuzzy sets" of age. These two parameters initially describe people's interest in age values around $18$ and have a greater impact on the social contact matrices than the other characteristics. The other selected parameters vary with countries and have different meanings considering the number of fuzzy sets. Thus to recreate realistic social networks, we tune the unique parameter set for each country to optimise the model performance, as measured by the similarity between real and simulated social contact matrices.

To conclude, the optimised parameters for each country all incorporate the $\mu_1^h$ and $\sigma_1^h$ values. They describe people's interest in age difference at around $0$ and are optimised to represent small age differences close to $0$. This implies that people in each country have more interest in age differences around these small values and captures the homophily effect presented in the real-world social contact matrices (See Fig.~\ref{FinlandMat} and Fig.~\ref{BelgiumMat},  Fig.~\ref{GermanyMat}, Fig.~\ref{LuxembourgMat}, Fig.~\ref{ItalyMat} and Fig.~\ref{PolandMat} in the appendix \ref{appMS}). In addition, the optimised parameters for Belgium and Finland both include the parameters for the second fuzzy set of age difference, but each centring around different values at around $24$ and $12$. They describe people's different interests in age differences and capture the social contacts across age groups (See Fig.~\ref{BelgiumMat} in the appendix \ref{appMS} and Fig.~\ref{FinlandMat}). The optimised parameters in case of Finland and Italy include the parameter for the third fuzzy set of age, each centring around the age of $28$ and $22$. They focus on people's preferences for age and captures the popularity of the age groups between $20$ and $40$, as presented in the respective social contact matrices (See Fig.~\ref{FinlandMat} and Fig.~\ref{ItalyMat} in the appendix \ref{appMS}). Overall, the above-mentioned selection of fuzzy sets and the selection and optimisation of parameters for fuzzy sets increases the accuracy of recreating realistic social networks, as indicated by a decreasing $EU$ distance between the target and real social contact matrices (See Tab.~\ref{summary}). This implies that the increasingly flexible set-ups of fuzzy representation principles increase the complexity of modelling DT-CNSs and enable to recreate more realistic social networks through the appropriate selection and optimisation of fuzzy sets.

\subsection{Network Representation}
In this section, we analyse the networks generated by the calibrated models from five perspectives, including the features and the related preferences (See Section \ref{1feature}), social contact matrices (See Section \ref{2mat}), degree distributions (See Section \ref{3deg}), clustering coefficient distributions (See Section \ref{4clus}) and shortest path length distributions (See Section \ref{5path}). Tab.~\ref{networkinfo} shows the number of connected and unconnected nodes and the information about the degree distributions, clustering coefficient distributions and the shortest path length distributions that vary depending on the values of the features and rules employed. Specifically, we also consider the non-existing shortest paths between the node pairs and assume their path lengths as $90$, which is the upper limit that cannot be achieved by any real path. Generally, all the nodes get connected with the paradigms of each country. Compared with other countries, Finland has a lower average node degree value, a lower clustering coefficient and more non-existing (fake) paths between the node pairs. Italy and Poland have higher average node degrees and fewer fake paths. These differences can result from different numbers of edges and heterogeneous features and rules. To better understand these characteristics, we calculate the social contact matrices of network simulations and explore the differences in degree distributions, clustering coefficient distributions and shortest path length distributions for males and females at different ages and for different countries.

\begin{table}[htp]
\centering
\scriptsize
\caption{Topological information of the generated networks.}
\label{networkinfo}
\setlength{\tabcolsep}{2pt}
\renewcommand{\arraystretch}{1.5}
\begin{tabular}{|c|c|c|c|c|c|c|c|c|c|c|c|c|c|c|c|c|}
\hline
\multirow{2}{*}{Country} &\multicolumn{2}{c|}{Nodes}& \multirow{2}{*}{Edges} &\multicolumn{4}{c|}{Node Degree} &\multicolumn{4}{c}{Clustering coefficient} & \multicolumn{5}{|c|}{Shortest path length}  \\
\cline{2-3}\cline{5-17}
&  Connected & Unconnected & & Avg. & Std. & Max. & Min.&Avg. & Std. & Max. & Min.& Fake Paths& Avg. & Std. & Max. & Min.\\
\hline
Belgium&90& 0& 614& 13.64& 11.84& 34& 0&0.54& 0.31& 1.00& 0.00& 264& 8.02& 21.79& 90&1\\
\hline
Finland & 90& 0& 336& 7.47& 6.94& 22& 0& 0.32& 0.27& 1.00& 0.00& 2287& 52.34& 43.46& 90& 1\\
\hline 
Germany & 90& 0& 462& 10.27& 9.8& 35& 0& 0.42& 0.30& 1.00& 0.00& 1155& 27.57& 39.75& 90& 1\\
\hline 
Italy & 90& 0& 770& 17.11& 13.67& 45& 0& 0.52& 0.27& 1.0& 0.00& 89& 4.14& 12.97& 90& 1\\
\hline 
Luxembourg &90& 0& 687& 15.27& 14.20& 43& 0& 0.51& 0.33& 1.00& 0.00& 1449& 33.81& 42.31& 90& 1\\
\hline
Poland &90& 0& 793& 17.62& 11.45& 45& 0& 0.51& 0.15& 0.83& 0.00& 89& 4.08& 12.98& 90& 1 \\
\hline 
\end{tabular}
\end{table}

\subsubsection{Features, Feature Differences and Social Preferences}
\label{1feature}

\begin{figure}[H] 
	\centering
	\subfigure[Feature Representation]{
		\begin{minipage}[b]{0.46\linewidth}
			\includegraphics[width=1\linewidth]{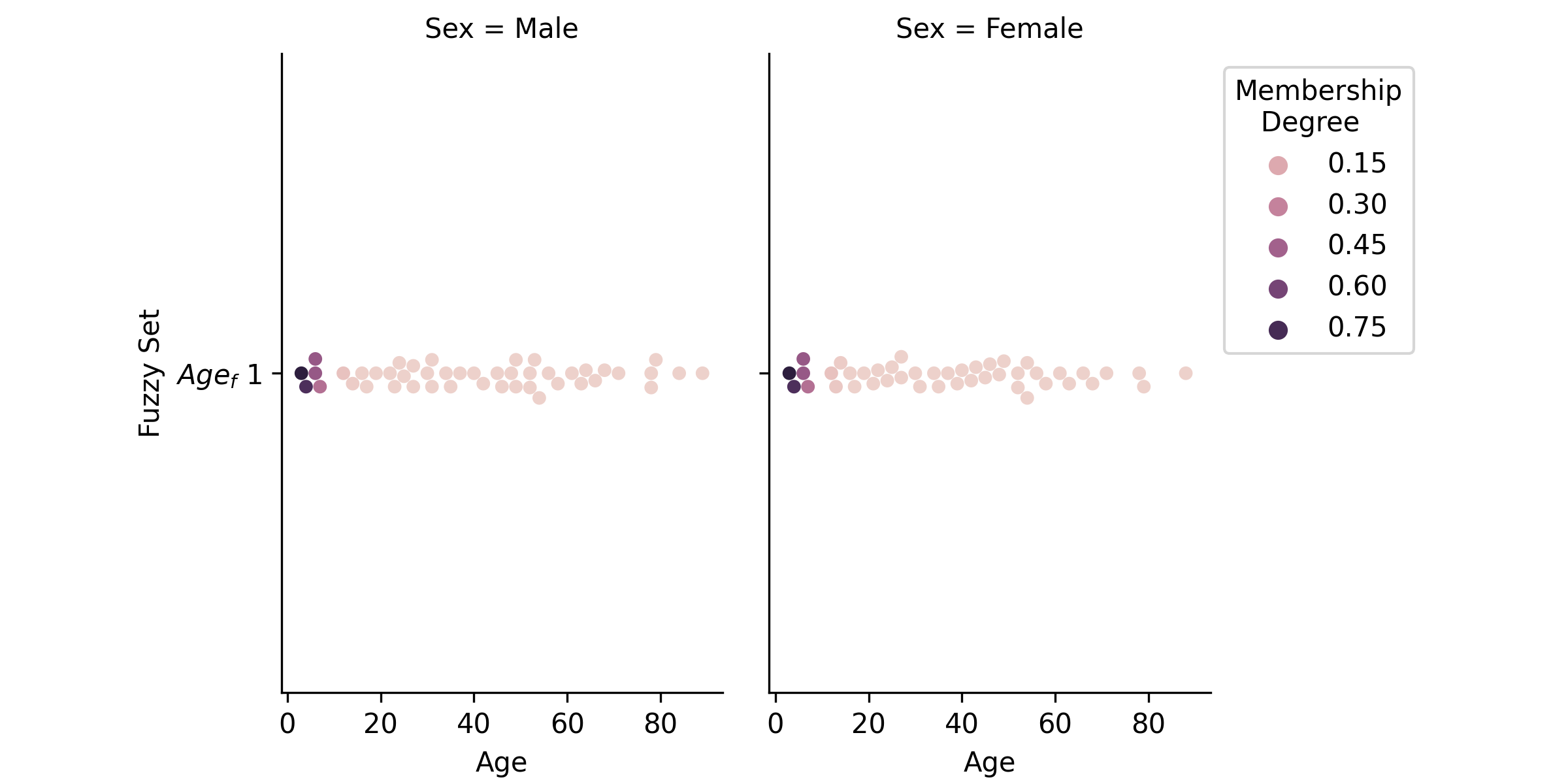}
	\end{minipage}}
 \hspace{-5mm}
	\subfigure[Feature Difference Representation]{
		\begin{minipage}[b]{0.46\linewidth}
			\includegraphics[width=1\linewidth]{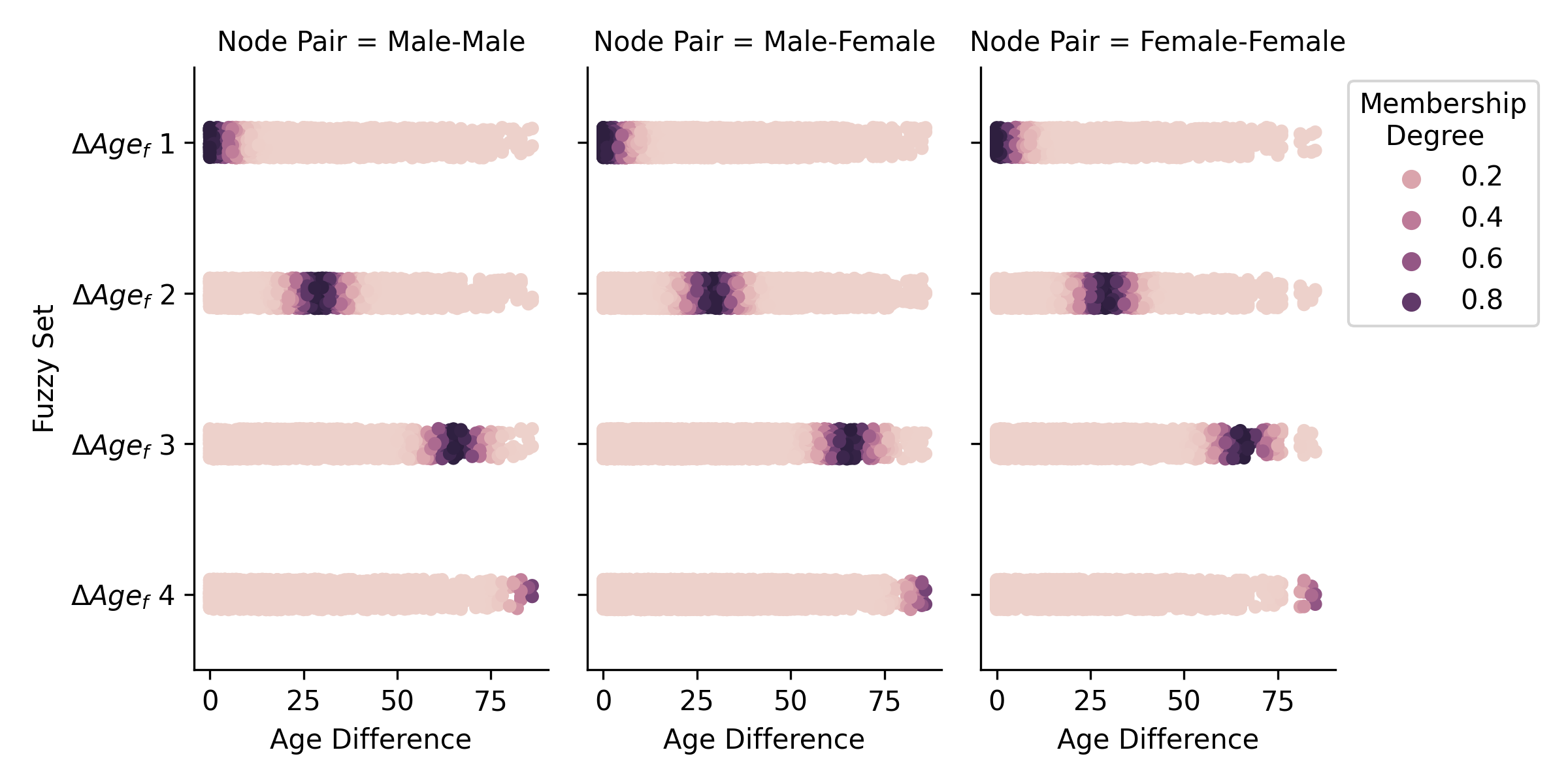}
	\end{minipage}}\\
 \vspace{-3mm}
	\subfigure[Feature Preference]{
		\begin{minipage}[b]{0.46\linewidth}
			\includegraphics[width=1\linewidth]{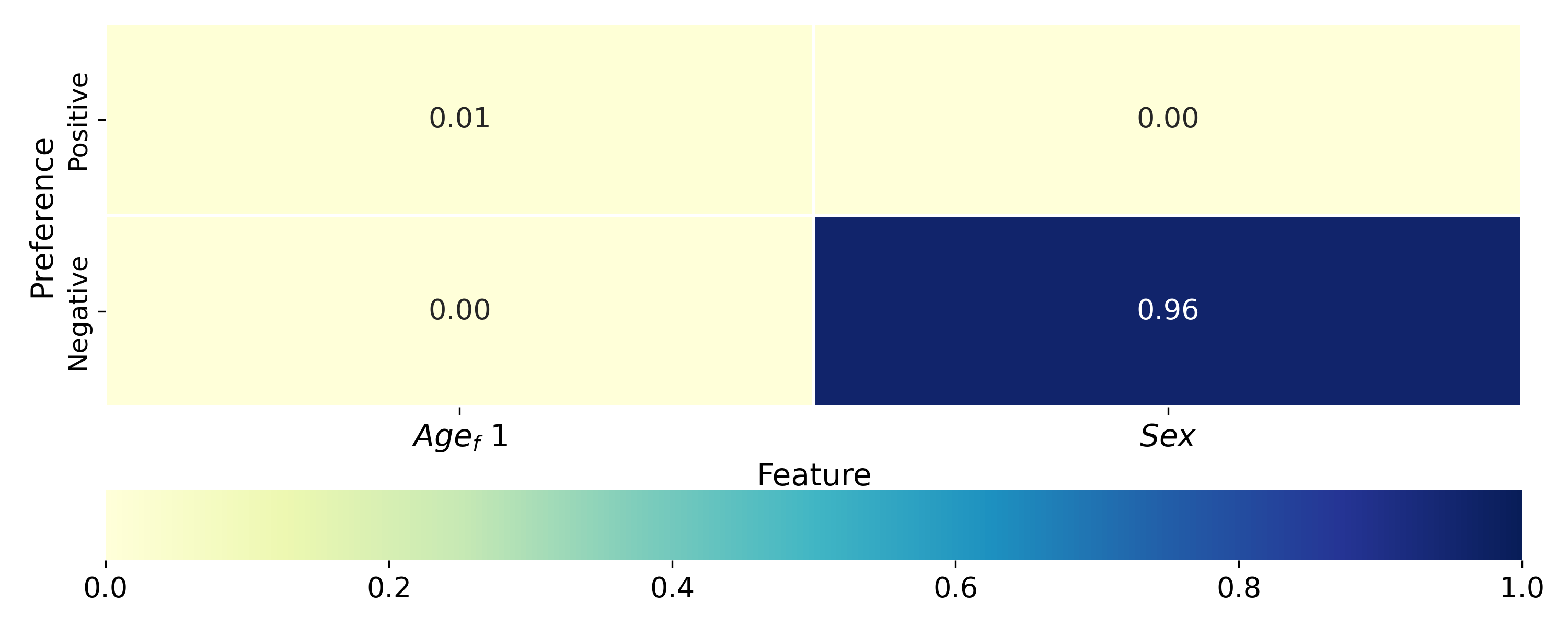}
	\end{minipage}}
 \hspace{-5mm}
	\subfigure[Feature Difference Preference]{
		\begin{minipage}[b]{0.46\linewidth}
			\includegraphics[width=1\linewidth]{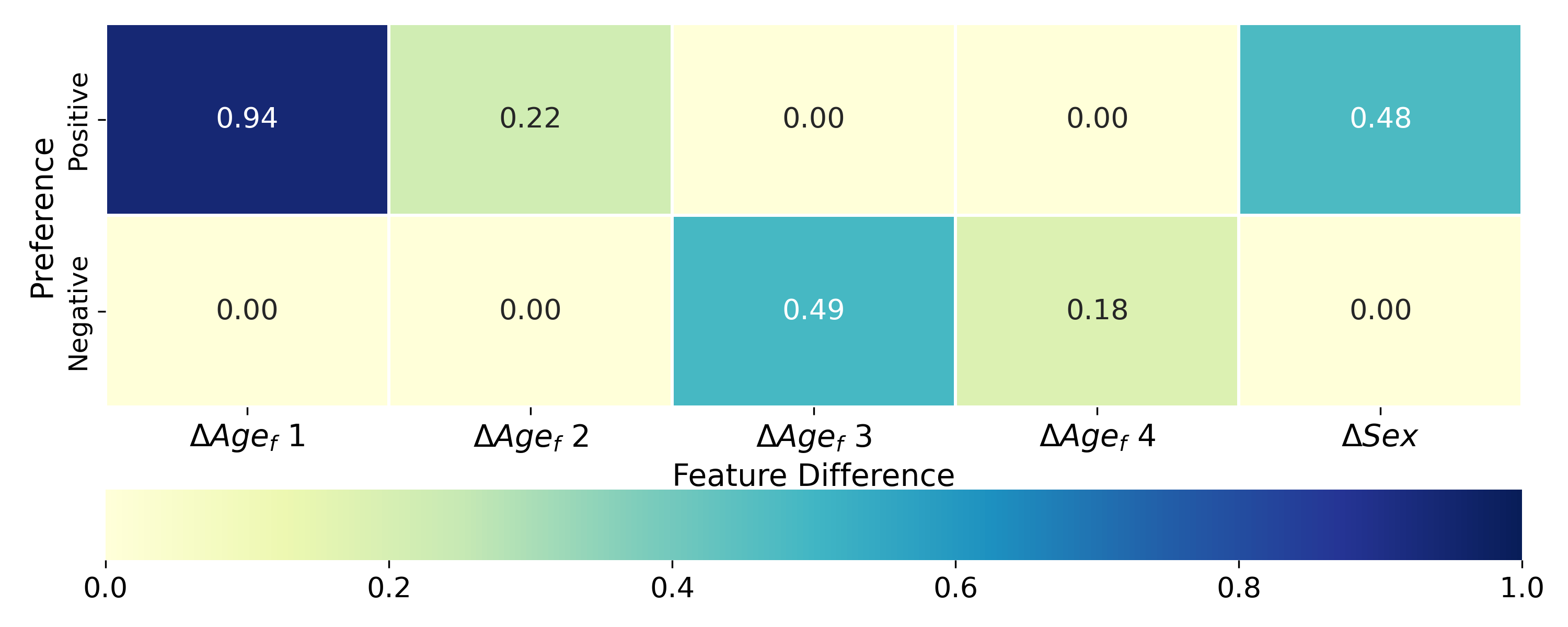}
	\end{minipage}}\\
	\caption{The representation of age features for males and females  (See Fig.~(a)) and the representation of age features between male-male, female-male and female-female nodes (See Fig.~(b)) in Belgium based on the membership degrees to fuzzy sets. The corresponding preferences (positive or negative) and the weights of preference (ranging within $[0,1]$) for each fuzzy membership degree is presented in Fig.~(c) and Fig.~(d). In Fig.~(c), $Age_f i$ and $Sex$ on the x-axis each denotes the $i_{th}$ fuzzy set for age and the crisp representation of sex with binary values ($1$ for male and $0$ for female). In Fig.~(d), $\Delta Age_f i$ and $\Delta Sex$ represents the $i_{th}$ fuzzy set for age difference and the crisp representation of sex differences with binary values ($1$ for the same sex feature and $0$ for different sex feature).}
 \label{BelgiumFeatPref}
\end{figure}

\begin{figure}[H] 
	\centering
	\subfigure[Feature Representation]{
		\begin{minipage}[b]{0.46\linewidth}
			\includegraphics[width=1\linewidth]{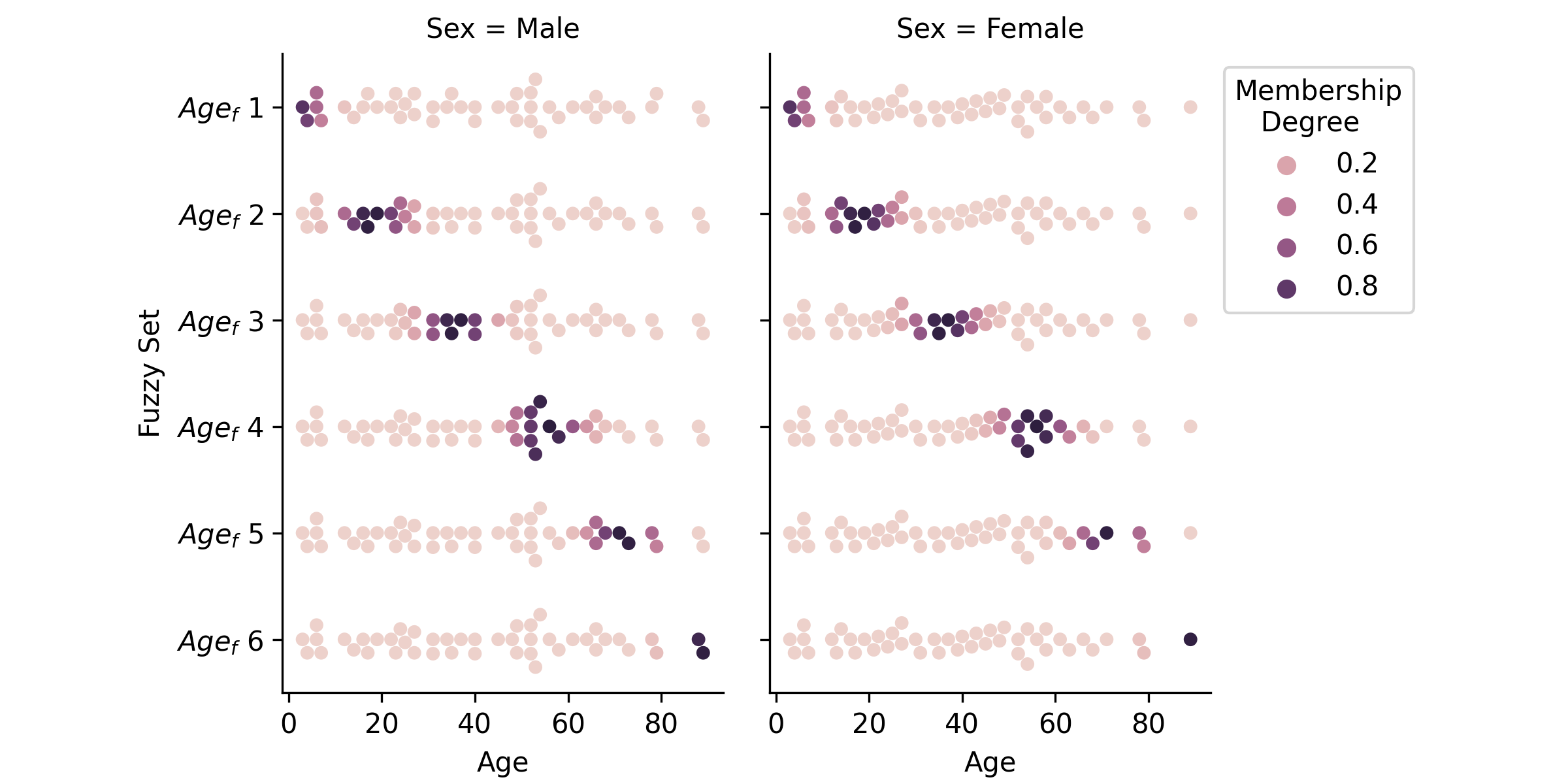}
	\end{minipage}}
 \hspace{-5mm}
	\subfigure[Feature Difference Representation]{
		\begin{minipage}[b]{0.46\linewidth}
			\includegraphics[width=1\linewidth]{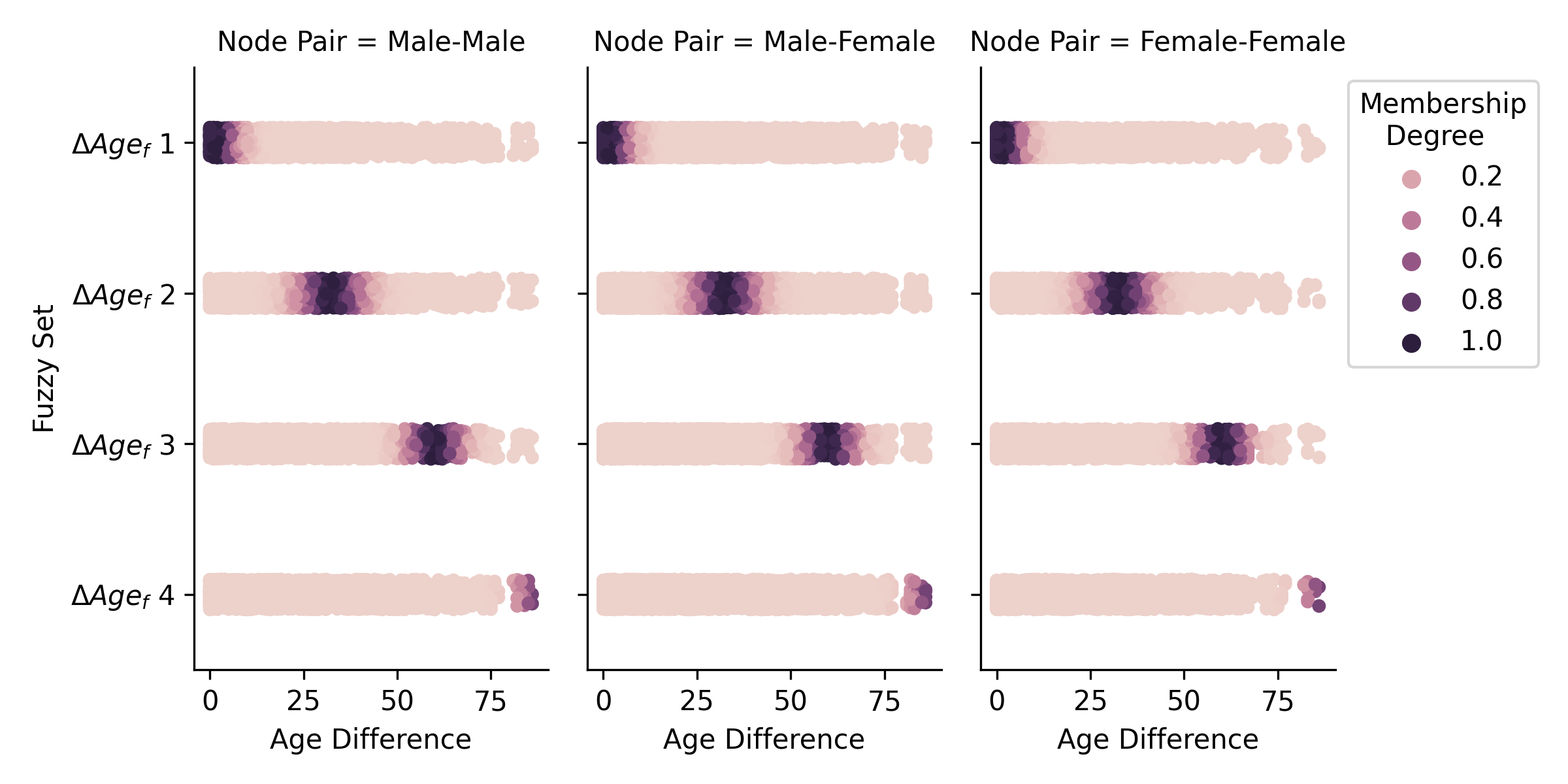}
	\end{minipage}}\\
 \vspace{-3mm}
	\subfigure[Feature Preference]{
		\begin{minipage}[b]{0.46\linewidth}
			\includegraphics[width=1\linewidth]{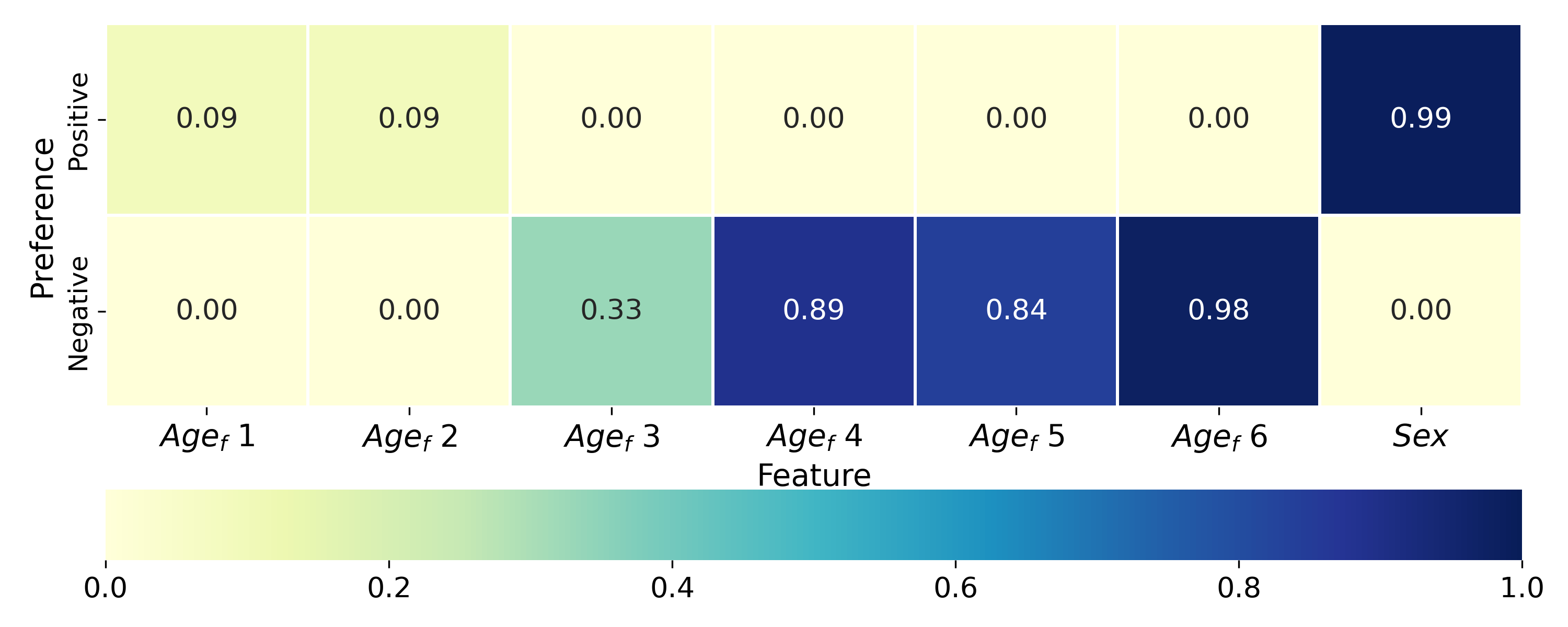}
	\end{minipage}}
 \hspace{-5mm}
	\subfigure[Feature Difference Preference]{
		\begin{minipage}[b]{0.46\linewidth}
			\includegraphics[width=1\linewidth]{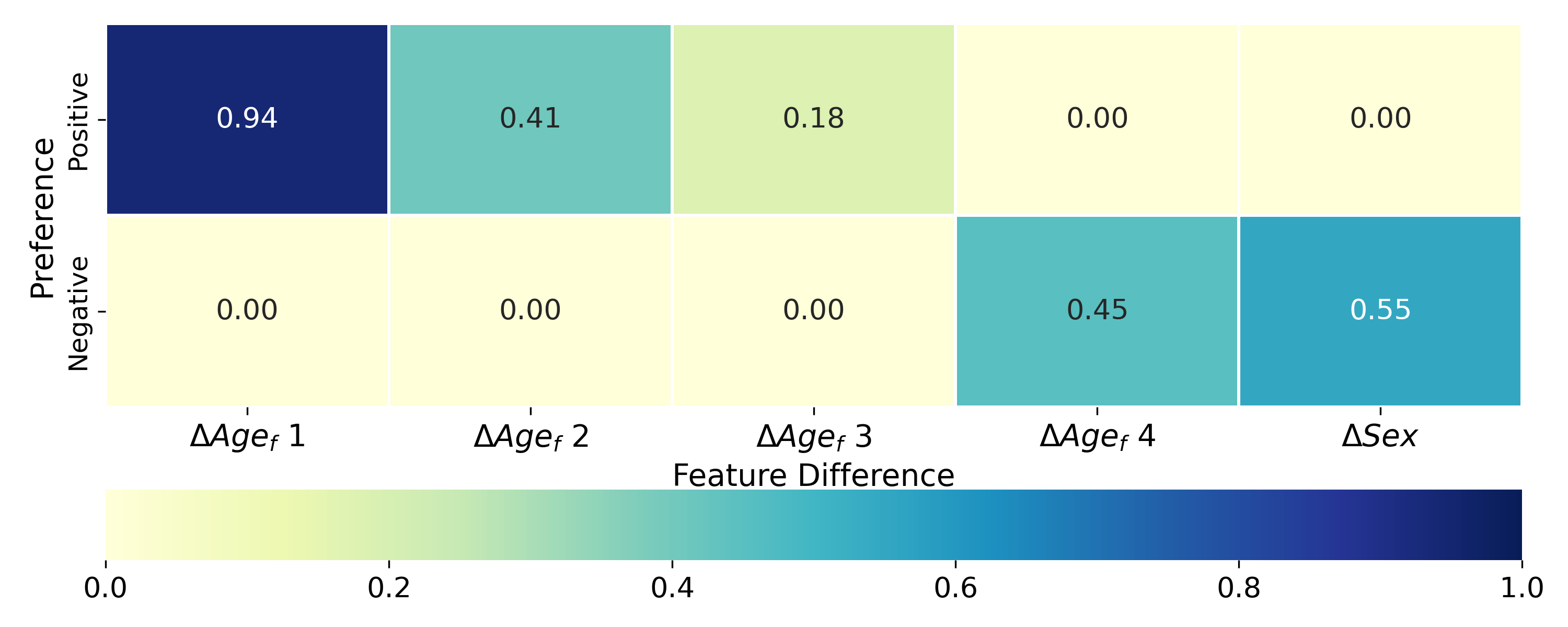}
	\end{minipage}}\\
	\caption{The representation of age features for males and females  (See Fig.~(a)) and the representation of age features between male-male, female-male and female-female nodes (See Fig.~(b)) in Finland based on the membership degrees to fuzzy sets. The corresponding preferences (positive or negative) and the weights of preference (ranging within $[0,1]$) for each fuzzy membership degree is presented in Fig.~(c) and Fig.~(d). In Fig.~(c), $Age_f i$ and $Sex$ on the x-axis each denotes the $i_{th}$ fuzzy set for age and the crisp representation of sex with binary values ($1$ for male and $0$ for female). In Fig.~(d), $\Delta Age_f i$ and $\Delta Sex$ represents the $i_{th}$ fuzzy set for age difference and the crisp representation of sex differences with binary values ($1$ for the same sex feature and $0$ for different sex feature).}
 \label{FinlandFeatPref}
\end{figure}

\begin{figure}[H] 
	\centering
	\subfigure[Feature Representation]{
		\begin{minipage}[b]{0.46\linewidth}
			\includegraphics[width=1\linewidth]{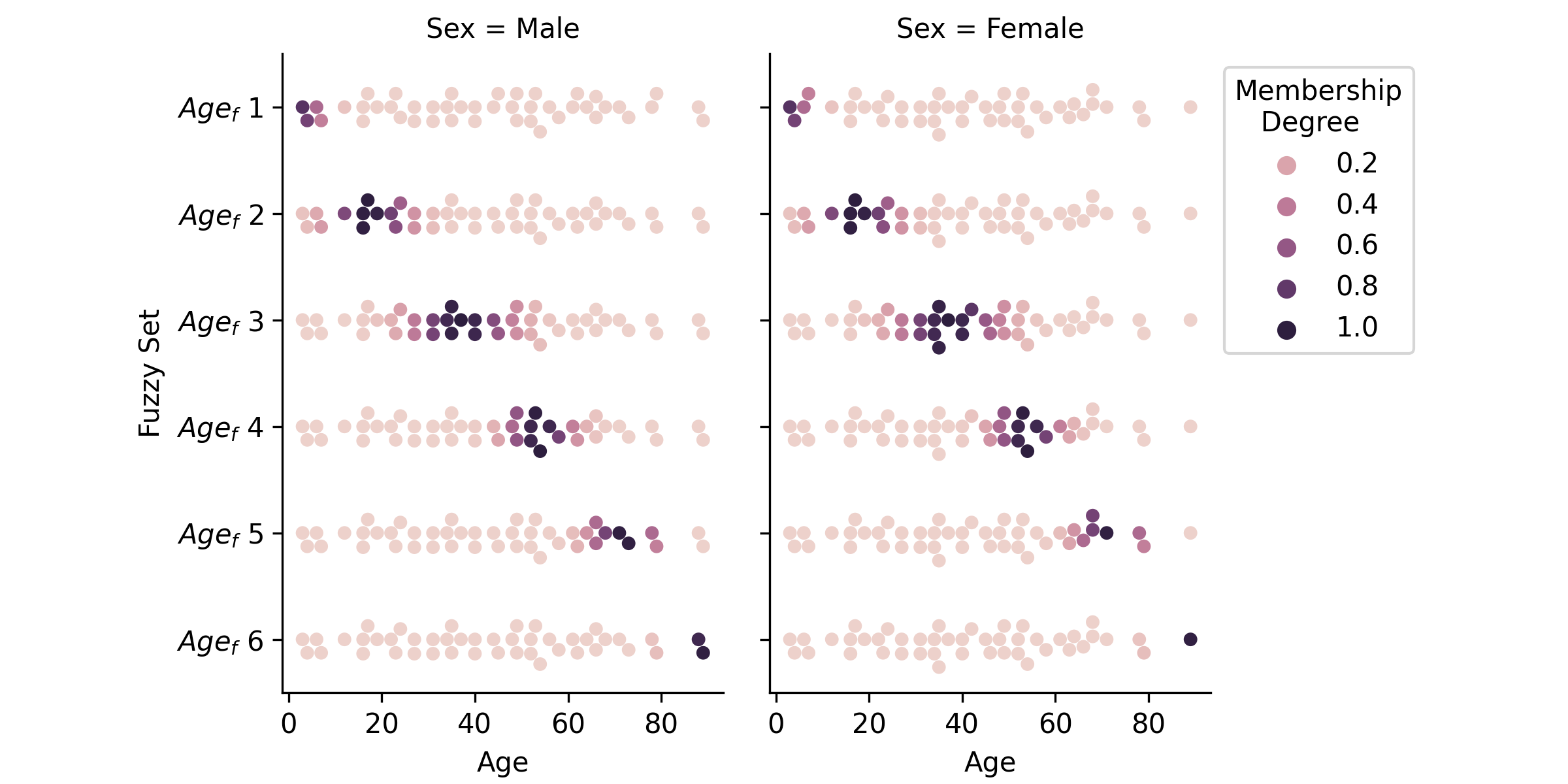}
	\end{minipage}}
 \hspace{-5mm}
	\subfigure[Feature Difference Representation]{
		\begin{minipage}[b]{0.46\linewidth}
			\includegraphics[width=1\linewidth]{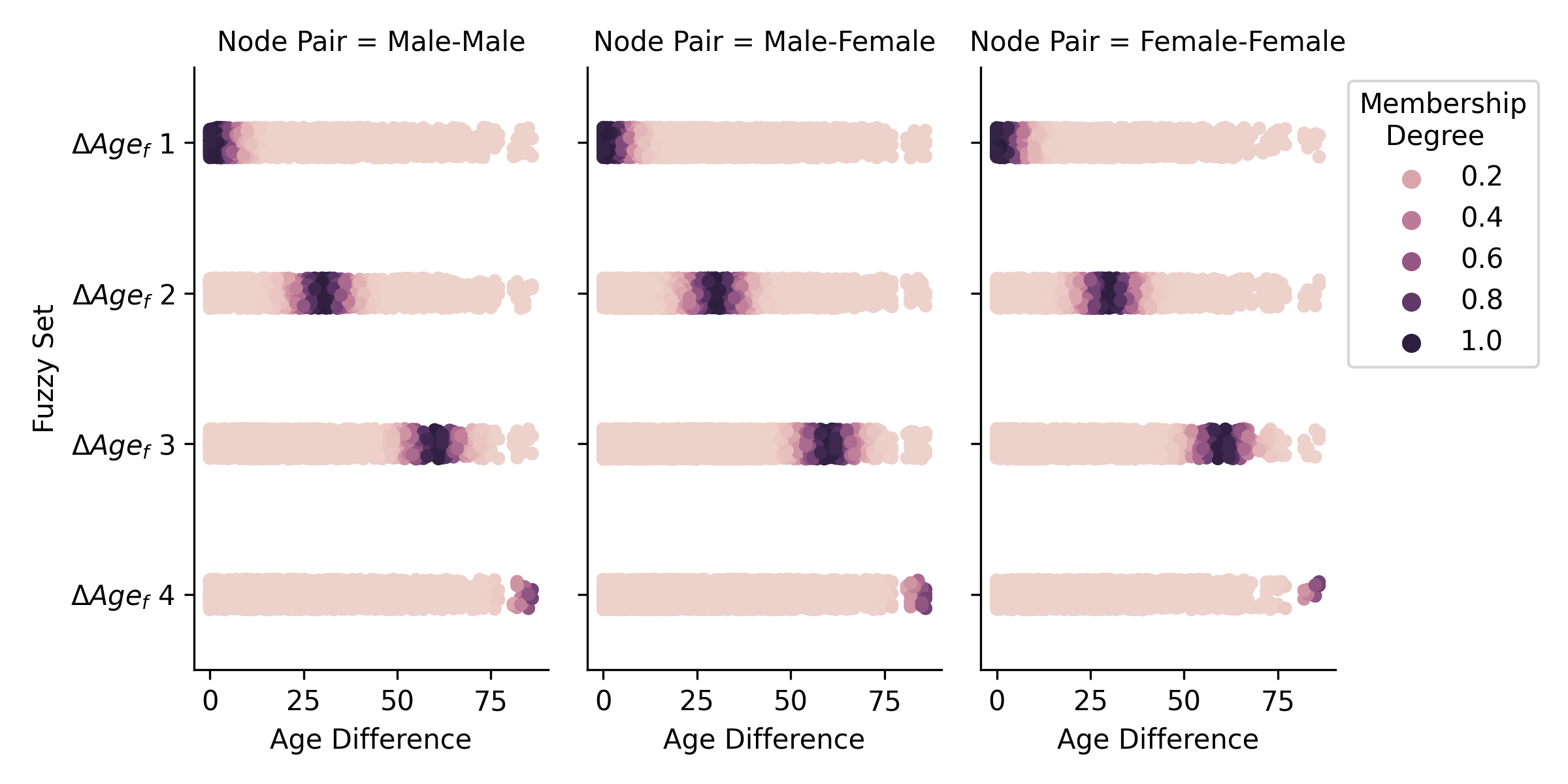}
	\end{minipage}}\\
 \vspace{-3mm}
	\subfigure[Feature Preference]{
		\begin{minipage}[b]{0.46\linewidth}
			\includegraphics[width=1\linewidth]{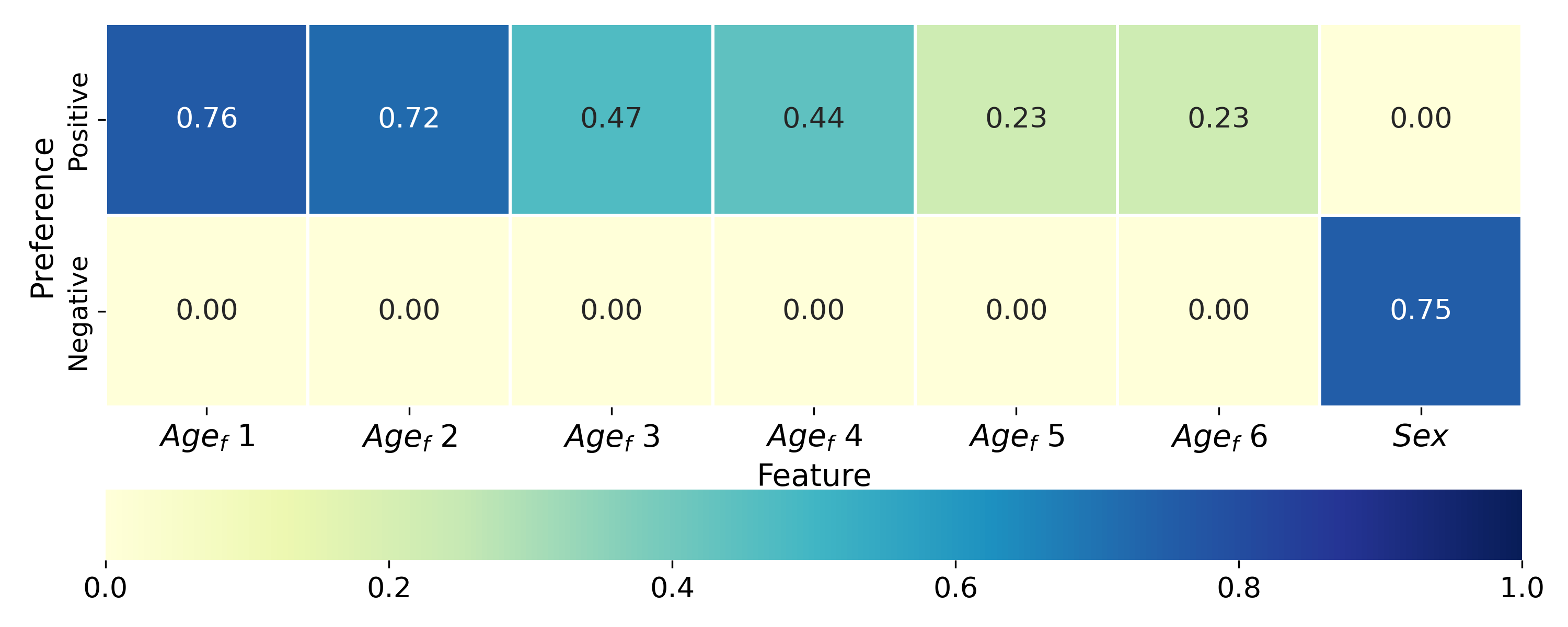}
	\end{minipage}}
 \hspace{-5mm}
	\subfigure[Feature Difference Preference]{
		\begin{minipage}[b]{0.46\linewidth}
			\includegraphics[width=1\linewidth]{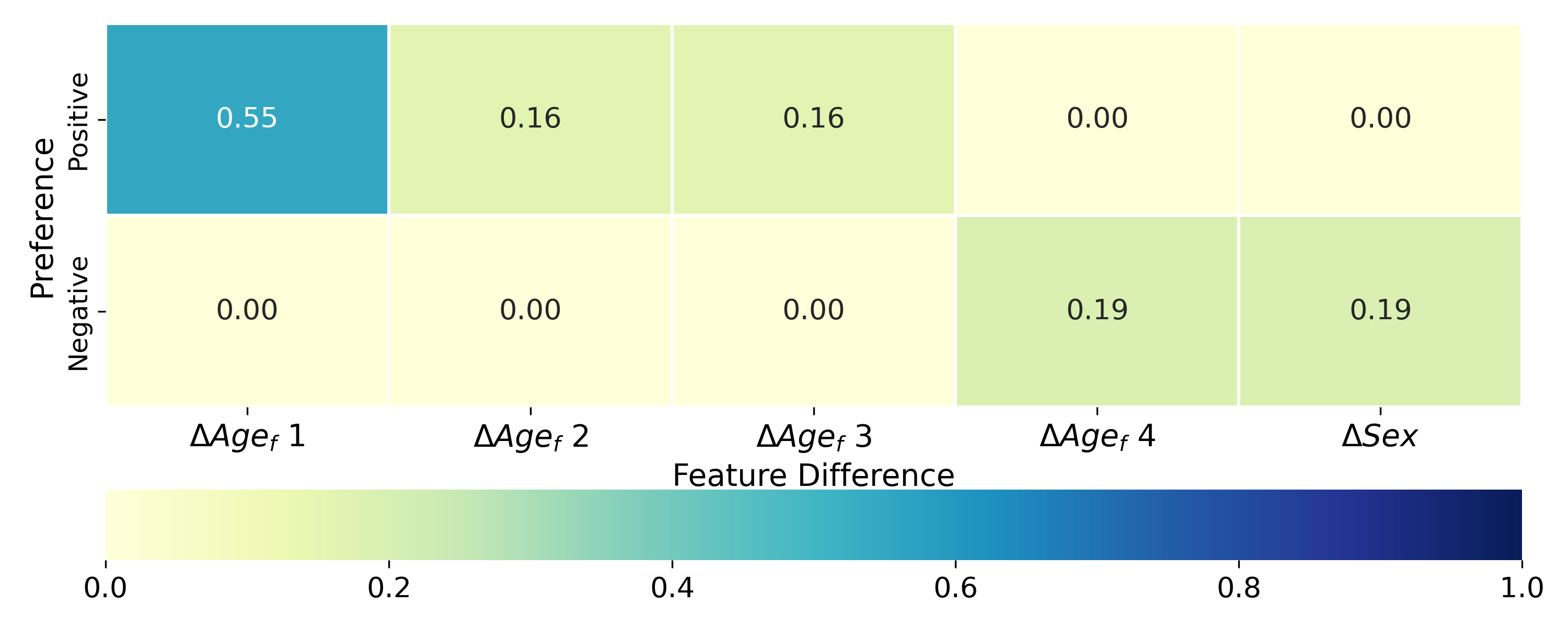}
	\end{minipage}}\\
	\caption{The representation of age features for males and females  (See Fig.~(a)) and the representation of age features between male-male, female-male and female-female nodes (See Fig.~(b)) in Germany based on the membership degrees to fuzzy sets. The corresponding preferences (positive or negative) and the weights of preference (ranging within $[0,1]$) for each fuzzy membership degree is presented in Fig.~(c) and Fig.~(d). In Fig.~(c), $Age_f i$ and $Sex$ on the x-axis each denotes the $i_{th}$ fuzzy set for age and the crisp representation of sex with binary values ($1$ for male and $0$ for female). In Fig.~(d), $\Delta Age_f i$ and $\Delta Sex$ represents the $i_{th}$ fuzzy set for age difference and the crisp representation of sex differences with binary values ($1$ for the same sex feature and $0$ for different sex feature)..}
 \label{GermanyFeatPref}
\end{figure}

\begin{figure}[H] 
	\centering
	\subfigure[Feature Representation]{
		\begin{minipage}[b]{0.46\linewidth}
			\includegraphics[width=1\linewidth]{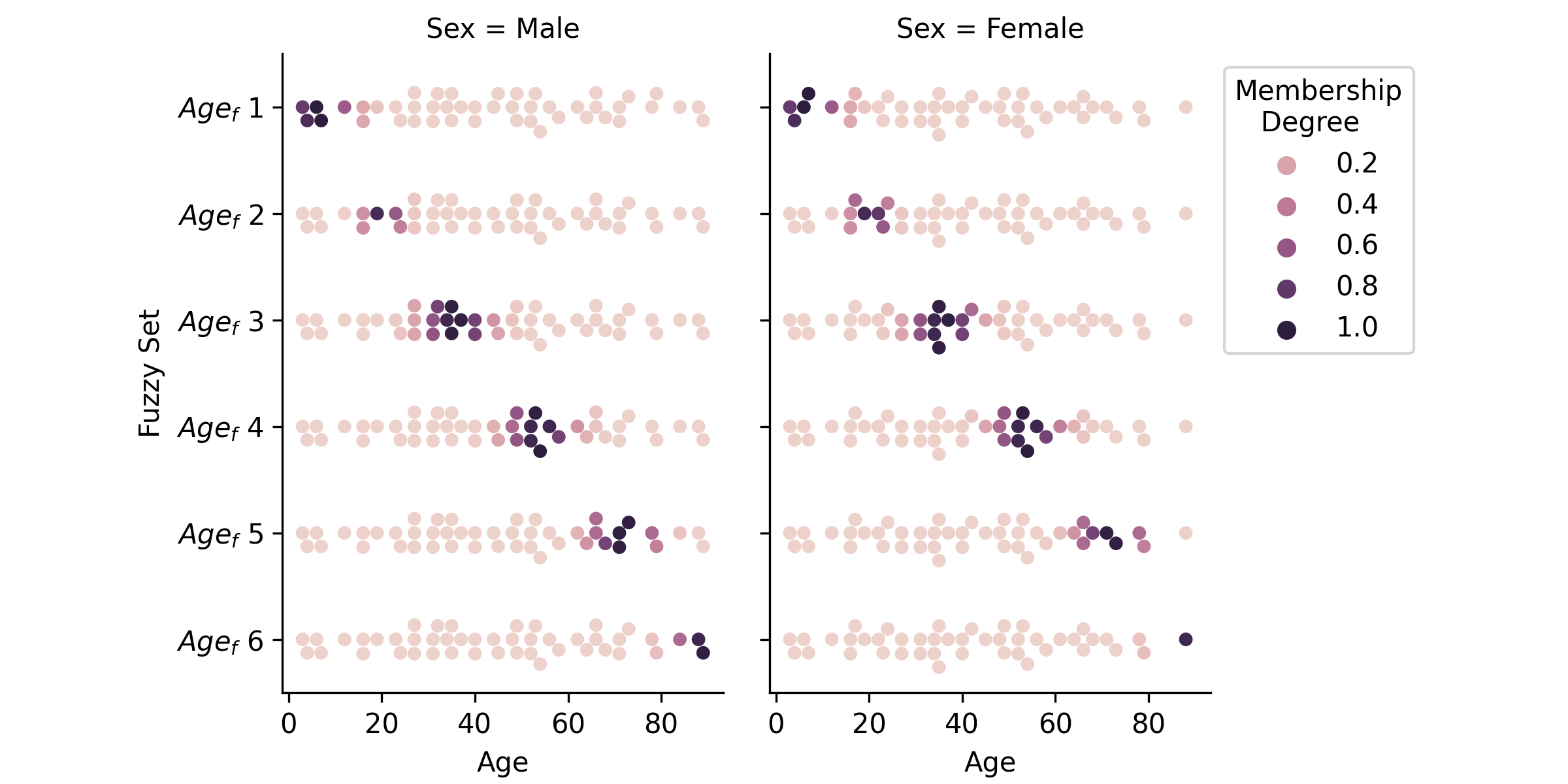}
	\end{minipage}}
 \hspace{-5mm}
	\subfigure[Feature Difference Representation]{
		\begin{minipage}[b]{0.46\linewidth}
			\includegraphics[width=1\linewidth]{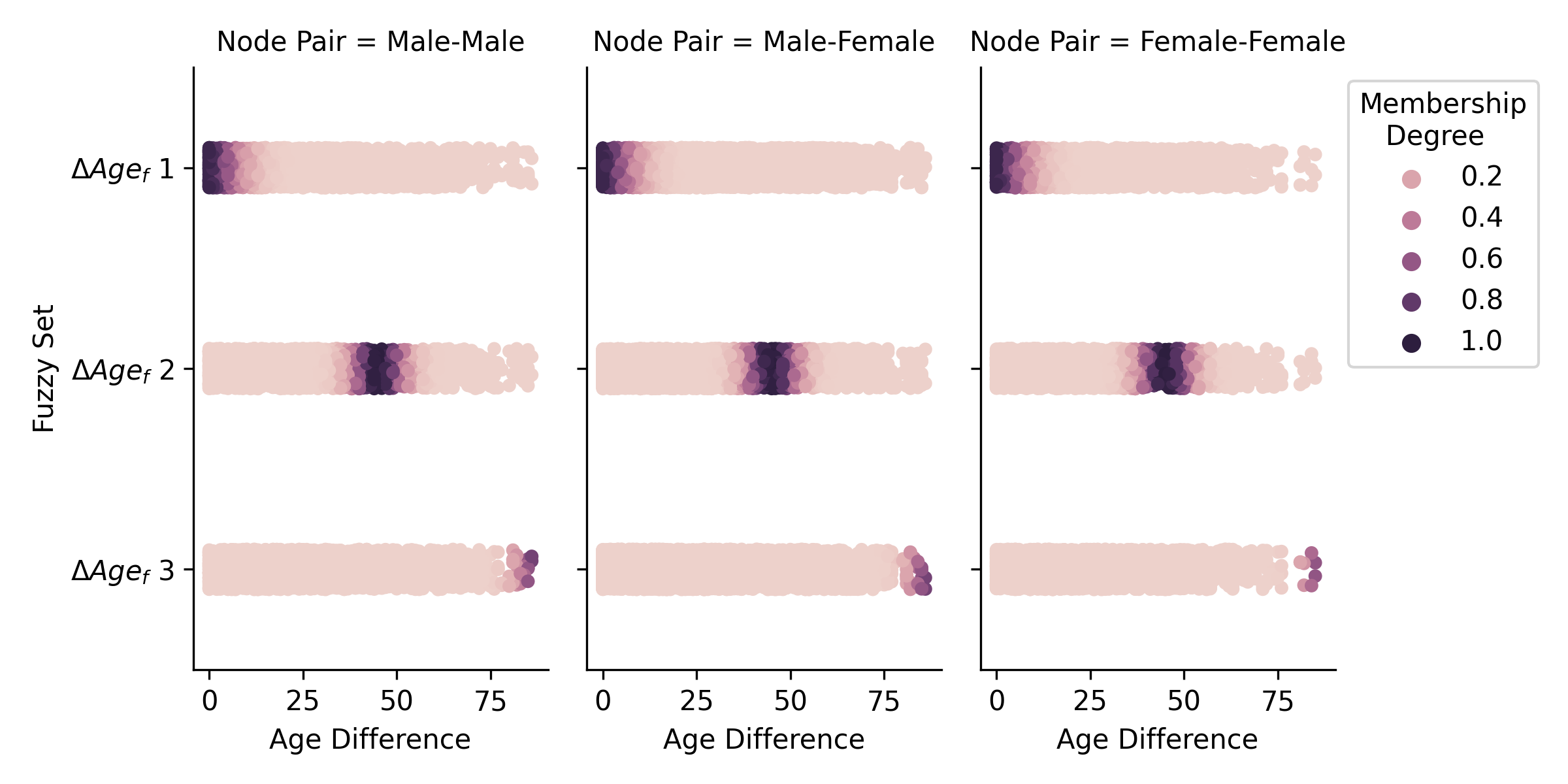}
	\end{minipage}}\\
 \vspace{-3mm}
	\subfigure[Feature Preference]{
		\begin{minipage}[b]{0.46\linewidth}
			\includegraphics[width=1\linewidth]{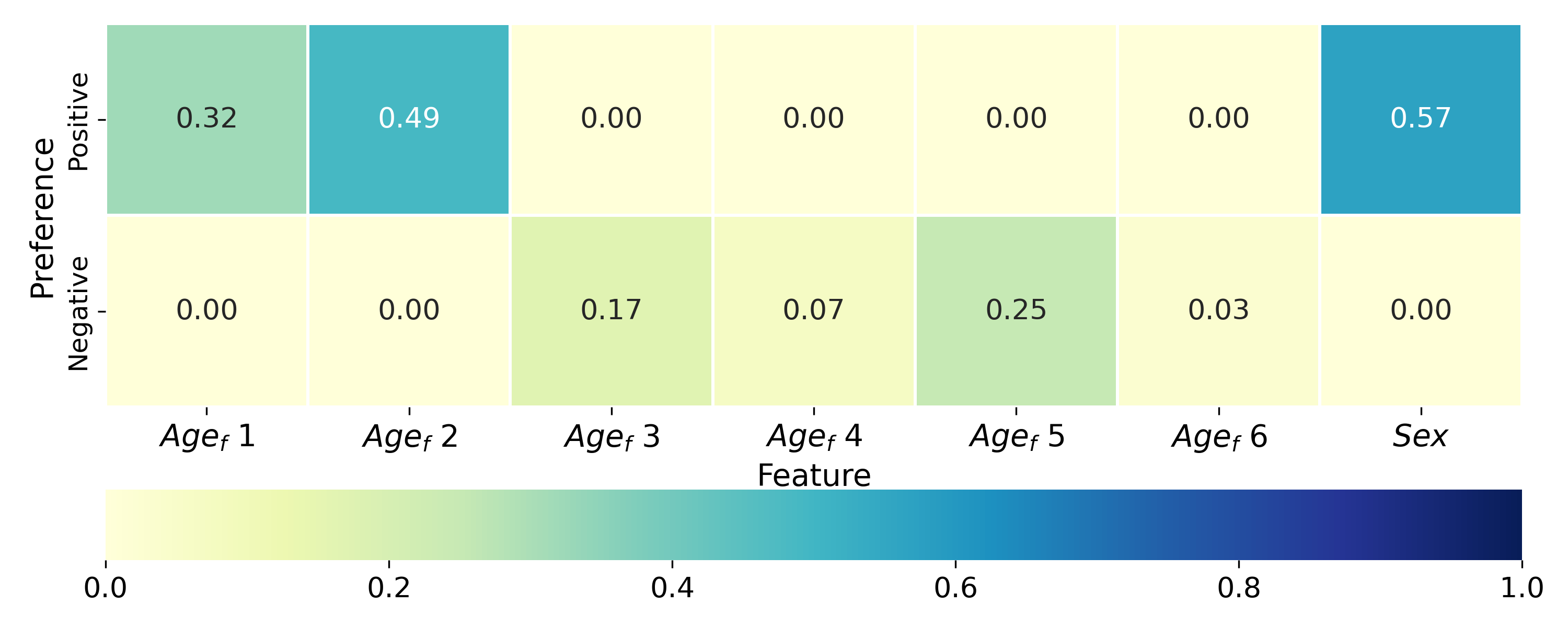}
	\end{minipage}}
 \hspace{-5mm}
	\subfigure[Feature Difference Preference]{
		\begin{minipage}[b]{0.46\linewidth}
			\includegraphics[width=1\linewidth]{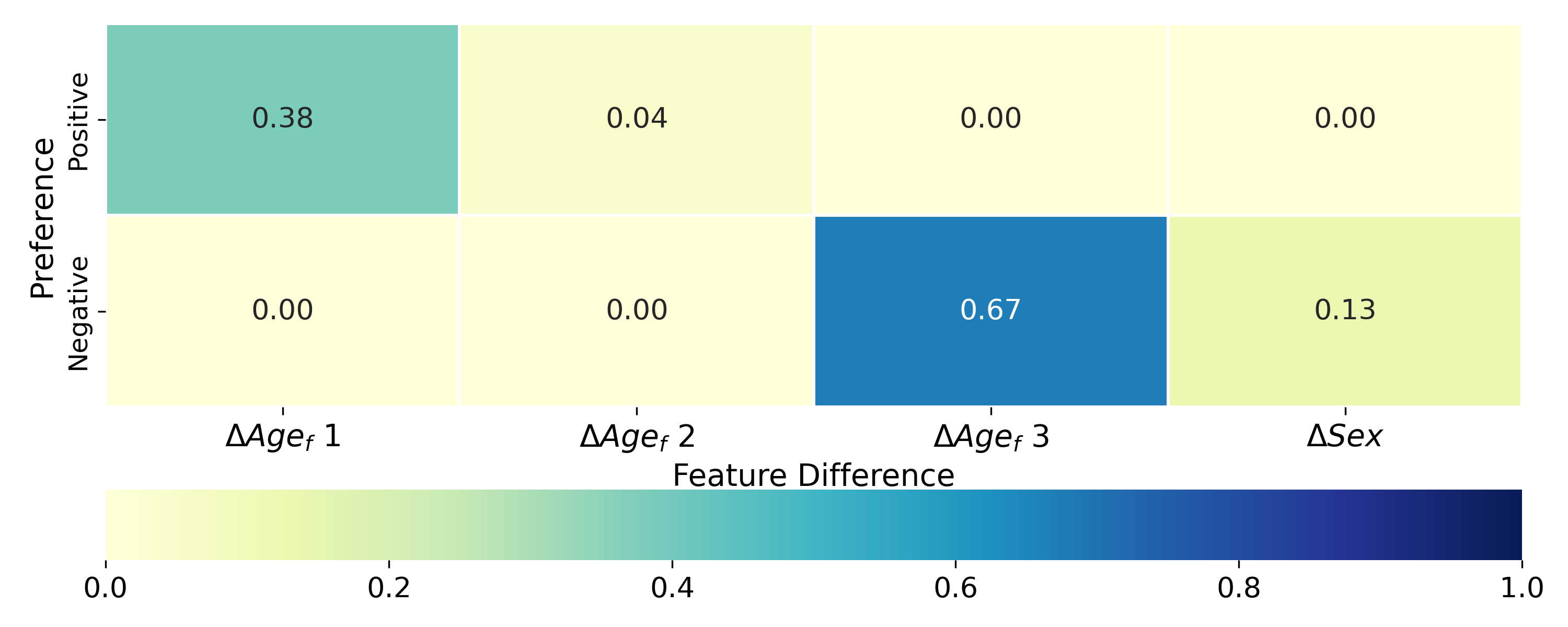}
	\end{minipage}}\\
	\caption{The representation of age features for males and females  (See Fig.~(a)) and the representation of age features between male-male, female-male and female-female nodes (See Fig.~(b)) in Italy based on the membership degrees to fuzzy sets. The corresponding preferences (positive or negative) and the weights of preference (ranging within $[0,1]$) for each fuzzy membership degree is presented in Fig.~(c) and Fig.~(d). In Fig.~(c), $Age_f i$ and $Sex$ on the x-axis each denotes the $i_{th}$ fuzzy set for age and the crisp representation of sex with binary values ($1$ for male and $0$ for female). In Fig.~(d), $\Delta Age_f i$ and $\Delta Sex$ represents the $i_{th}$ fuzzy set for age difference and the crisp representation of sex differences with binary values ($1$ for the same sex feature and $0$ for different sex feature)..}
 \label{ItalyFeatPref}
\end{figure}

\begin{figure}[H] 
	\centering
	\subfigure[Feature Representation]{
		\begin{minipage}[b]{0.46\linewidth}
			\includegraphics[width=1\linewidth]{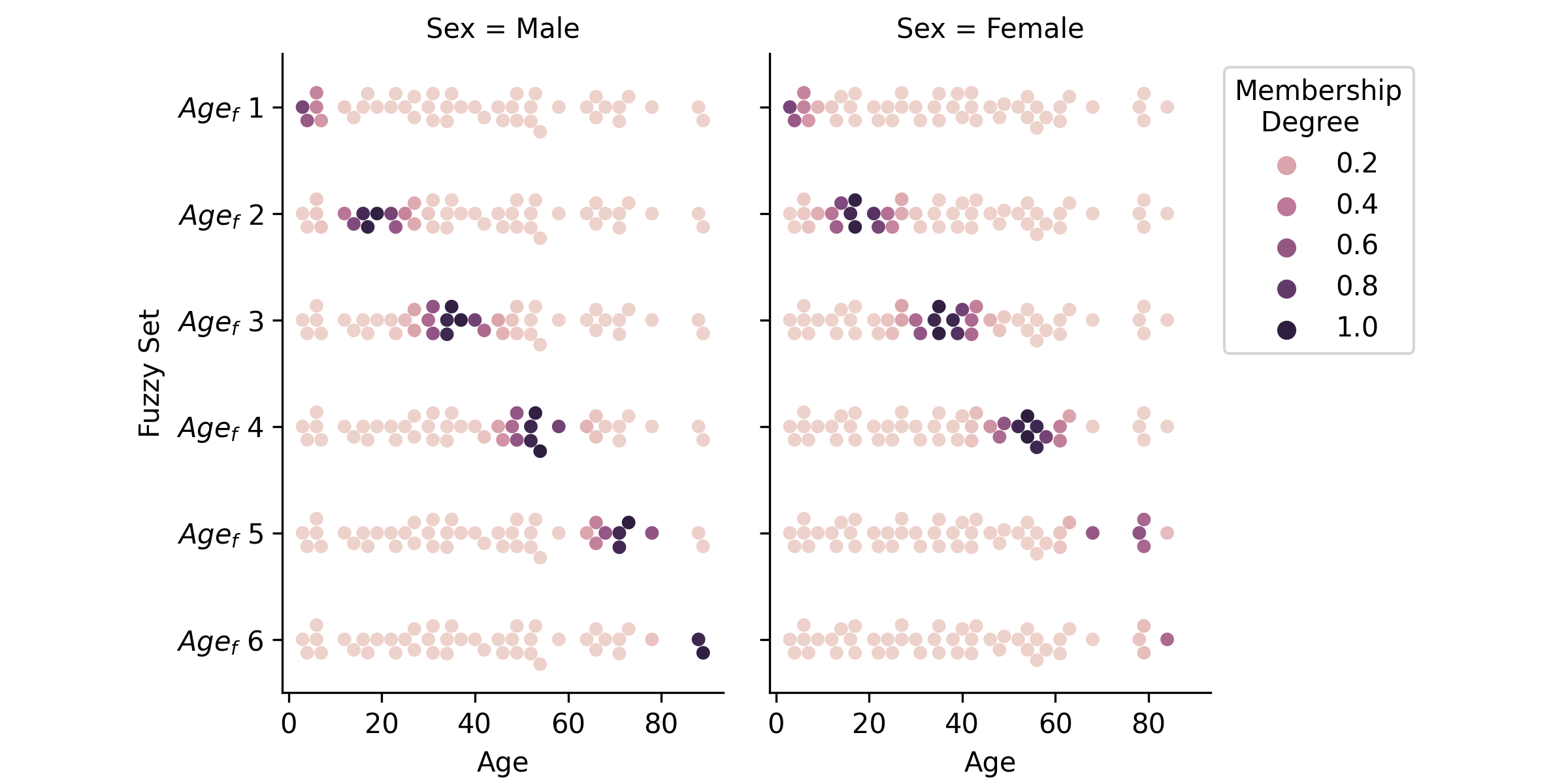}
	\end{minipage}}
 \hspace{-5mm}
	\subfigure[Feature Difference Representation]{
		\begin{minipage}[b]{0.46\linewidth}
			\includegraphics[width=1\linewidth]{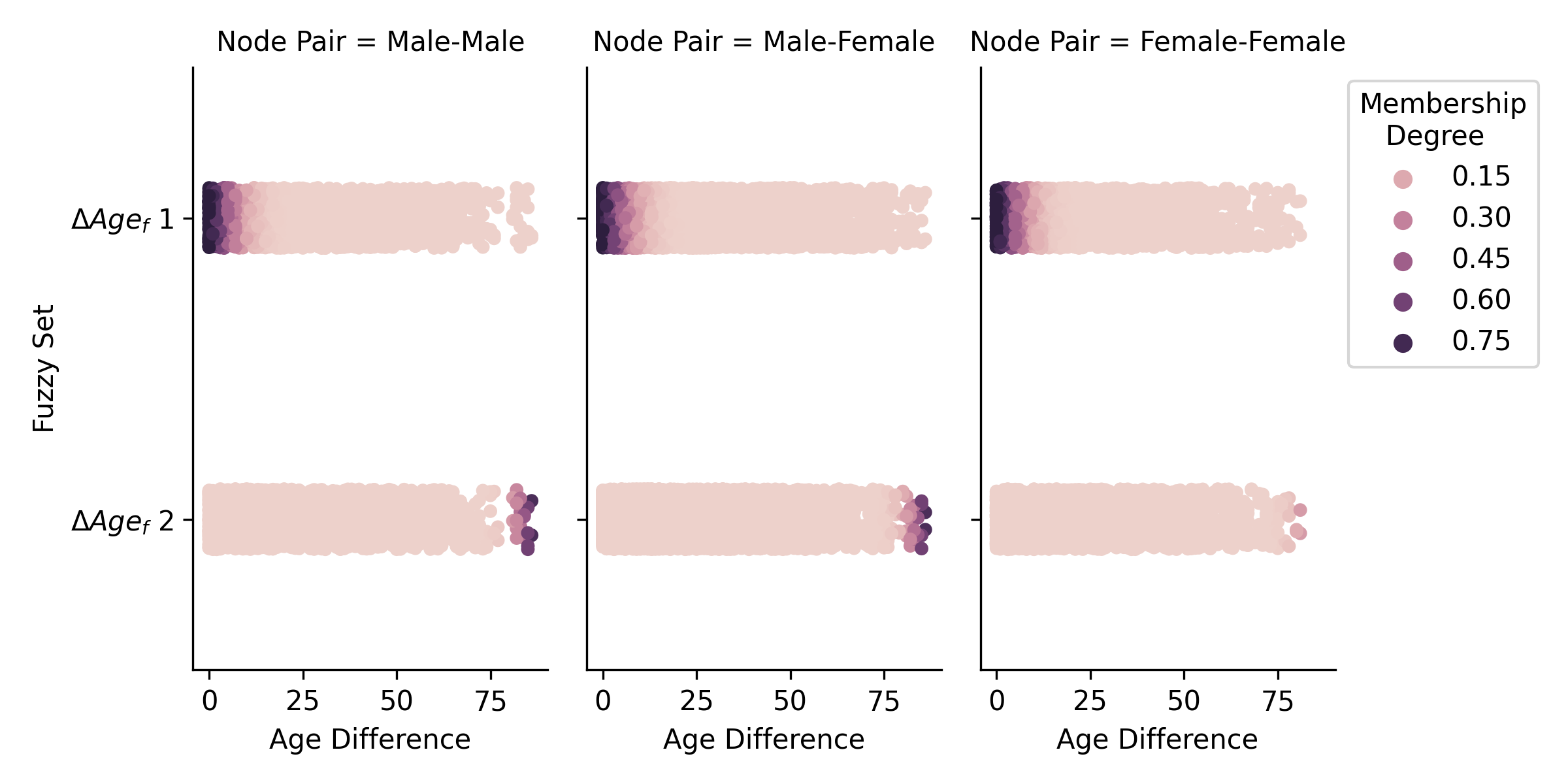}
	\end{minipage}}\\
 \vspace{-3mm}
	\subfigure[Feature Preference]{
		\begin{minipage}[b]{0.46\linewidth}
			\includegraphics[width=1\linewidth]{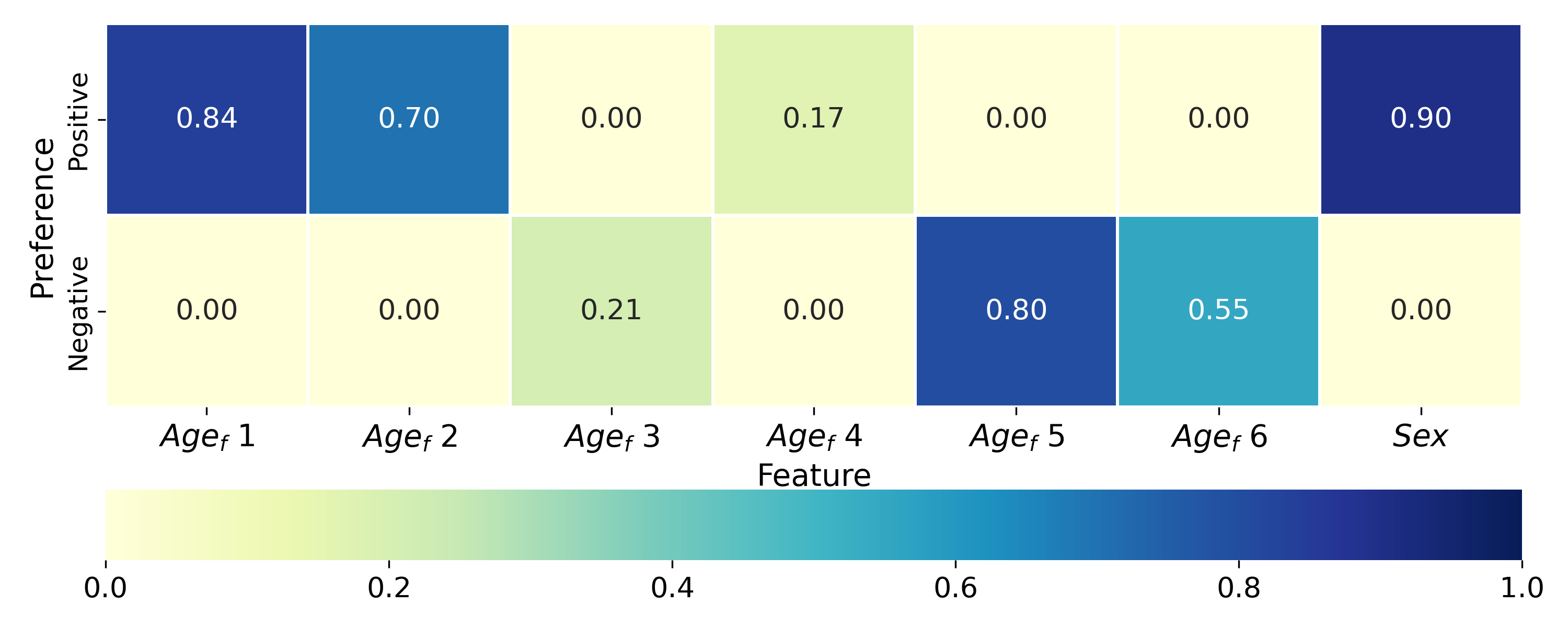}
	\end{minipage}}
 \hspace{-5mm}
	\subfigure[Feature Difference Preference]{
		\begin{minipage}[b]{0.46\linewidth}
			\includegraphics[width=1\linewidth]{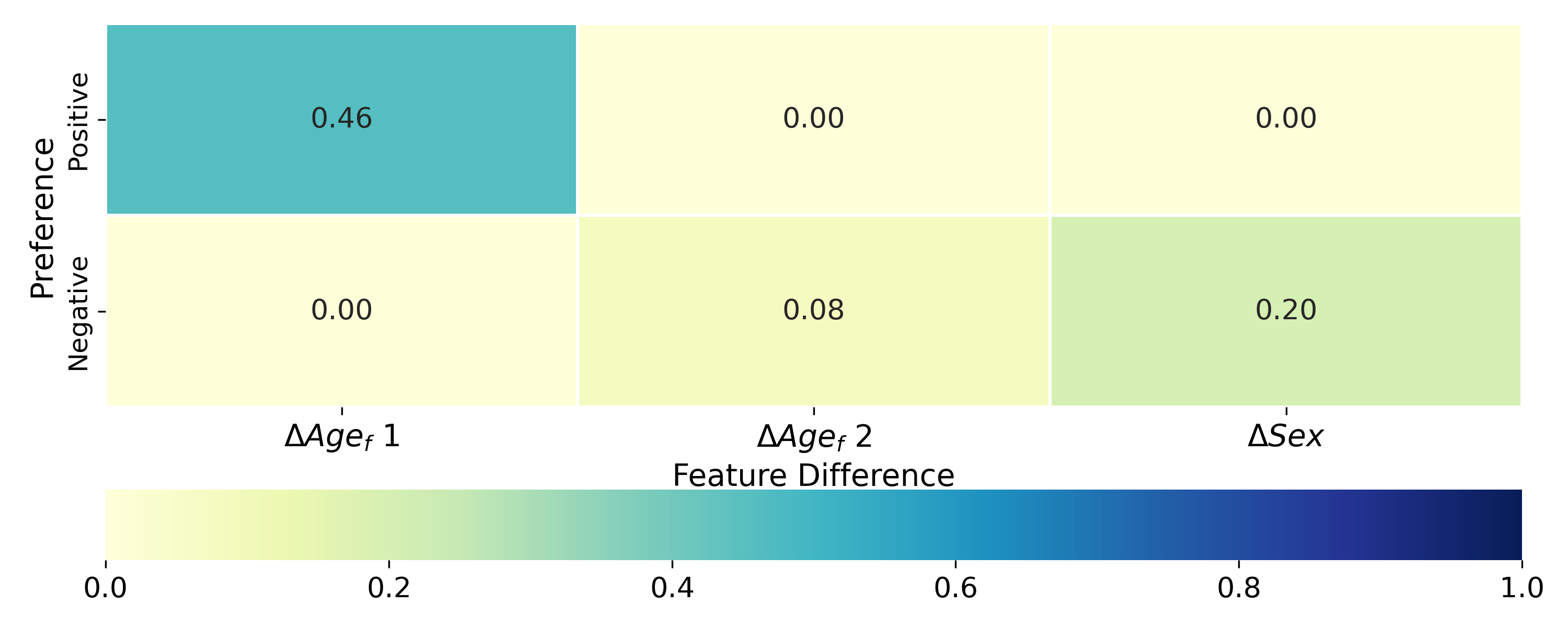}
	\end{minipage}}\\
	\caption{The representation of age features for males and females  (See Fig.~(a)) and the representation of age features between male-male, female-male and female-female nodes (See Fig.~(b)) in Luxembourg based on the membership degrees to fuzzy sets. The corresponding preferences (positive or negative) and the weights of preference (ranging within $[0,1]$) for each fuzzy membership degree is presented in Fig.~(c) and Fig.~(d). In Fig.~(c), $Age_f i$ and $Sex$ on the x-axis each denotes the $i_{th}$ fuzzy set for age and the crisp representation of sex with binary values ($1$ for male and $0$ for female). In Fig.~(d), $\Delta Age_f i$ and $\Delta Sex$ represents the $i_{th}$ fuzzy set for age difference and the crisp representation of sex differences with binary values ($1$ for the same sex feature and $0$ for different sex feature)..}
 \label{LuxembourgFeatPref}
\end{figure}

\begin{figure}[H] 
	\centering
	\subfigure[Feature Representation]{
		\begin{minipage}[b]{0.46\linewidth}
			\includegraphics[width=1\linewidth]{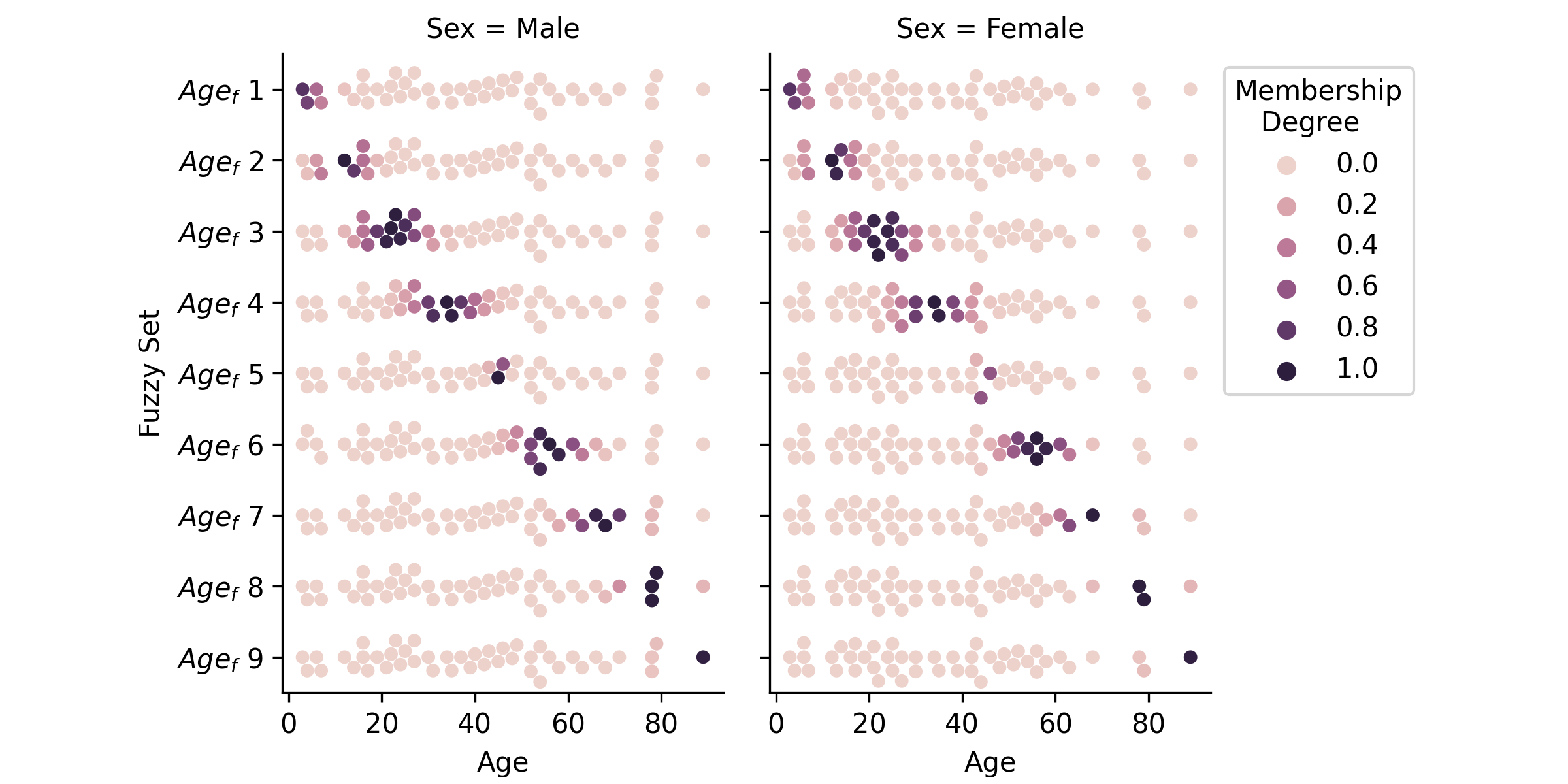}
	\end{minipage}}
 \hspace{-5mm}
	\subfigure[Feature Difference Representation]{
		\begin{minipage}[b]{0.46\linewidth}
			\includegraphics[width=1\linewidth]{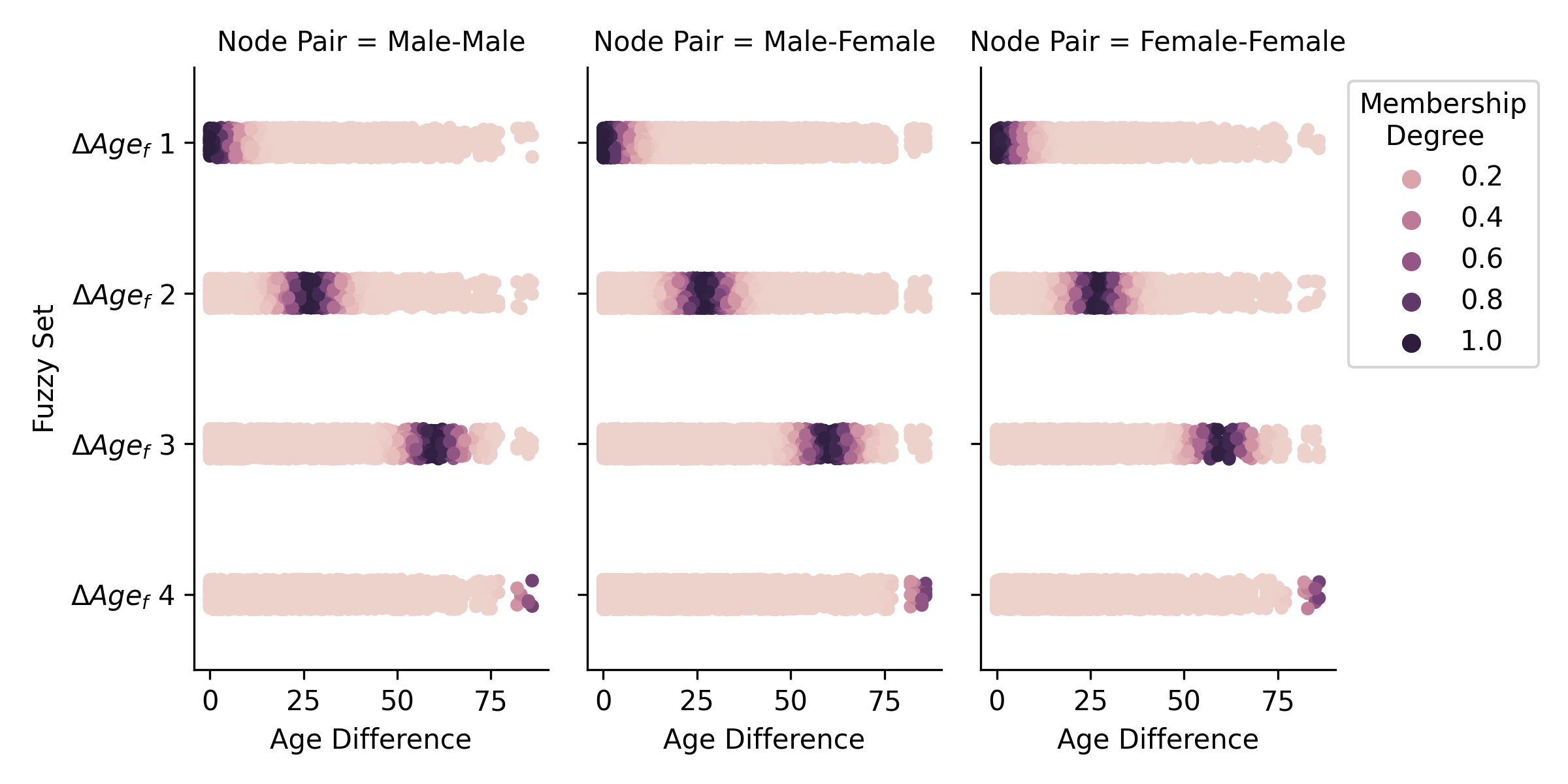}
	\end{minipage}}\\
 \vspace{-3mm}
	\subfigure[Feature Preference]{
		\begin{minipage}[b]{0.46\linewidth}
			\includegraphics[width=1\linewidth]{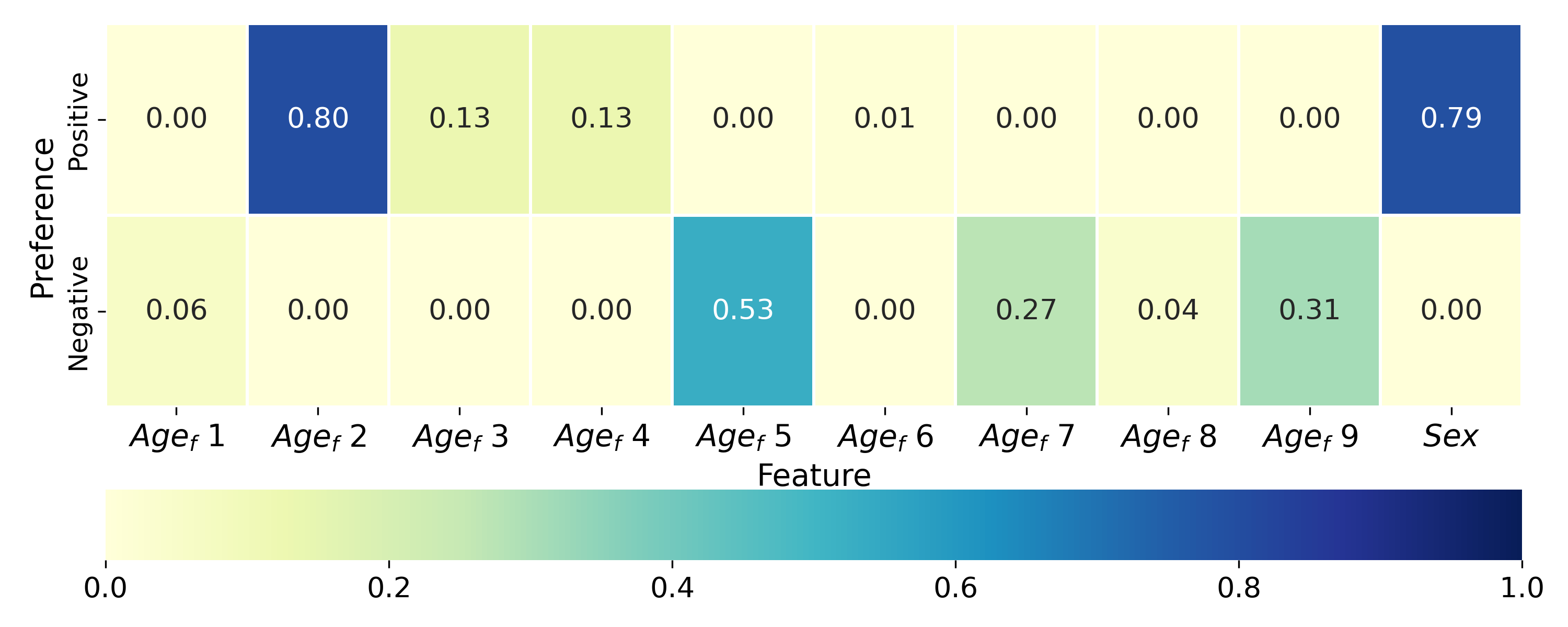}
	\end{minipage}}
 \hspace{-5mm}
	\subfigure[Feature Difference Preference]{
		\begin{minipage}[b]{0.46\linewidth}
			\includegraphics[width=1\linewidth]{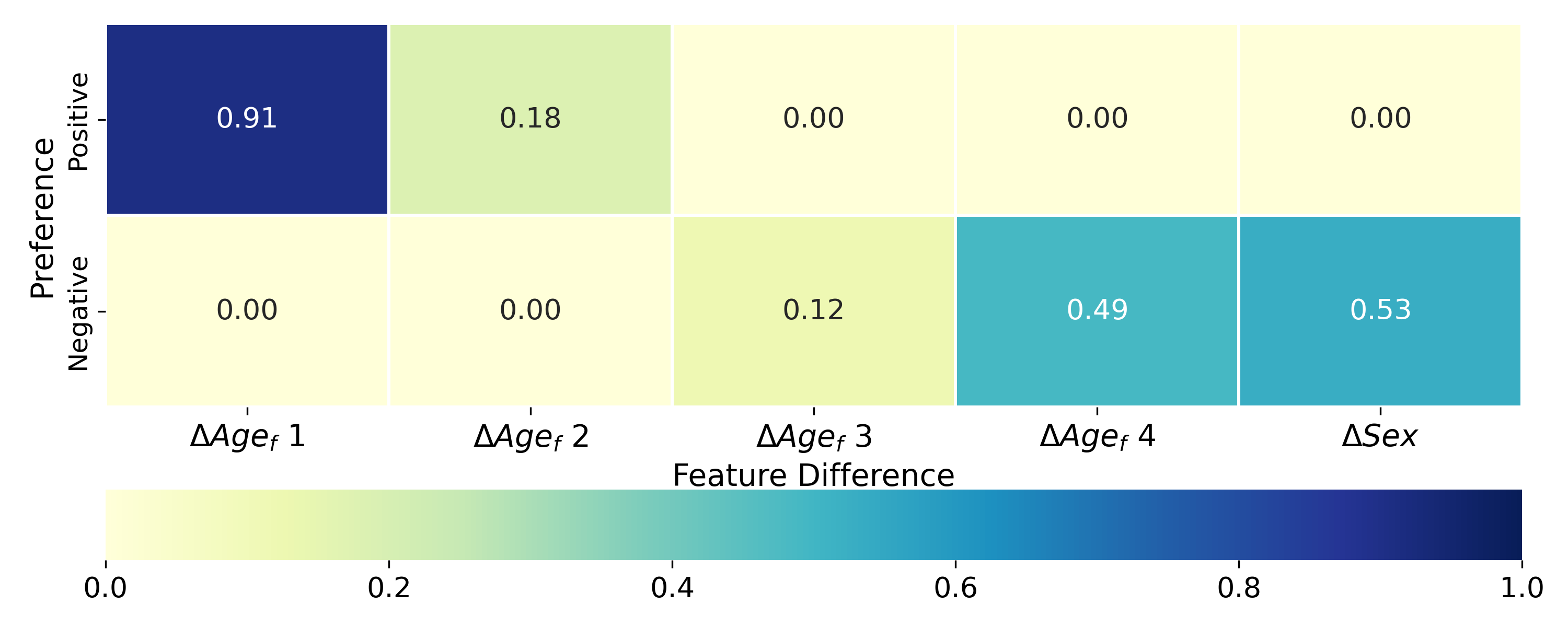}
	\end{minipage}}\\
	\caption{The representation of age features for males and females  (See Fig.~(a)) and the representation of age features between male-male, female-male and female-female nodes (See Fig.~(b)) in Poland based on the membership degrees to fuzzy sets. The corresponding preferences (positive or negative) and the weights of preference (ranging within $[0,1]$) for each fuzzy membership degree is presented in Fig.~(c) and Fig.~(d). In Fig.~(c), $Age_f i$ and $Sex$ on the x-axis each denotes the $i_{th}$ fuzzy set for age and the crisp representation of sex with binary values ($1$ for male and $0$ for female). In Fig.~(d), $\Delta Age_f i$ and $\Delta Sex$ represents the $i_{th}$ fuzzy set for age difference and the crisp representation of sex differences with binary values ($1$ for the same sex feature and $0$ for different sex feature)..}
 \label{PolandFeatPref}
\end{figure}

In the experiments, we represent the features and feature differences based on the calibrated fuzzy representation principles in section \ref{paramselectsection} for each country, including Belgium (Fig.~\ref{BelgiumFeatPref}), Finland (Fig.~\ref{FinlandFeatPref}), Germany (Fig.~\ref{GermanyFeatPref}), Italy (Fig.~\ref{ItalyFeatPref}), Luxembourg (Fig.~\ref{LuxembourgFeatPref}) and Poland (Fig.~\ref{PolandFeatPref}).

As shown in Fig.~\ref{BelgiumFeatPref} (a) and (b), people's interest in age features and age differences can be respectively represented by one fuzzy set and four fuzzy sets. A single age value, either male or female, is represented by a membership degree to the first fuzzy set of age ($Age_f 1$), where smaller ages around $0$ can have higher membership values. In addition, the representation of the age difference between any node pair can also be unfolded into a vector that is composed of four membership degrees to the four fuzzy sets (Denoted by $\Delta Age i$ in Fig.~\ref{BelgiumFeatPref} (b)); See Fig.~\ref{fuzzyBselectparam} (e)). A large number of nodes, centring around the age of $22$ and $65$ are respectively assigned with higher membership degrees to the second and third fuzzy set of age ($\Delta Age_f 2$ and $\Delta Age_f3$), which captures people's interests and is in line with the parameter rankings in the model calibration process (See Fig.~\ref{fuzzyBselectparam}(d)). The representation of age features and age differences, considering "males/females" and "male-male/female-female/male-female" relations, does not vary much, except for a few differences in density due to the differences in age distributions of males and females (See Fig.~\ref{DNfeat22} (a)). In Fig.~\ref{BelgiumFeatPref} (c) and (d), the age features and age differences, unfolded by these fuzzy sets, are preferred either positively or negatively based on a preference weight. Females and the age difference around $0$ (represented by the first fuzzy set) are strongly preferred with a preference weight that is close to $1$. In addition, people in Belgium also have a positive preference for age differences around $29$ ($\Delta Age 2$, the second fuzzy set of age difference; See Fig.~\ref{fuzzyBselectparam} (d)) and a negative preference for age differences around $65$ ($\Delta Age 2$, the second fuzzy set of age difference; See Fig.~\ref{fuzzyBselectparam} (d)) and $90$. The corresponding preference weights are over $0.15$, significantly higher than the preference for age features shown in Fig.~\ref{BelgiumFeatPref} (c). This indicates that people in Belgium have more interests in age differences and sex features, where they strongly prefer to be connected with females at a similar age in network formation.

As shown in Fig.~\ref{FinlandFeatPref} (a) and (b), people's interest in age features and age differences can be each represented by six fuzzy sets and four fuzzy sets. A greater number of nodes are distributed around the age of $50$ and are assigned with higher membership degrees to the fourth age fuzzy set ($Age_f 4$, centring around the age of $54$). In addition, a significant number of node pairs have an age difference around the age of $30$ and $30$, which have higher membership degrees to the second and the third fuzzy sets of age difference ($\Delta Age_f 2$ and $\Delta Age_f 3$; See Fig.~\ref{fuzzyFselectparam} (e)). The abovementioned phenomenon indicates people's interests in age values around $50$ and the age differences around $0$, $30$ and $60$. Similar to the case of Belgium, the distributions of the membership degrees do not vary much with the sex features due to the similar age distributions of males and females (See Fig.~\ref{DNfeat22} (b)). In Fig.~\ref{FinlandFeatPref} (c) and (d), males and age differences around $0$ are strongly preferred with a preference weight over $0.90$. In contrast, people in Finland have a strong negative preference for ages around $54$, $72$ and $90$, with a preference weight over $0.85$. This indicates that people have more interest in age differences around $0$ and seek connections with people of a similar age, in contrast with the attention and negative preference for age values around $54$. In addition, people in Finland also show positive preference for age difference around $32$ ($\Delta Age_f^2$, the second fuzzy set of age difference) and age difference around $60$ ($\Delta Age_f^3$, the third fuzzy set of age difference) and negative preference for age around $36$ ($Age_f^3$, the third fuzzy set of age). They are characterised with a preference weight over $0.15$, which, multiplied with greater membership degrees to the corresponding fuzzy set, can also have a great impact on the interactions.

As shown in Fig.~\ref{GermanyFeatPref} (a) and (b), people's interest in age features and age differences can also be represented by six fuzzy sets and four fuzzy sets. Similar to the cases of Finland, a greater number of nodes are distributed around the age range of $[25-60]$ and have higher membership degrees to the corresponding fuzzy set, including the second and the third fuzzy set of age ($Age_f 2$ and $Age_f 3$, each centring around $37$ and $54$; See Fig.~\ref{DNfeat22} (e)). There are also a significant number of node pairs allocated around the age difference of $30$ and $60$, characterised with higher membership degrees to the second and the third fuzzy sets of age difference ($\Delta Age_f 2$ and $\Delta Age_f 3$; See Fig.~\ref{fuzzyGselectparam} (e)). The abovementioned phenomenon indicates people's interests in age values and age differences. Similar to the cases of the abovementioned countries, the distributions of the membership degrees do not vary much with the sex features due to the similar age distributions of males and females (See Fig.~\ref{DNfeat22} (b)). In Fig.~\ref{GermanyFeatPref} (c) and (d), females and age values around $0$, $17$ and $37$ (each related to the first, second and third fuzzy sets of age, denoted by $Age_f 1$, $Age_f 2$ and $Age_f 3$) are strongly preferred with a preference weight over $0.70$. This is followed by relatively weaker positive preferences for other age values, positive preferences for age differences around $30$ and $60$ (each related to the second and third fuzzy sets of age difference, denoted by $\Delta Age_f 2$ and $\Delta Age_f 3$) and the negative preference for age difference around $90$ (related to $\Delta Age_f 4$, the fourth fuzzy set of age) and the same sex features. Both the heterogeneous patterns of feature representation and preferences can influence the network formation process, which makes it hard to find a decisive interaction rule and detriments the interpretability of resulting network patterns.

As shown in Fig.~\ref{ItalyFeatPref} (a) and (b), people's interest in age features and age differences can also be represented by six fuzzy sets and three fuzzy sets. Similar to the cases of Finland and Germany, a greater number of nodes are distributed around the age range of $[25-60]$ and have higher membership degrees to the corresponding fuzzy set, including the second and the third fuzzy set of age ($Age_f 2$ and $Age_f 3$, each centring around $36$ and $54$; See Fig.~\ref{DNfeat22} (d)). In contrast with the cases of Finland and Germany, people in Italy pay more attention to the age difference of $45$ and thus assign higher membership degrees to the second fuzzy set of age difference $\Delta Age_f 2$ (See Fig.~\ref{fuzzyIselectparam} (e)). In Fig.~\ref{ItalyFeatPref} (c) and (d), people in Italy have a stronger positive preference, characterised by a preference weight around $0.50$ for males, age values around $20$ ($Age_f 2$, the second fuzzy set of age) and age differences around $0$ ($\Delta Age_f 1$, the first fuzzy set of age difference). In addition, they also have a negative preference for age difference around $90$ ($\Delta Age_f 3$, the third fuzzy set of age difference) with a preference weight at $0.57$, which, however, have limited impact on the network formation because smaller number of node pairs have higher membership degrees to the third fuzzy set of age difference. People in Italy show negative preferences for age values around $36$ and $72$ with slightly weaker preference weights around $0.20$, which, multiplied with people's interests and feature distributions for the third and fifth fuzzy set of age ($Age_f 3$ and $Age_f 5$), can also have a significant impact on network formation.

As shown in Fig.~\ref{LuxembourgFeatPref} (a) and (b), people's interest in age features and age differences can also be represented by six fuzzy sets and two fuzzy sets. Similar to the cases of Finland, Germany and Italy, a significant number of nodes are distributed around the age $40$ and have higher membership degrees to the corresponding fuzzy sets, including the third and fourth fuzzy set of age (each denoted by $Age_f 3$ and $Age_f 4$). In contrast with the cases of the abovementioned countries, people in Luxembourg pay more attention to the age difference around $0$ and thus assign higher membership degrees to the corresponding first fuzzy set of age difference $\Delta Age_f 1$ (See Fig.~\ref{fuzzyLselectparam} (e)). In Fig.~\ref{LuxembourgFeatPref} (c) and (d), people in Luxembourg have a strong positive preference, characterised by a preference weight over $0.70$, for males, age values around $0$ ($Age_f 1$, the first fuzzy set of age) and age values around $18$ ($Age_f 2$, the second fuzzy set of age). They have a strong negative preference around $72$ ($Age_f 5$, the fifth fuzzy set of age), with a preference weight of $0.80$. In addition, they also have a positive preference for age difference around $0$ ($\Delta Age_f 1$, the first fuzzy set of age difference), which, despite its slightly lower preference weight at $0.46$, can also have a significant impact on the network formation considering people's interests in similar ages (as shown in Fig.~\ref{LuxembourgFeatPref} (b)).

As shown in Fig.~\ref{PolandFeatPref} (a) and (b), people's interest in age features and age differences can be represented by nine fuzzy sets and four fuzzy sets. Different from the abovementioned countries, people in Poland have a more extensive interest in age values over various ranges. Based on the distribution of age features (See Fig.~\ref{DNfeat22} (f)), a significant number of nodes have higher membership degrees to the third, fourth, sixth and seventh fuzzy sets (each denoted by $Age_f 3$, $Age_f 4$, $Age_f 6$ and $Age_f 7$, centring around $22$, $33$, $56$ and $67$). Similar to the cases of Belgium, Finland and Germany, a significant number of people have an age difference of around $26$ and $60$, characterised by higher membership degrees to the second and the third fuzzy set of the age difference (denoted by $\Delta Age_f 2$ and $\Delta Age_f 3$; See Fig.~\ref{fuzzyPselectparam} (f)). In Fig.~\ref{PolandFeatPref} (c) and (d), people in Poland have a stronger positive preference, characterised by a preference weight over $0.75$, for males, age values around $0$ ($Age_f 1$, the first fuzzy set of age) and age differences around $0$ ($\Delta Age_f 1$, the first fuzzy set of age difference). They also have a significant negative preference, with a preference weight around $0.50$, for age around $45$, ($Age_f 5$, the fifth fuzzy set of age) and age difference around $90$ ($ \Delta Age_4 6$, the sixth fuzzy set of age difference). In addition, people in Poland have a slightly positive preference for age difference around $30$ ($\Delta Age_f 2$, the second fuzzy set of age difference) and similar sex features. These two preferences can also have a significant impact on the network formation, considering people's great interests in the similar ages and the specific differences between males and females (as shown in Fig.~\ref{PolandFeatPref} (b)).

To conclude, people in each country have different interests and preferences in social contact. They generally have great interests or strong preferences in the sex features, age differences around $0$, $30$ and $60$. The variety of heterogeneous features and interaction rules add to the complexity of network formation but gives us a clue to decipher the complex social network patterns. 

\subsubsection{Social Contact Matrices}
\label{2mat}
\begin{figure}[H] 
	\centering
	\subfigure[Target]{
		\begin{minipage}[b]{0.32\linewidth}
			\includegraphics[width=1\linewidth]{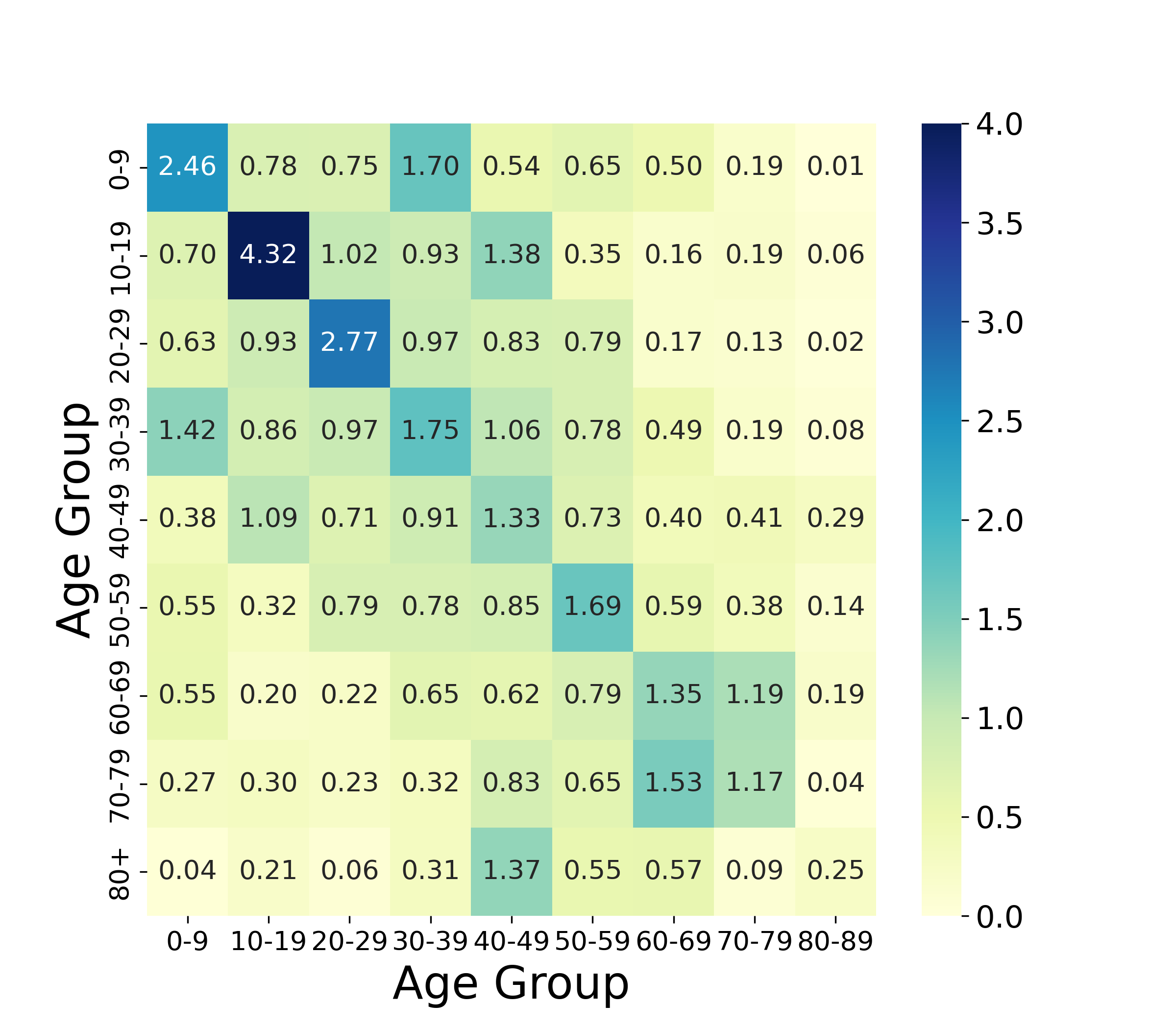}
	\end{minipage}}
 \hspace{-5mm}
		\subfigure[Simulation]{
		\begin{minipage}[b]{0.32\linewidth}
			\includegraphics[width=1\linewidth]{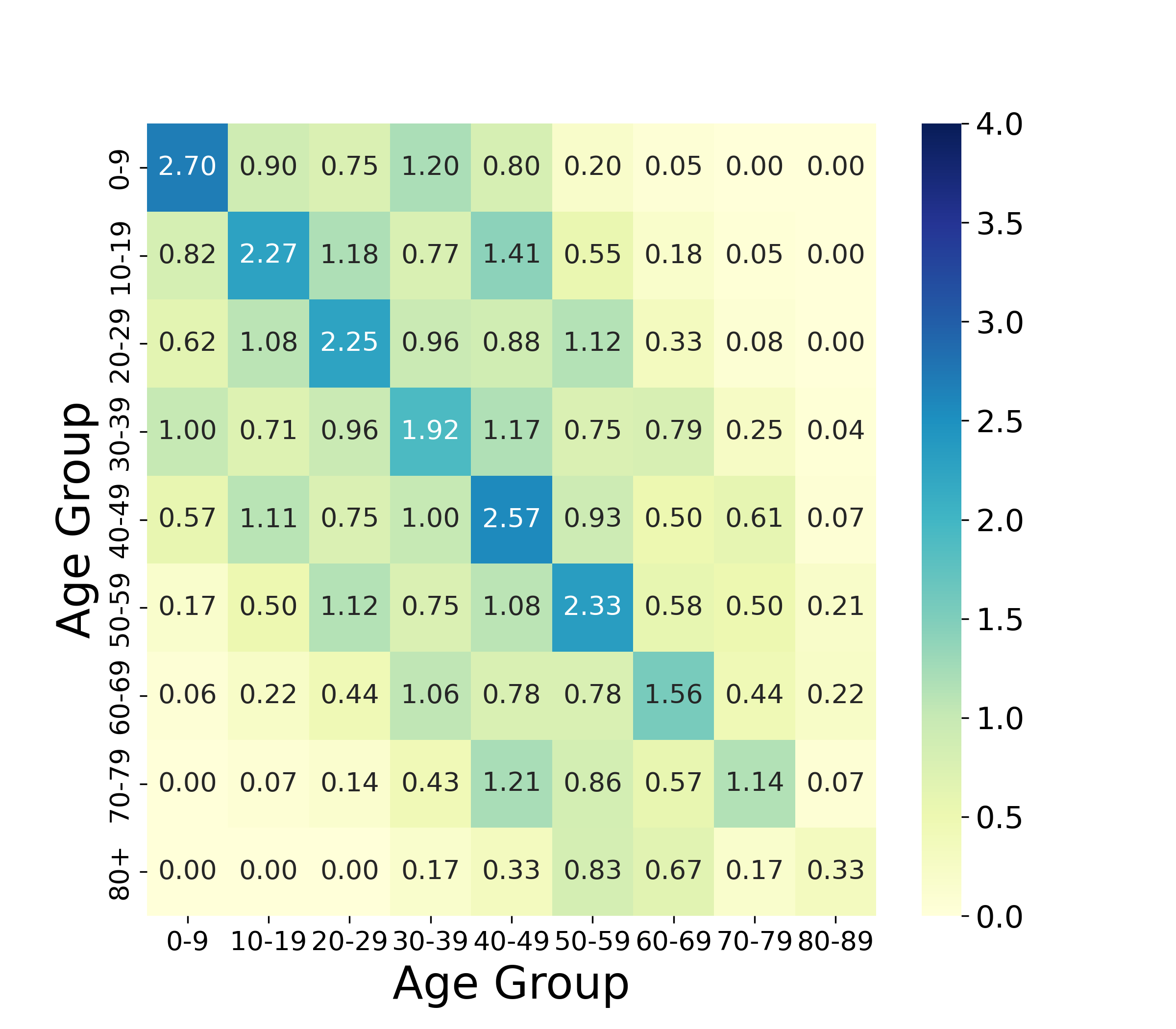}
	\end{minipage}}\\
	\caption{The target social contact matrix ((a) target) and the recreated social contact matrices of Belgium by (b) the $HN-A_f-S$ built on calibrated fuzzy sets.}
 \label{BelgiumExtension}
\end{figure}

\begin{figure}[H]
	\centering
	\subfigure[Target]{
		\begin{minipage}[b]{0.32\linewidth}
			\includegraphics[width=1\linewidth]{FinlandTarget.png}
	\end{minipage}}
 \hspace{-5mm}
		\subfigure[Simulation]{
		\begin{minipage}[b]{0.32\linewidth}
			\includegraphics[width=1\linewidth]{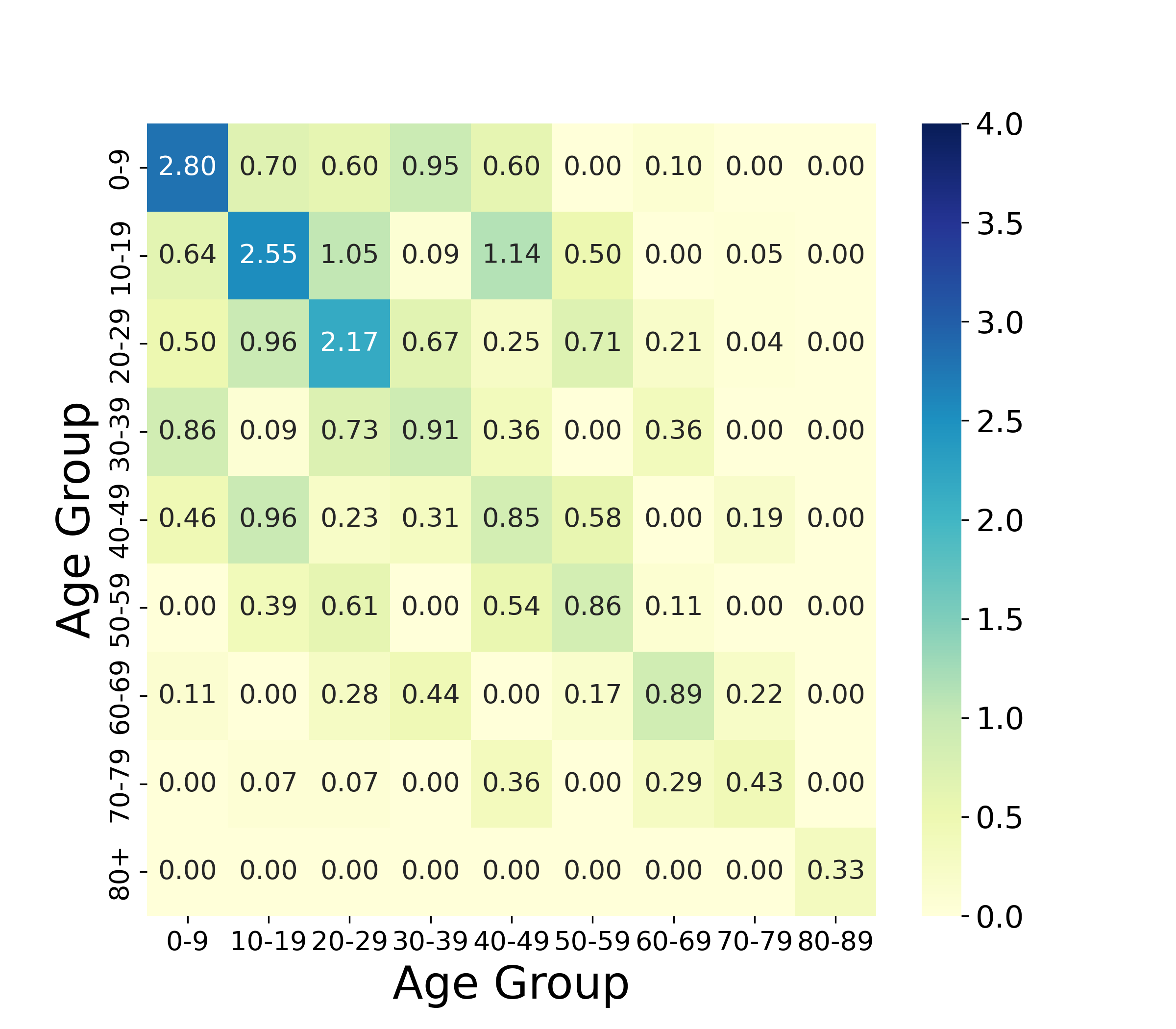}
	\end{minipage}}\\
	\caption{The target social contact matrix ((a) target) and the recreated social contact matrices of Finland by (b) the $HN-A_f-S$ built on calibrated fuzzy sets.}
 \label{FinlandExtension}
\end{figure}

\begin{figure}[H]
	\centering
	\subfigure[Target]{
		\begin{minipage}[b]{0.32\linewidth}
			\includegraphics[width=1\linewidth]{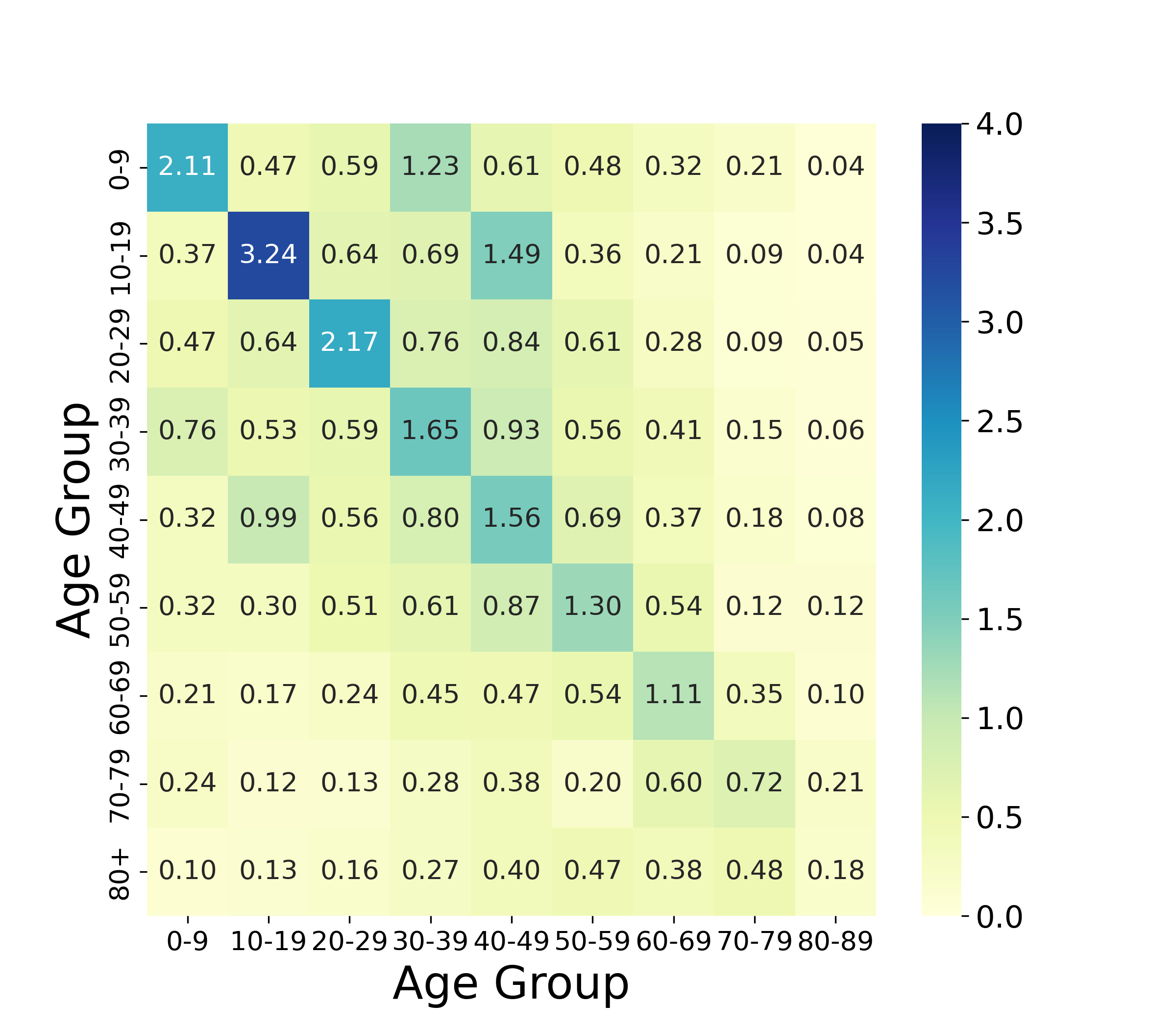}
	\end{minipage}}
 \hspace{-5mm}
		\subfigure[Simulation]{
		\begin{minipage}[b]{0.32\linewidth}
			\includegraphics[width=1\linewidth]{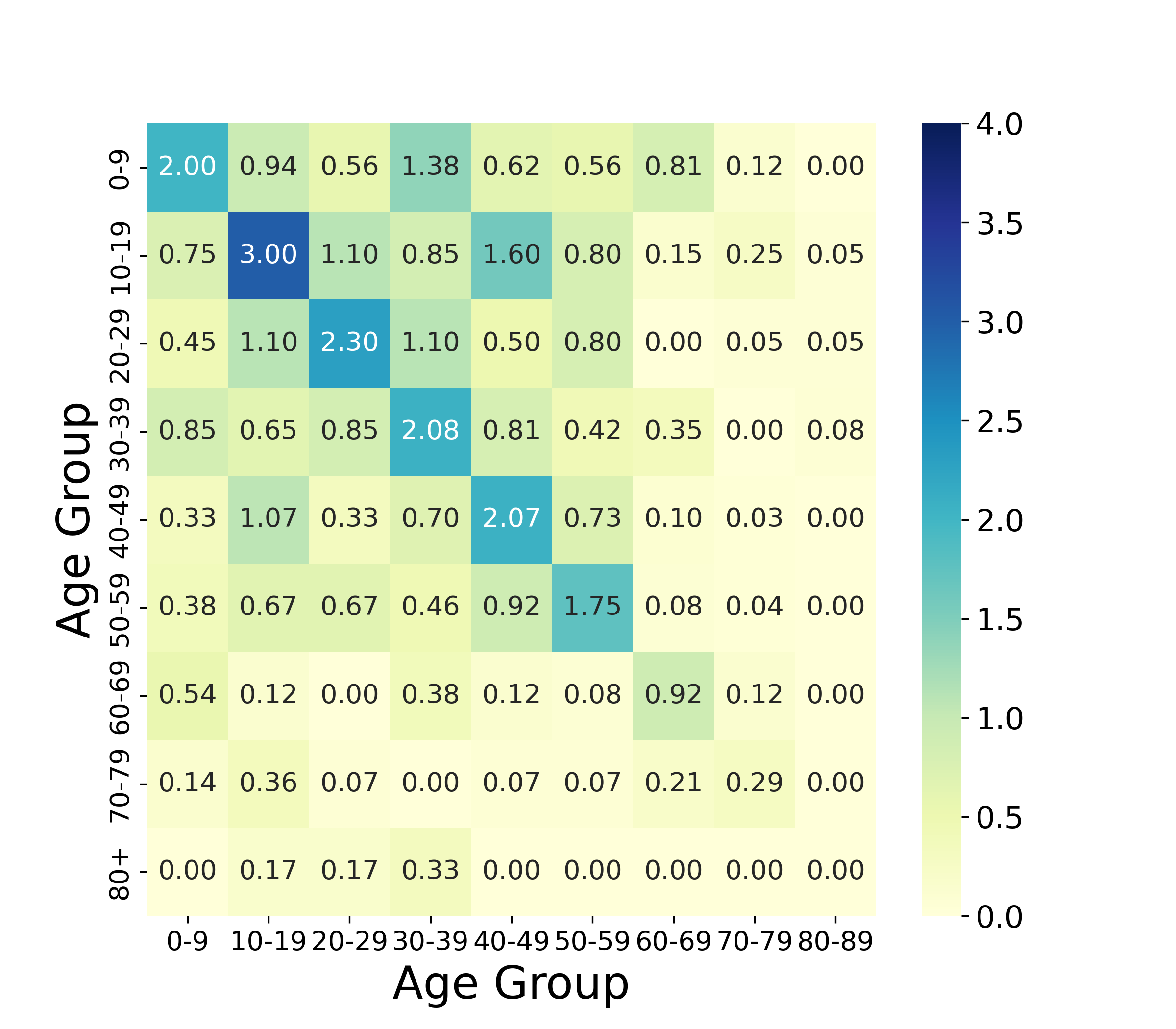}
	\end{minipage}}\\
	\caption{The target social contact matrix ((a) target) and the recreated social contact matrices of Germany by (b) the $HN-A_f-S$ built on calibrated fuzzy sets.}
 \label{GermanyExtension}
\end{figure}

\begin{figure}[H]
	\centering
	\subfigure[Target]{
		\begin{minipage}[b]{0.32\linewidth}
			\includegraphics[width=1\linewidth]{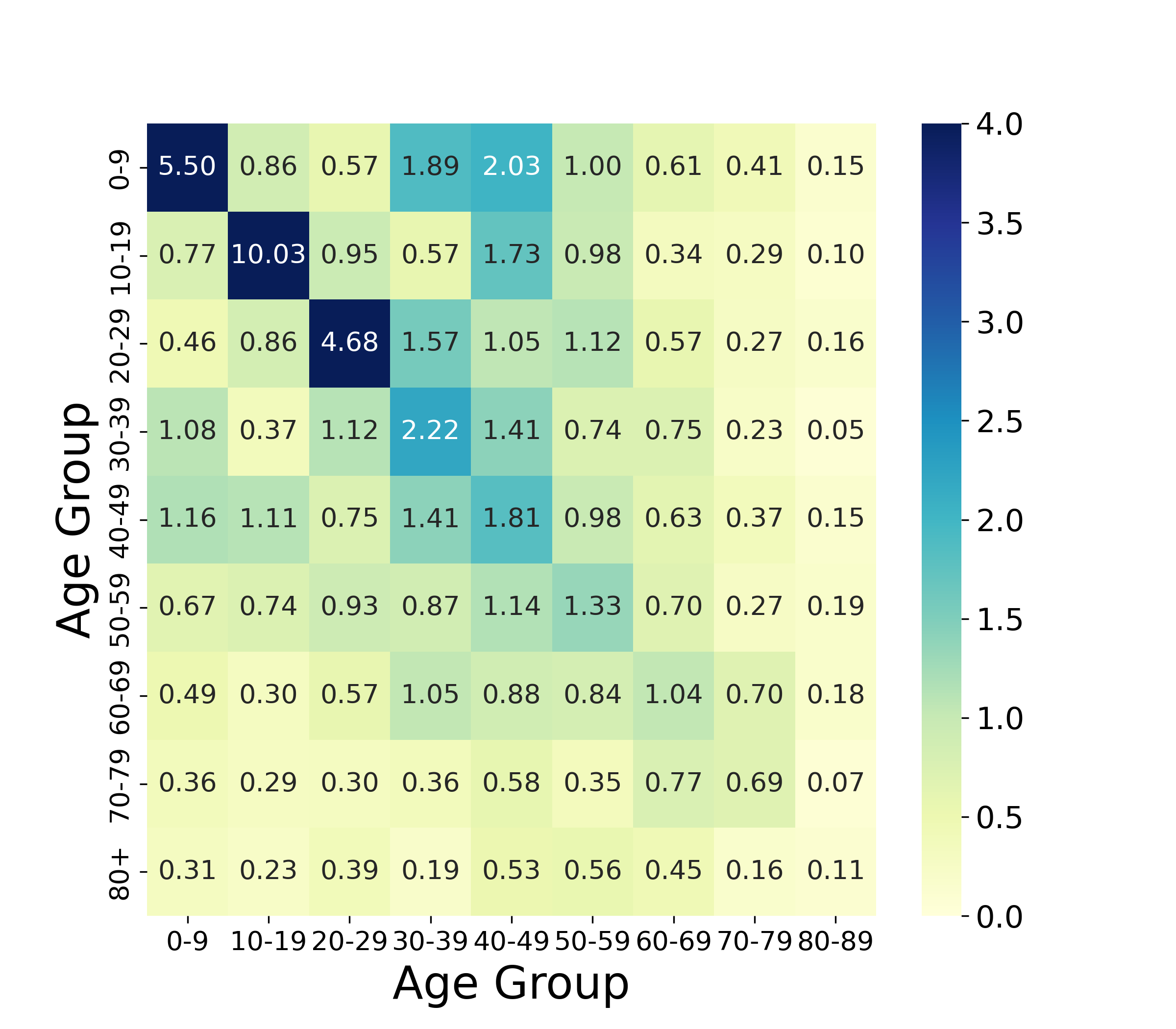}
	\end{minipage}}
 \hspace{-5mm}
		\subfigure[Simulation]{
		\begin{minipage}[b]{0.32\linewidth}
			\includegraphics[width=1\linewidth]{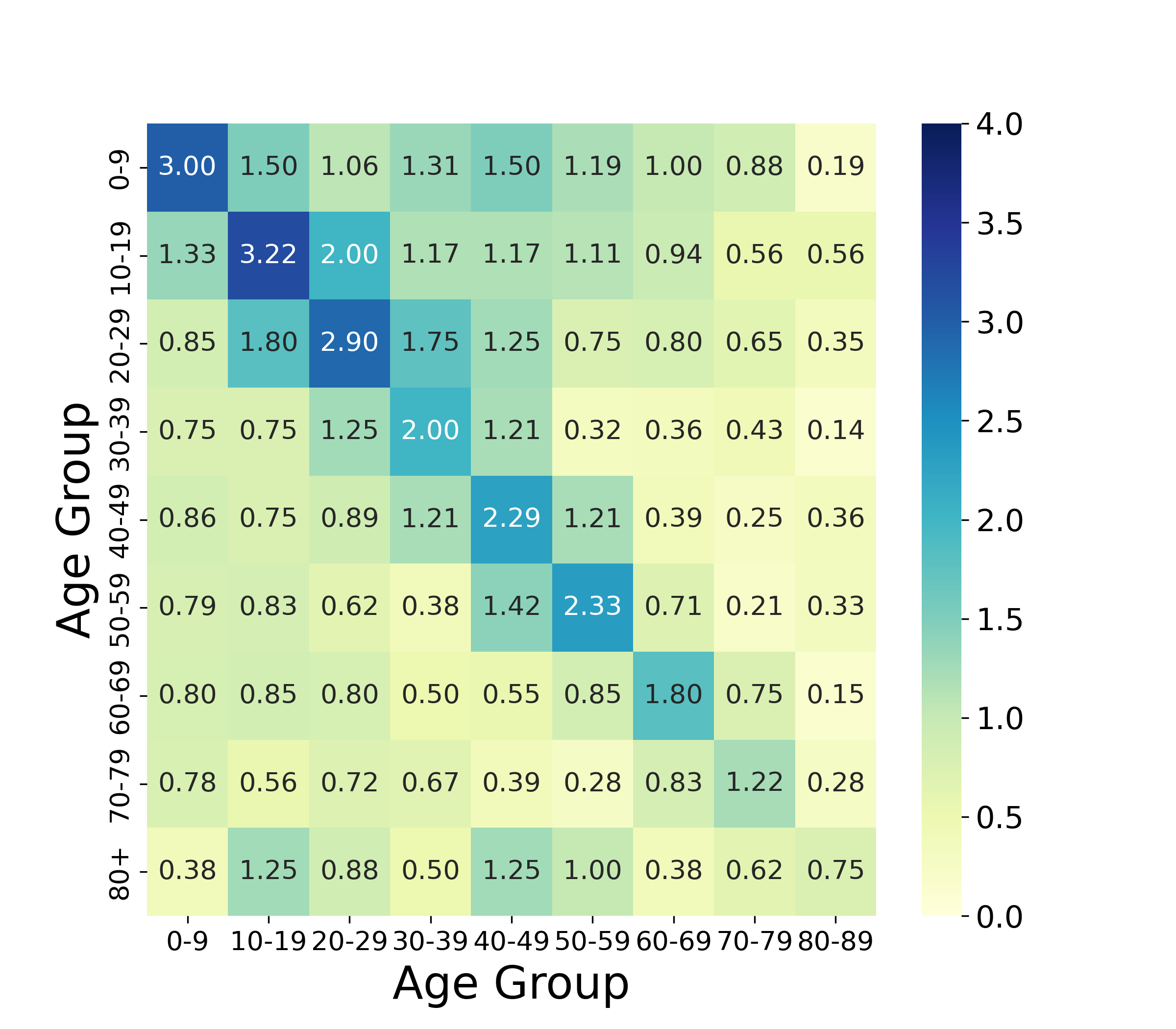}
	\end{minipage}}\\
	\caption{The target social contact matrix ((a) target) and the recreated social contact matrices of Italy by (b) the $HN-A_f-S$ built on calibrated fuzzy sets.}
 \label{ItalyExtension}
\end{figure}

\begin{figure}[H]
	\centering
	\subfigure[Target]{
		\begin{minipage}[b]{0.32\linewidth}
			\includegraphics[width=1\linewidth]{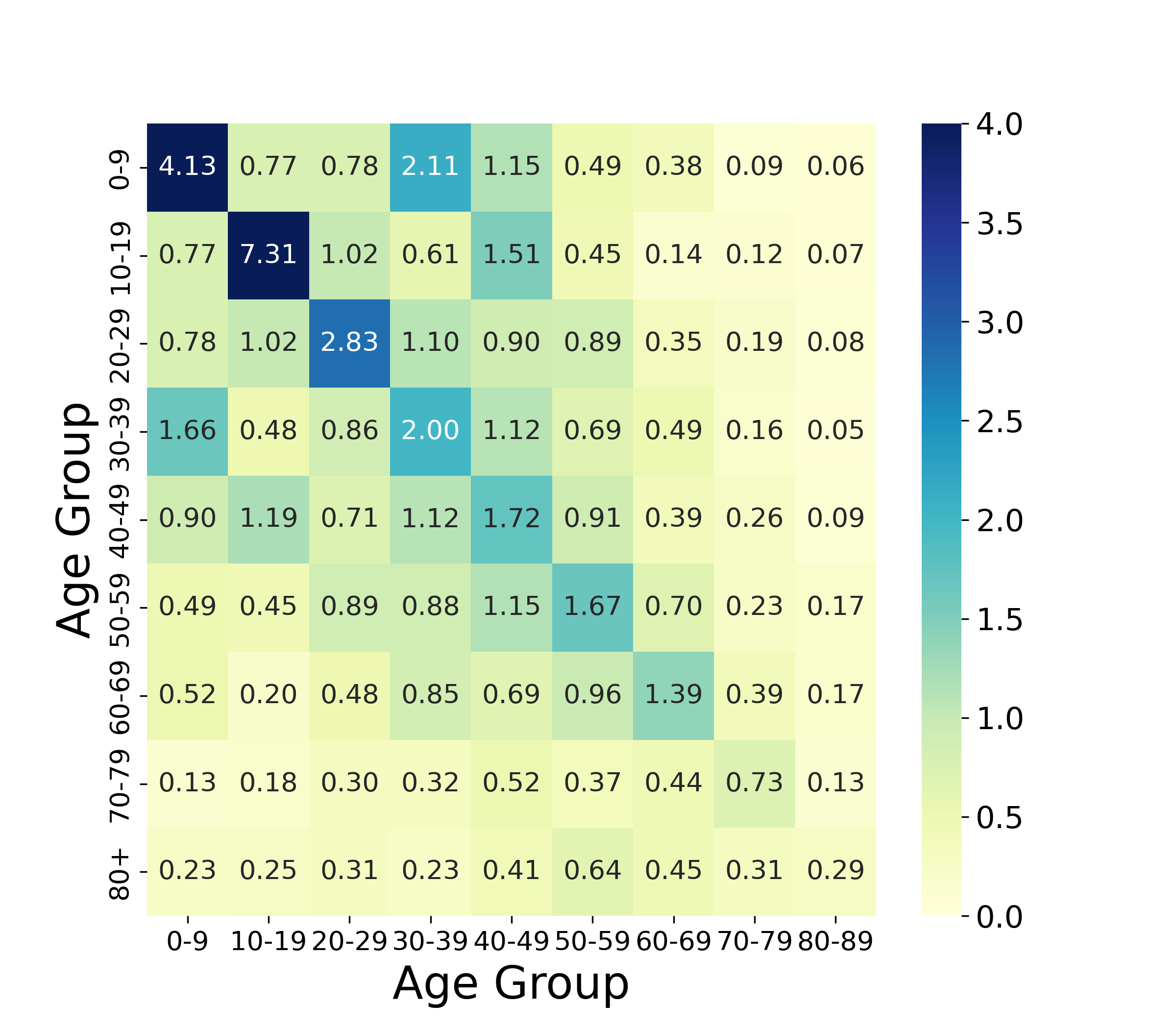}
	\end{minipage}}
 \hspace{-5mm}
		\subfigure[Simulation]{
		\begin{minipage}[b]{0.32\linewidth}
			\includegraphics[width=1\linewidth]{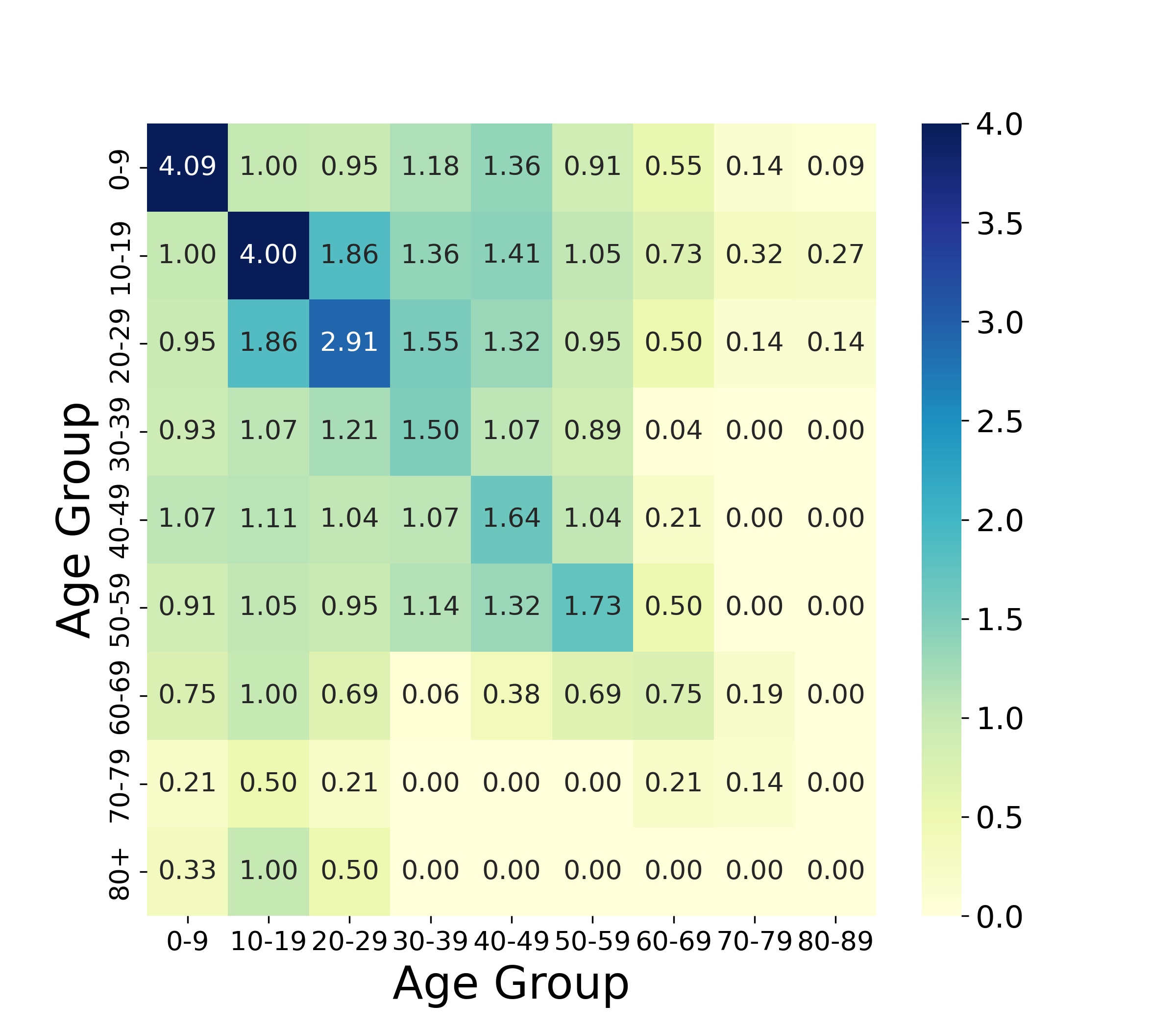}
	\end{minipage}}\\
	\caption{The target social contact matrix ((a) target) and the recreated social contact matrices of Luxembourg by (b) the $HN-A_f-S$ built on calibrated fuzzy sets.}
 \label{LuxembourgExtension}
\end{figure}

\begin{figure}[H]
	\centering
	\subfigure[Target]{
		\begin{minipage}[b]{0.32\linewidth}
			\includegraphics[width=1\linewidth]{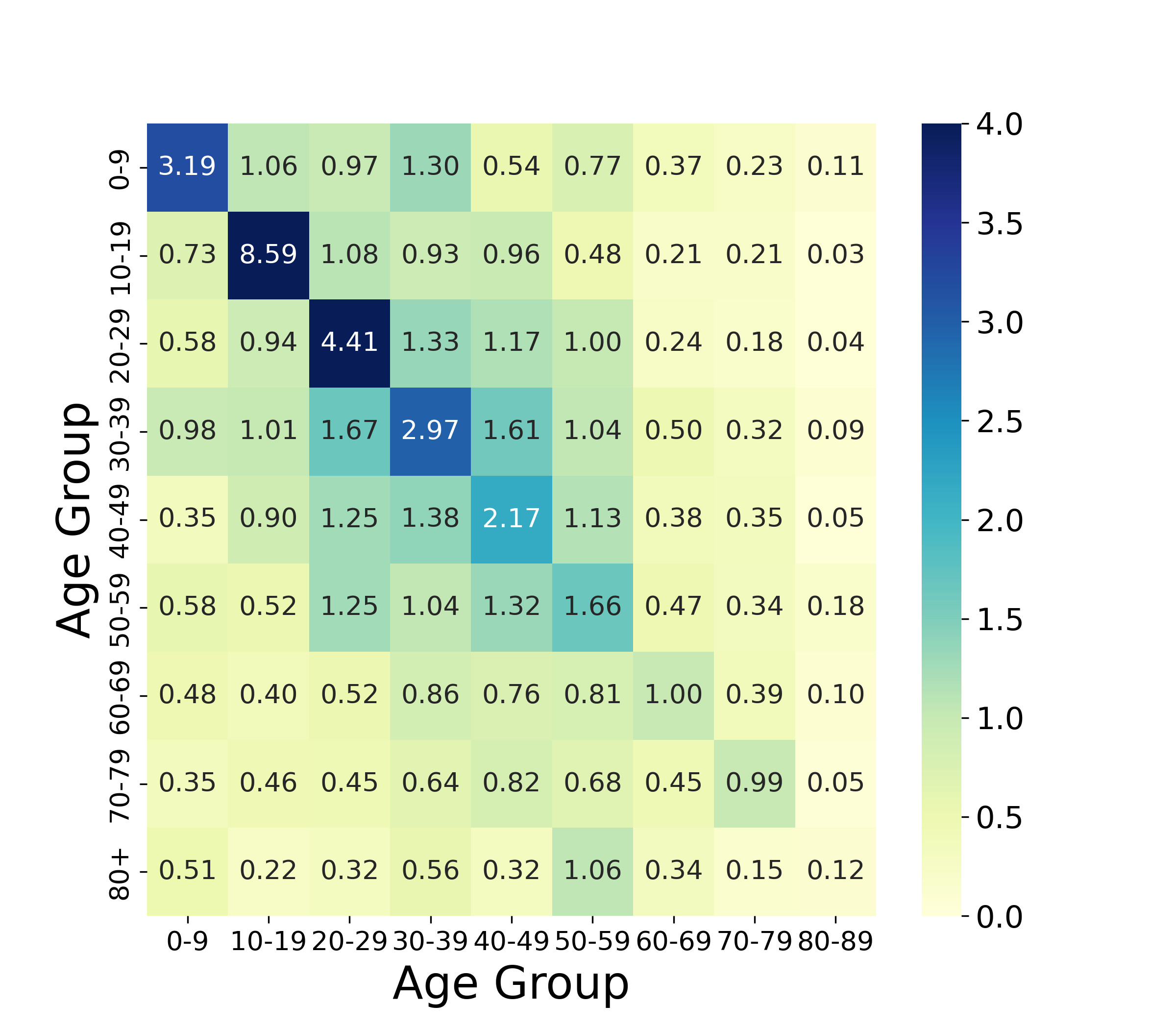}
	\end{minipage}}
 \hspace{-5mm}
		\subfigure[Simulation]{
		\begin{minipage}[b]{0.32\linewidth}
			\includegraphics[width=1\linewidth]{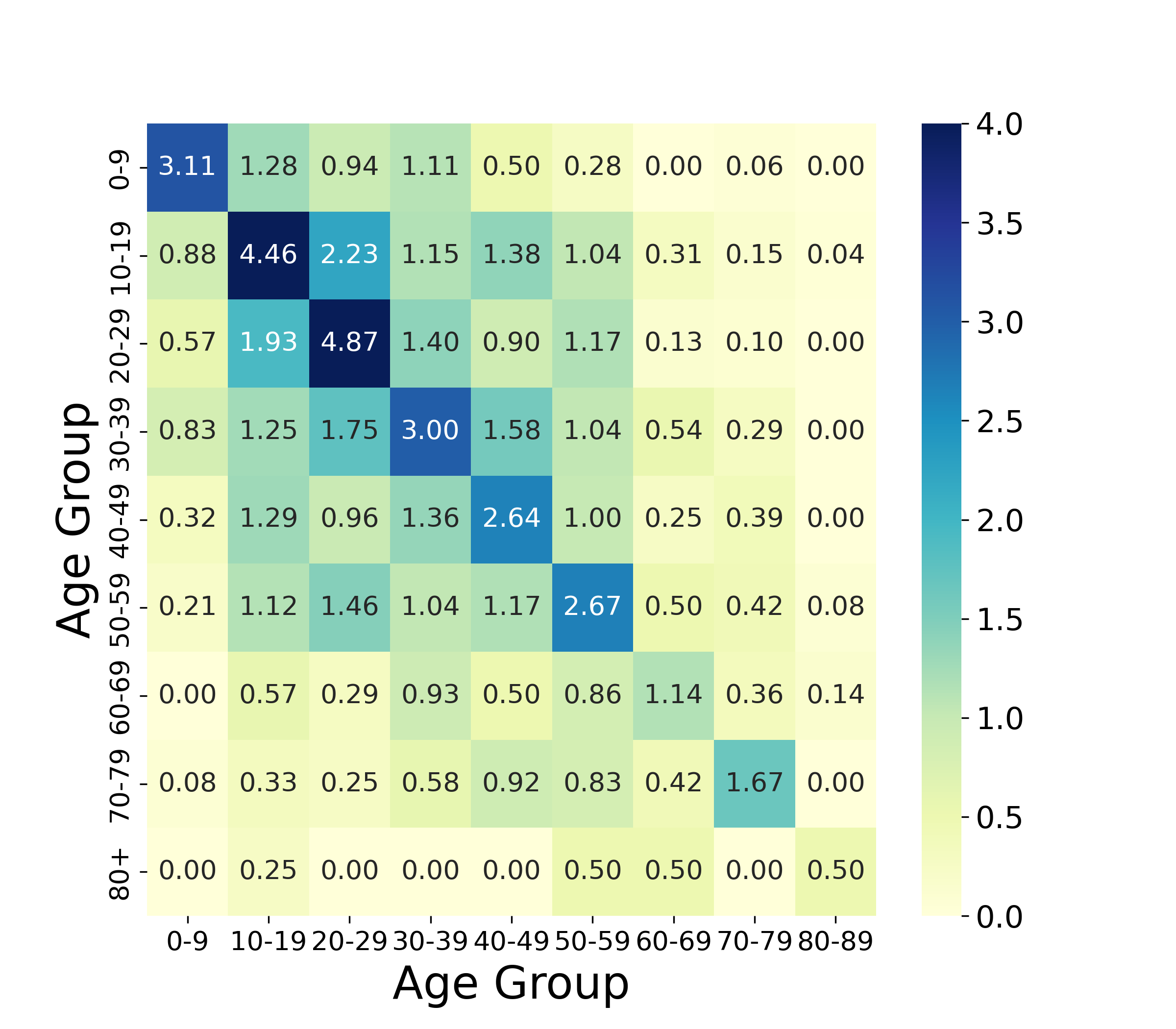}
	\end{minipage}}\\
	\caption{The target social contact matrix ((a) target) and the recreated social contact matrices of Poland by (b) the $HN-A_f-S$ built on calibrated fuzzy sets.}
 \label{PolandExtension}
\end{figure}

Based on the features and preferences presented in Section \ref{1feature}, we can generate social network simulations and calculate the corresponding social contact matrices based on the age features for Belgium (See Fig.~\ref{BelgiumExtension}), Finland (See Fig.~\ref{FinlandExtension}, Germany (See  Fig.~\ref{GermanyExtension}), Italy (See Fig.~\ref{ItalyExtension}), Luxembourg (See Fig.~\ref{LuxembourgExtension}) and Poland (See Fig.~\ref{PolandExtension}). The social contact matrices generally capture the characteristics of the real ones, including people's connections with similar ages and people with an age difference of around $30$. This results from the great interest and strong preference for the corresponding fuzzy representation of age differences (e.g. See $\Delta Age_f 1$ for the first fuzzy set of age difference around $0$ and $\Delta Age_f 2$ for the second fuzzy set of age difference around $30$ in Belgium (Fig.~\ref{BelgiumFeatPref}), Finland (Fig.~\ref{FinlandFeatPref}), Germany (Fig.~\ref{BelgiumFeatPref}) and Poland (Fig.~\ref{BelgiumFeatPref})). Among these countries, Finland has less average social contacts between different ages than the other countries. In addition, the average social contact number approaches zero for people over $80$ in Finland, which leads to lower average node degrees and clustering coefficient and a greater number of non-existing paths between nodes. In the social contact matrices of the simulated social networks, there is no social contact between people over $80$ and people younger than $80$, even though there is a smaller number of social contacts in the real contact matrix. This is because people have a strong negative preference for older nodes around the age of $55$, $72$ and $90$ (each related to the fourth, fifth and sixth fuzzy set of age differences; See Fig.~\ref{FinlandFeatPref} (c)). However, this is unrealistic and more detailed set-ups related to social contacts are required to model the limited number of connections developed by specific node pairs.

\subsubsection{Degree Distribution}
\label{3deg}
\begin{figure}[H]
	\centering
	\subfigure[Males\&Females]{
		\begin{minipage}[b]{0.32\linewidth}
			\includegraphics[width=1\linewidth]{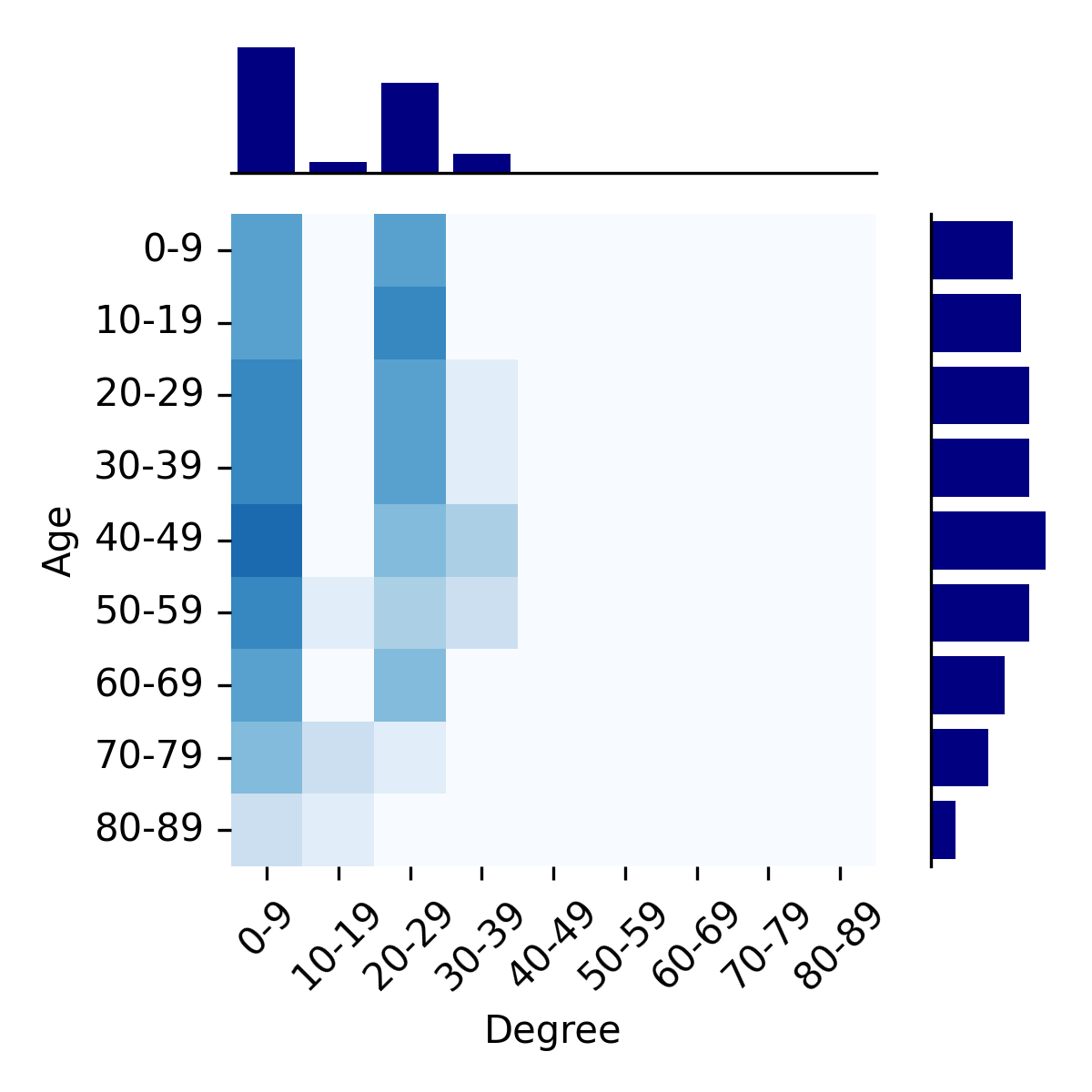}
	\end{minipage}}
	\subfigure[Males]{
		\begin{minipage}[b]{0.32\linewidth}
			\includegraphics[width=1\linewidth]{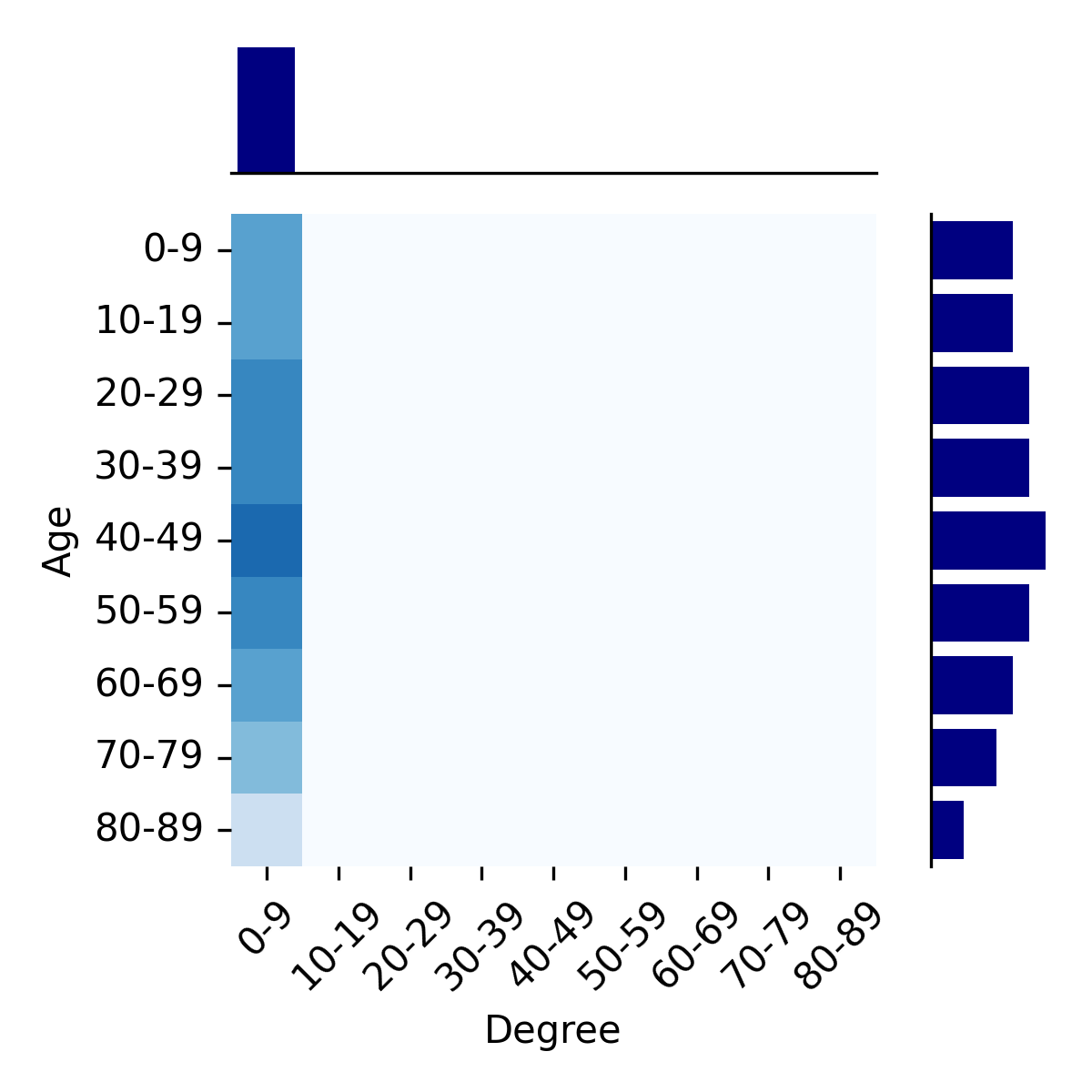}
	\end{minipage}}
	\subfigure[Females]{
		\begin{minipage}[b]{0.32\linewidth}
			\includegraphics[width=1\linewidth]{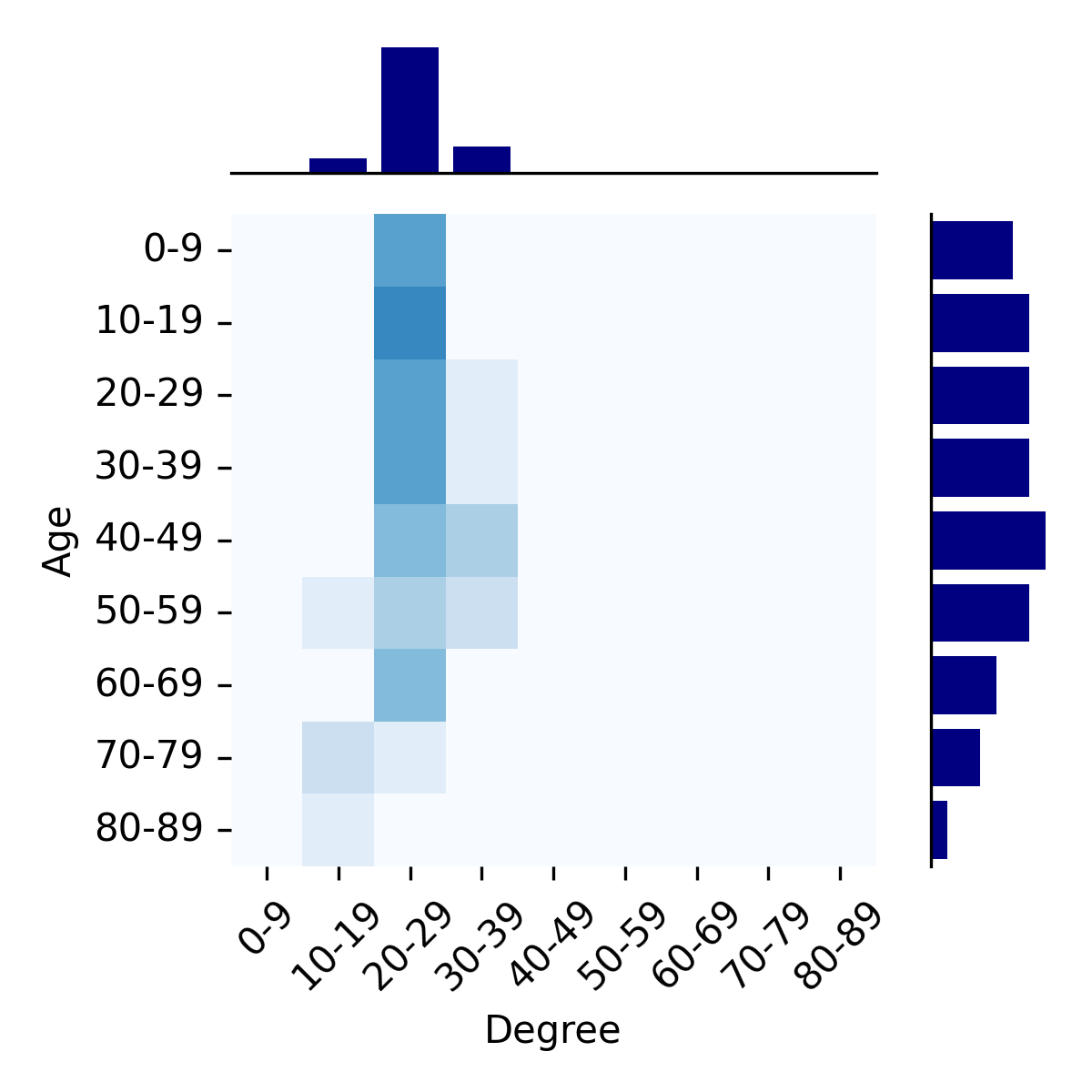}
	\end{minipage}}\\
	\caption{The age and degree distributions in Belgium for (a) males and females, (b) males and (c) females.}
 \label{BelgiumDegree}
\end{figure}

\begin{figure}[H]
	\centering
	\subfigure[Males\&Females]{
		\begin{minipage}[b]{0.32\linewidth}
			\includegraphics[width=1\linewidth]{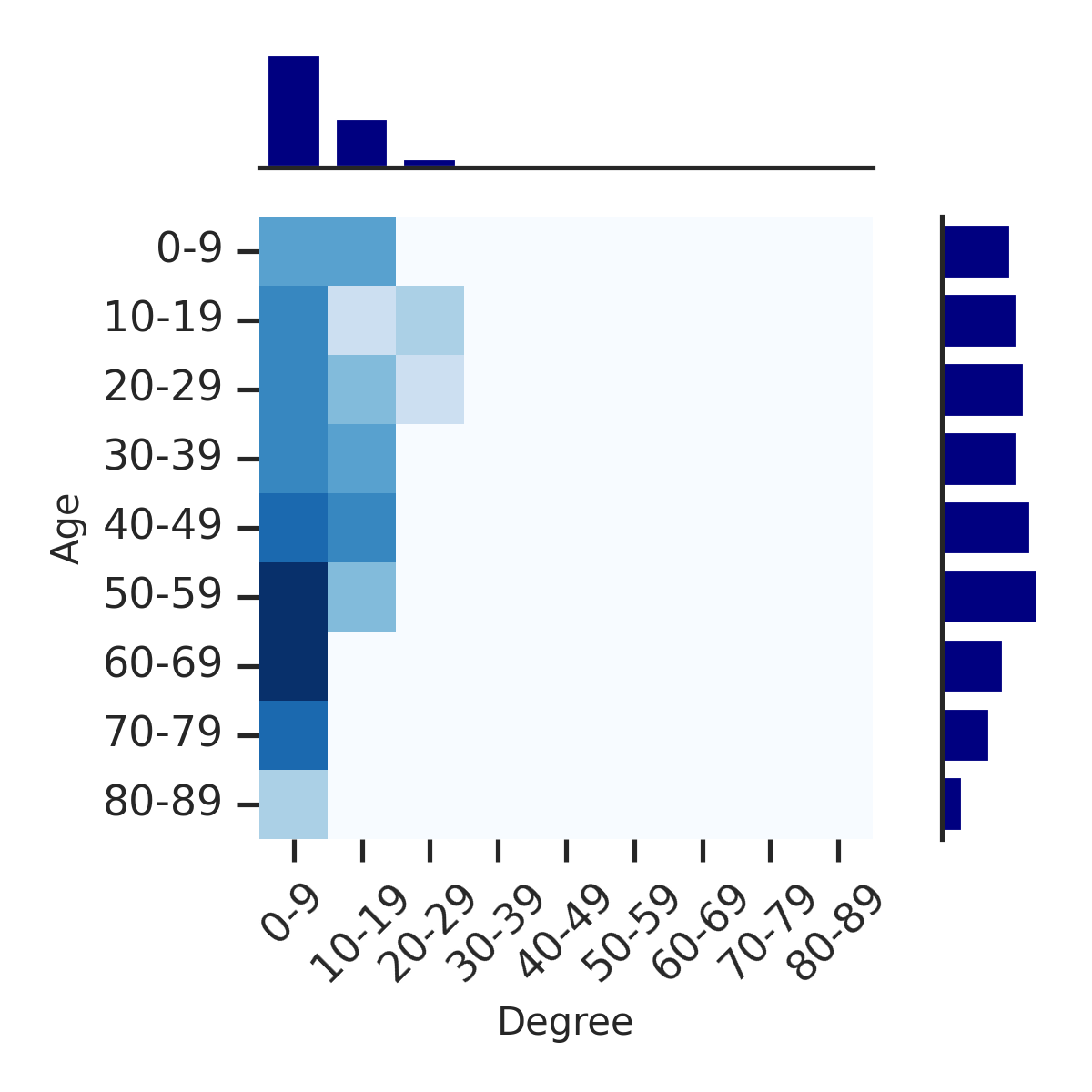}
	\end{minipage}}
	\subfigure[Males]{
		\begin{minipage}[b]{0.32\linewidth}
			\includegraphics[width=1\linewidth]{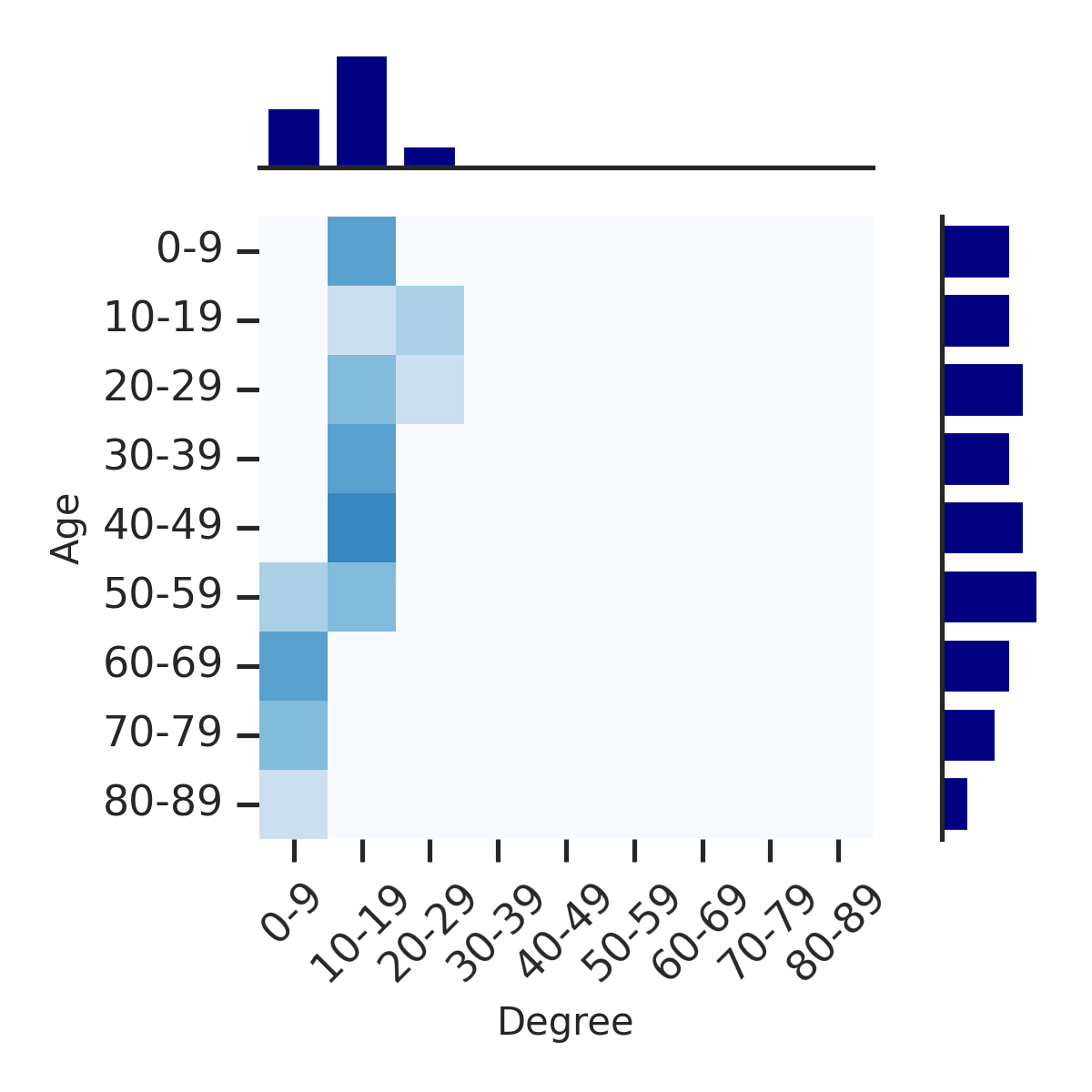}
	\end{minipage}}
	\subfigure[Females]{
		\begin{minipage}[b]{0.32\linewidth}
			\includegraphics[width=1\linewidth]{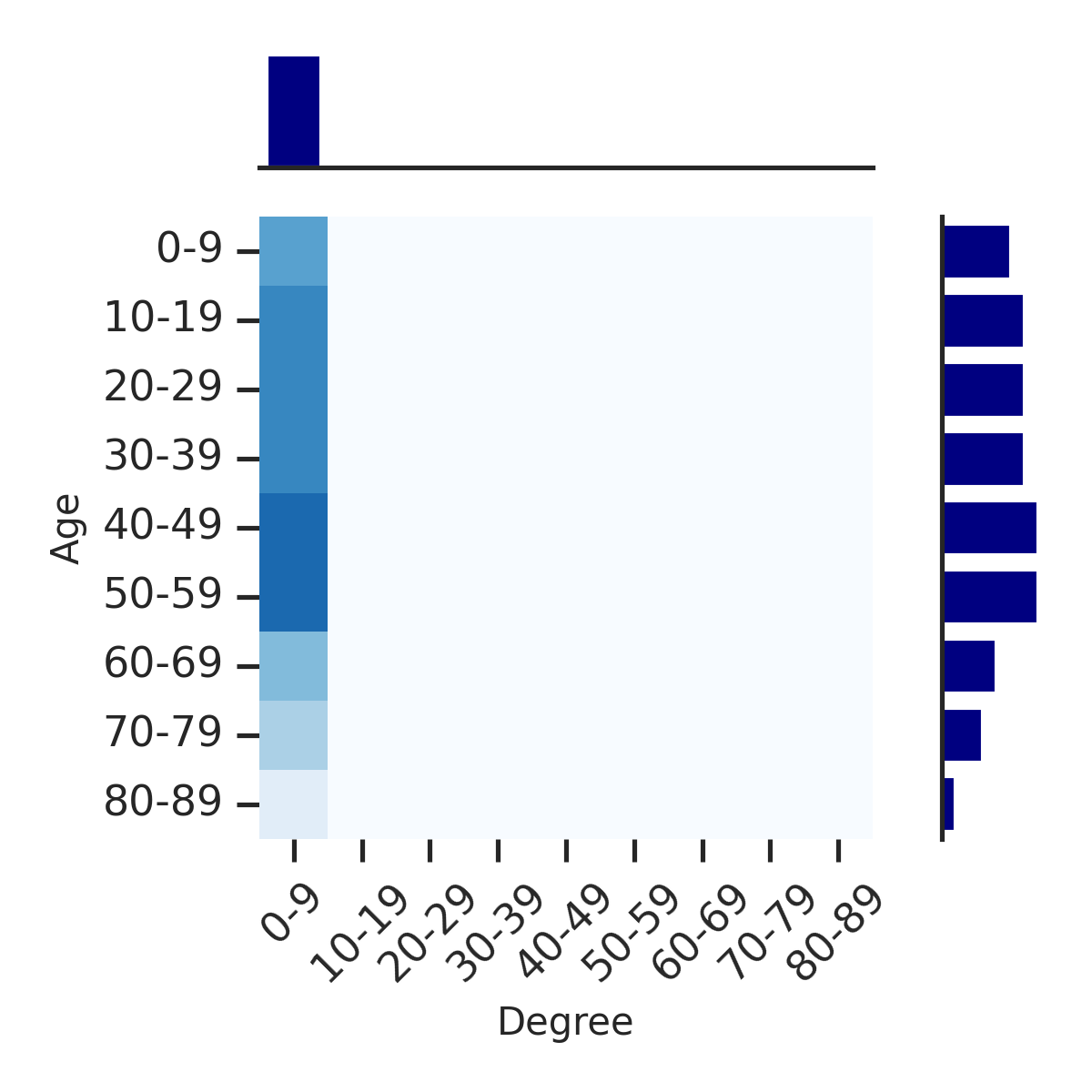}
	\end{minipage}}\\
	\caption{The age and degree distributions in Finland for (a) males and females, (b) males and (c) females.}
 \label{FinlandDegree}
\end{figure}

\begin{figure}[H]
	\centering
	\subfigure[Males\&Females]{
		\begin{minipage}[b]{0.32\linewidth}
			\includegraphics[width=1\linewidth]{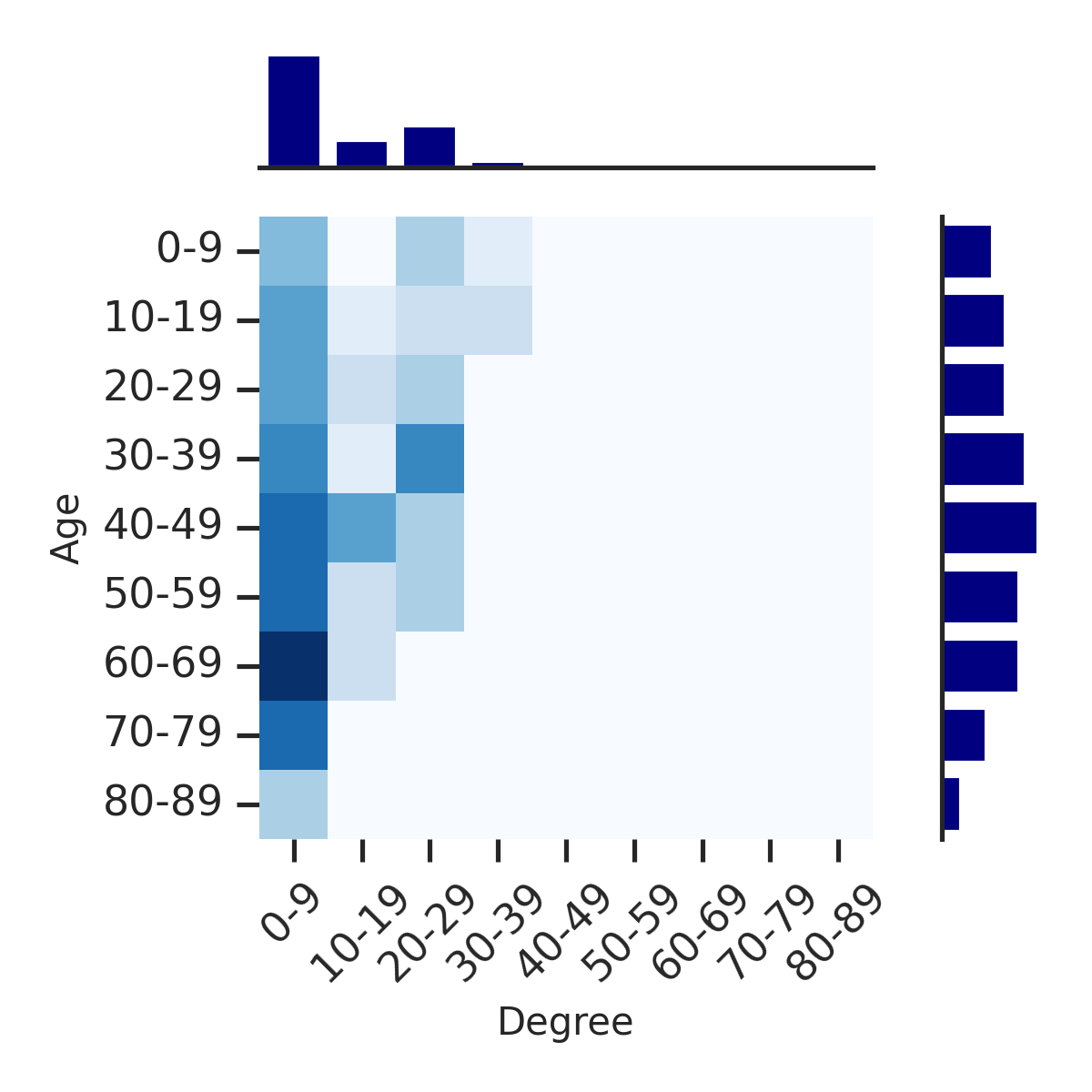}
	\end{minipage}}
	\subfigure[Males]{
		\begin{minipage}[b]{0.32\linewidth}
			\includegraphics[width=1\linewidth]{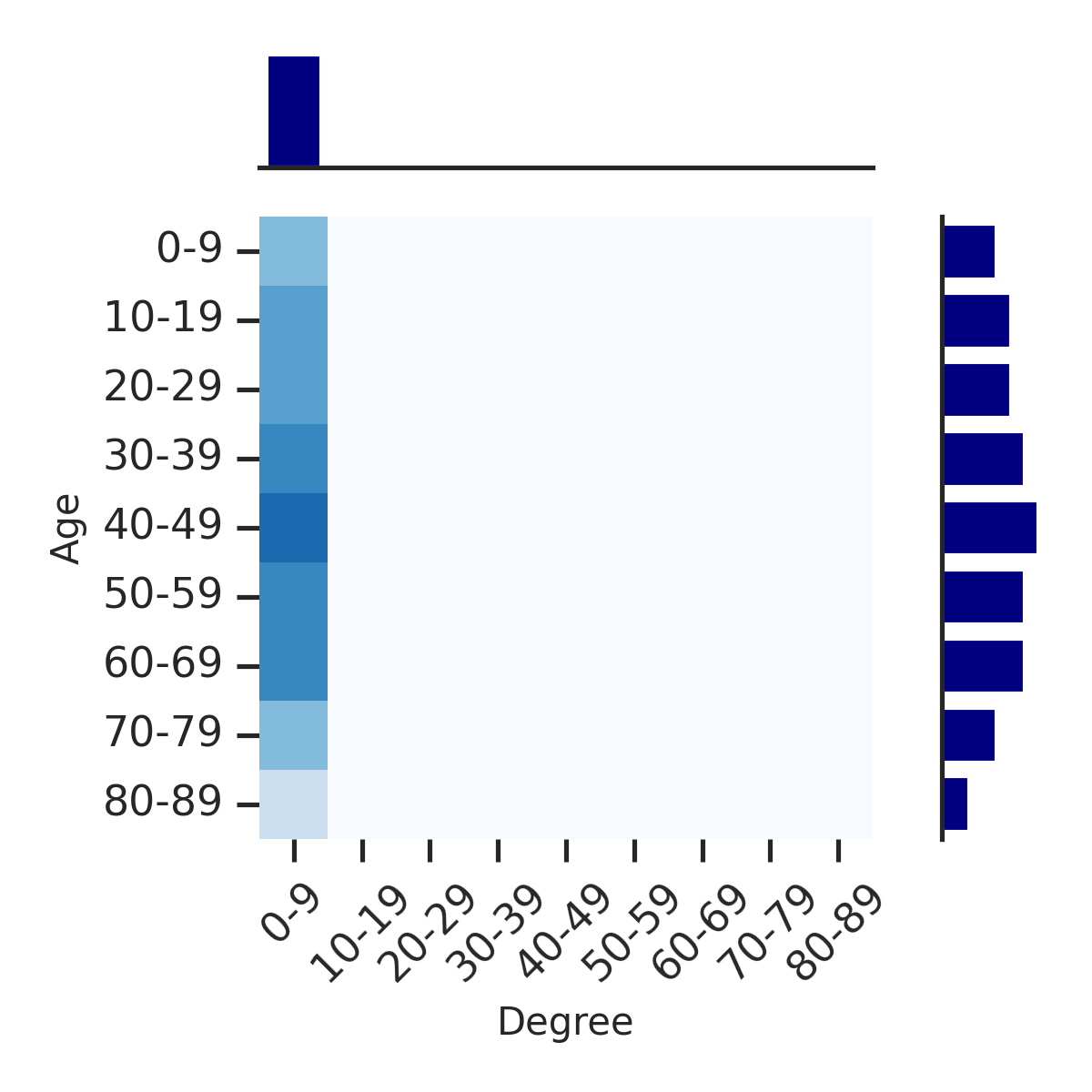}
	\end{minipage}}
	\subfigure[Females]{
		\begin{minipage}[b]{0.32\linewidth}
			\includegraphics[width=1\linewidth]{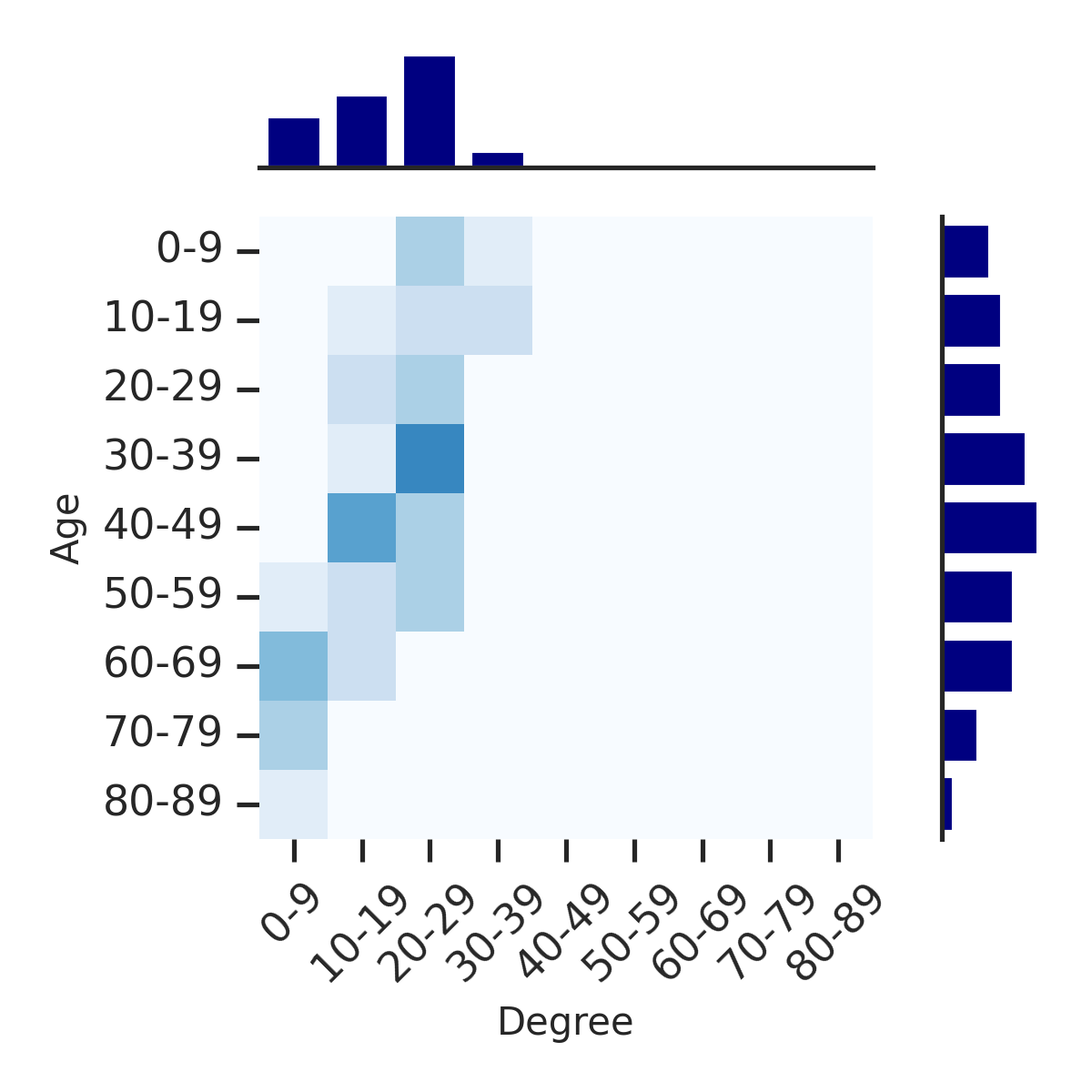}
	\end{minipage}}\\
	\caption{The age and degree distributions in Germany for (a) males and females, (b) males and (c) females.}
 \label{GermanyDegree}
\end{figure}

\begin{figure}[H]
	\centering
	\subfigure[Males\&Females]{
		\begin{minipage}[b]{0.32\linewidth}
			\includegraphics[width=1\linewidth]{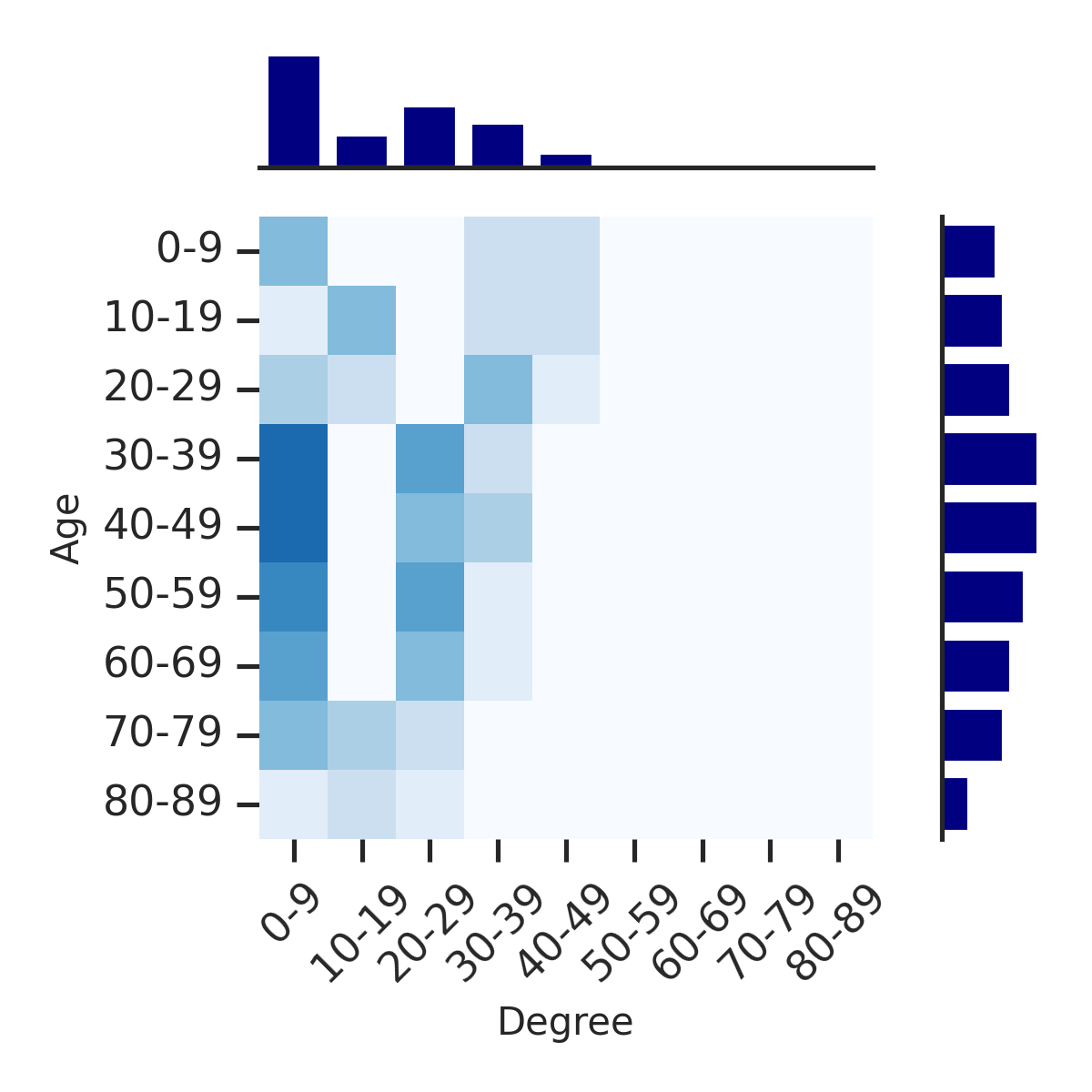}
	\end{minipage}}
	\subfigure[Males]{
		\begin{minipage}[b]{0.32\linewidth}
			\includegraphics[width=1\linewidth]{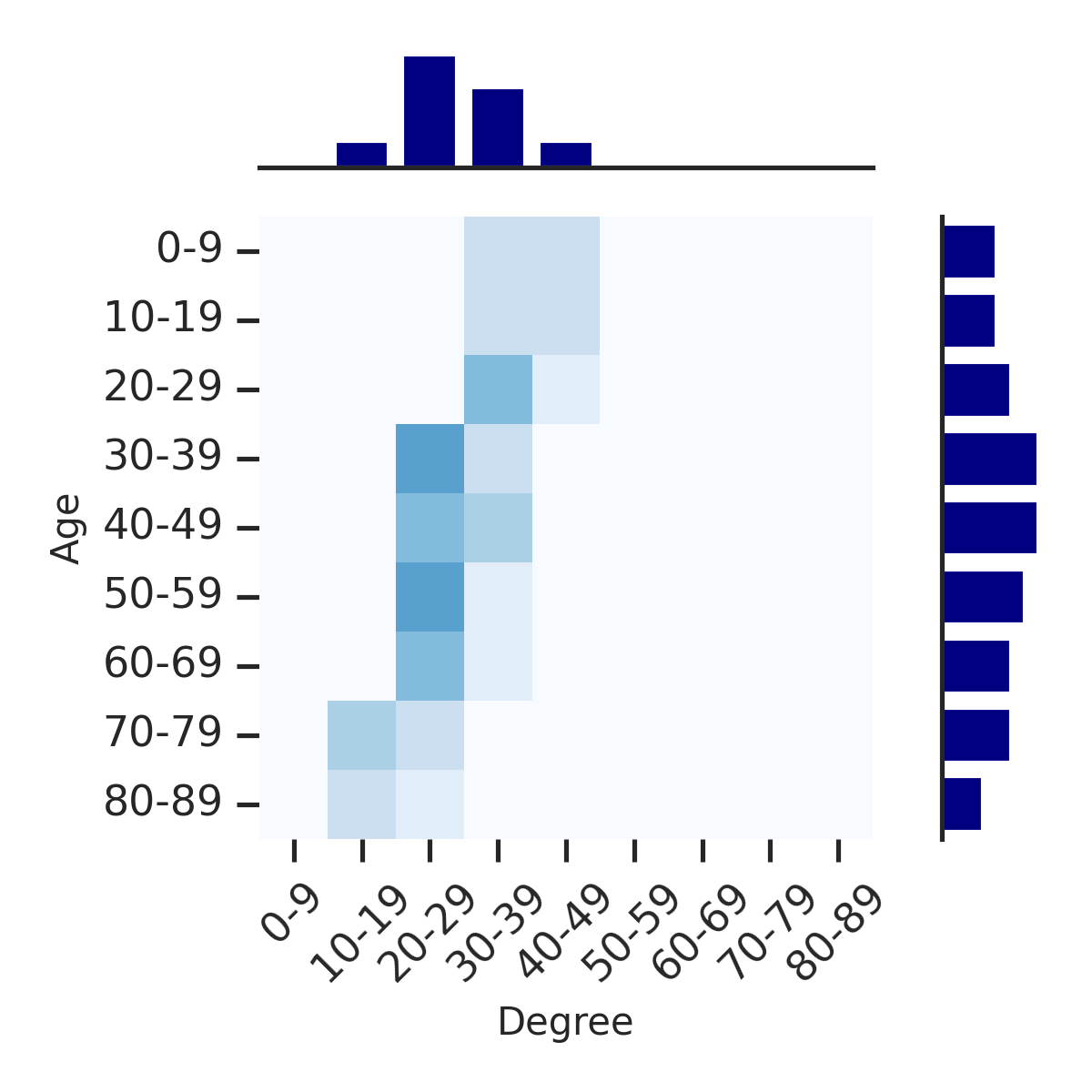}
	\end{minipage}}
	\subfigure[Females]{
		\begin{minipage}[b]{0.32\linewidth}
			\includegraphics[width=1\linewidth]{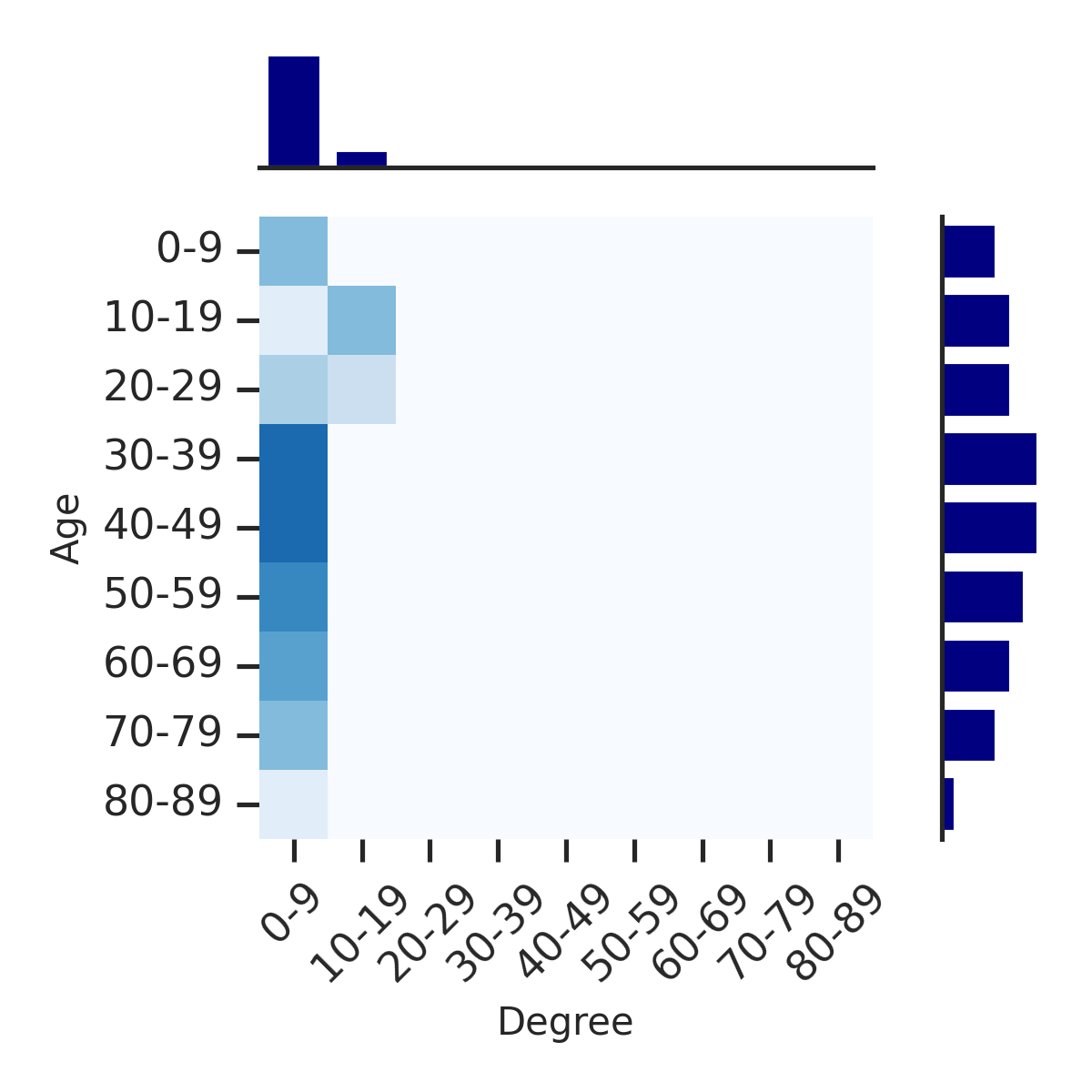}
	\end{minipage}}\\
	\caption{The age and degree distributions in Italy for (a) males and females, (b) males and (c) females.}
 \label{ItalyDegree}
\end{figure}

\begin{figure}[H]
	\centering
	\subfigure[Males\&Females]{
		\begin{minipage}[b]{0.32\linewidth}
			\includegraphics[width=1\linewidth]{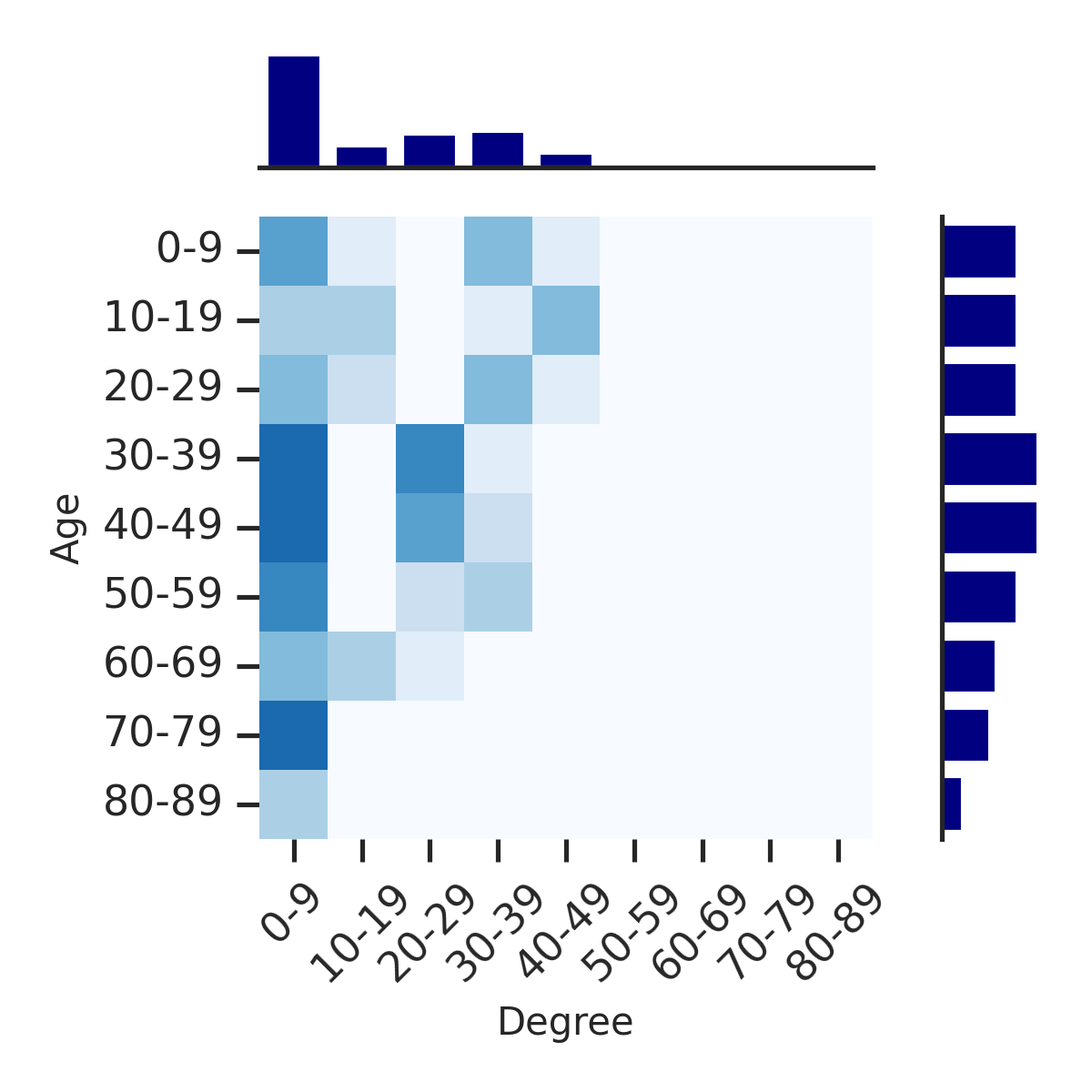}
	\end{minipage}}
	\subfigure[Males]{
		\begin{minipage}[b]{0.32\linewidth}
			\includegraphics[width=1\linewidth]{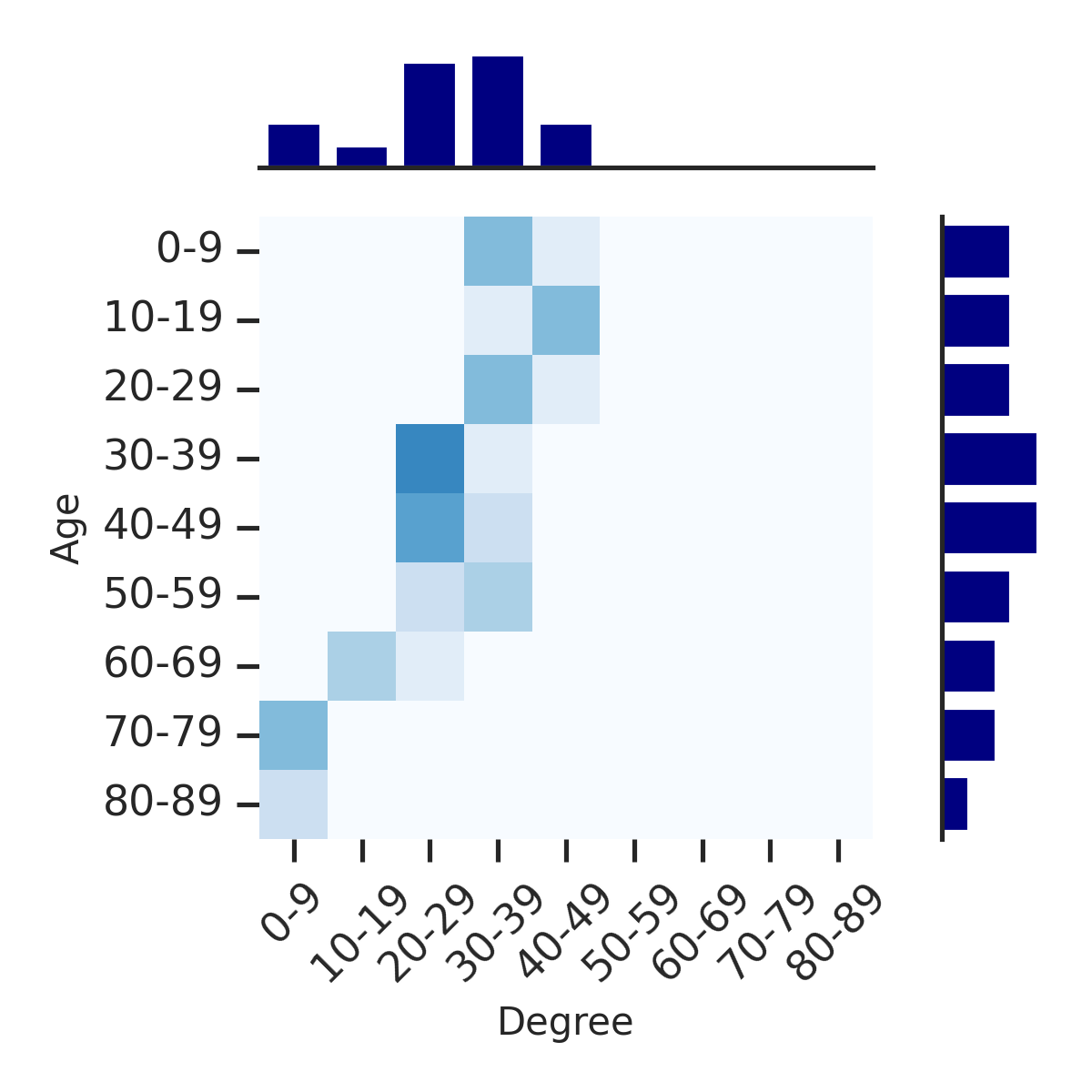}
	\end{minipage}}
	\subfigure[Females]{
		\begin{minipage}[b]{0.32\linewidth}
			\includegraphics[width=1\linewidth]{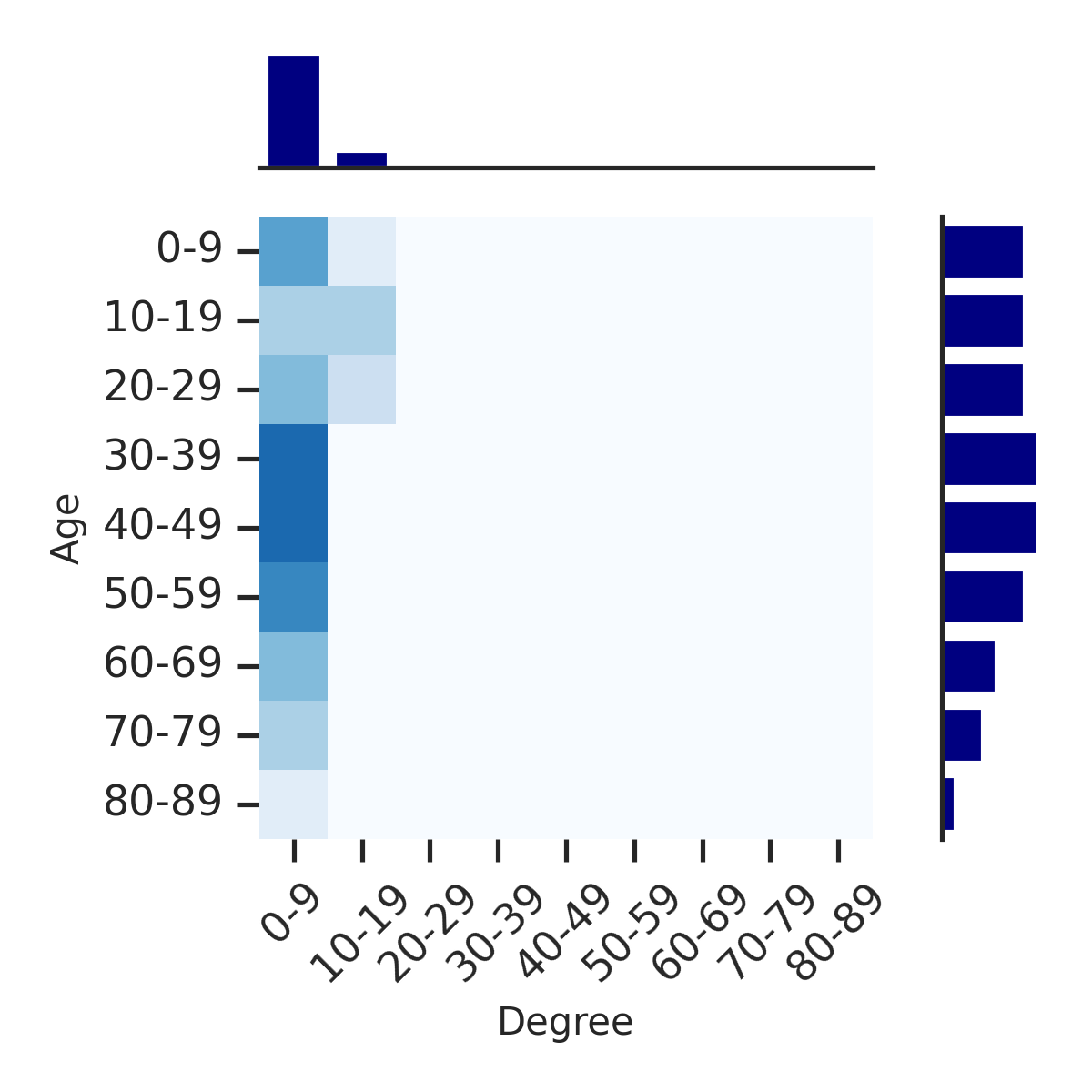}
	\end{minipage}}\\
	\caption{The age and degree distributions in Luxembourg for (a) males and females, (b) males and (c) females.}
 \label{LuxembourgDegree}
\end{figure}

\begin{figure}[H]
	\centering
	\subfigure[Males\&Females]{
		\begin{minipage}[b]{0.32\linewidth}
			\includegraphics[width=1\linewidth]{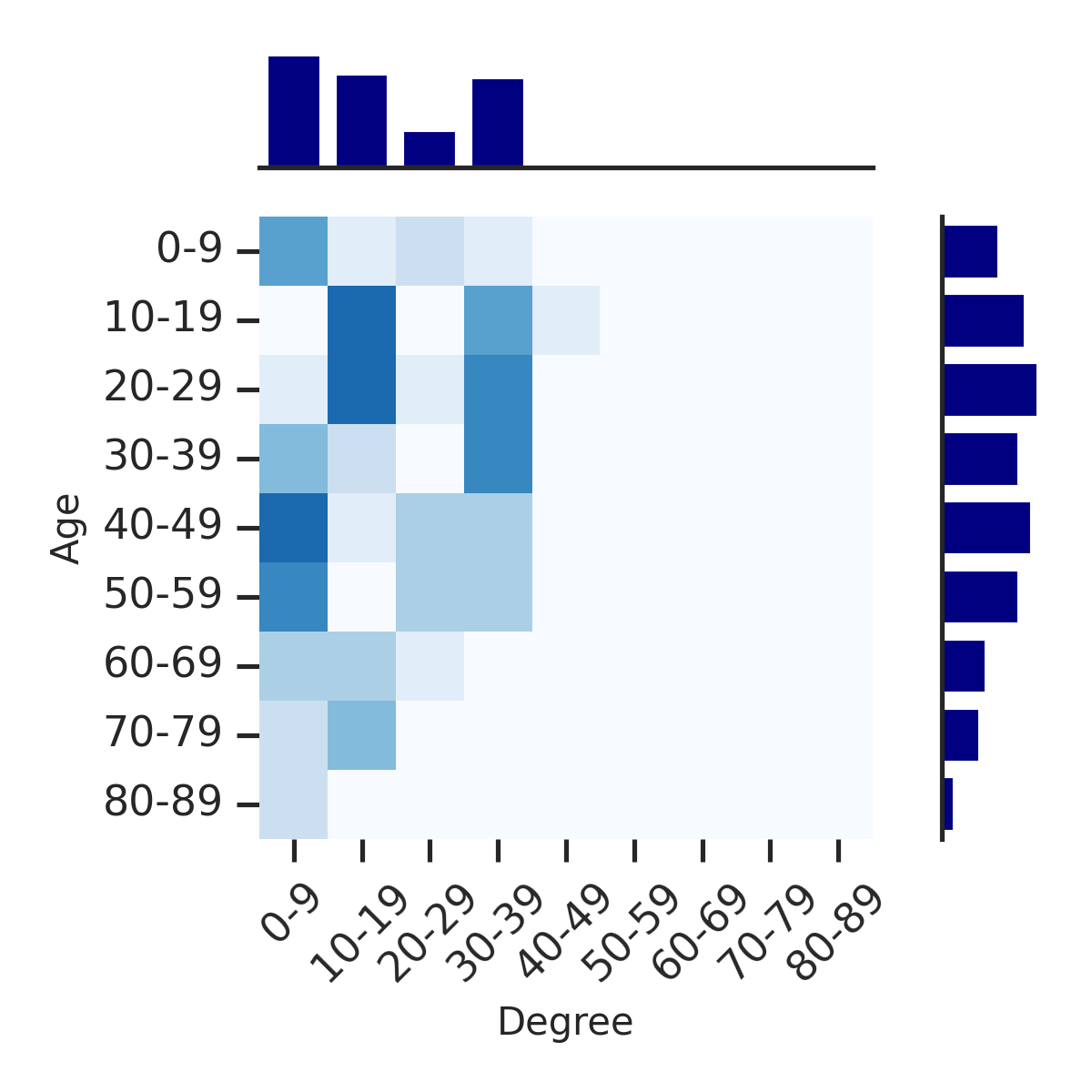}
	\end{minipage}}
	\subfigure[Males]{
		\begin{minipage}[b]{0.32\linewidth}
			\includegraphics[width=1\linewidth]{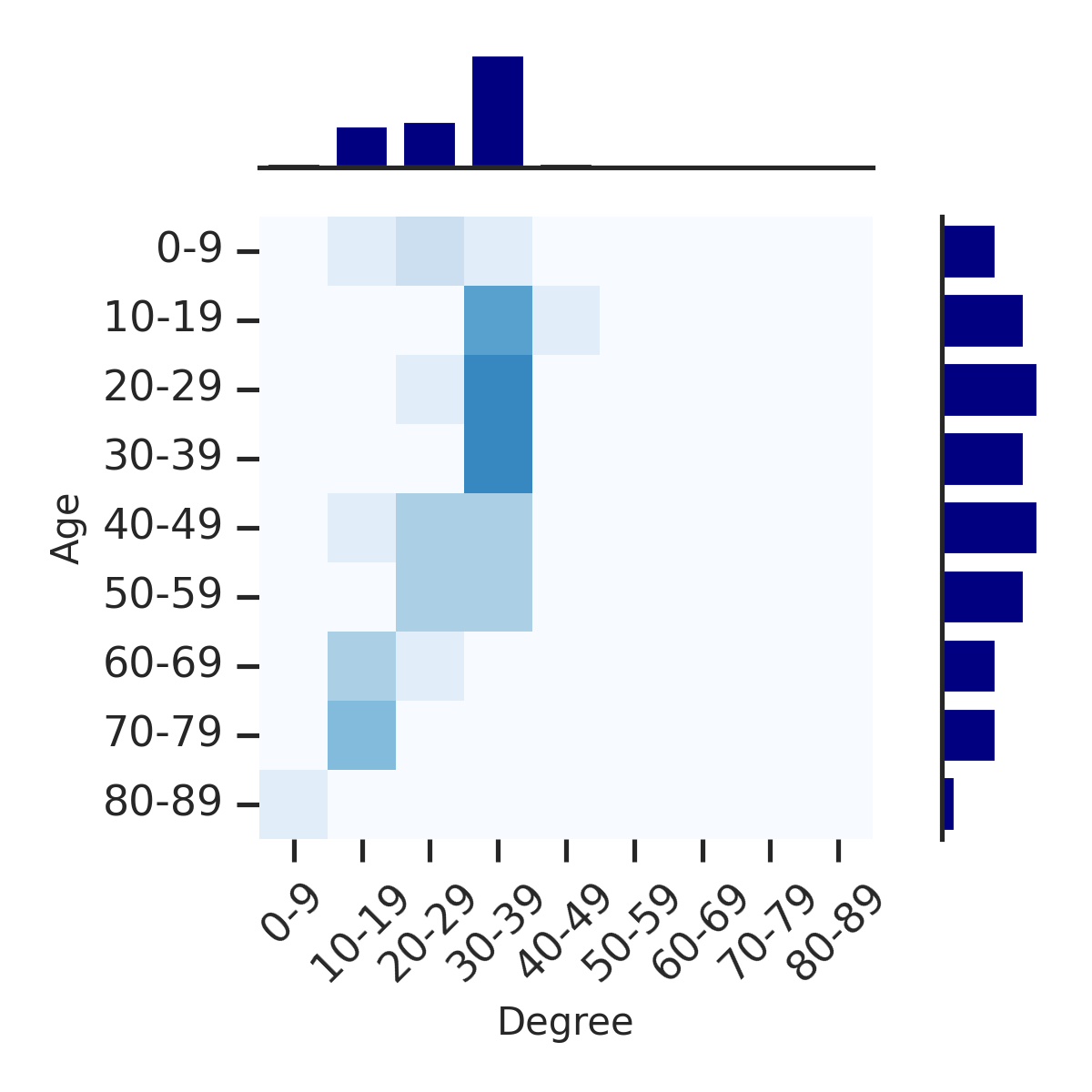}
	\end{minipage}}
	\subfigure[Females]{
		\begin{minipage}[b]{0.32\linewidth}
			\includegraphics[width=1\linewidth]{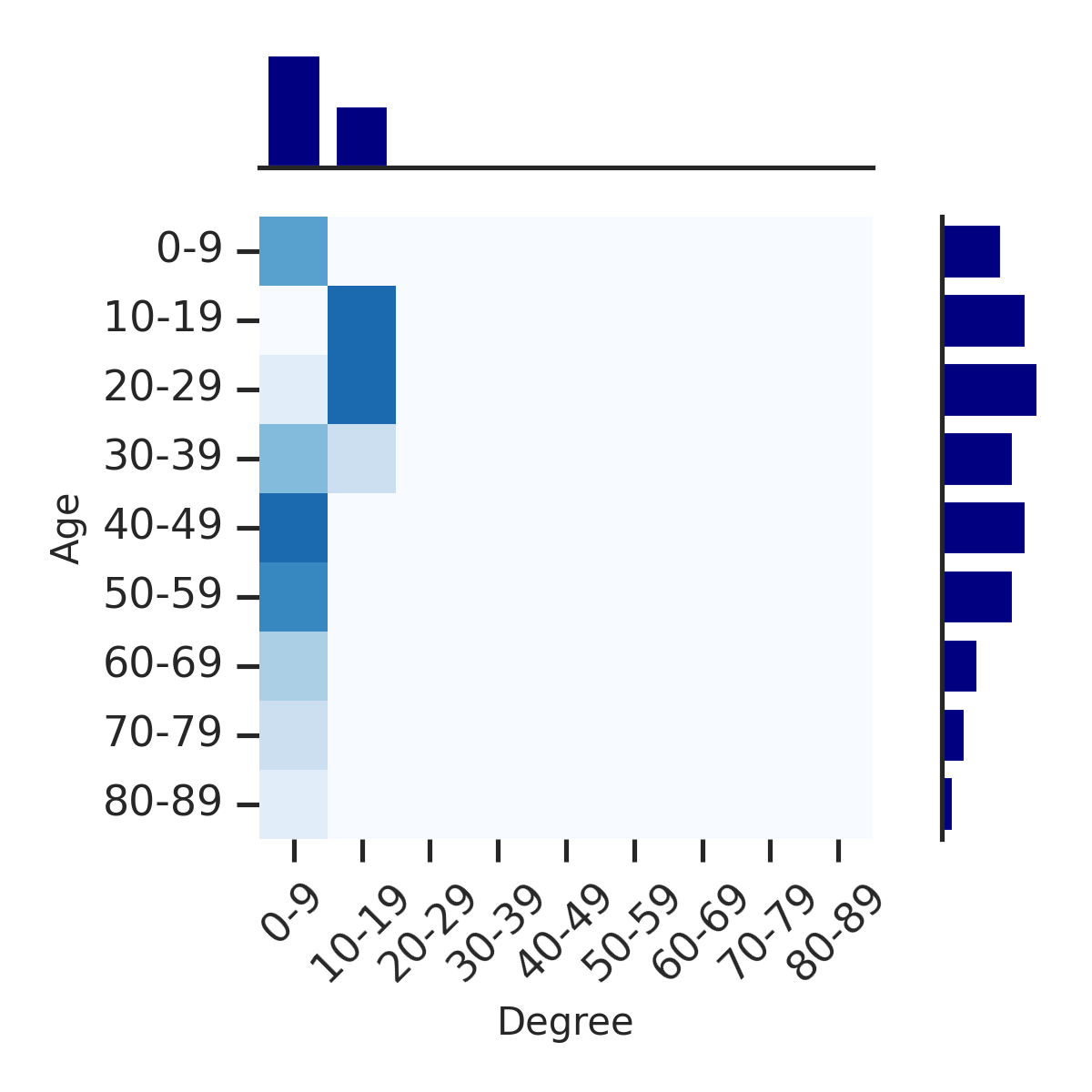}
	\end{minipage}}\\
	\caption{The age and degree distributions in Poland for (a) males and females, (b) males and (c) females.}
 \label{PolandDegree}
\end{figure}

We explore the distributions of node degrees for males, females and both males and females in different ages based on the network simulations of each country, including Belgium (See Fig.~\ref{BelgiumDegree}), Finland (See Fig.~\ref{FinlandDegree}, Germany (See  Fig.~\ref{GermanyDegree}), Italy (See Fig.~\ref{ItalyDegree}), Luxembourg (See Fig.~\ref{LuxembourgDegree}) and Poland (See Fig.~\ref{PolandDegree}). The degree distributions generally have a left tail shape, with a significant number of nodes' degrees fall in [0-10]. The degree distributions of the age groups, shown in each cell of the matrix, form a pyramid shape, where younger nodes generally have a higher node degrees than the older ones. This results from the interests and preferences for young ages and smaller age differences. 

We explore the differences in degrees distributions in each country. We find that the people with preferred sex and age features have higher node degrees. For example, in Belgium and Germany, young females around the age of $[0-50]$ have higher node degrees (at around $10-30$) than males, whose node degrees fall in $[0-9]$. This can be caused by the strong preference for females and young age close to $0$ (See Fig.~\ref{BelgiumFeatPref} (c) and Fig.~\ref{GermanyFeatPref} (c)). This is in contrast with the case of Finland, young males around the age of $[0-50]$ have higher node degrees than females due to the preference for young and male features (See Fig.~\ref{FinlandFeatPref} (c)). In addition, people in dense age groups, with preferred features or preferred similarity, have larger node degrees. For example, in Poland, people around the age of $[10-30]$ have higher node degrees due to people's strong preference for age differences around $0$ and young ages around $17$ (See Fig.~\ref{PolandFeatPref} for $\Delta Age_f 1$ and $Age_f 2$). 


\subsubsection{Clustering Coefficient Distribution}
\label{4clus}

\begin{figure}[H]
	\centering
	\subfigure[Males\&Females]{
		\begin{minipage}[b]{0.32\linewidth}
			\includegraphics[width=1\linewidth]{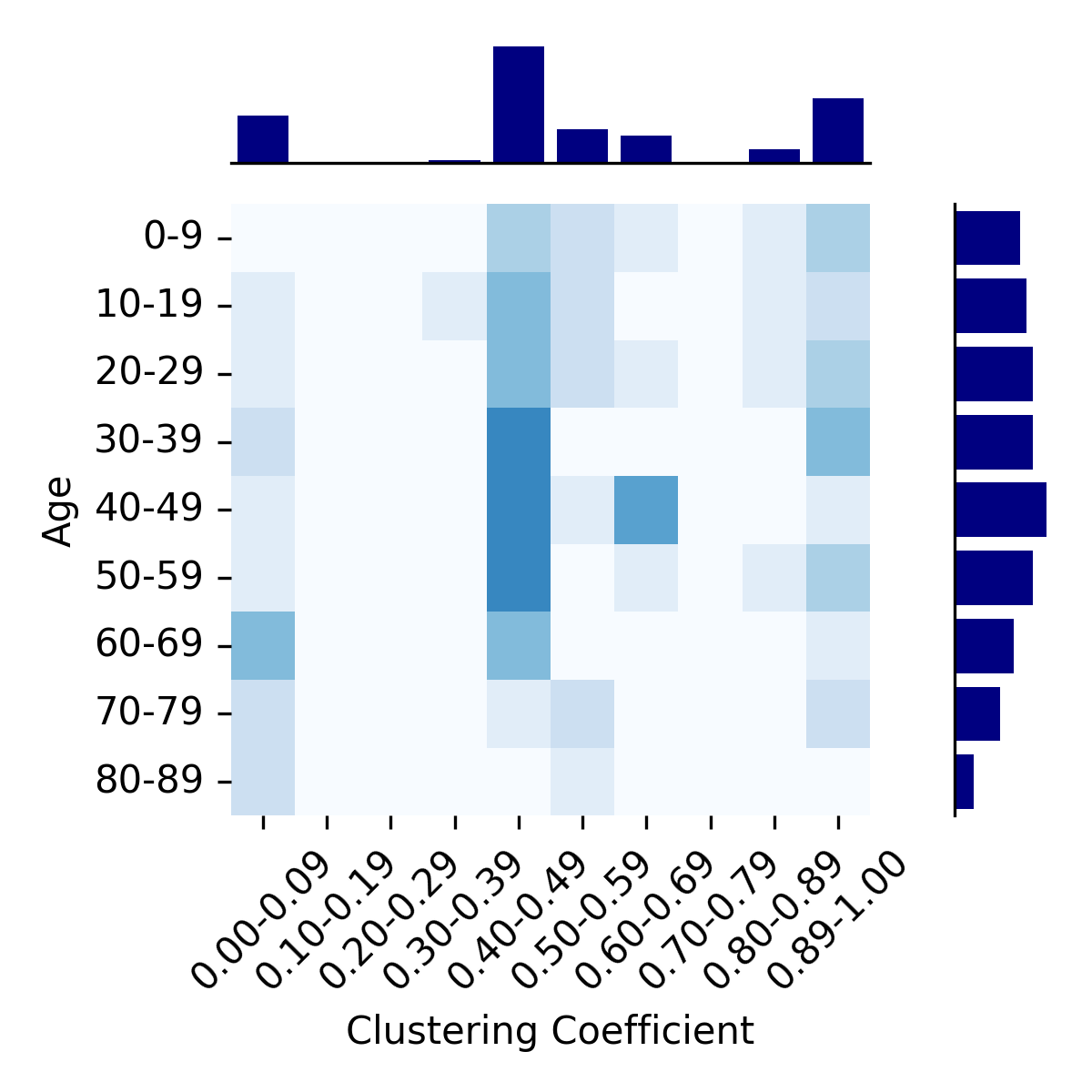}
	\end{minipage}}
	\subfigure[Males]{
		\begin{minipage}[b]{0.32\linewidth}
			\includegraphics[width=1\linewidth]{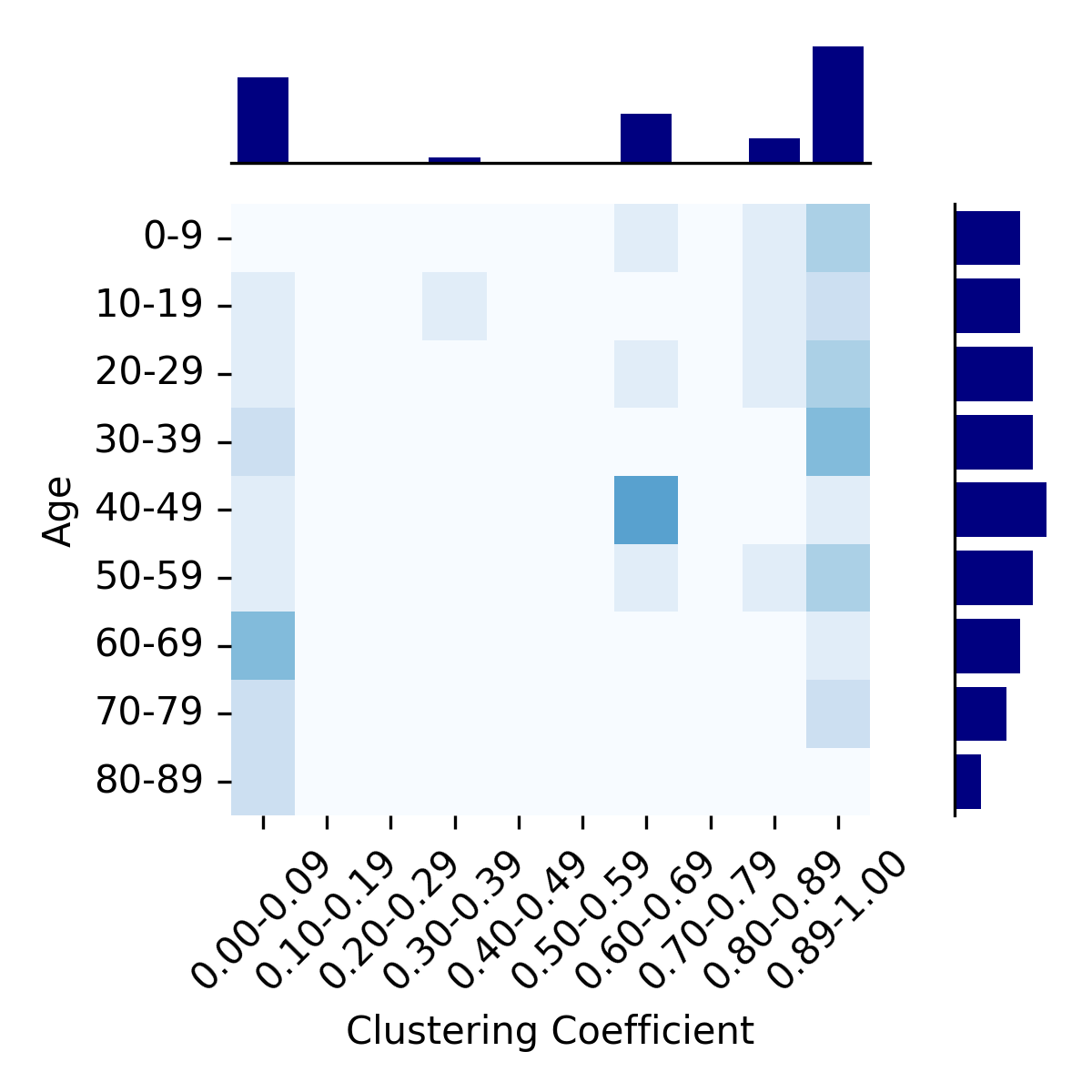}
	\end{minipage}}
	\subfigure[Females]{
		\begin{minipage}[b]{0.32\linewidth}
			\includegraphics[width=1\linewidth]{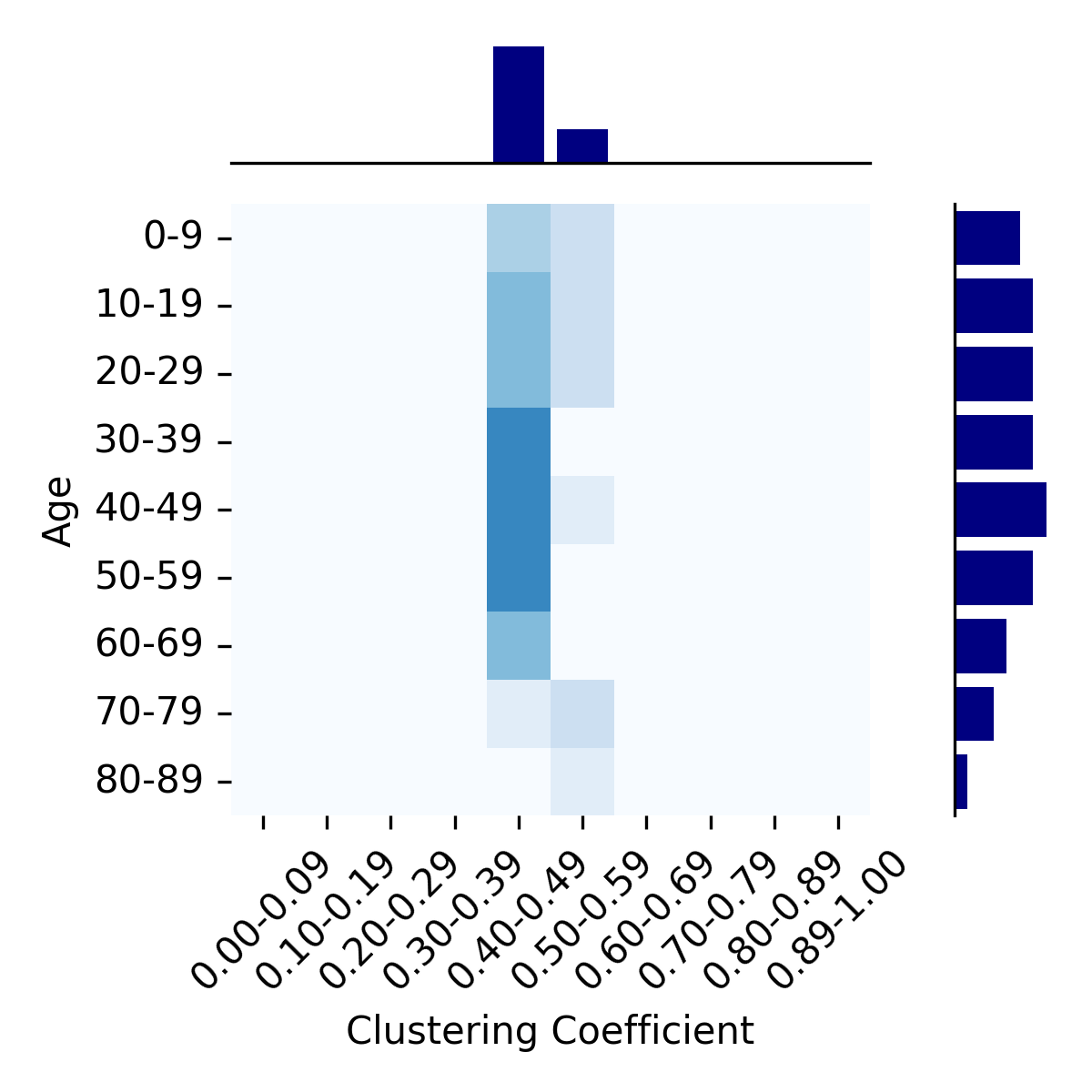}
	\end{minipage}}\\
	\caption{The age and clustering coefficient distributions in Belgium for (a) males and females, (b) males and (c) females.}
 \label{BelgiumClus}
\end{figure}

\begin{figure}[H]
	\centering
	\subfigure[Males\&Females]{
		\begin{minipage}[b]{0.32\linewidth}
			\includegraphics[width=1\linewidth]{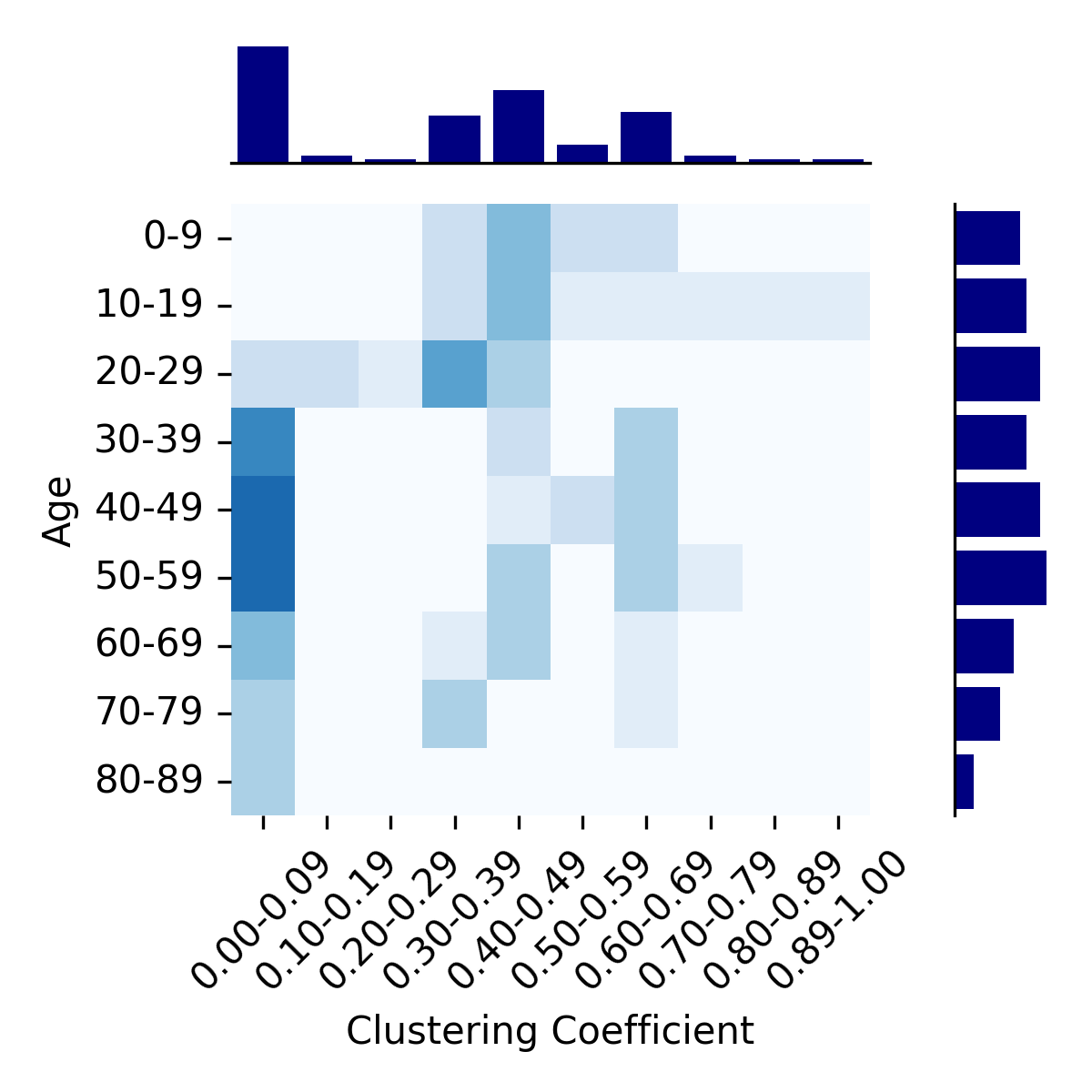}
	\end{minipage}}
	\subfigure[Males]{
		\begin{minipage}[b]{0.32\linewidth}
			\includegraphics[width=1\linewidth]{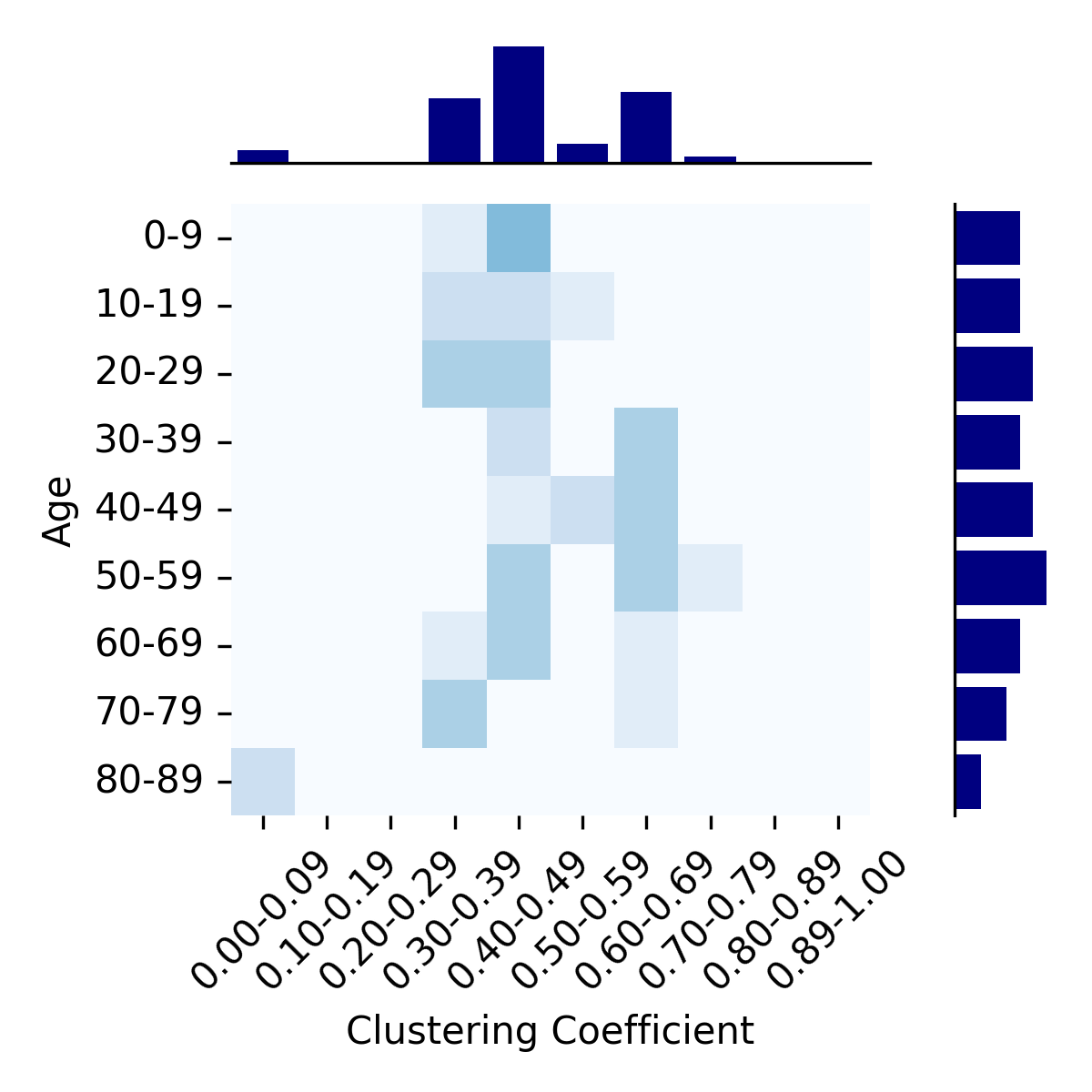}
	\end{minipage}}
	\subfigure[Females]{
		\begin{minipage}[b]{0.32\linewidth}
			\includegraphics[width=1\linewidth]{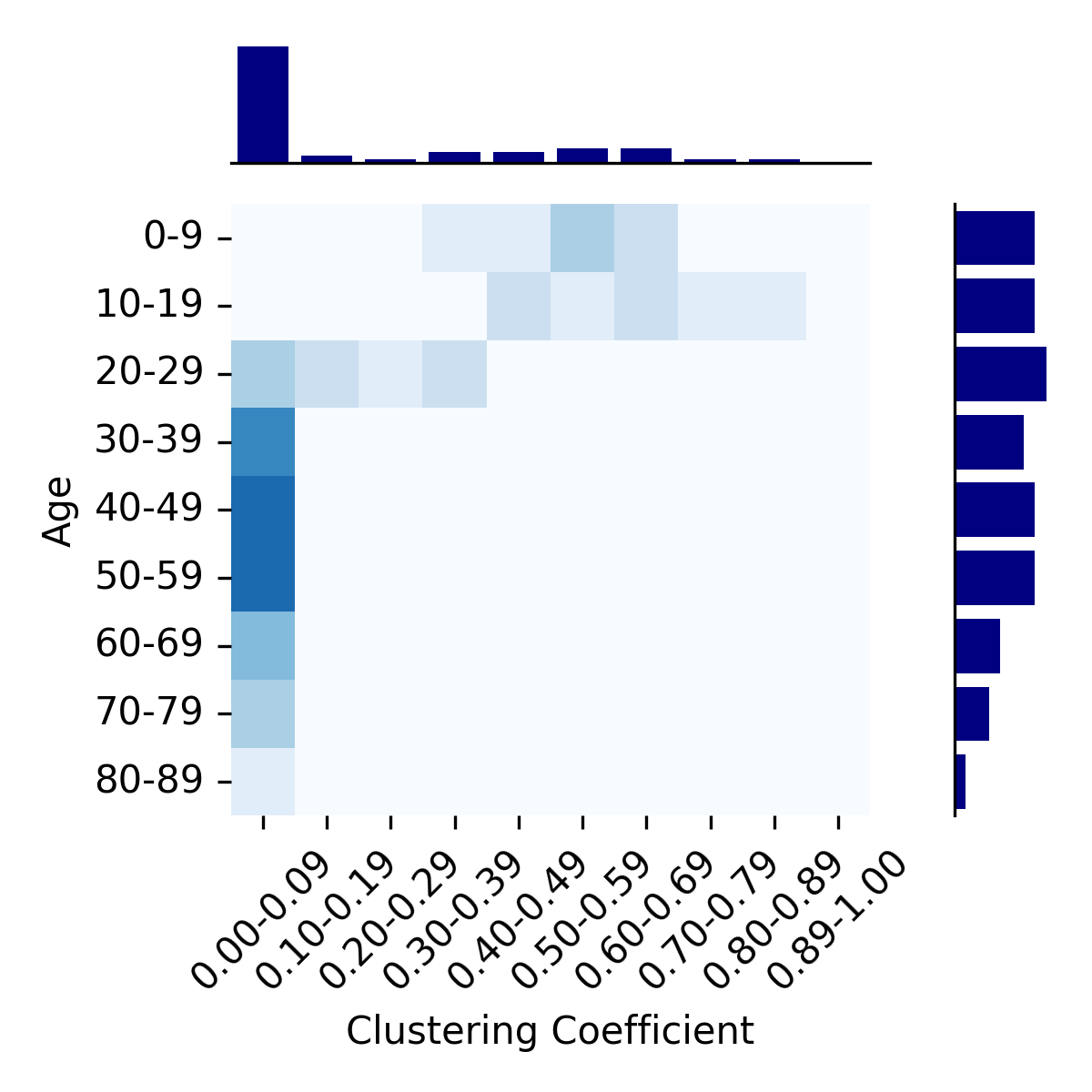}
	\end{minipage}}\\
	\caption{The age and clustering coefficient distributions in Finland for (a) males and females, (b) males and (c) females.}
 \label{FinlandClus}
\end{figure}

\begin{figure}[H]
	\centering
	\subfigure[Males\&Females]{
		\begin{minipage}[b]{0.32\linewidth}
			\includegraphics[width=1\linewidth]{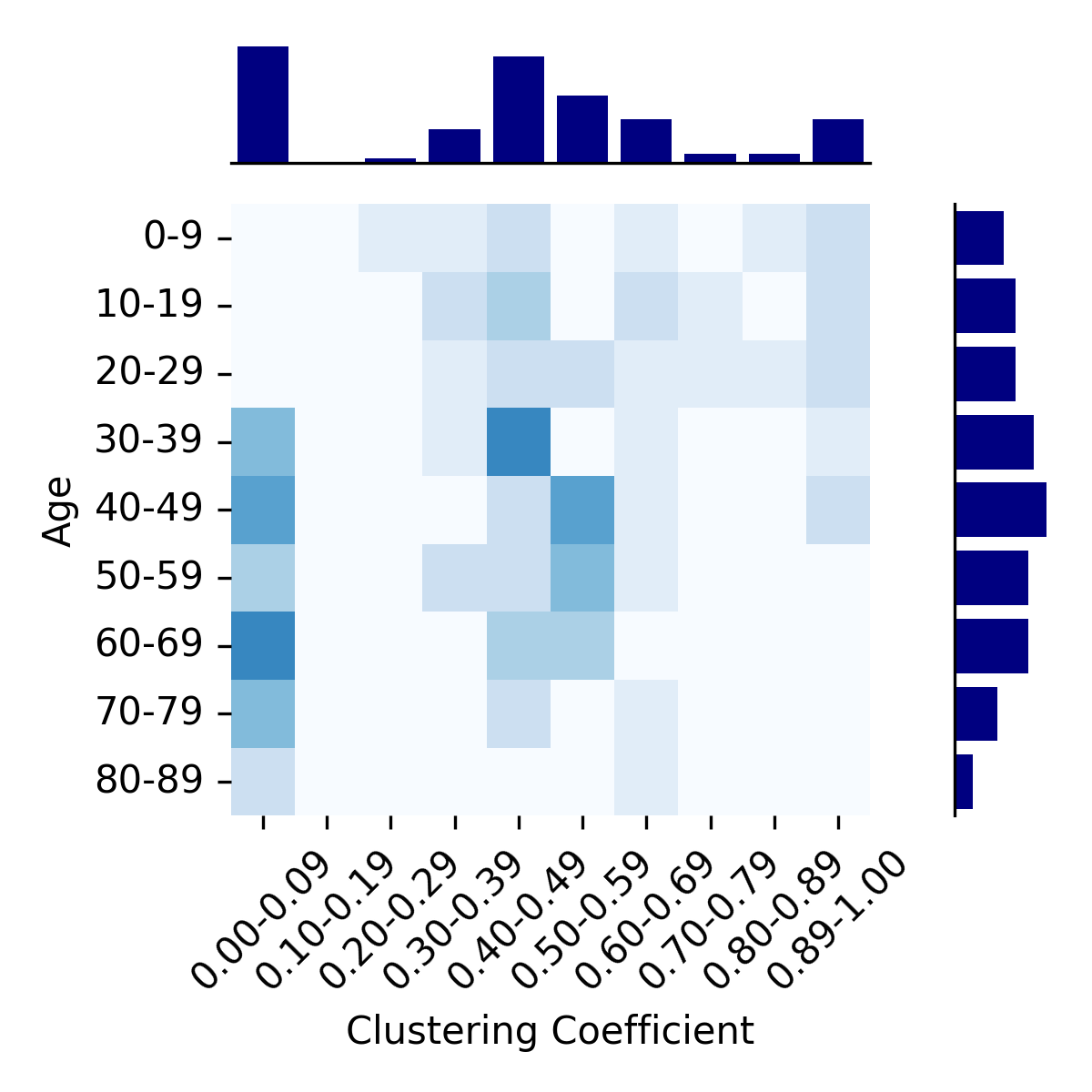}
	\end{minipage}}
	\subfigure[Males]{
		\begin{minipage}[b]{0.32\linewidth}
			\includegraphics[width=1\linewidth]{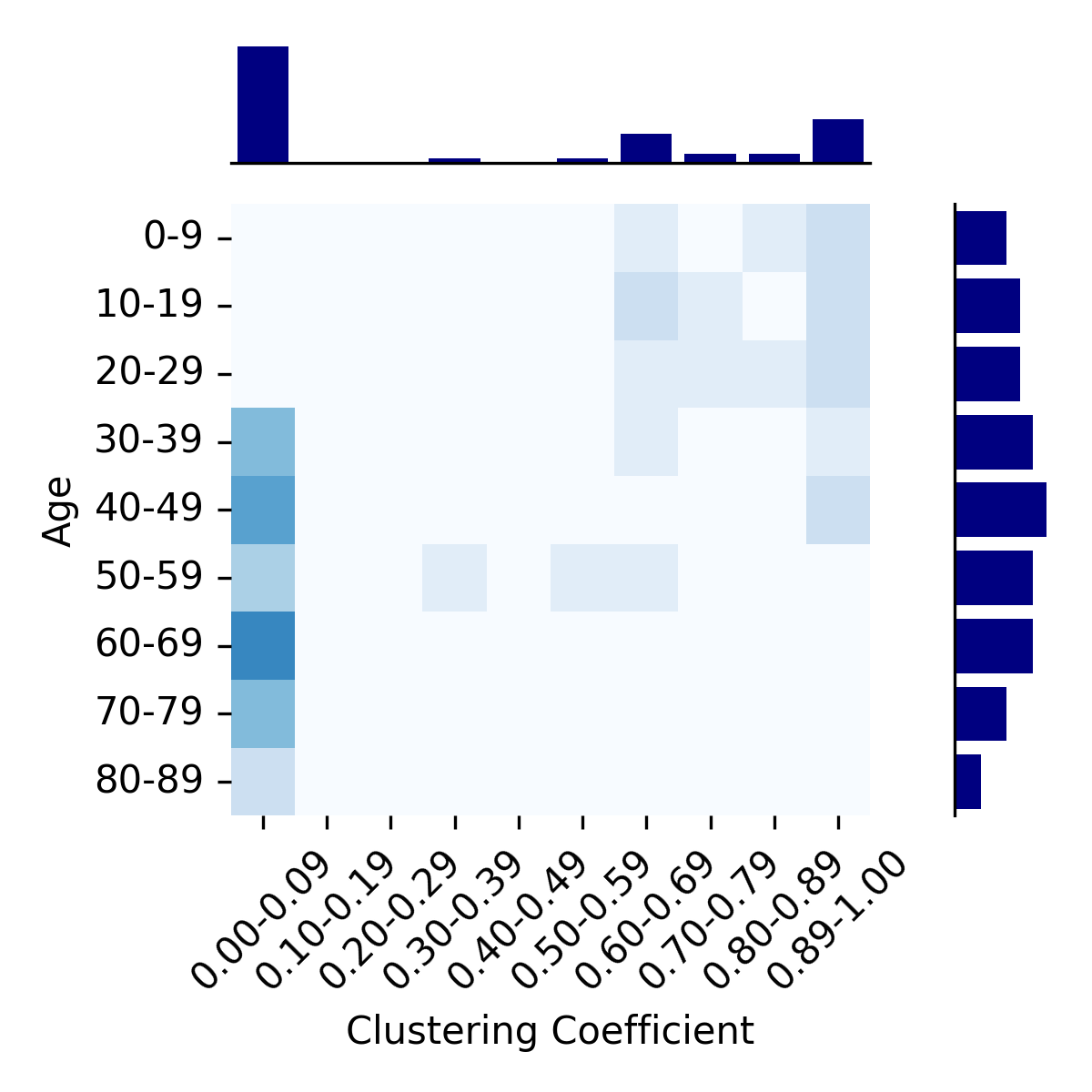}
	\end{minipage}}
	\subfigure[Females]{
		\begin{minipage}[b]{0.32\linewidth}
			\includegraphics[width=1\linewidth]{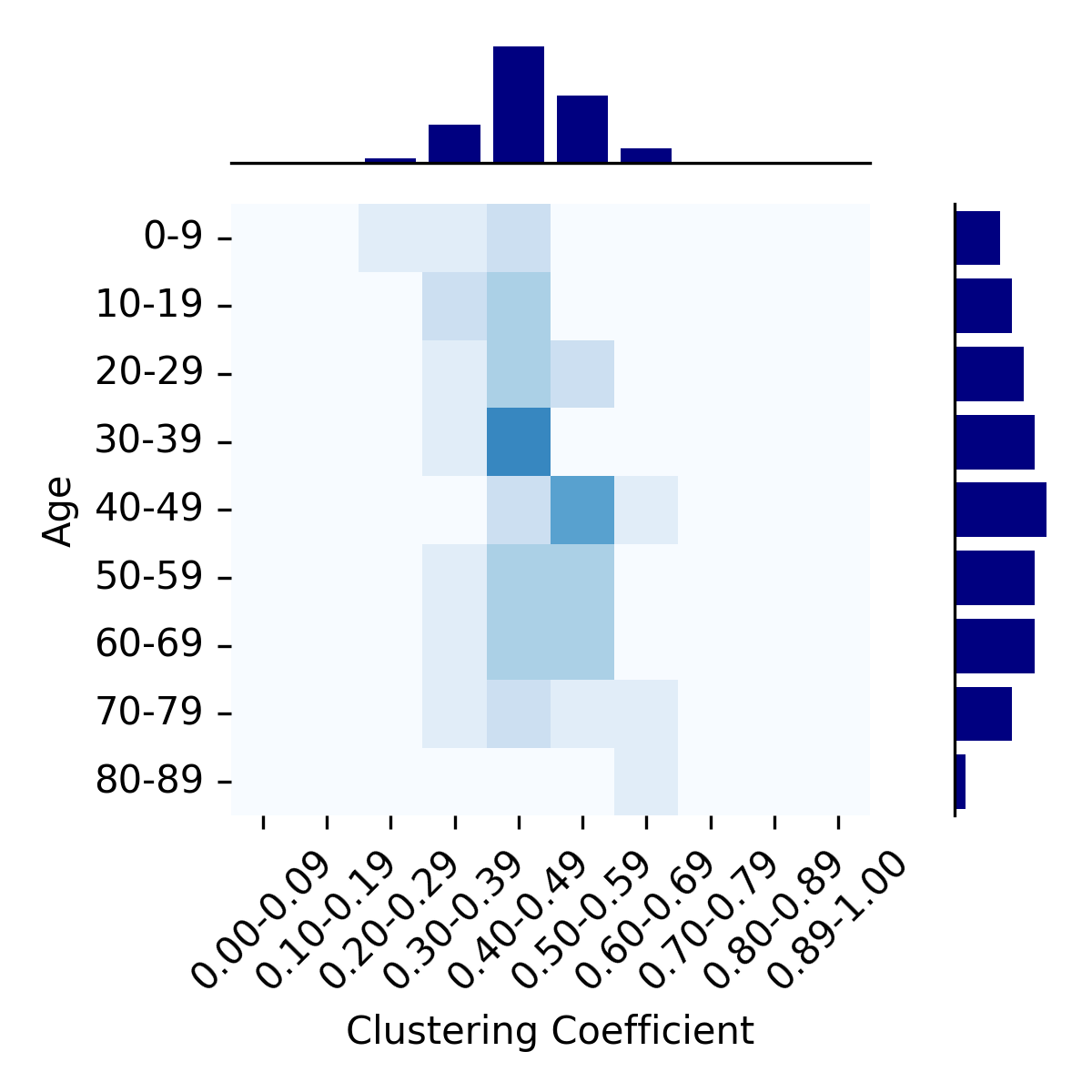}
	\end{minipage}}\\
	\caption{The age and clustering coefficient distributions in Germany for (a) males and females, (b) males and (c) females.}
 \label{GermanyClus}
\end{figure}

\begin{figure}[H]
	\centering
	\subfigure[Males\&Females]{
		\begin{minipage}[b]{0.32\linewidth}
			\includegraphics[width=1\linewidth]{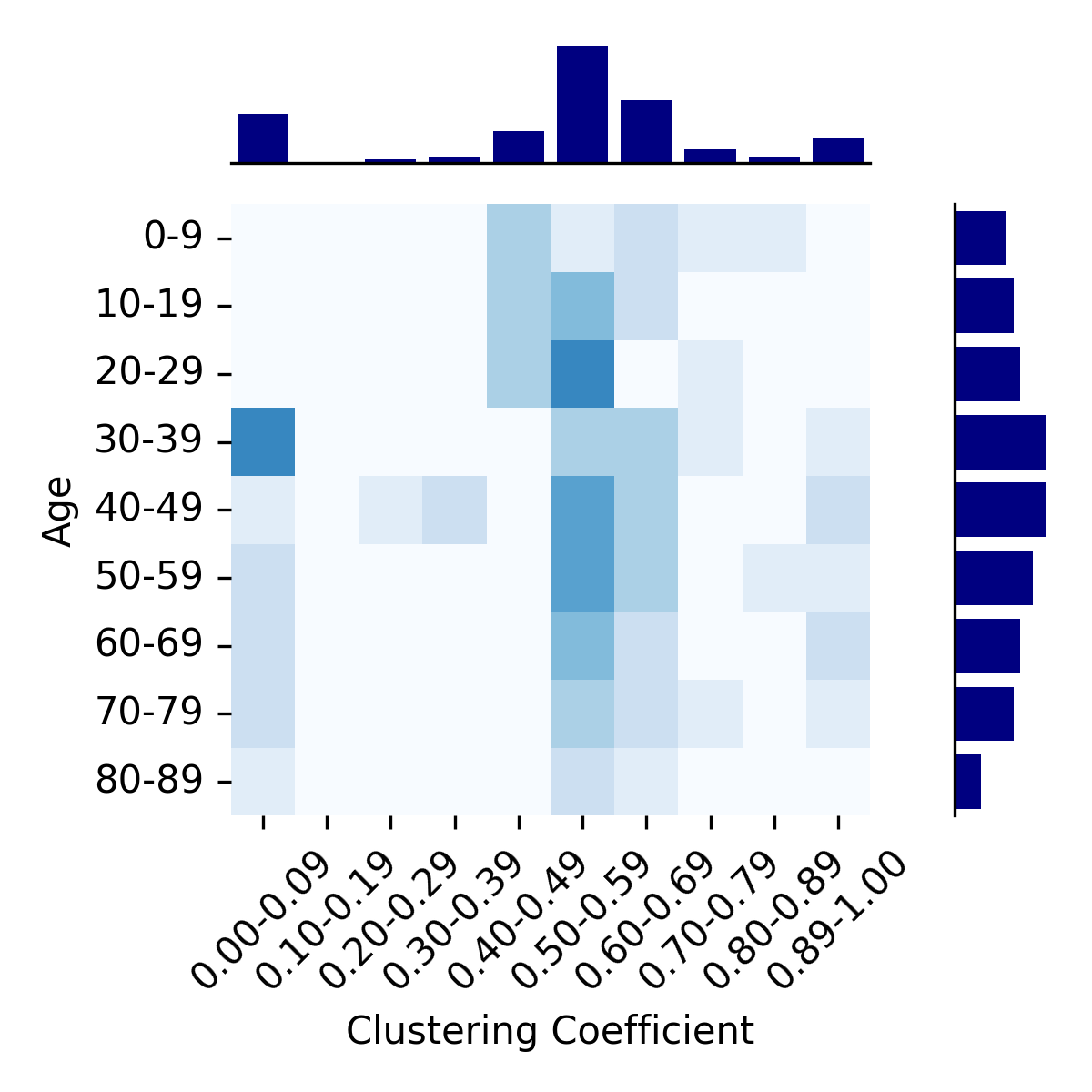}
	\end{minipage}}
	\subfigure[Males]{
		\begin{minipage}[b]{0.32\linewidth}
			\includegraphics[width=1\linewidth]{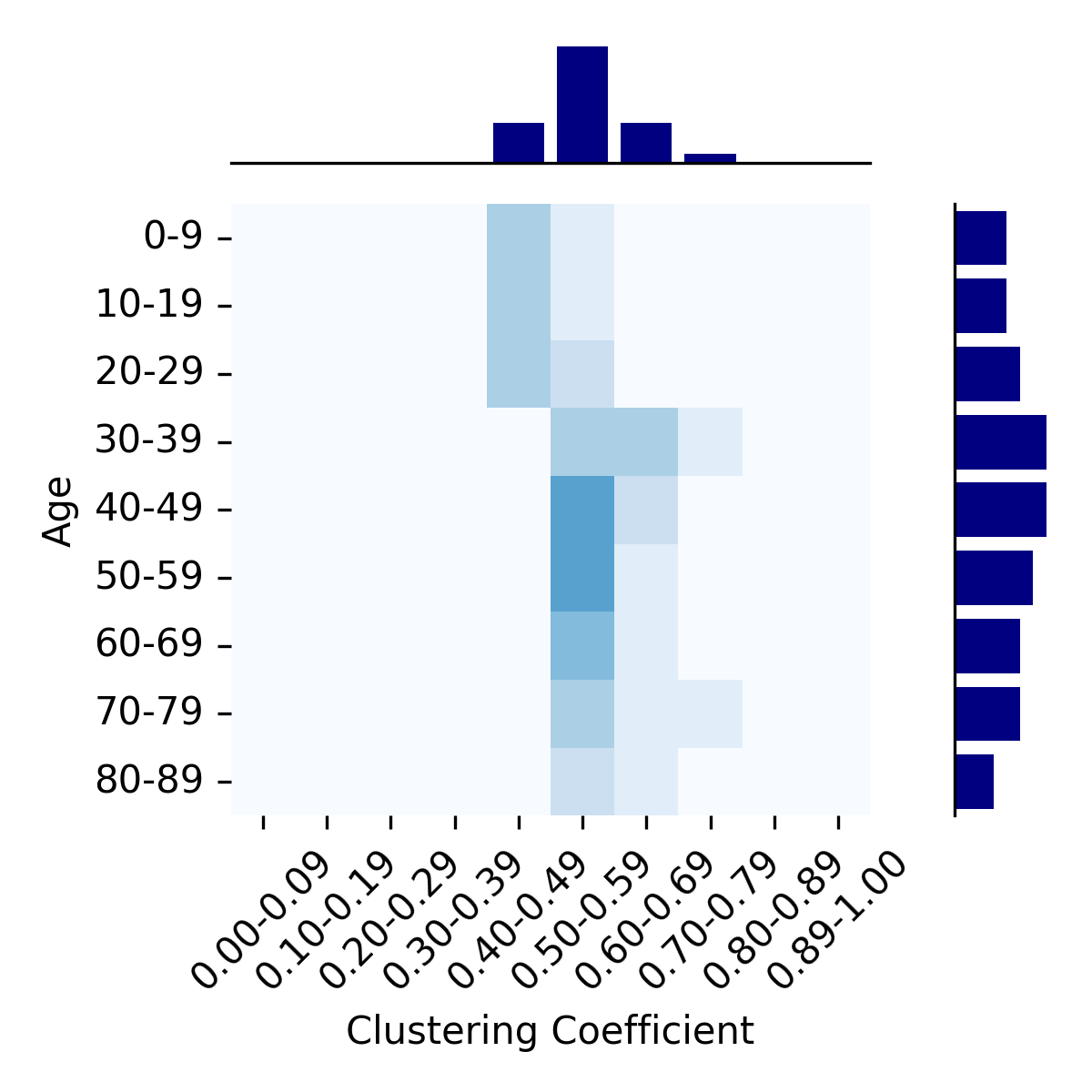}
	\end{minipage}}
	\subfigure[Females]{
		\begin{minipage}[b]{0.32\linewidth}
			\includegraphics[width=1\linewidth]{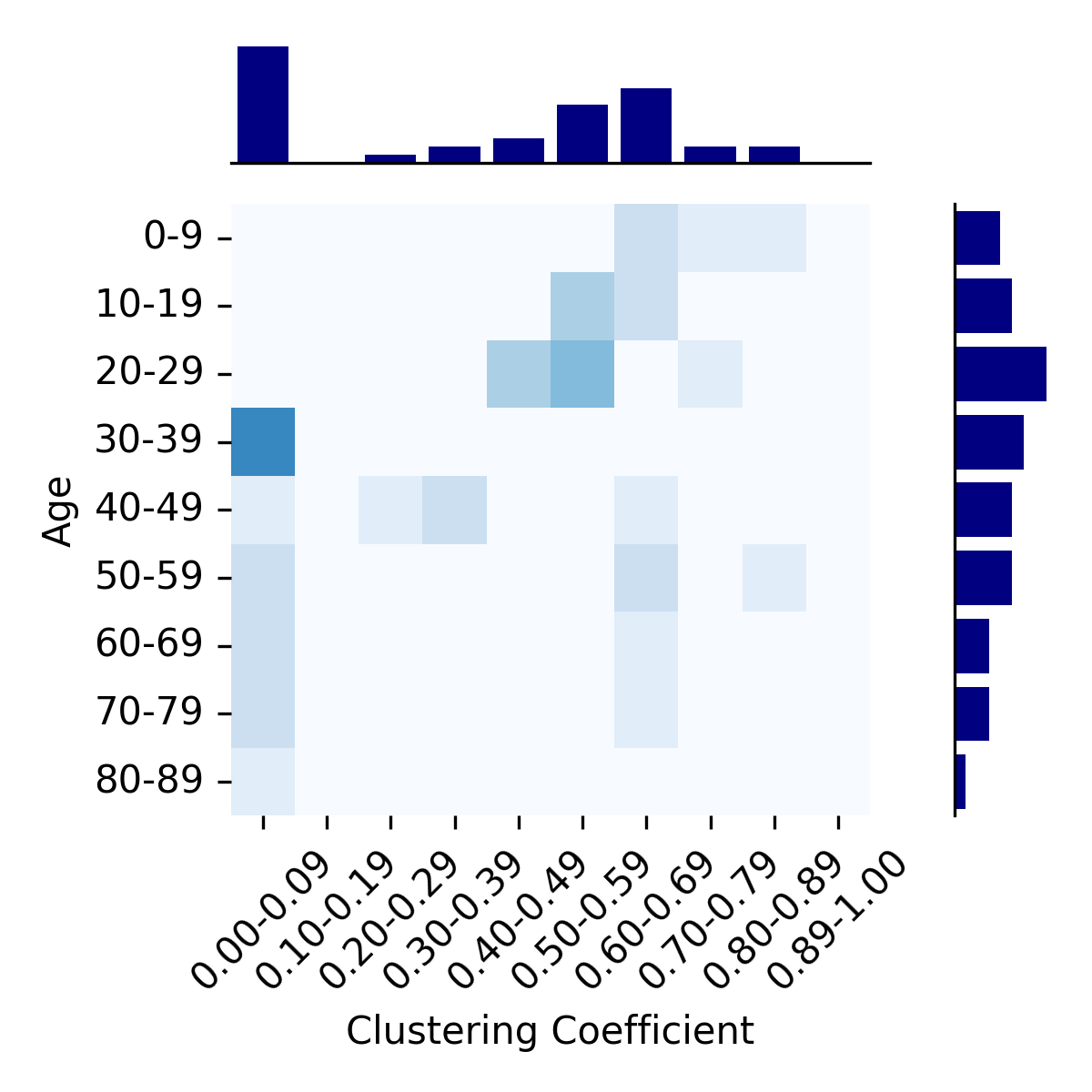}
	\end{minipage}}\\
	\caption{The age and clustering coefficient distributions in Italy for (a) males and females, (b) males and (c) females.}
 \label{ItalyClus}
\end{figure}

\begin{figure}[H]
	\centering
	\subfigure[Males\&Females]{
		\begin{minipage}[b]{0.32\linewidth}
			\includegraphics[width=1\linewidth]{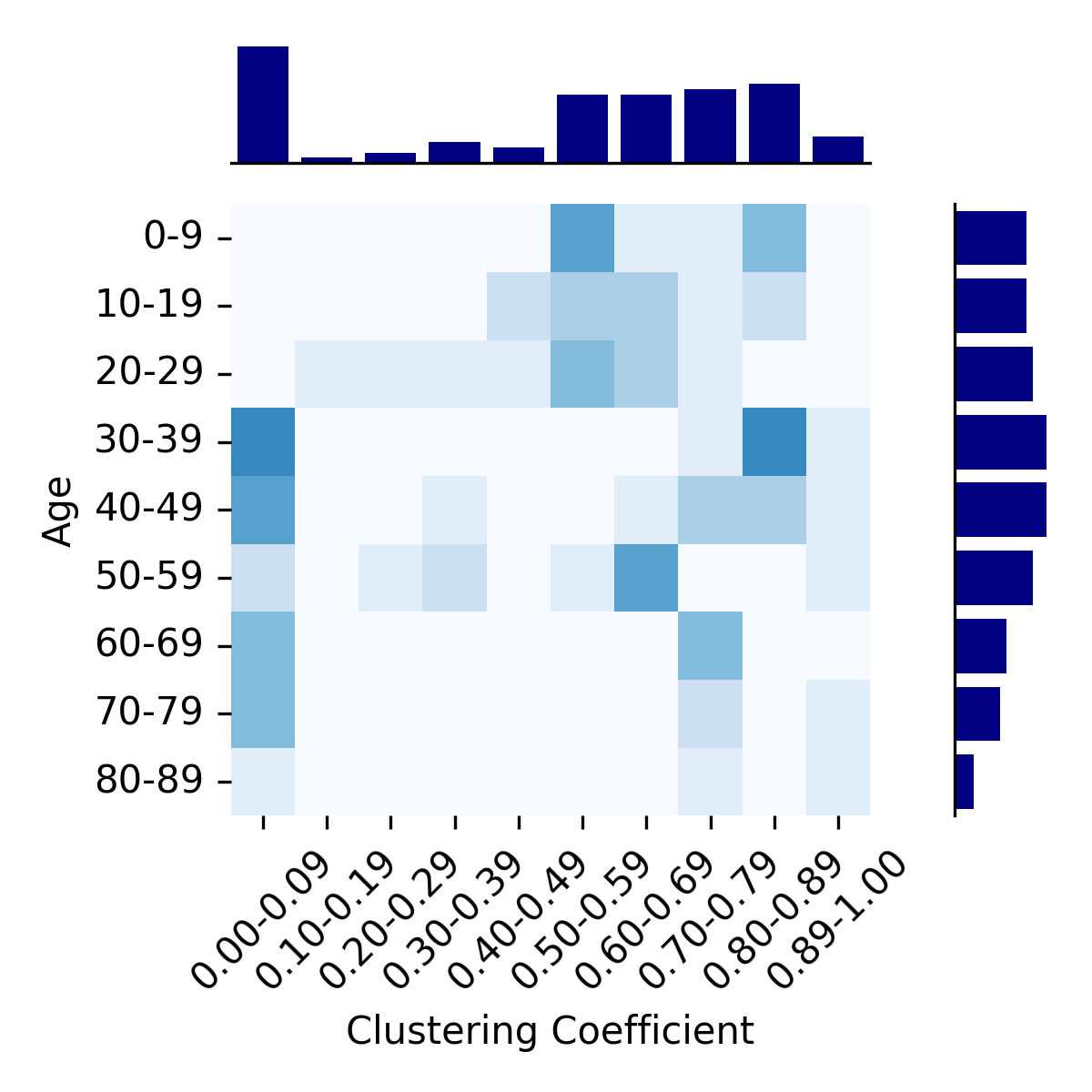}
	\end{minipage}}
	\subfigure[Males]{
		\begin{minipage}[b]{0.32\linewidth}
			\includegraphics[width=1\linewidth]{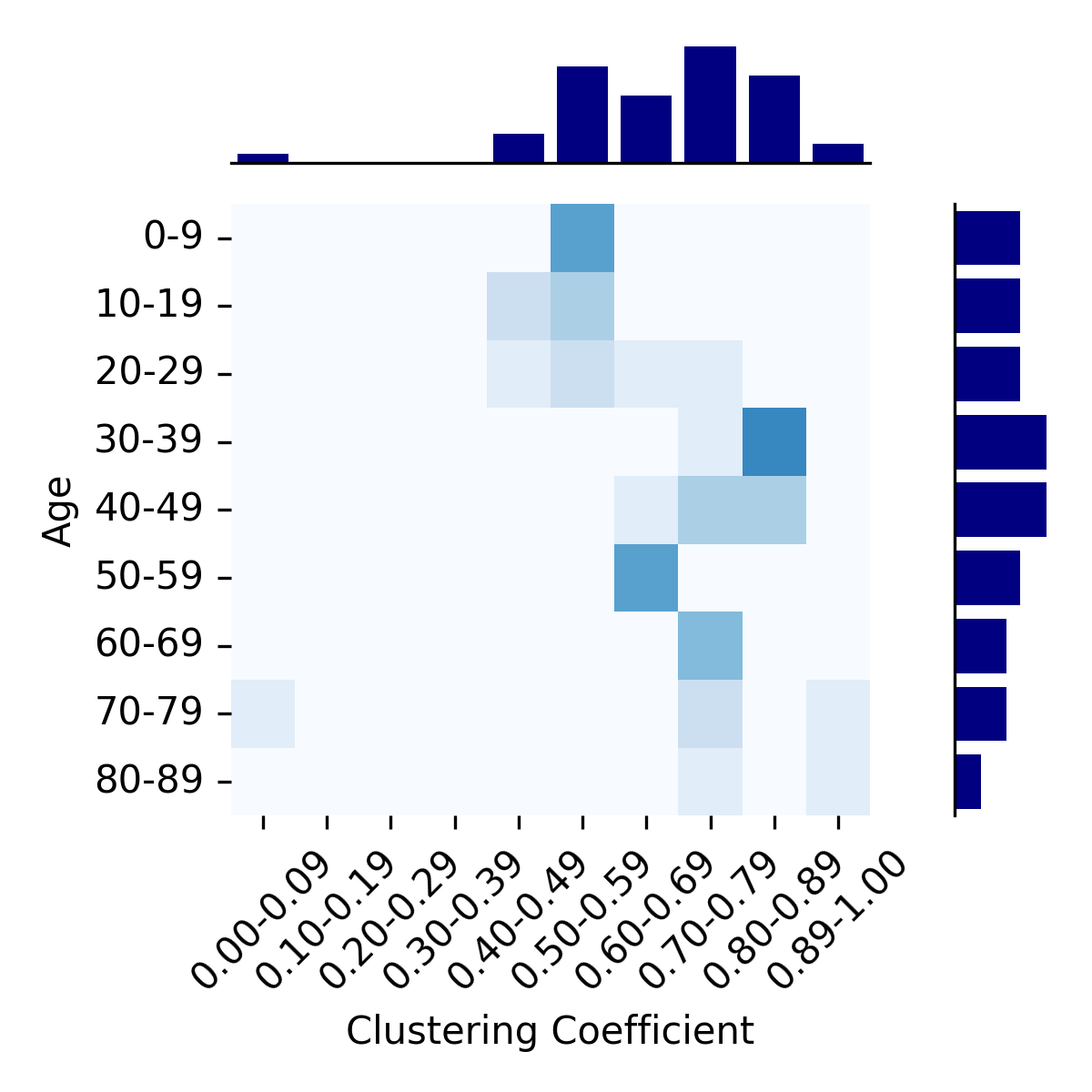}
	\end{minipage}}
	\subfigure[Females]{
		\begin{minipage}[b]{0.32\linewidth}
			\includegraphics[width=1\linewidth]{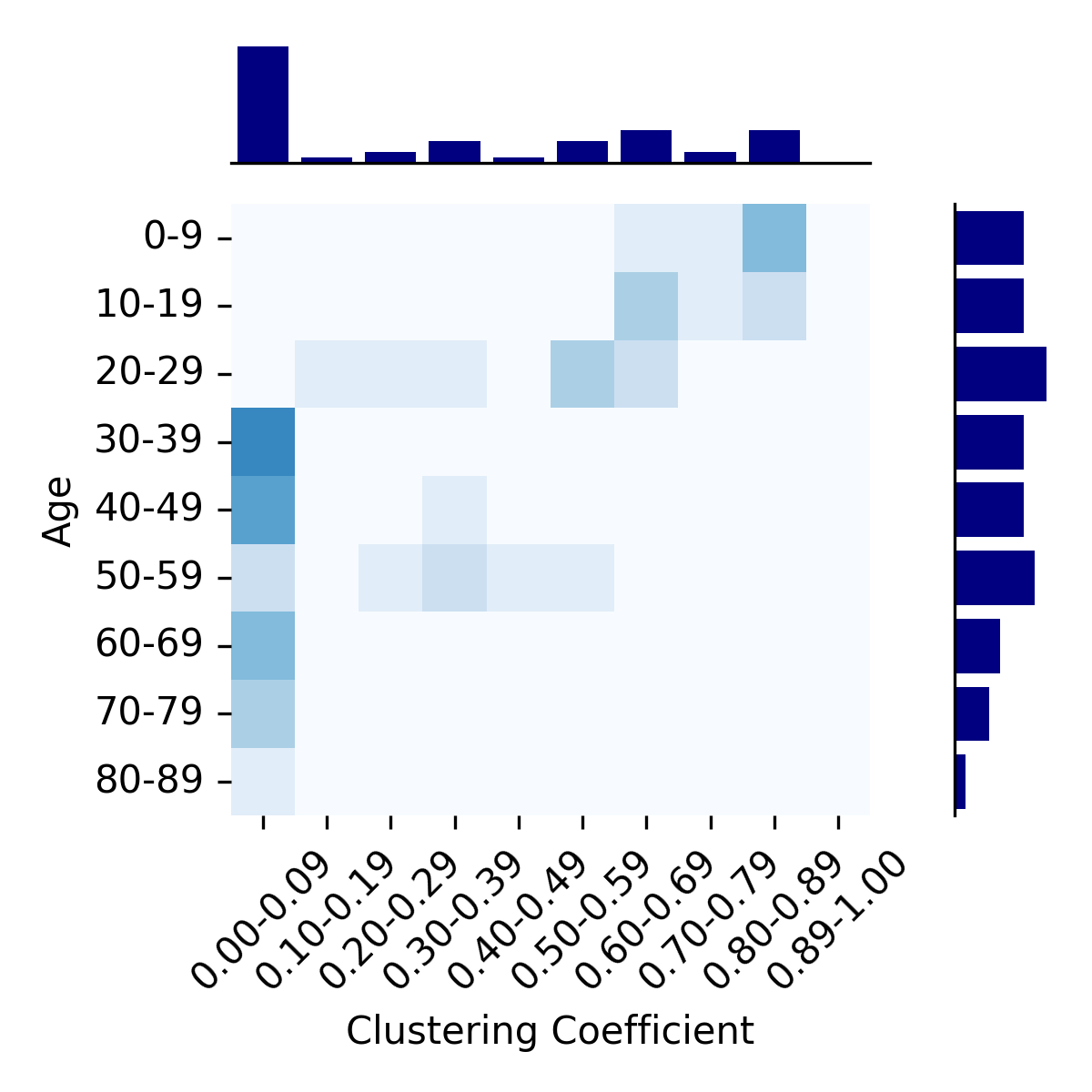}
	\end{minipage}}\\
	\caption{The age and clustering coefficient distributions in Luxembourg for (a) males and females, (b) males and (c) females.}
 \label{LuxembourgClus}
\end{figure}

\begin{figure}[H]
	\centering
	\subfigure[Males\&Females]{
		\begin{minipage}[b]{0.32\linewidth}
			\includegraphics[width=1\linewidth]{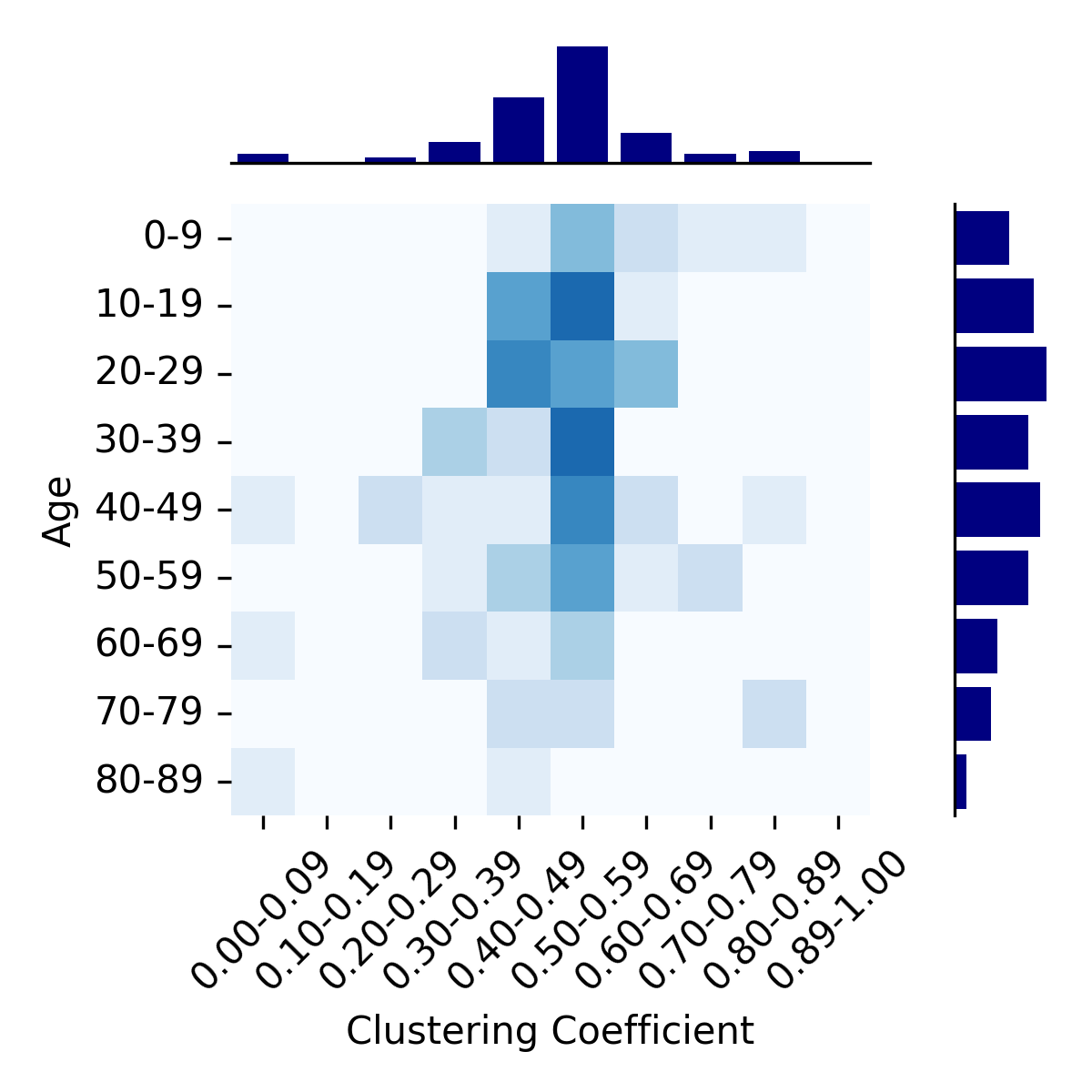}
	\end{minipage}}
	\subfigure[Males]{
		\begin{minipage}[b]{0.32\linewidth}
			\includegraphics[width=1\linewidth]{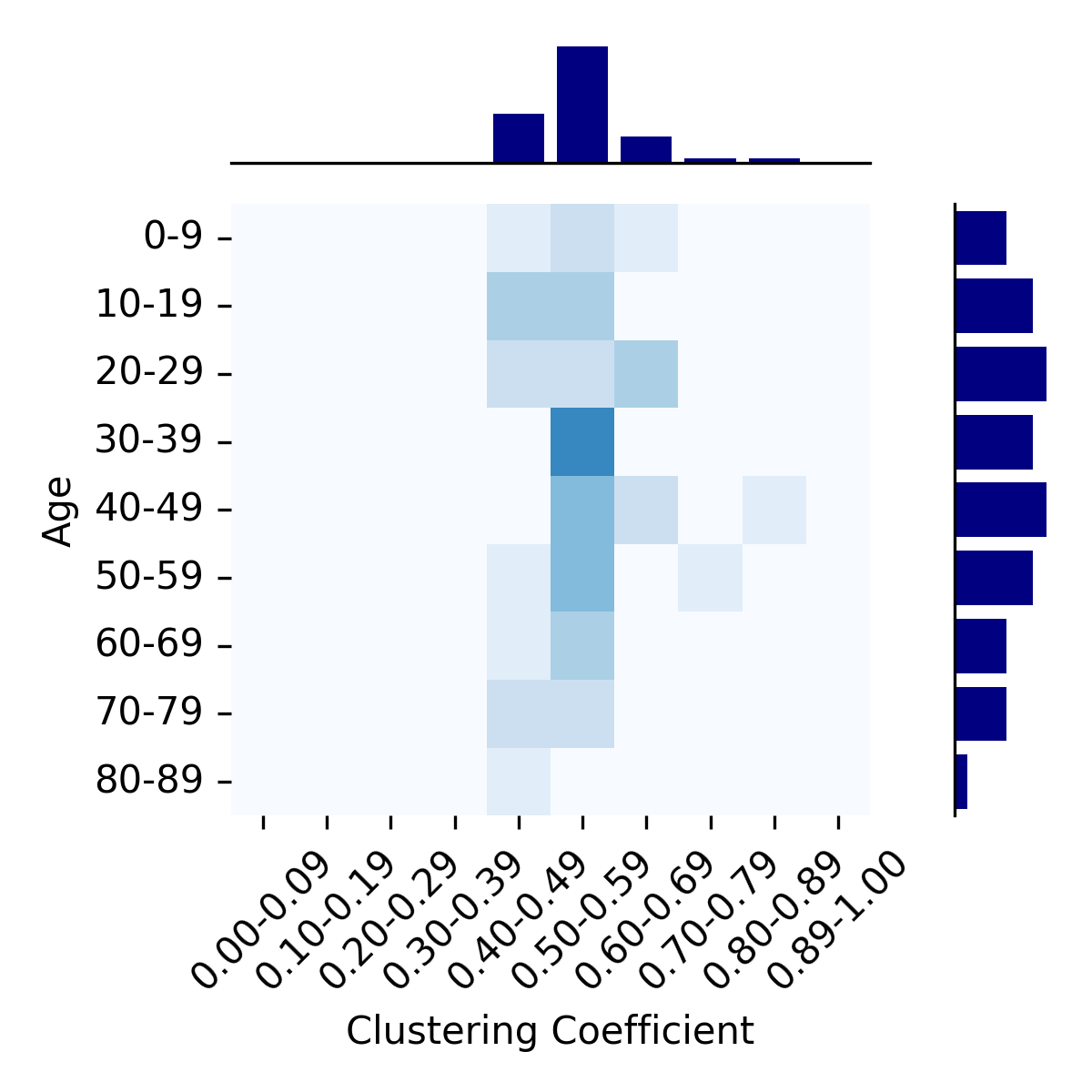}
	\end{minipage}}
	\subfigure[Females]{
		\begin{minipage}[b]{0.32\linewidth}
			\includegraphics[width=1\linewidth]{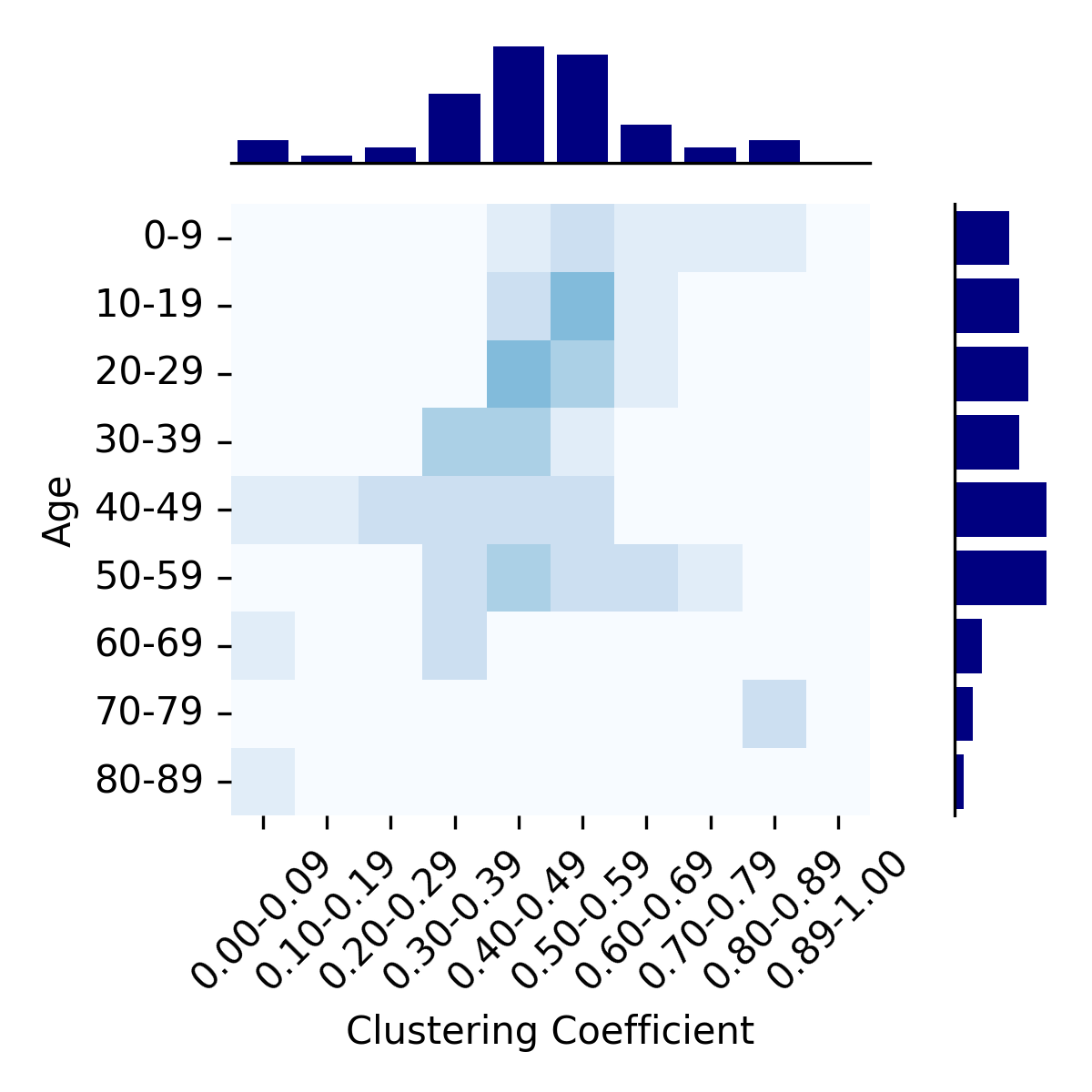}
	\end{minipage}}\\
	\caption{The age and clustering distributions in Belgium for (a) males and females, (b) males and (c) females.}
 \label{PolandClus}
\end{figure}

We explore the distributions of clustering coefficient for males, females and both males and females of different ages based on the network simulations of each country, including Belgium (See Fig.~\ref{BelgiumClus}), Finland (See Fig.~\ref{FinlandClus}, Germany (See  Fig.~\ref{GermanyClus}), Italy (See Fig.~\ref{ItalyClus}), Luxembourg (See Fig.~\ref{LuxembourgClus}) and Poland (See Fig.~\ref{PolandClus}). The clustering coefficient is unevenly distributed, resulting from the heterogeneous preferences of nodes and the corresponding connections with various nodes. This challenges the interpretation of results. However, we can still find from the case of Belgium and Finland that the strongly preferred nodes have higher clustering coefficients than others. In Belgium, all females have a positive clustering coefficient due to the stronger preference for the crisp female features (See Fig.~\ref{BelgiumClus}); this contrasts with the clustering coefficient of males, where some of them have significantly lower clustering coefficients between $0$ and $0.10$. Similarly, in Finland, people around $[0-20]$ have a higher clustering coefficient than other nodes. This is because people in Finland have a strong interest and preference for age differences around $0$ and $32$. In this way, young people around the age of $[0-20]$ are preferred by a significant number of peers and people at their parents' age $[32-52]$ (See Fig.~\ref{FinlandClus}).

\subsubsection{Shortest Path Length Distribution}
\label{5path}

\begin{figure}[H]
	\centering
	\subfigure[People-People]{
		\begin{minipage}[b]{0.22\linewidth}
			\includegraphics[width=1\linewidth]{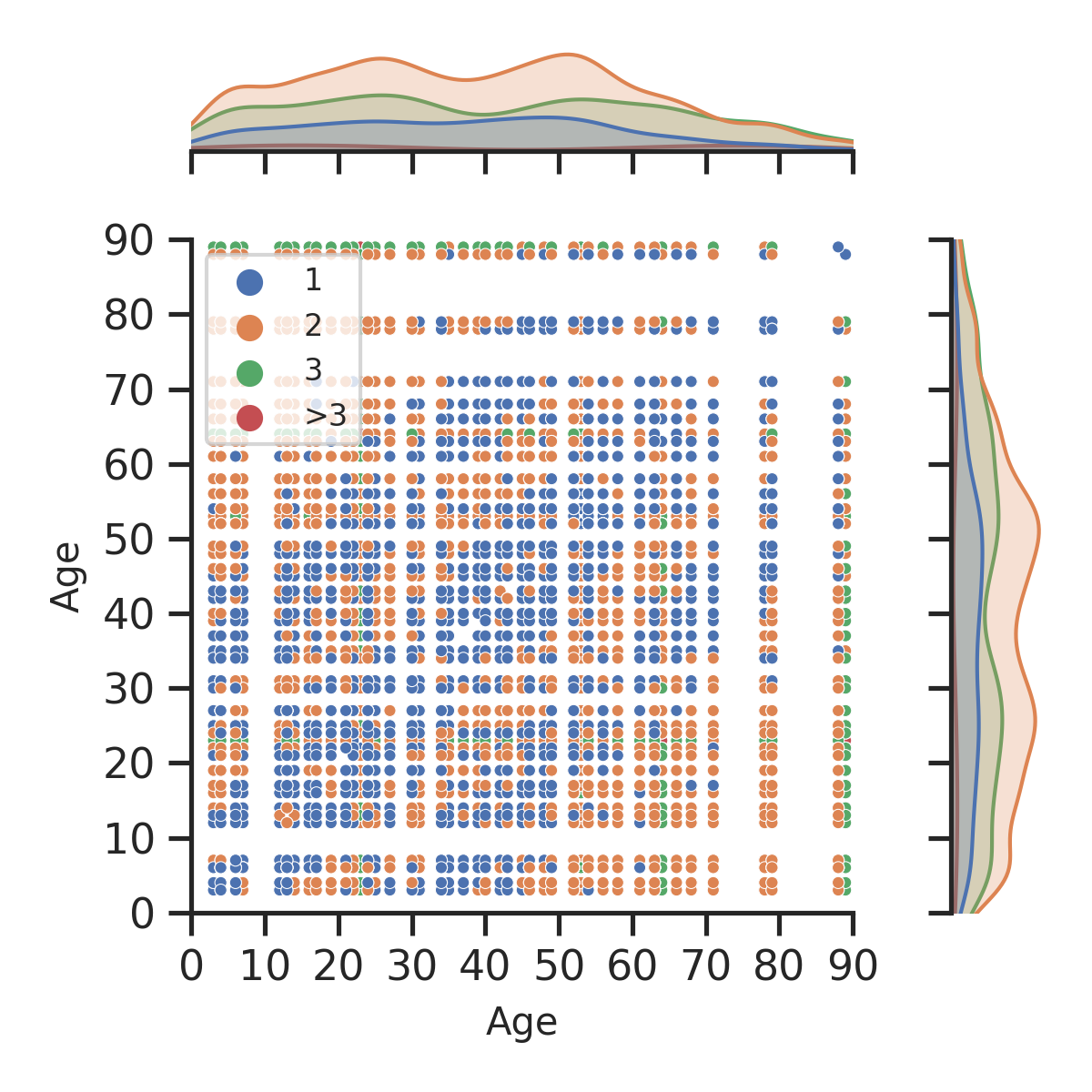}
	\end{minipage}}
	\subfigure[Male-Female]{
		\begin{minipage}[b]{0.22\linewidth}
			\includegraphics[width=1\linewidth]{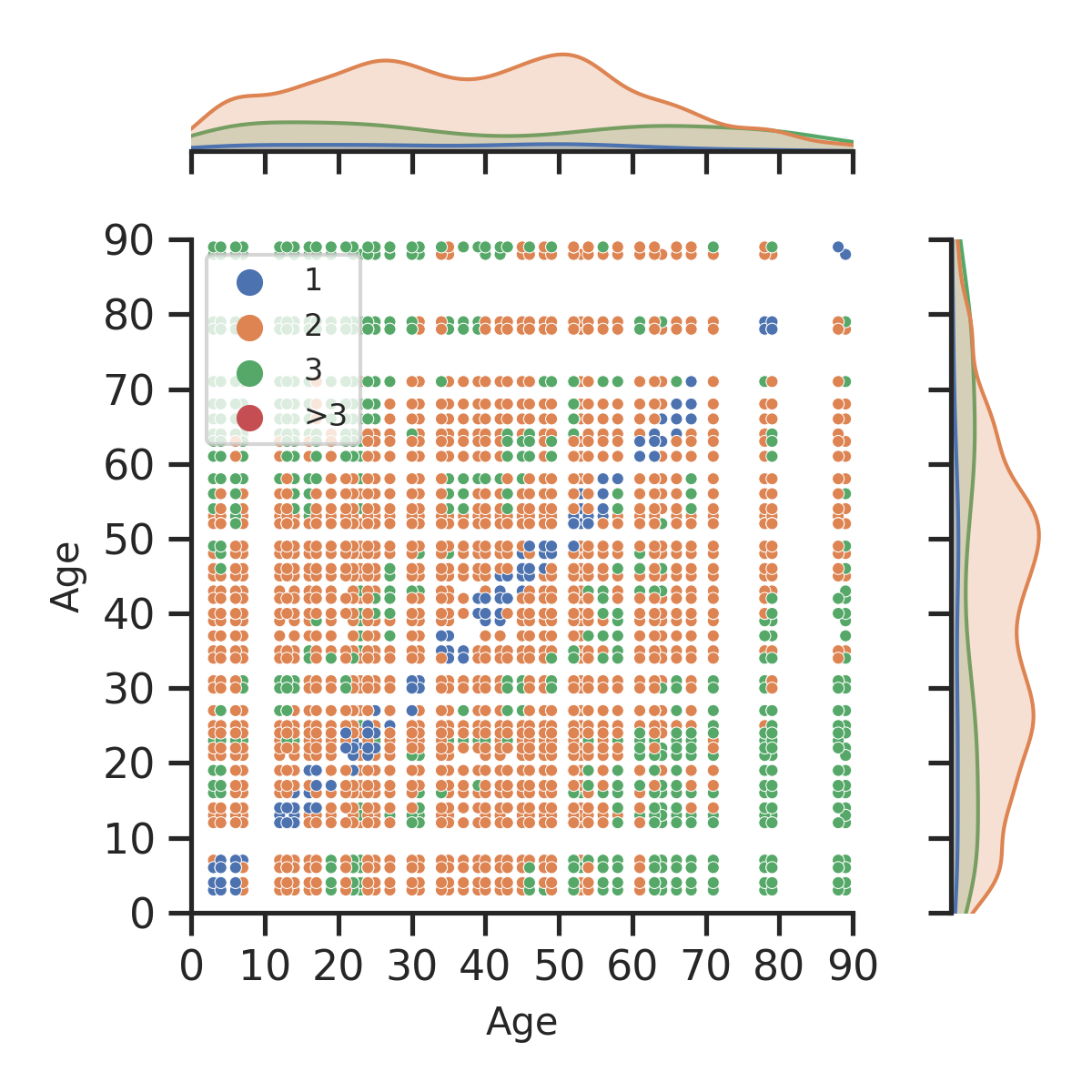}
	\end{minipage}}
	\subfigure[Male-Male]{
		\begin{minipage}[b]{0.22\linewidth}
			\includegraphics[width=1\linewidth]{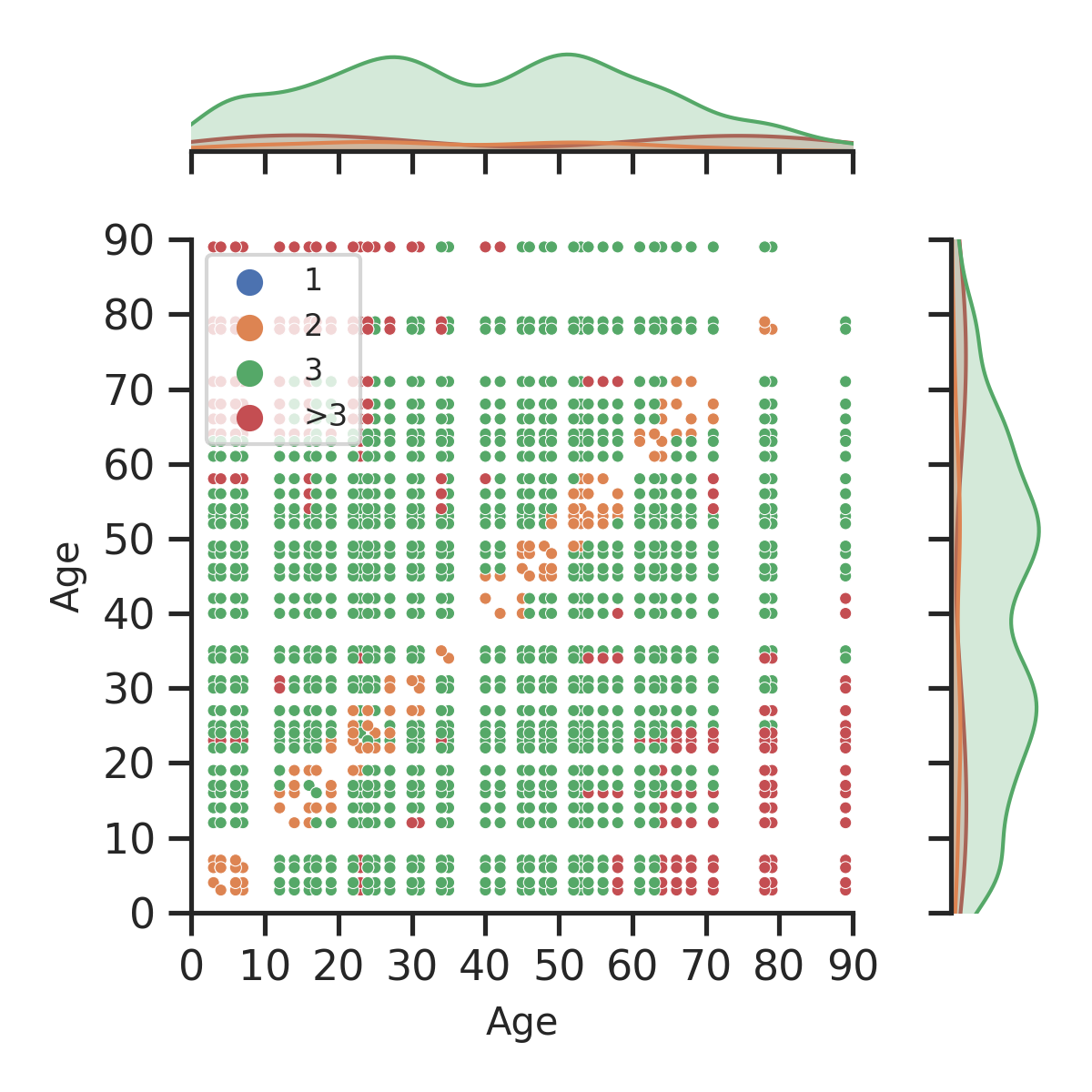}
	\end{minipage}}
 	\subfigure[Female-Female]{
		\begin{minipage}[b]{0.22\linewidth}
			\includegraphics[width=1\linewidth]{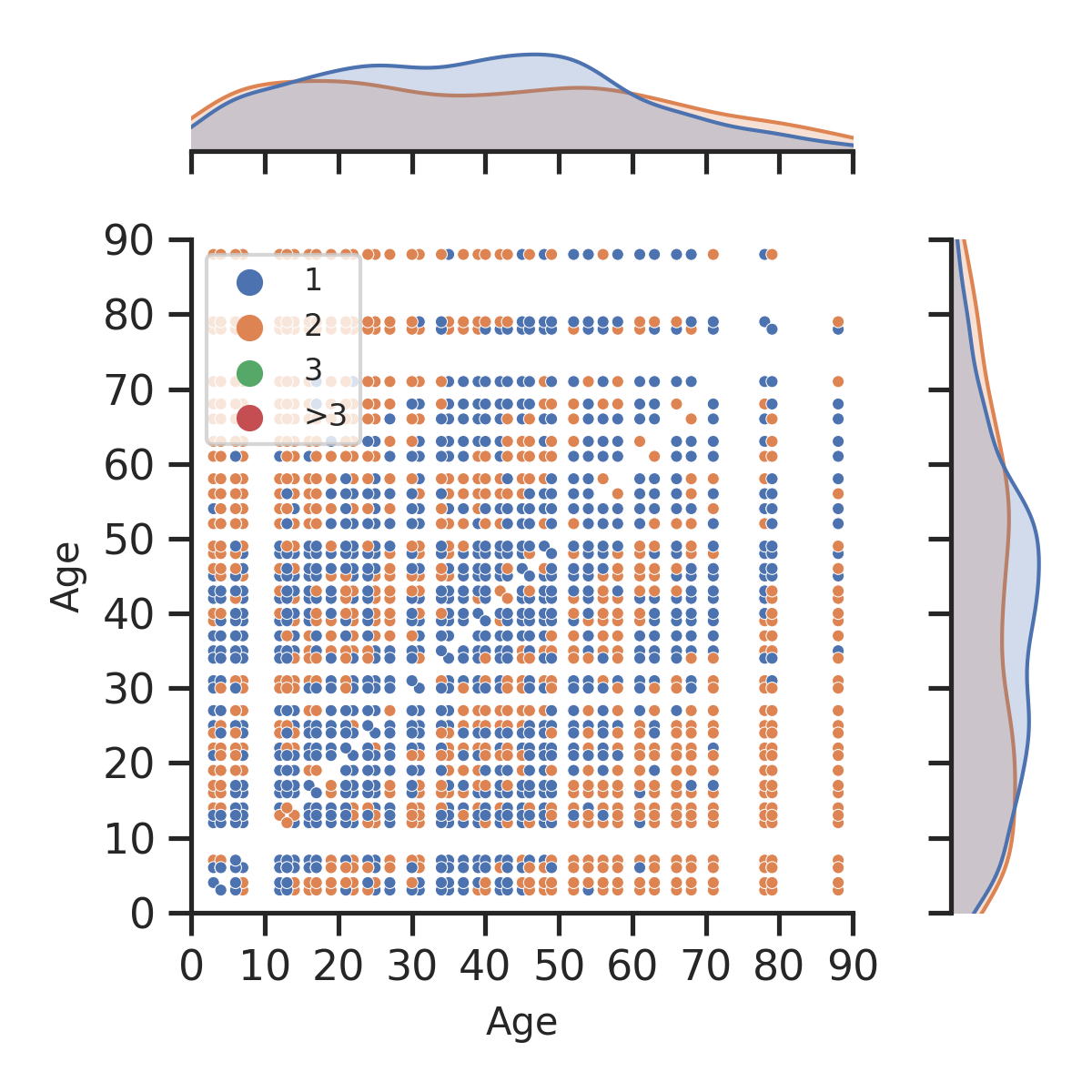}
	\end{minipage}}\\
	\caption{The age and shortest path length distributions in Belgium for contact between (a) all people, (b) females and males, (c) males and males, and (d) females and females.}
 \label{BelgiumPath}
\end{figure}

\begin{figure}[H]
	\centering
	\subfigure[People-People]{
		\begin{minipage}[b]{0.22\linewidth}
			\includegraphics[width=1\linewidth]{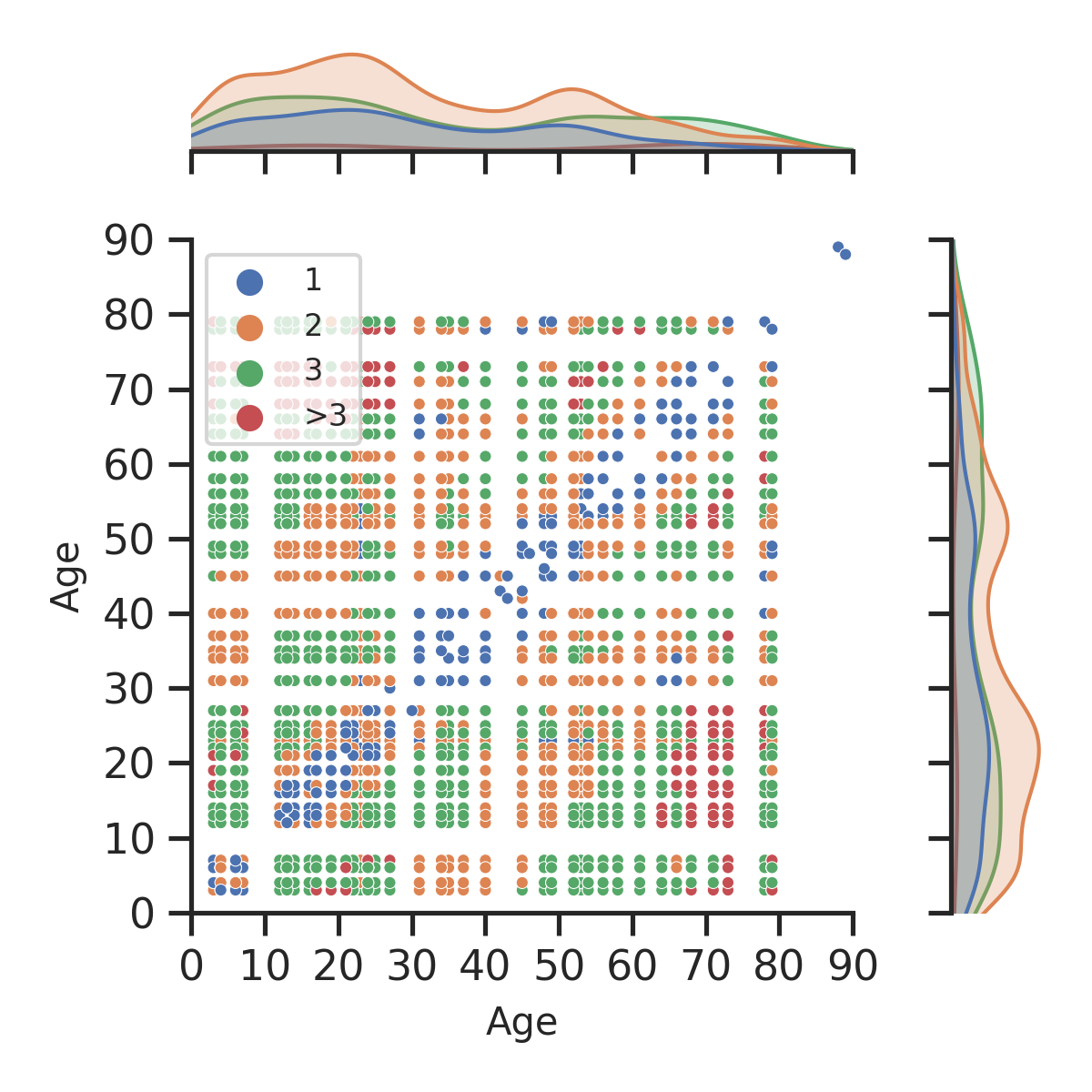}
	\end{minipage}}
	\subfigure[Male-Female]{
		\begin{minipage}[b]{0.22\linewidth}
			\includegraphics[width=1\linewidth]{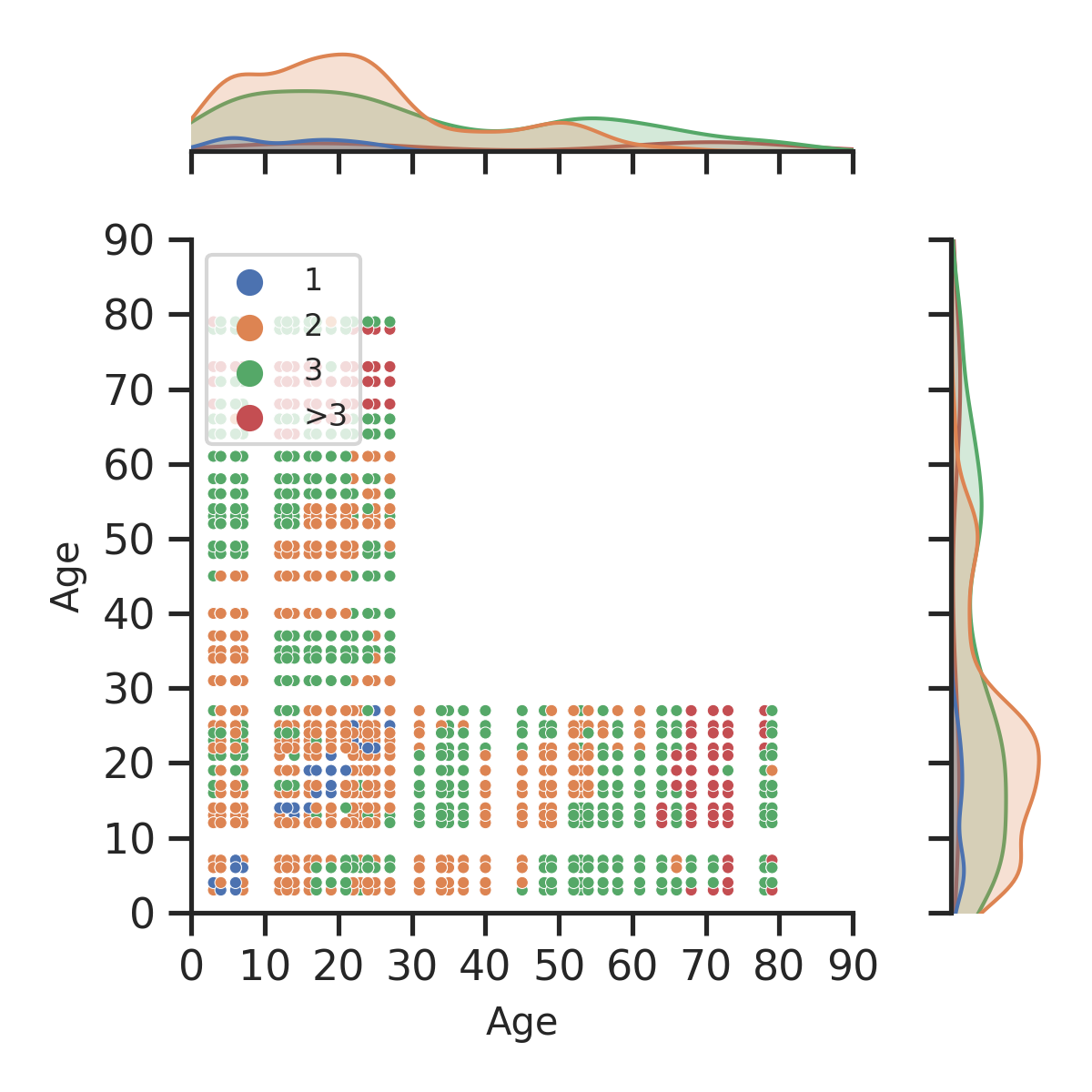}
	\end{minipage}}
	\subfigure[Male-Male]{
		\begin{minipage}[b]{0.22\linewidth}
			\includegraphics[width=1\linewidth]{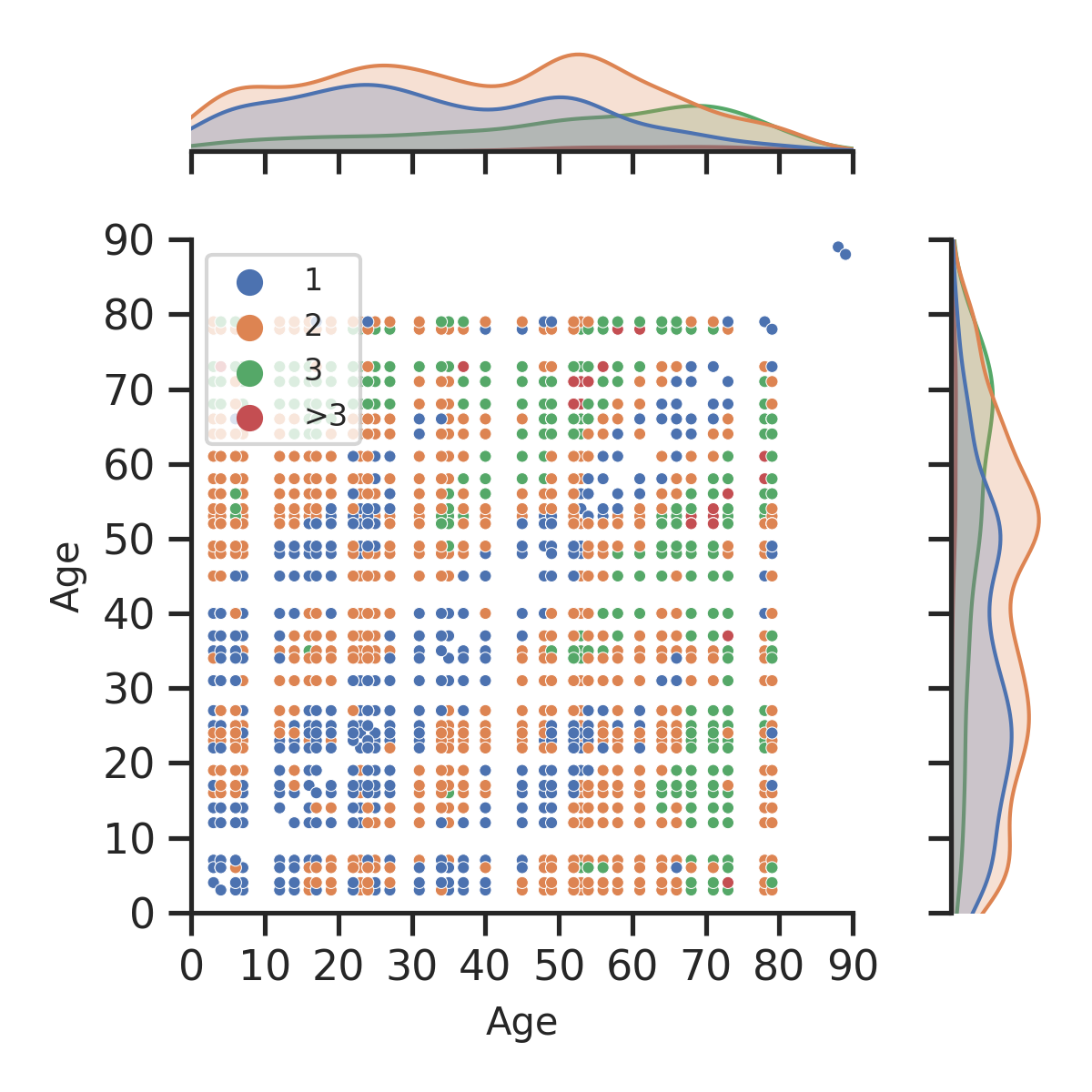}
	\end{minipage}}
 	\subfigure[Female-Female]{
		\begin{minipage}[b]{0.22\linewidth}
			\includegraphics[width=1\linewidth]{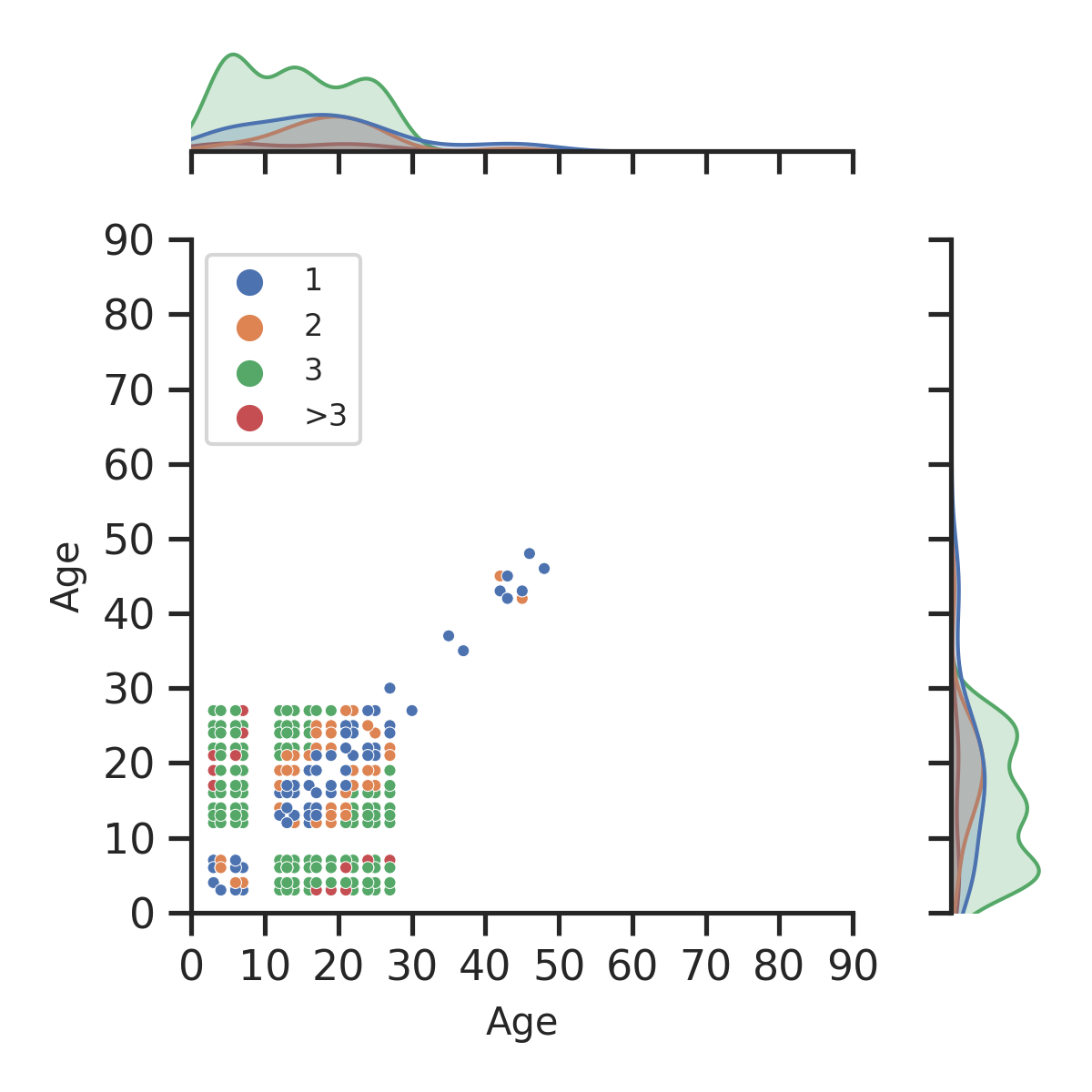}
	\end{minipage}}\\
	\caption{The age and shortest path length distributions in Finland for contact between (a) all people, (b) females and males, (c) males and males, and (d) females and females.}
 \label{FinlandPath}
\end{figure}

\begin{figure}[H]
	\centering
	\subfigure[People-People]{
		\begin{minipage}[b]{0.22\linewidth}
			\includegraphics[width=1\linewidth]{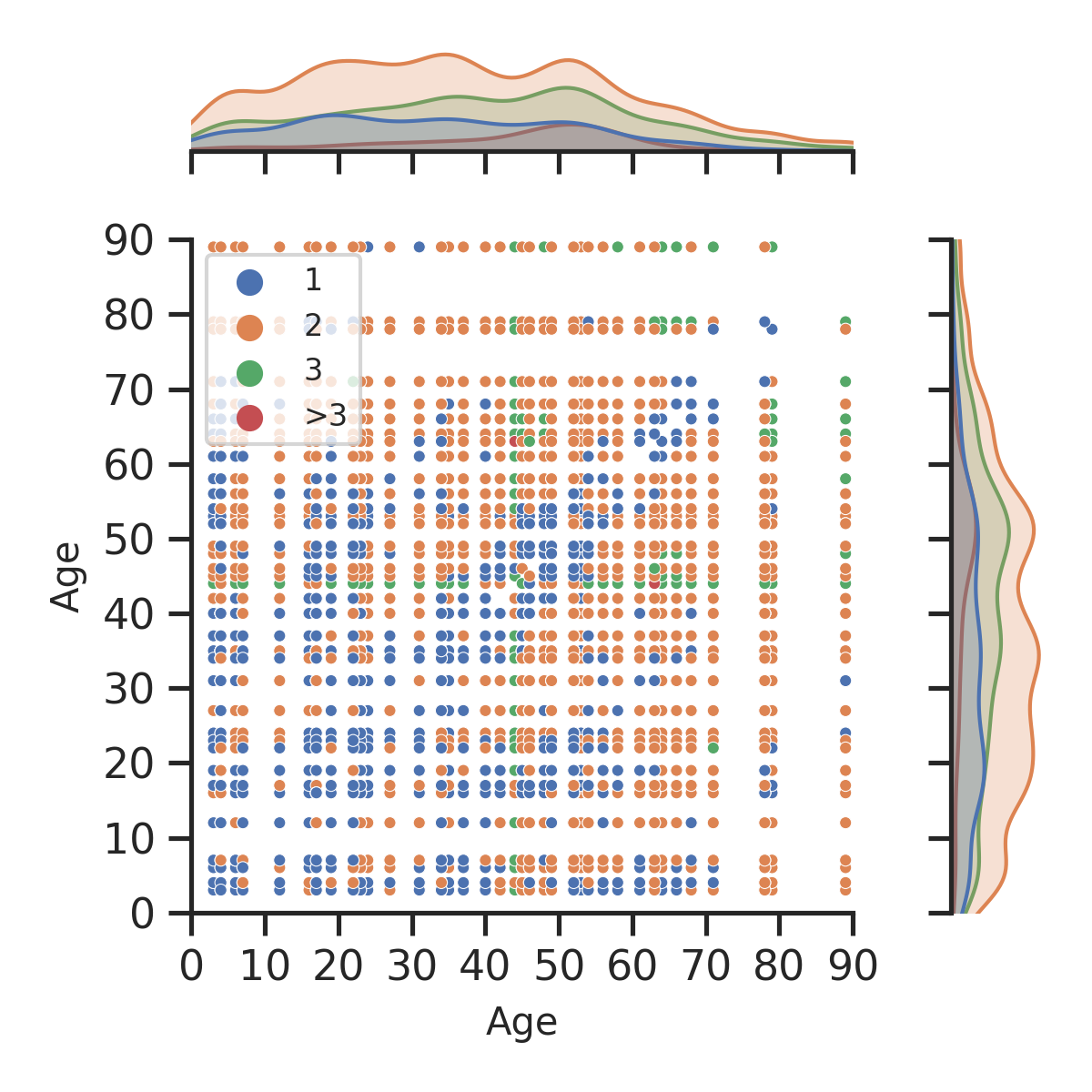}
	\end{minipage}}
	\subfigure[Male-Female]{
		\begin{minipage}[b]{0.22\linewidth}
			\includegraphics[width=1\linewidth]{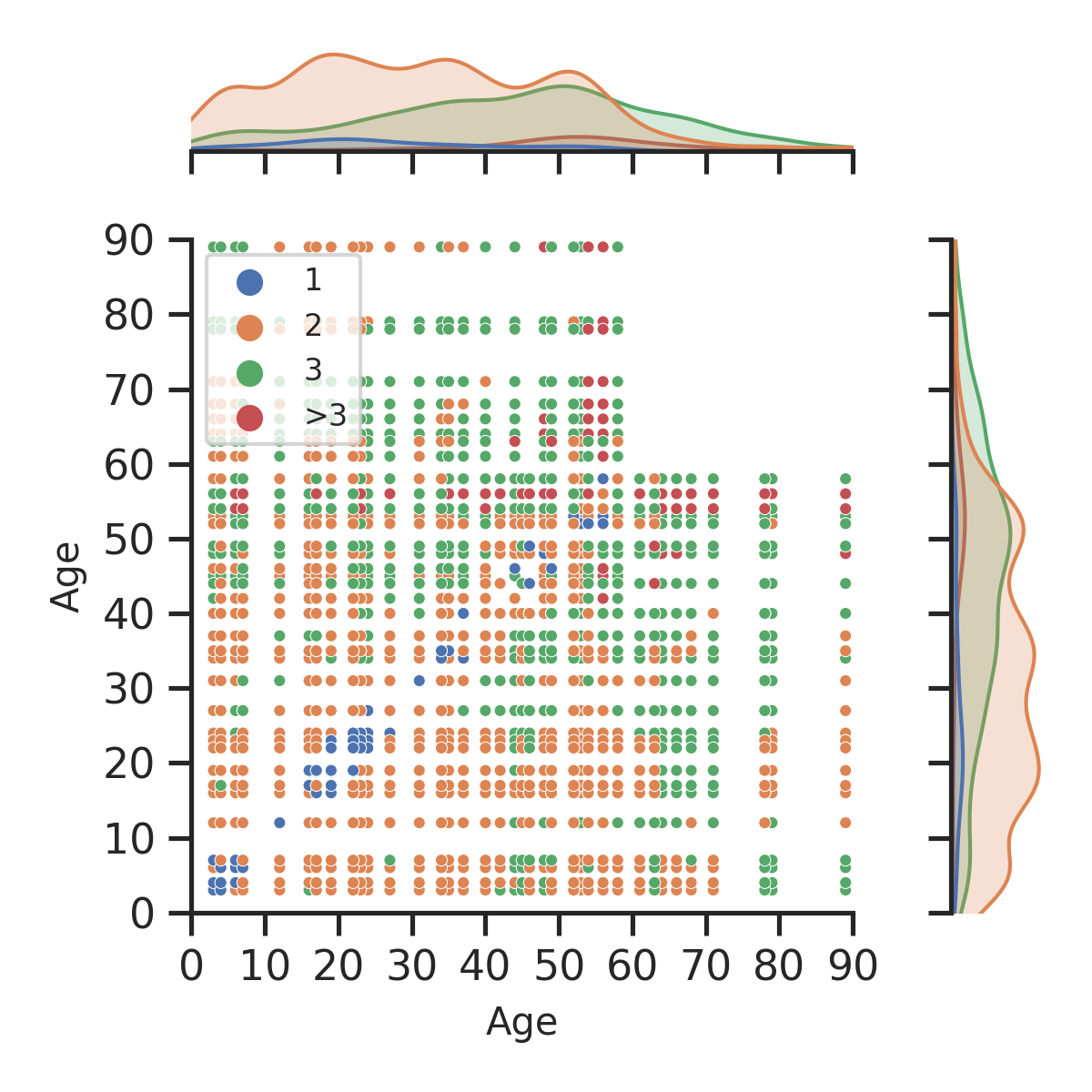}
	\end{minipage}}
	\subfigure[Male-Male]{
		\begin{minipage}[b]{0.22\linewidth}
			\includegraphics[width=1\linewidth]{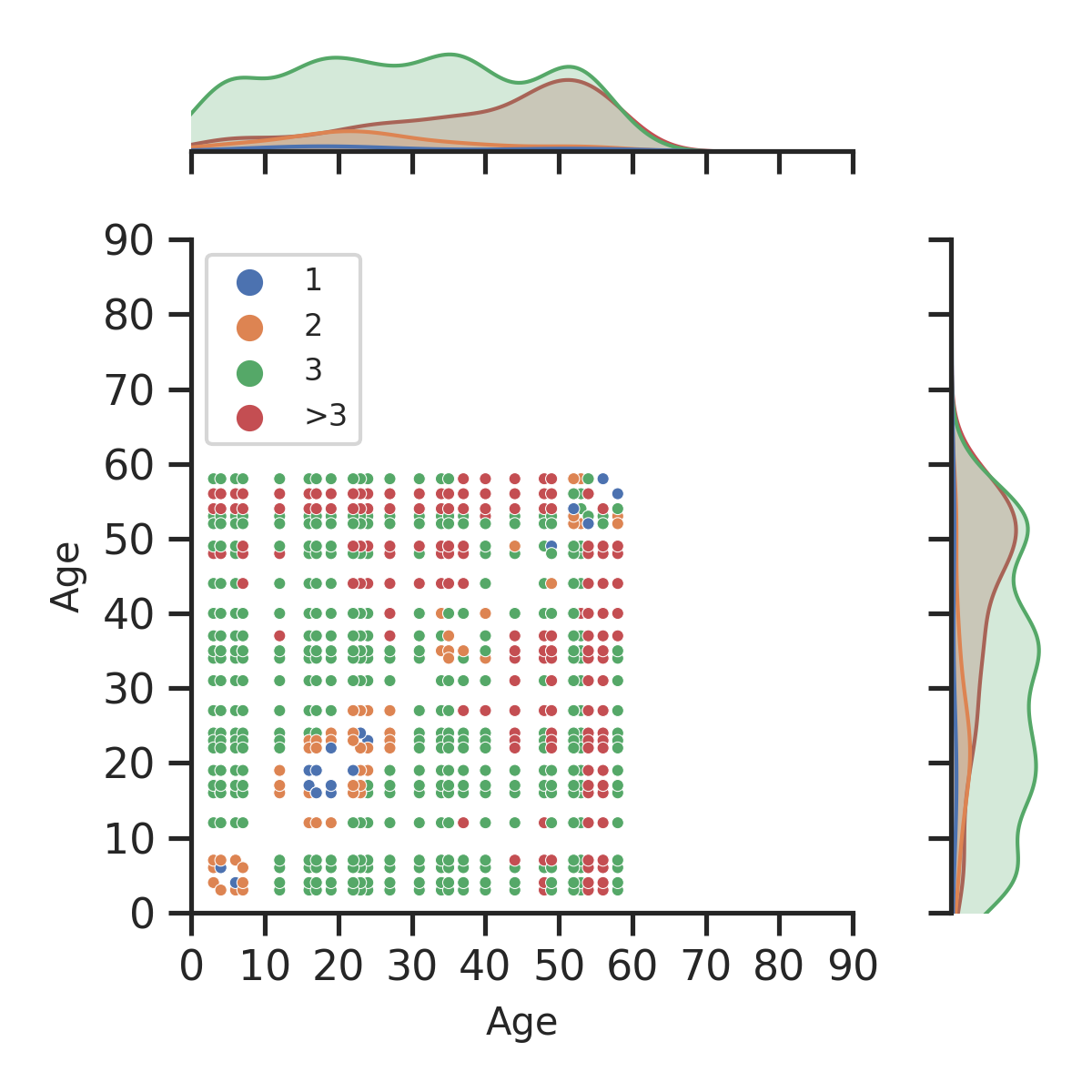}
	\end{minipage}}
 	\subfigure[Female-Female]{
		\begin{minipage}[b]{0.22\linewidth}
			\includegraphics[width=1\linewidth]{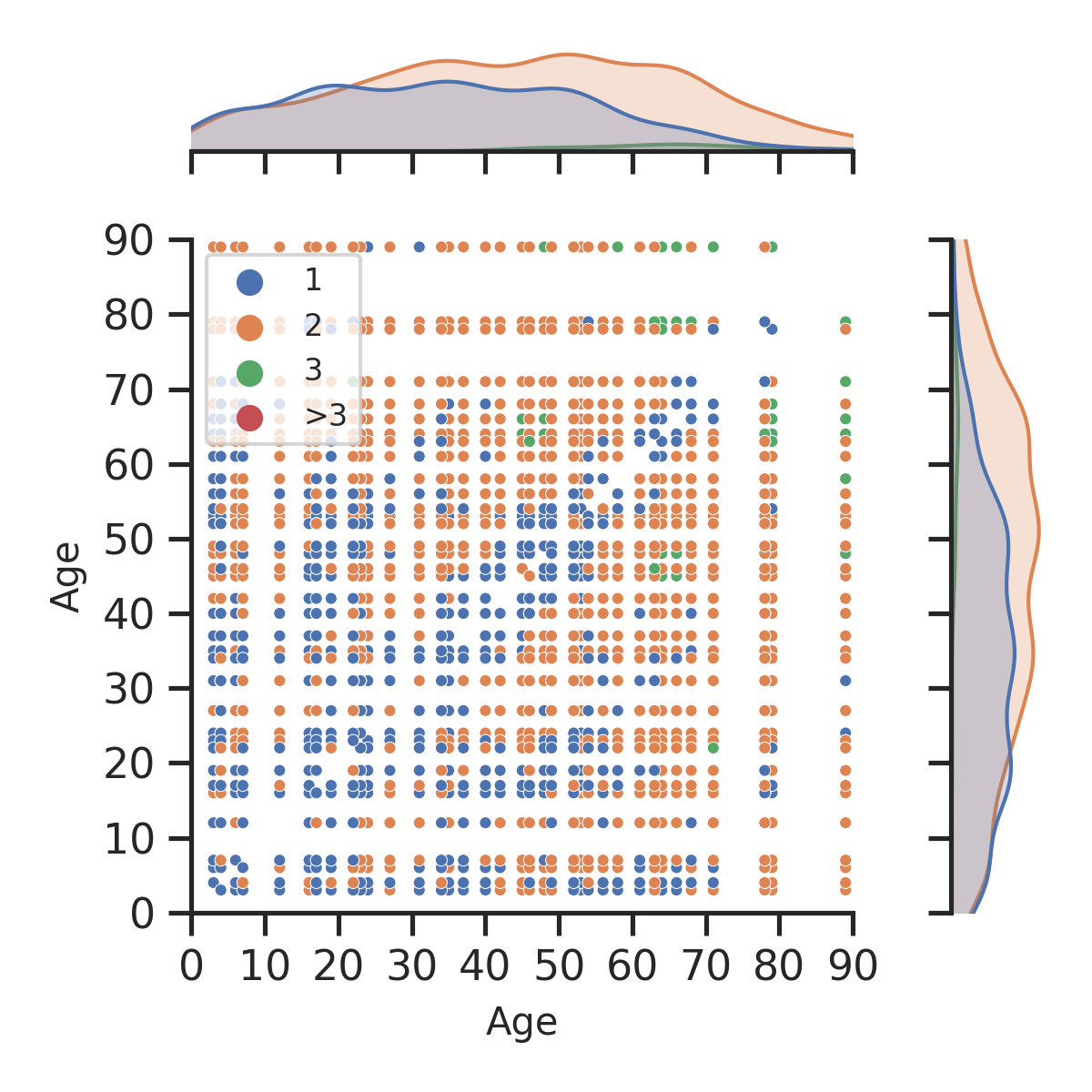}
	\end{minipage}}\\
	\caption{The age and shortest path length distributions in Germany for contact between (a) all people, (b) females and males, (c) males and males, and (d) females and females.}
 \label{GermanyPath}
\end{figure}

\begin{figure}[H]
	\centering
	\subfigure[People-People]{
		\begin{minipage}[b]{0.22\linewidth}
			\includegraphics[width=1\linewidth]{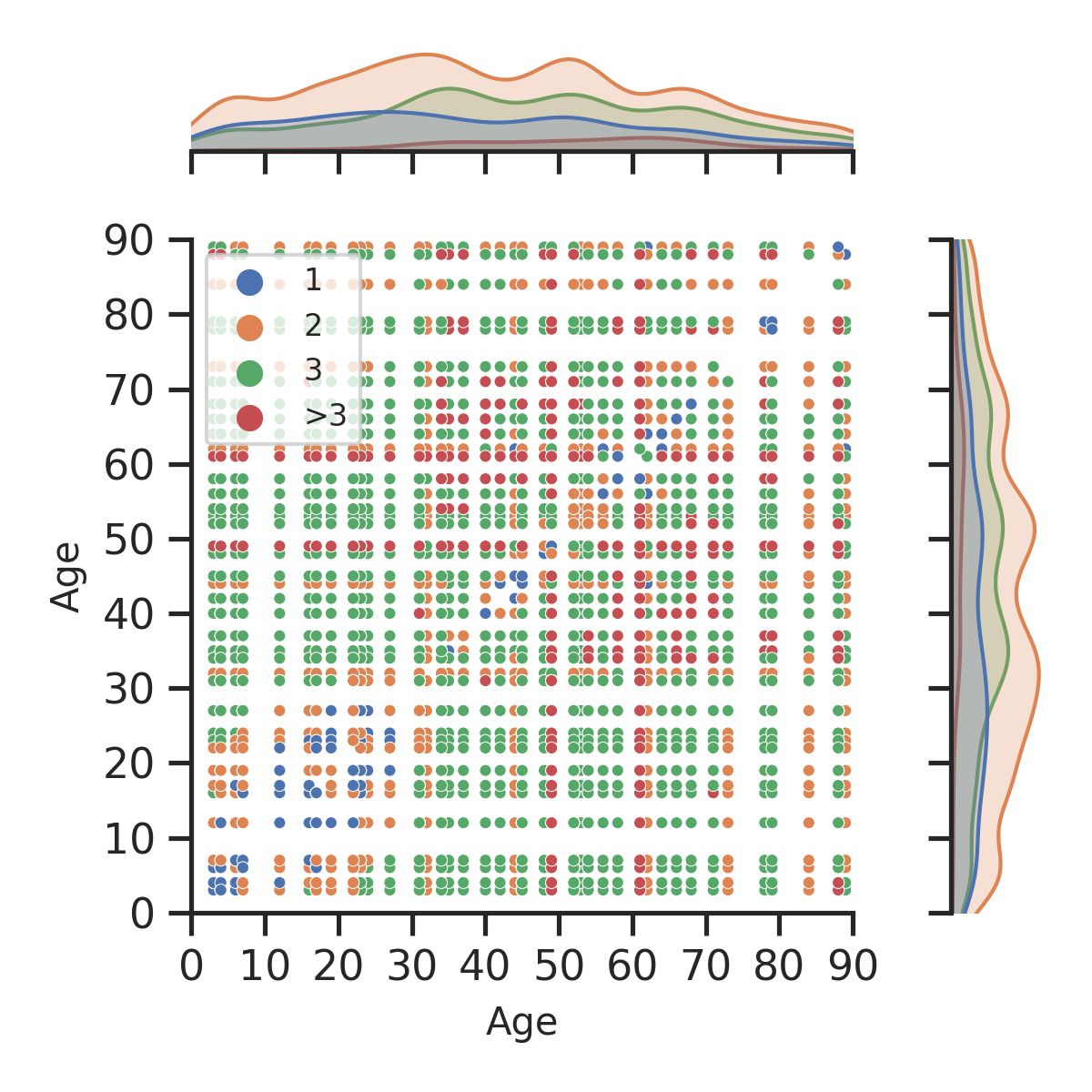}
	\end{minipage}}
	\subfigure[Male-Female]{
		\begin{minipage}[b]{0.22\linewidth}
			\includegraphics[width=1\linewidth]{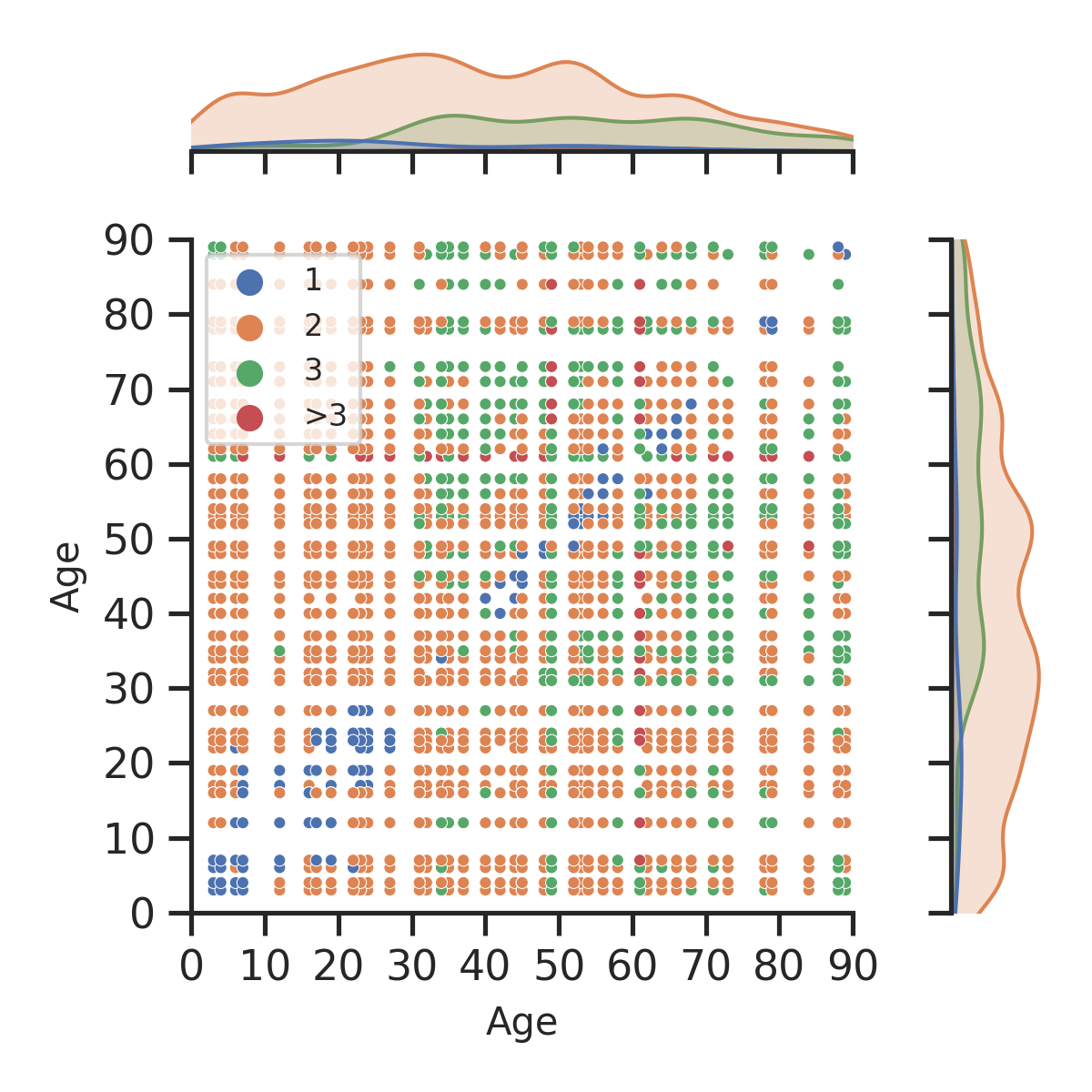}
	\end{minipage}}
	\subfigure[Male-Male]{
		\begin{minipage}[b]{0.22\linewidth}
			\includegraphics[width=1\linewidth]{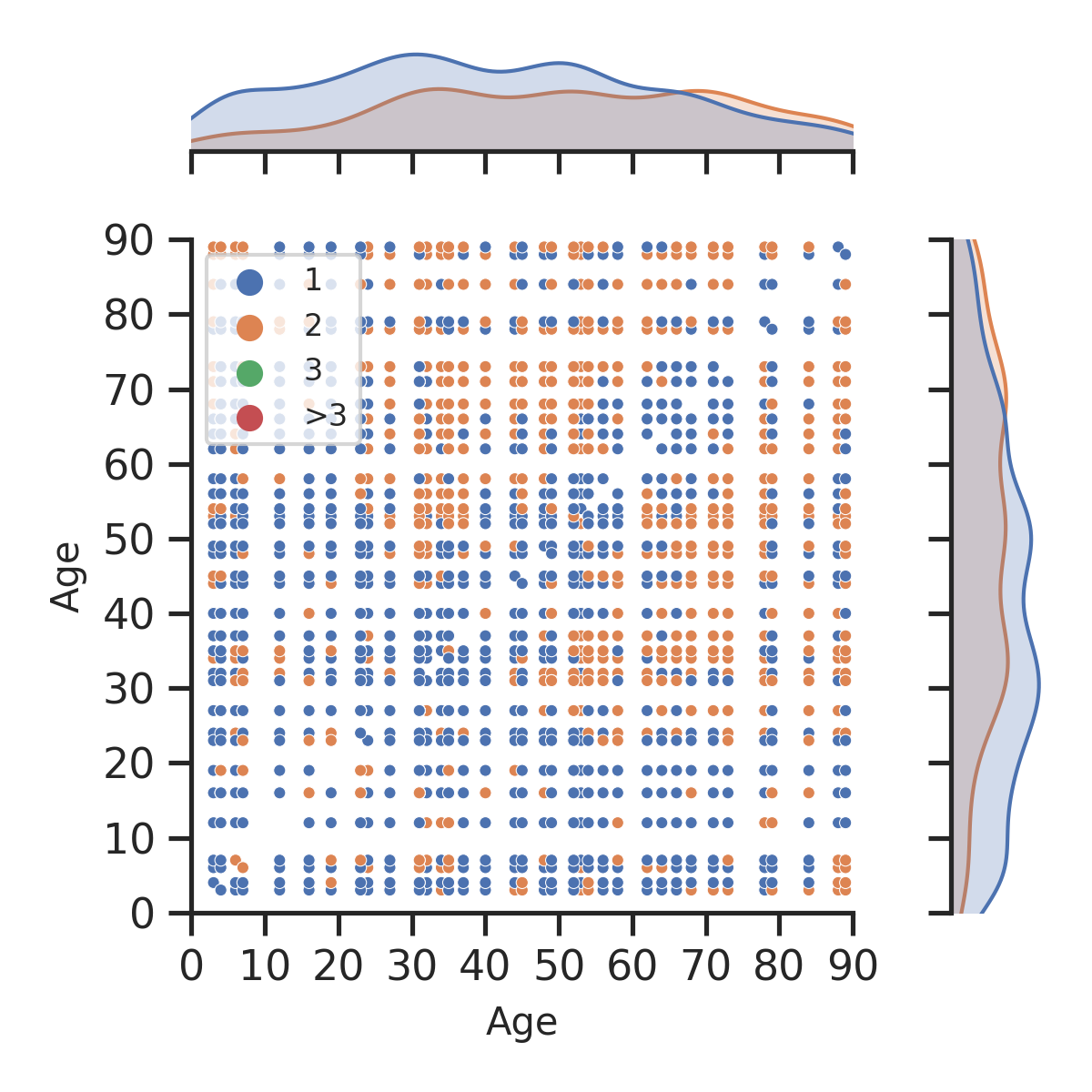}
	\end{minipage}}
 	\subfigure[Female-Female]{
		\begin{minipage}[b]{0.22\linewidth}
			\includegraphics[width=1\linewidth]{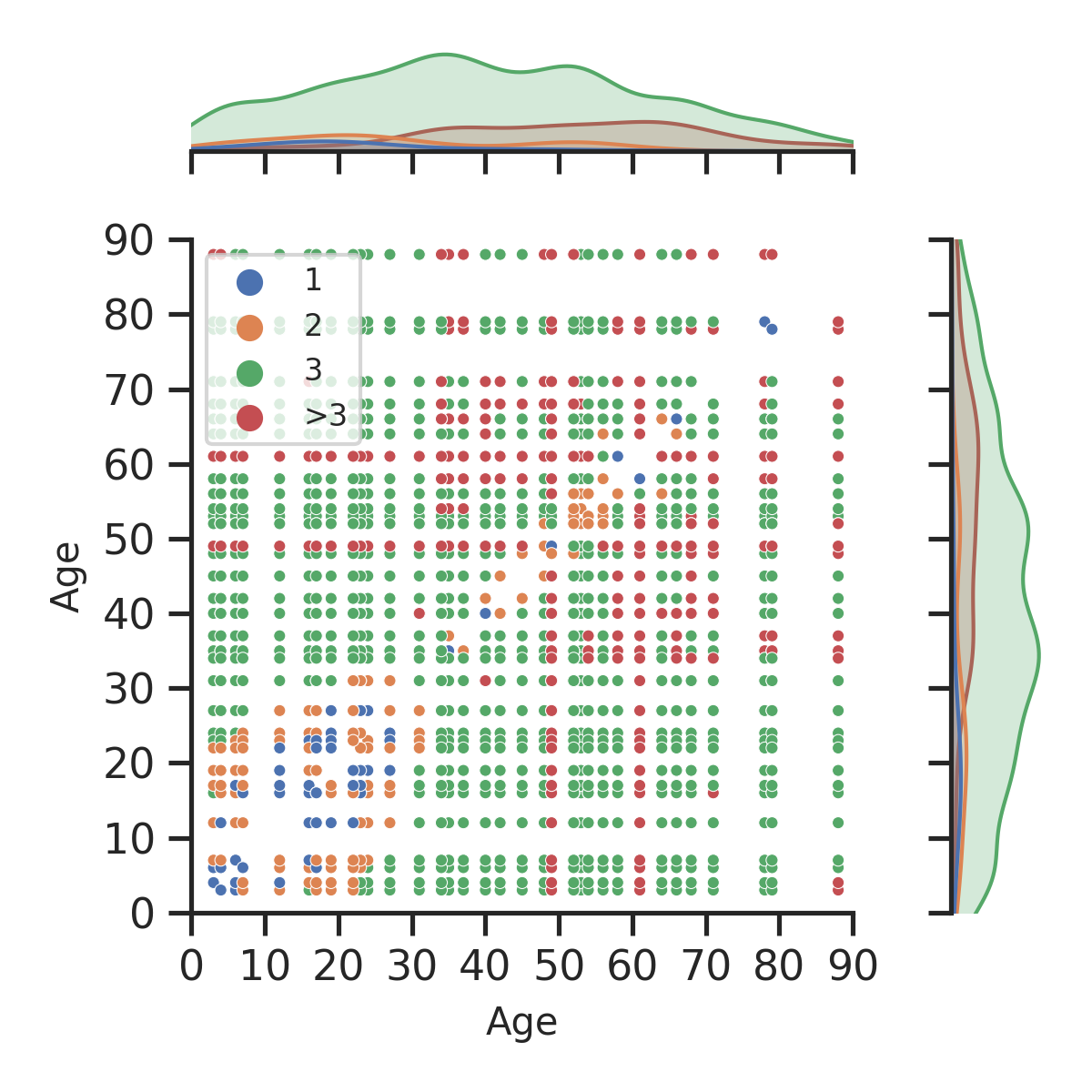}
	\end{minipage}}\\
	\caption{The age and shortest path length distributions in Italy for contact between (a) all people, (b) females and males, (c) males and males, and (d) females and females.}
 \label{ItalyPath}
\end{figure}

\begin{figure}[H]
	\centering
	\subfigure[People-People]{
		\begin{minipage}[b]{0.22\linewidth}
			\includegraphics[width=1\linewidth]{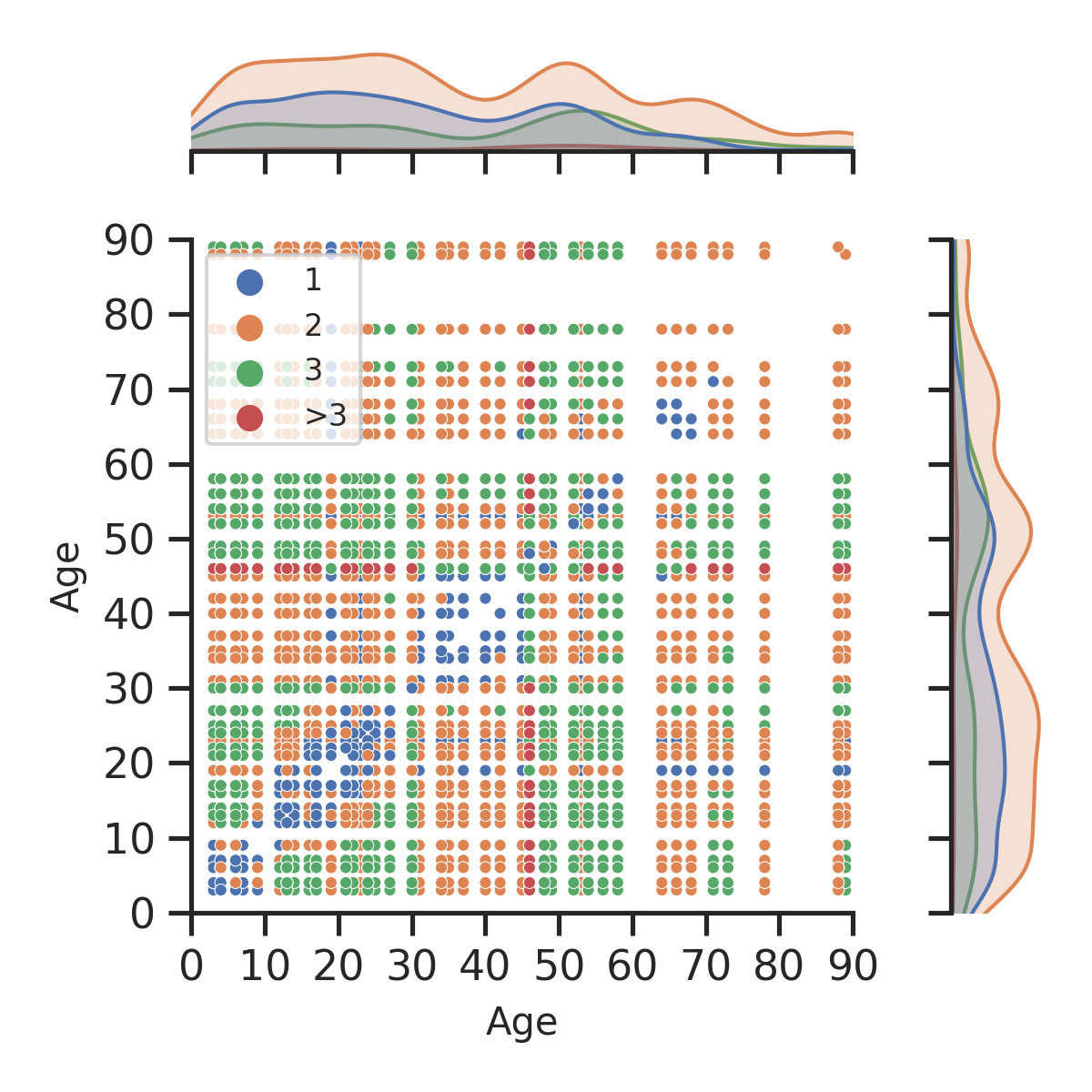}
	\end{minipage}}
	\subfigure[Male-Female]{
		\begin{minipage}[b]{0.22\linewidth}
			\includegraphics[width=1\linewidth]{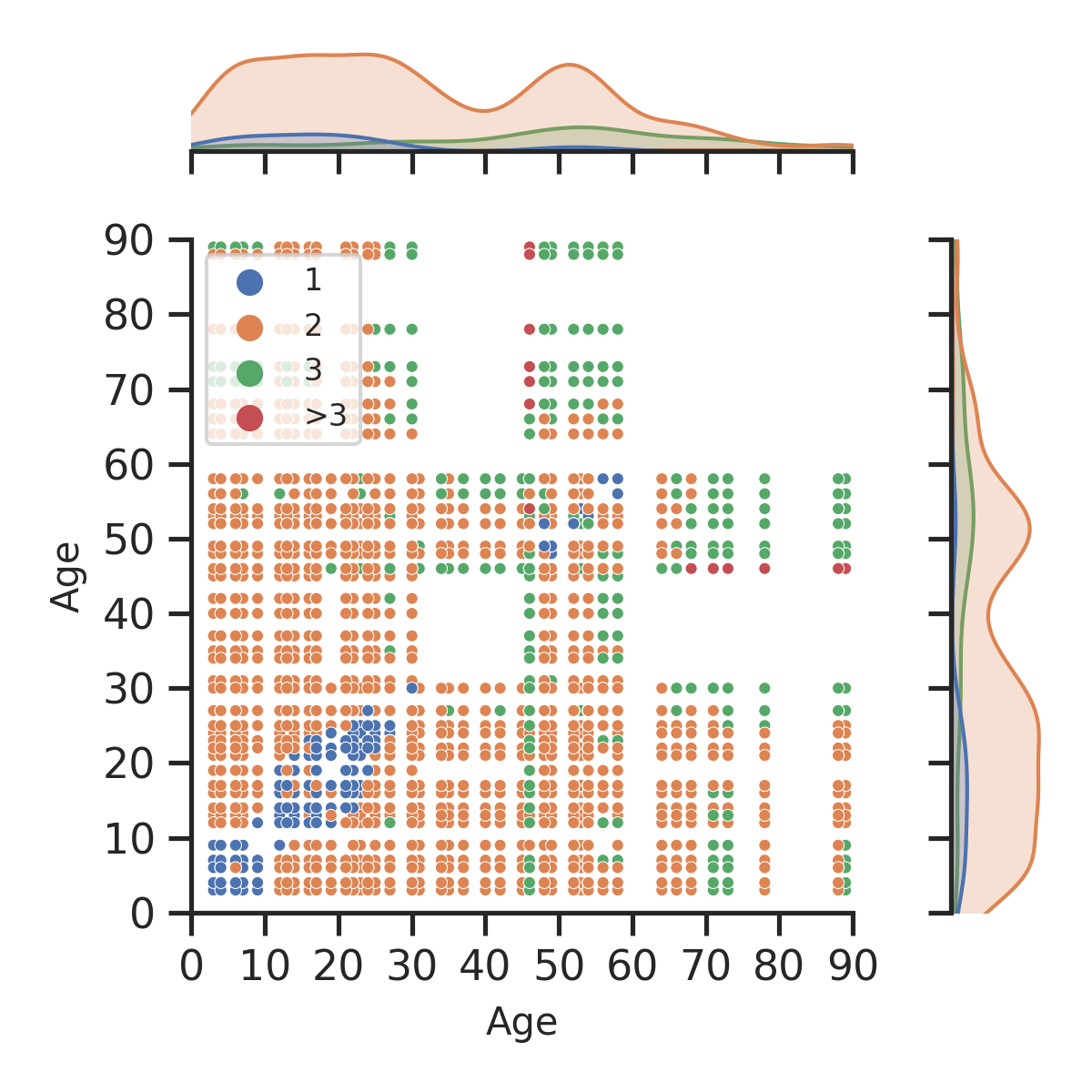}
	\end{minipage}}
	\subfigure[Male-Male]{
		\begin{minipage}[b]{0.22\linewidth}
			\includegraphics[width=1\linewidth]{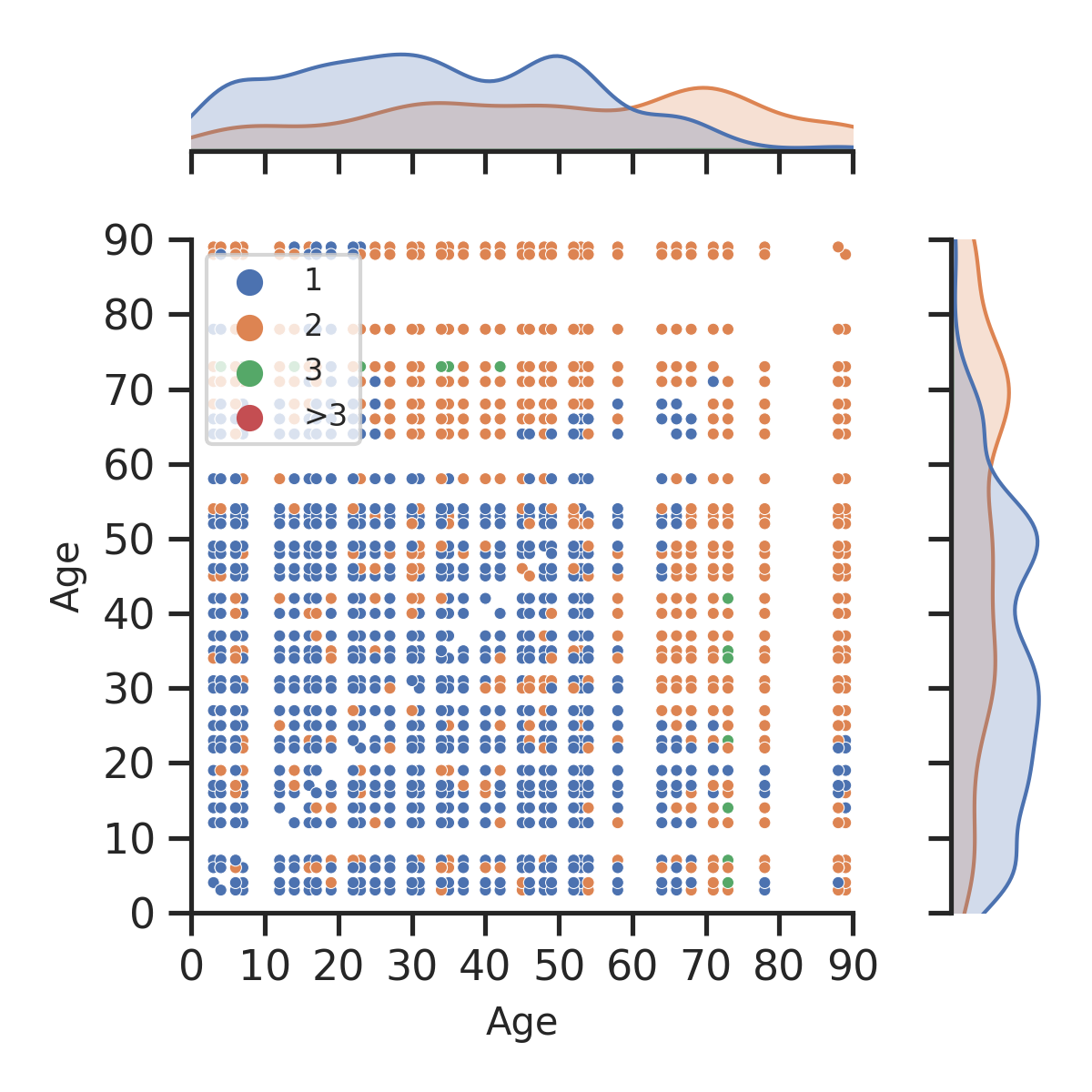}
	\end{minipage}}
 	\subfigure[Female-Female]{
		\begin{minipage}[b]{0.22\linewidth}
			\includegraphics[width=1\linewidth]{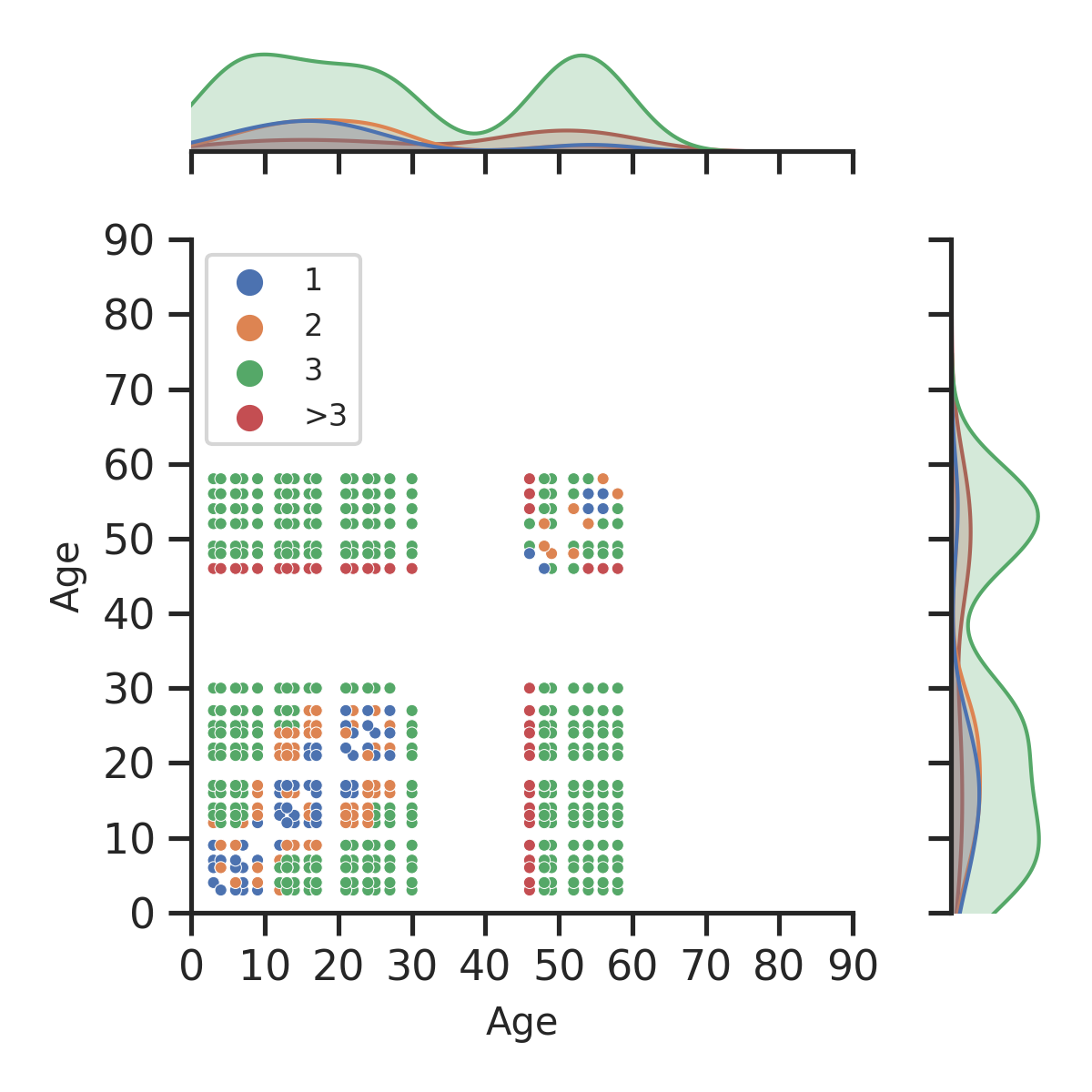}
	\end{minipage}}\\
	\caption{The age and shortest path length distributions in Luxembourg for contact between (a) all people, (b) females and males, (c) males and males, and (d) females and females.}
 \label{LuxembourgPath}
\end{figure}

\begin{figure}[H]
	\centering
	\subfigure[People-People]{
		\begin{minipage}[b]{0.22\linewidth}
			\includegraphics[width=1\linewidth]{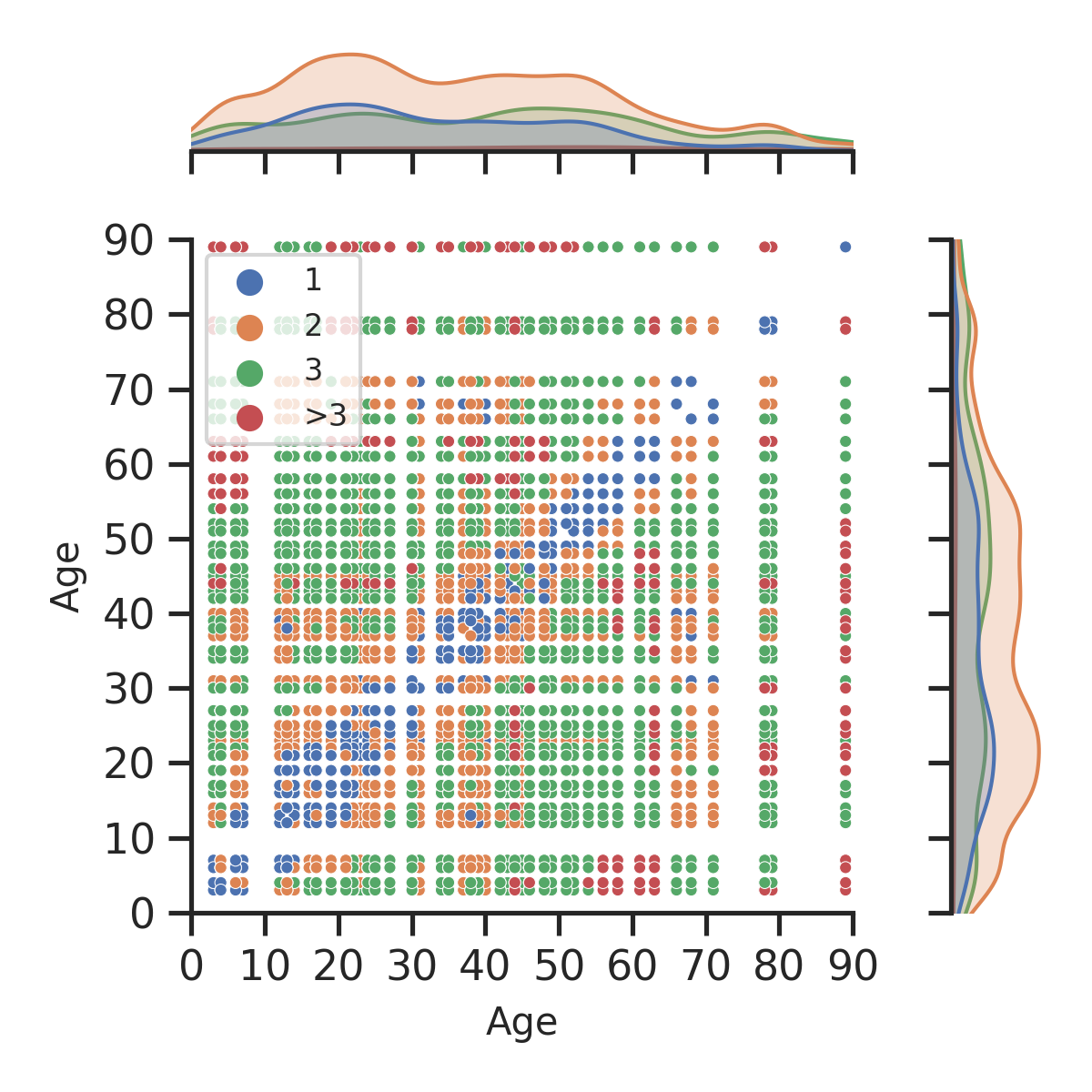}
	\end{minipage}}
	\subfigure[Male-Female]{
		\begin{minipage}[b]{0.22\linewidth}
			\includegraphics[width=1\linewidth]{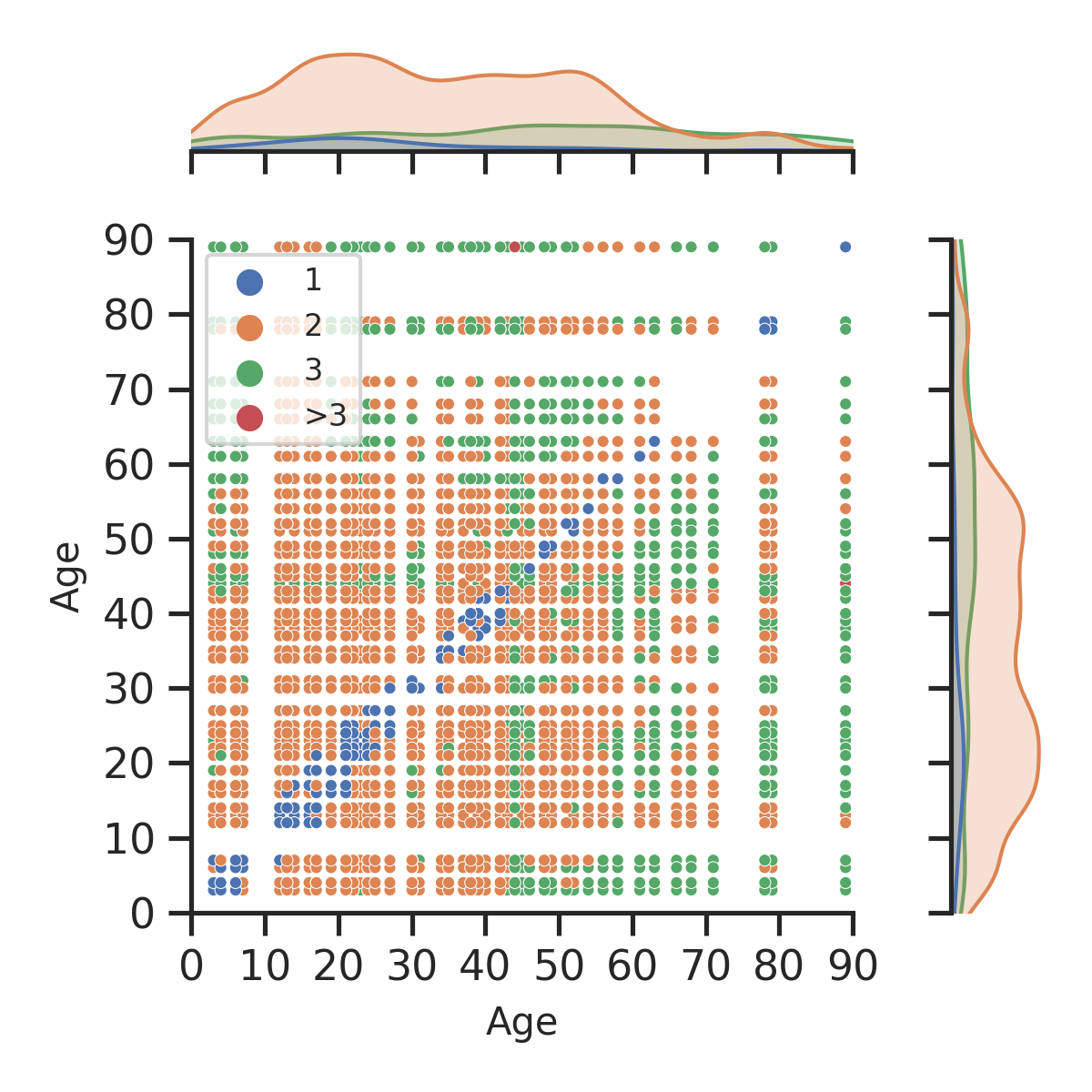}
	\end{minipage}}
	\subfigure[Male-Male]{
		\begin{minipage}[b]{0.22\linewidth}
			\includegraphics[width=1\linewidth]{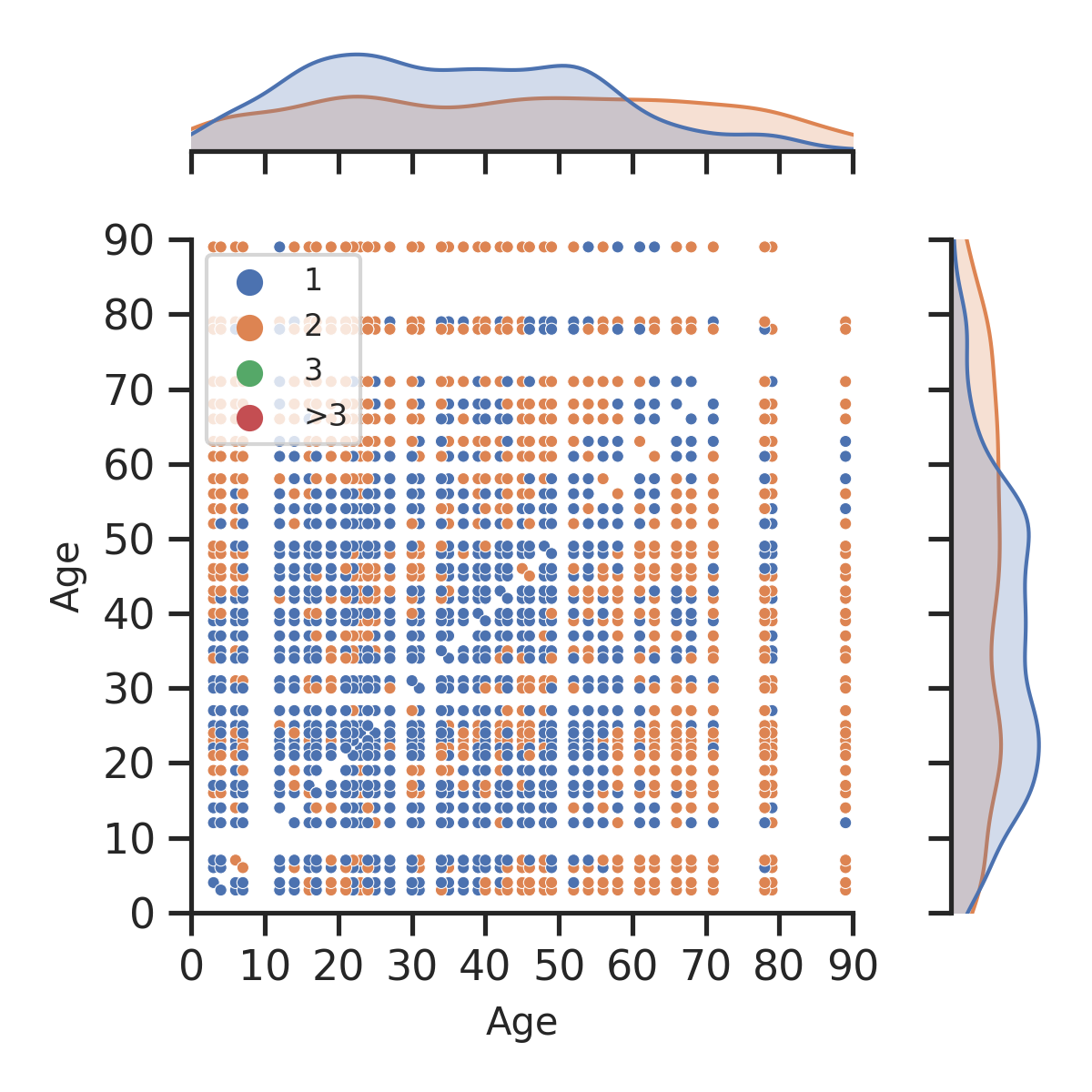}
	\end{minipage}}
 	\subfigure[Female-Female]{
		\begin{minipage}[b]{0.22\linewidth}
			\includegraphics[width=1\linewidth]{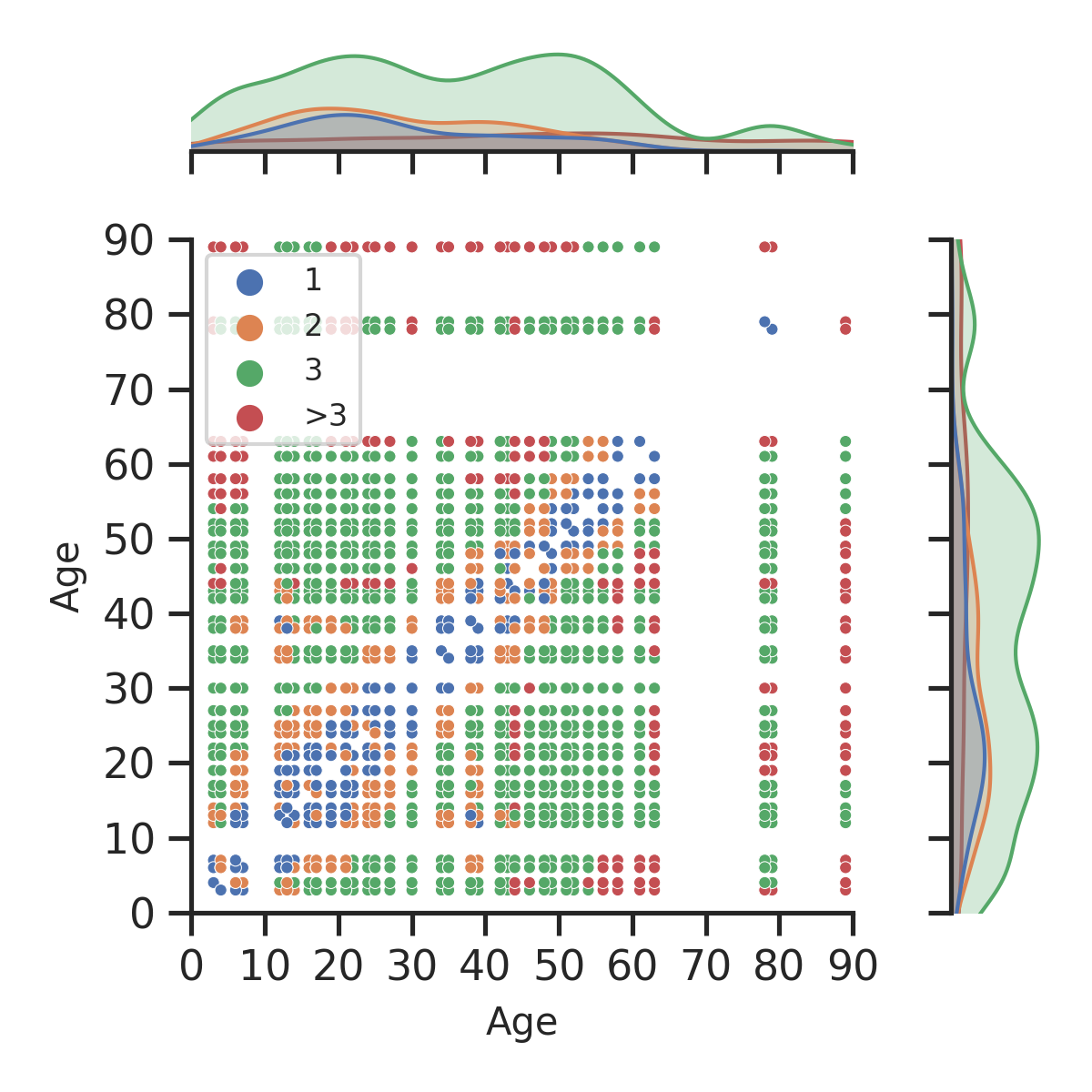}
	\end{minipage}}\\
	\caption{The age and shortest path length distributions in Poland for contact between (a) all people, (b) females and males, (c) males and males, and (d) females and females.}
 \label{PolandPath}
\end{figure}

We explore the distributions of shortest path lengths between males and males, females and females, males and females, and all people based on the network simulations of each country, including Belgium (See Fig.~\ref{BelgiumPath}), Finland (See Fig.~\ref{FinlandPath}), Germany (See  Fig.~\ref{GermanyPath}), Italy (See Fig.~\ref{ItalyPath}), Luxembourg (See Fig.~\ref{LuxembourgPath}) and Poland (See Fig.~\ref{PolandPath}). We identify different lengths of the shortest path with different colours (blue for $1$, orange for $2$, green for $3$ and red for over $3$; the fake (non-existing) paths are left vacant for clarity). We find that nodes with those unpopular features do not have a path between each other. For example, old males in Germany do not have direct or indirect connections to each other because they have a positive preference for females and dissimilar sex features (See Fig.~\ref{GermanyPath} and Fig.~\ref{GermanyFeatPref} (c) and (d)). This contrasts with the cases of Luxembourg and Finland, where old females over the age of $60$ nearly do not have any direct or indirect interactions with others due to people's strong positive preference for young males and negative preferences for older females (See Fig.~\ref{FinlandFeatPref} (c) and (d); See Fig.~\ref{LuxembourgFeatPref} (c) and (d)). We also find that most of the direct interactions take place between people who have preferred sex features and preferred age differences. As an example, in Belgium and Germany, a significant number of direct interactions take place between females of similar ages or having an age difference of around $30$ (See Fig.~\ref{BelgiumFeatPref} (c) and Fig.~\ref{GermanyFeatPref} (c) for preference information). This is different from the cases of Finland, Italy, Luxembourg and Poland, where most direct interactions take place between males around similar ages or having an age difference around $30$ (See Fig.~\ref{FinlandFeatPref} (c) and (d), Fig.~\ref{ItalyFeatPref} (c) and (d), Fig.~\ref{LuxembourgFeatPref} (c) and (d), and Fig.~\ref{PolandFeatPref} (c) and (d)). In addition, we also find a limited number of direct interactions between the nodes with young and similar ages. For example, in Finland, Belgium, Germany, Italy, Luxembourg and Poland, we can find a few direct interactions between females and males at a similar young age despite their negative preference for different sex features. Moreover, in Belgium, there are more interactions between females and males over the age range of $[0-90]$ due to their preferences for different sex features and similar ages.

There is a significant number of indirect interactions, as indicated by a shortest path length of $2$, distributed around the above-mentioned direct interactions. This takes place between the nodes and their neighbour's neighbours. 
For example, in Poland, males at the age of $50$ prefer to be connected with males at the age of $20$, and in the meantime, males around the age of $20$ may be connected with people at the age around $[10-20]$ due to similar age differences. These two direct interactions lead to an indirect shortest path between the age around $50$ and the age around $10$ (See Fig.~\ref{PolandFeatPref} (b)). In all the countries, we can find a significant number of indirect interactions (with the shortest path length of $2$) between male and female people and a longer length of shortest path between people with unpreferred sex features (e.g. the female-female contact in Fig.~\ref{LuxembourgFeatPref} (d)). 

\subsection{Disaster Resilience}
\label{rep1-2section34}
We investigate the disaster resilience in by introducing a simulation-based epidemic spreading process in the recreated social networks. We assume that (i) the disease transmissibility given exposure is $0.80$, without change over time, (ii) the epidemic spreads with a single seed selection in the initial stage based on the highest node degree, (iii) the epidemic spread on the network propagates one step (edge) away for each time step and (iv) the susceptible nodes get infected without recovery (SI process). The transmissibility at $0.8$ is a high transmissibility which speeds up the epidemic spread and leads to a significant number of infections within $6$ time steps. Under this transmissibility, the different network patterns of each country lead to heterogeneous infection patterns.
\subsubsection{Seed Selection}

The epidemic spreading process starts from the most popular node in each country (with the highest node degree), leading to the biggest number of infections within a limited time and space. Tab.~\ref{startnode} shows the features of the most popular node in each country's social network. The most popular node is a male age around $20$ in Finland, Italy, Luxembourg and Poland.
In contrast, the most popular node is a female age $19$ in Germany, and  a 48-year-old female is the most popular node in Belgium's social networks. These differences result from different social networks in each country and lead to various spreads. More specifically, the differences in the sex features of the most popular node result from people's preferences. People in Belgium and Germany have strong positive preference for females (See Fig.~\ref{BelgiumFeatPref} (c) and Fig.~\ref{GermanyFeatPref} (c)), which contrasts the cases of Finland, Italy, Luxembourg and Poland (See Fig.~\ref{FinlandFeatPref} (c), Fig.~\ref{ItalyFeatPref} (c), Fig.~\ref{LuxembourgFeatPref} (c), Fig.~\ref{LuxembourgFeatPref} (c) and Fig.~\ref{PolandFeatPref} (c)). In addition, people in all the countries, except for Belgium, have a strong positive preference for young ages (or a significant negative preference for old ages), which makes young nodes popular. In Belgium, people have a greater interest and positive preference for similar ages (See Fig.~\ref{BelgiumFeatPref} (c)), and this results in a significant number of connections between nodes in a dense age group such as $[40-49]$. 

\begin{table}[h]
\centering
\small
\caption{The starting node of the epidemic spreading process in each country.}
\label{startnode}
\setlength{\tabcolsep}{3pt}
\renewcommand{\arraystretch}{1.5}
\begin{tabular}{|c|c|c|c|c|c|c|}
\hline
Country &Belgium &Finland&Germany&Italy&Luxembourg&Poland \\
\hline
Age of the starting node& 48& 19 & 19&19&22&12\\
\hline
Sex of the starting node&Female&Male&  Female&Male& Male& Male \\
\hline
\end{tabular}
\end{table}

\subsubsection{Infection Occurrence}

\begin{figure}[H] 
	\centering
	\subfigure[]{
		\begin{minipage}[b]{0.45\linewidth}
			\includegraphics[width=1\linewidth]{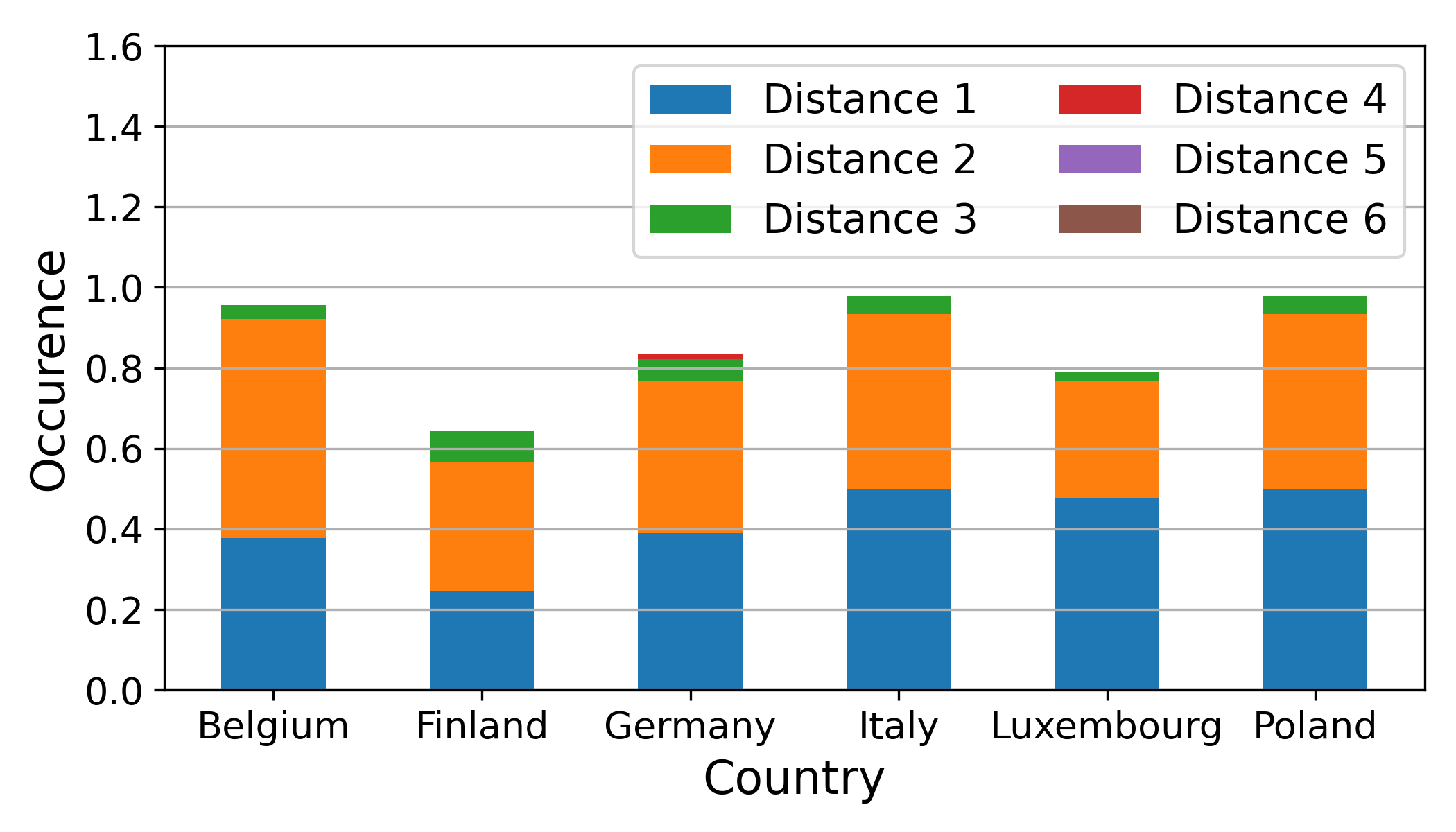}
	\end{minipage}}
	\subfigure[]{
		\begin{minipage}[b]{0.45\linewidth}
			\includegraphics[width=1\linewidth]{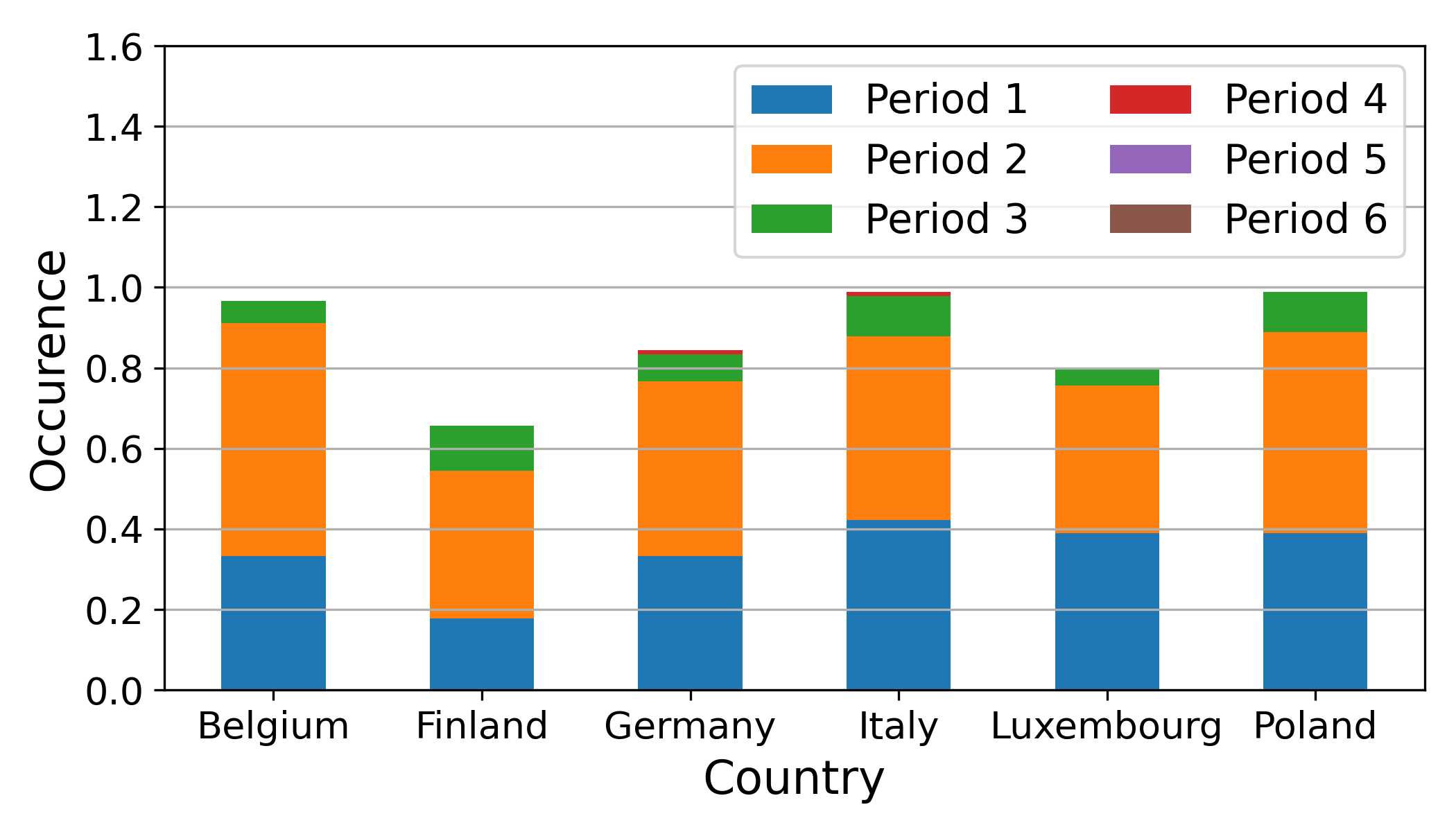}
	\end{minipage}}\\
	\caption{The proportion of infections to the overall population ($90$ nodes) within the specific distance (see Fig.~(a)) to the seeded node within specific simulation steps (see Fig.~(b)).}
\label{HNPeriodDistance}
\end{figure}

We simulate the epidemic spreading process for each country over time until all the connected nodes get infected.
Fig.~\ref{HNPeriodDistance} shows the number of infections within specific distance to the seed node (Fig.~\ref{HNPeriodDistance}(a)) and specific time steps of the epidemic transmission (Fig.~\ref{HNPeriodDistance}(b)). 

As shown in Fig.~\ref{HNPeriodDistance}, generally, all the countries, except for Germany, obtain the most infections within a distance of around $3$ and after $3$ time steps. In contrast, Germany achieves the largest number of infections within the distance of $4$ over $4$ time steps, where one more infection takes place in the fourth time step and four edges away from the seed node. This can be caused by the existence of long shortest path lengths between the nodes and the uncertainty of epidemic transmission given multiple exposures. Compared with the other countries, Finland has the fewest infections in the first step (less than $30
\%$) and overall infections (less than $70\%$, in contrasts with the other countries' infection occurrence over $80\%$.). This results from the isolated characteristics of people with unprefered features and the limited number of social contacts (See the node degree distribution in Fig.~\ref{FinlandDegree} and the shortest path length dsitribution in Fig.~\ref{FinlandPath}). Moreover, Belgium and Germany have fewer infections (less than $40\%$) within the distance of 1 in the first step but more widespread infections afterwards (over $80\%$) than Italy, Luxembourg and Poland. This is because, in the social network simulations, people in Belgium and Germany have a smaller number of direct interactions and a larger number of indirect interactions than people in Italy, Luxembourg and Poland. This implies that isolation policies imposed on the seed node and its direct infections at the epidemic's beginning in countries like Belgium and Germany can effectively reduce the infection numbers.

\subsubsection{People at Risk}

We describe the People at Risk ($PaR(T,D)$) of getting infected within the simulation steps of $[0,T]$ and the distance of $D$ from the seed node based on the equation \ref{pare}, as proposed in our previous study \cite{wen2023dtcns}. 
\begin{equation}
\label{pare}
    PaR(T,D) = \frac{\sum\limits_{v_{i,t}\in V_t} \mathbf{r}(v_{i,t})\delta_{\mathrm{l}(v_{i,t},\mathbf{s}_{t}),D}\delta_{t,T}}{N}, \quad D\leq T
\end{equation}
where $PaR(T,D)$ represents the proportion of infections within a  distance $D$ and $T$ time steps since the epidemic outbreak from the seed node. $\delta_{t,T}$ and $\delta_{\mathrm{l}(v_{i,t},\mathbf{s}_{t}),D}$ each represents the Kronecker functions related to time $t$ and the distance $\mathrm{l}(v_{i,t},\mathbf{s}_{t})$ between node $v_{i,t}$ and seed $\mathbf{s}_t$. $N$ represents the number of nodes.

\begin{figure}[H] 
	\centering
	\subfigure[Belgium]{
		\begin{minipage}[b]{0.3\linewidth}
			\includegraphics[width=1\linewidth]{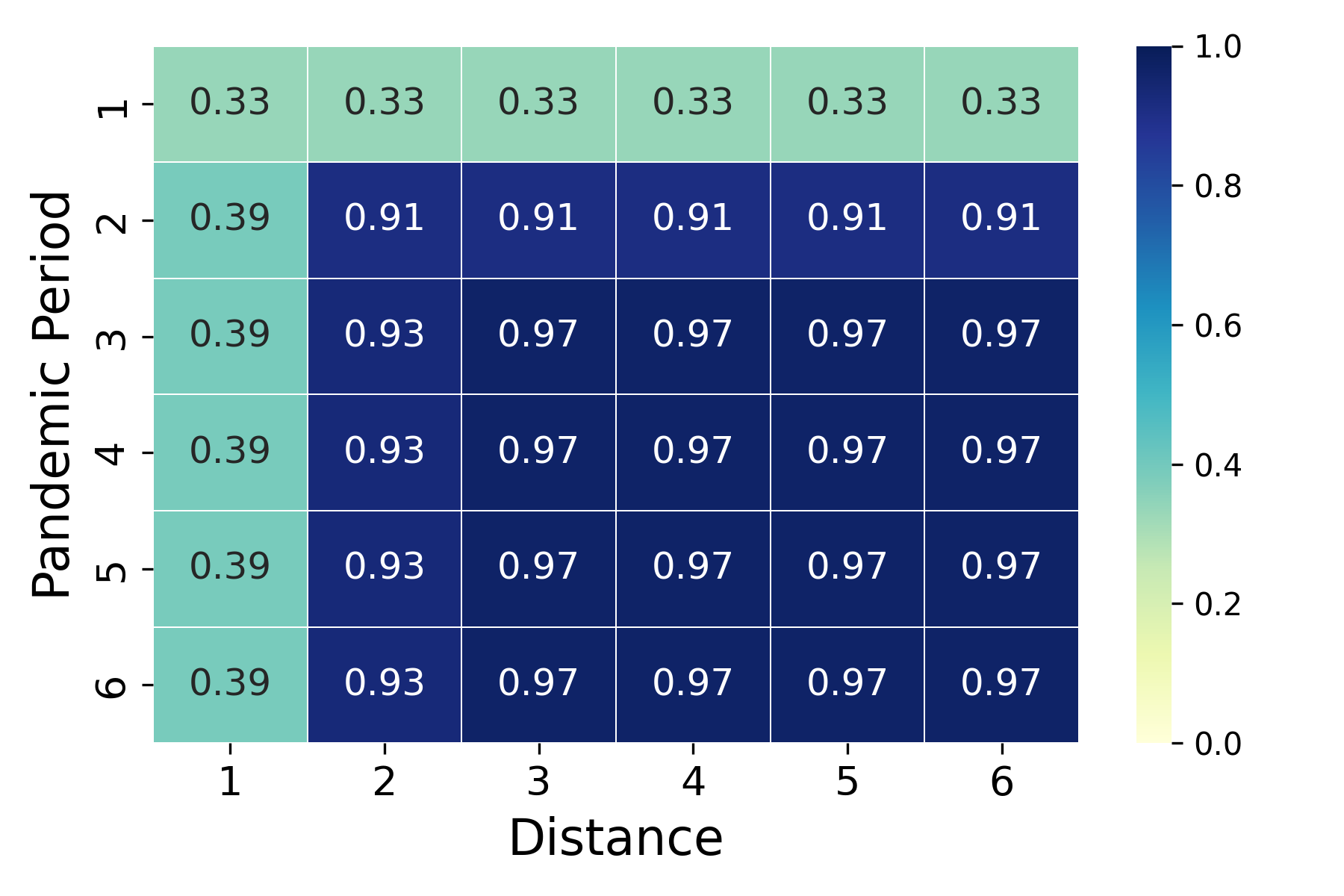}
	\end{minipage}}
	\subfigure[Finland]{
		\begin{minipage}[b]{0.3\linewidth}
			\includegraphics[width=1\linewidth]{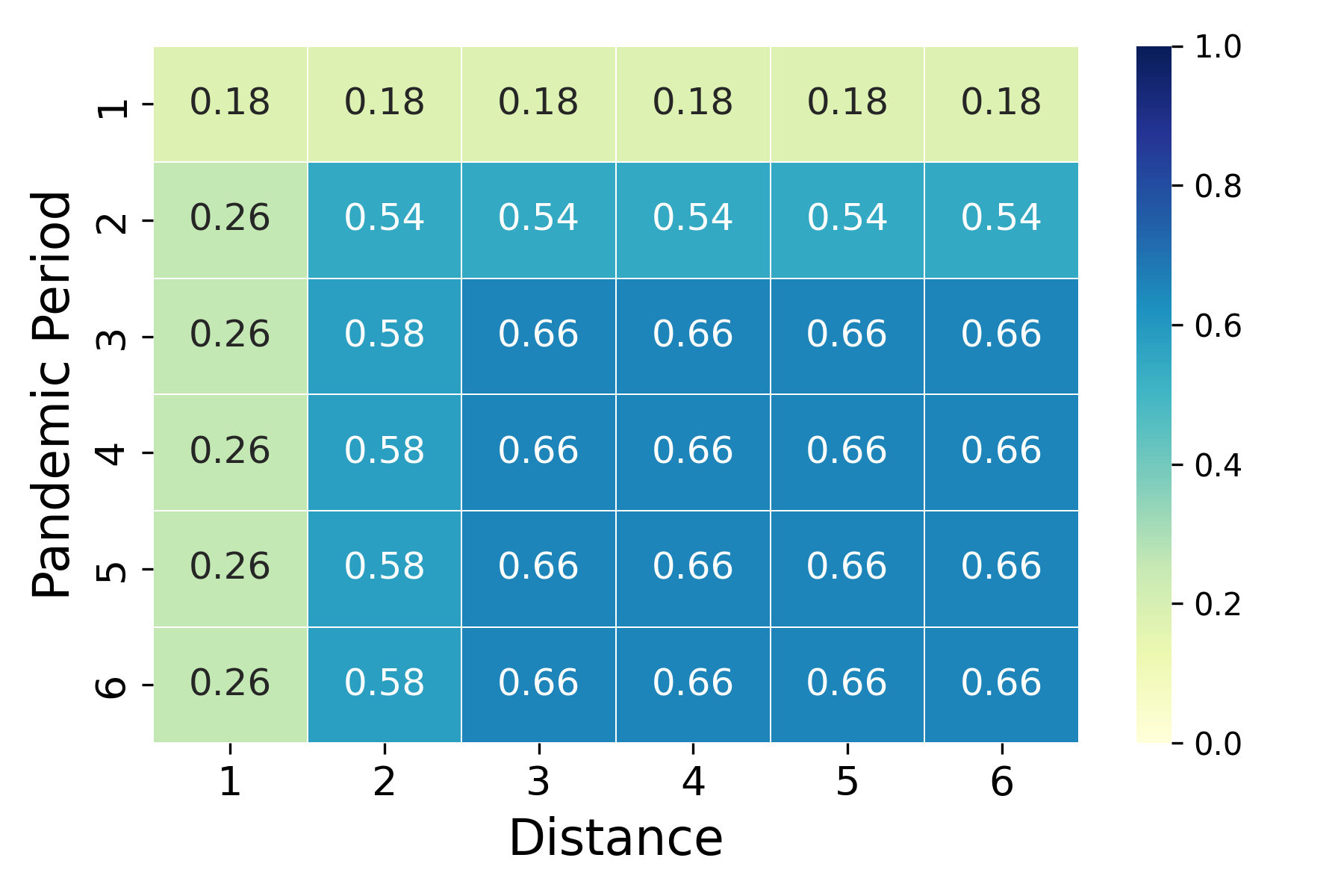}
	\end{minipage}}
		\subfigure[Germany]{
		\begin{minipage}[b]{0.3\linewidth}
			\includegraphics[width=1\linewidth]{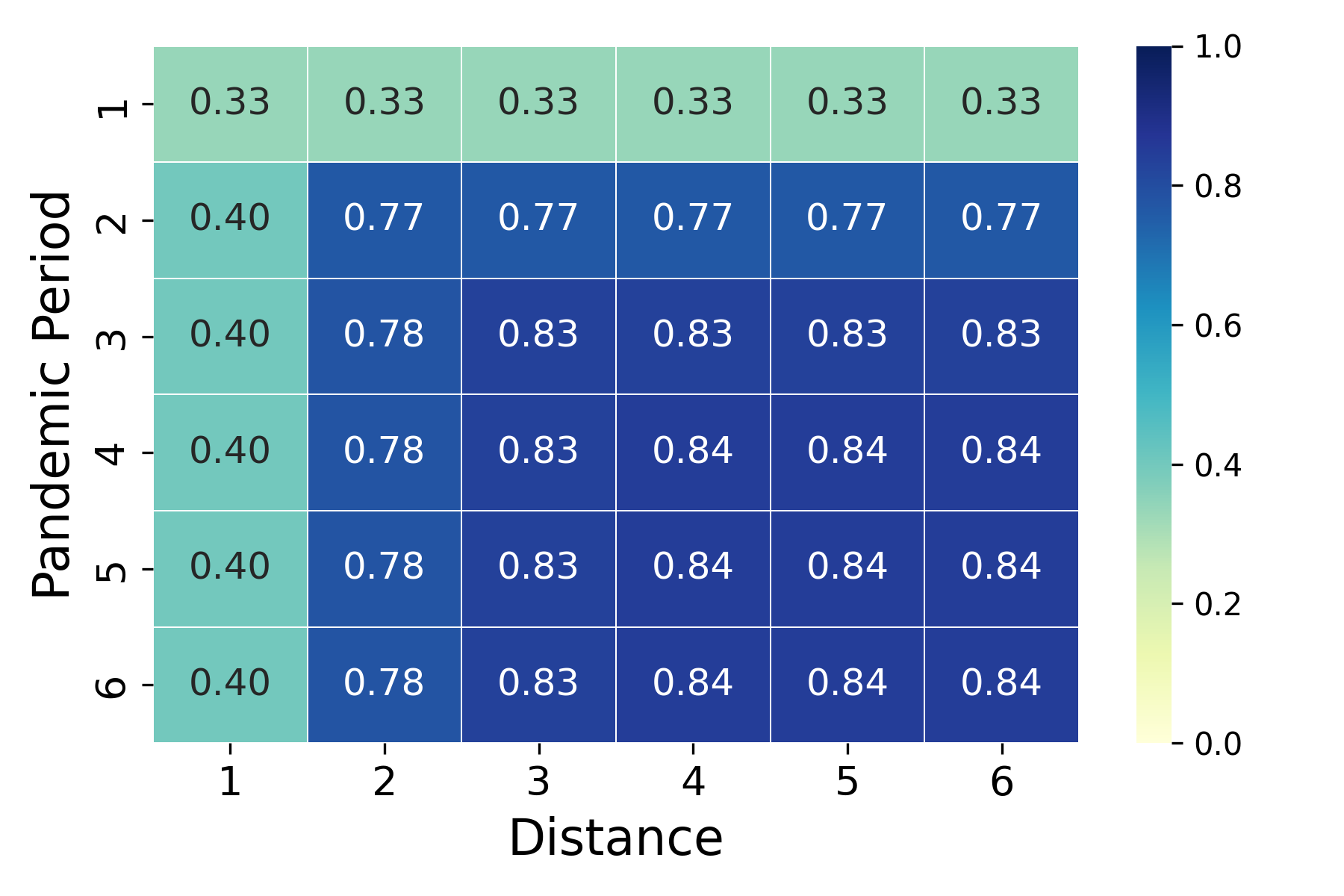}
	\end{minipage}}\\
		\subfigure[Italy]{
		\begin{minipage}[b]{0.3\linewidth}
			\includegraphics[width=1\linewidth]{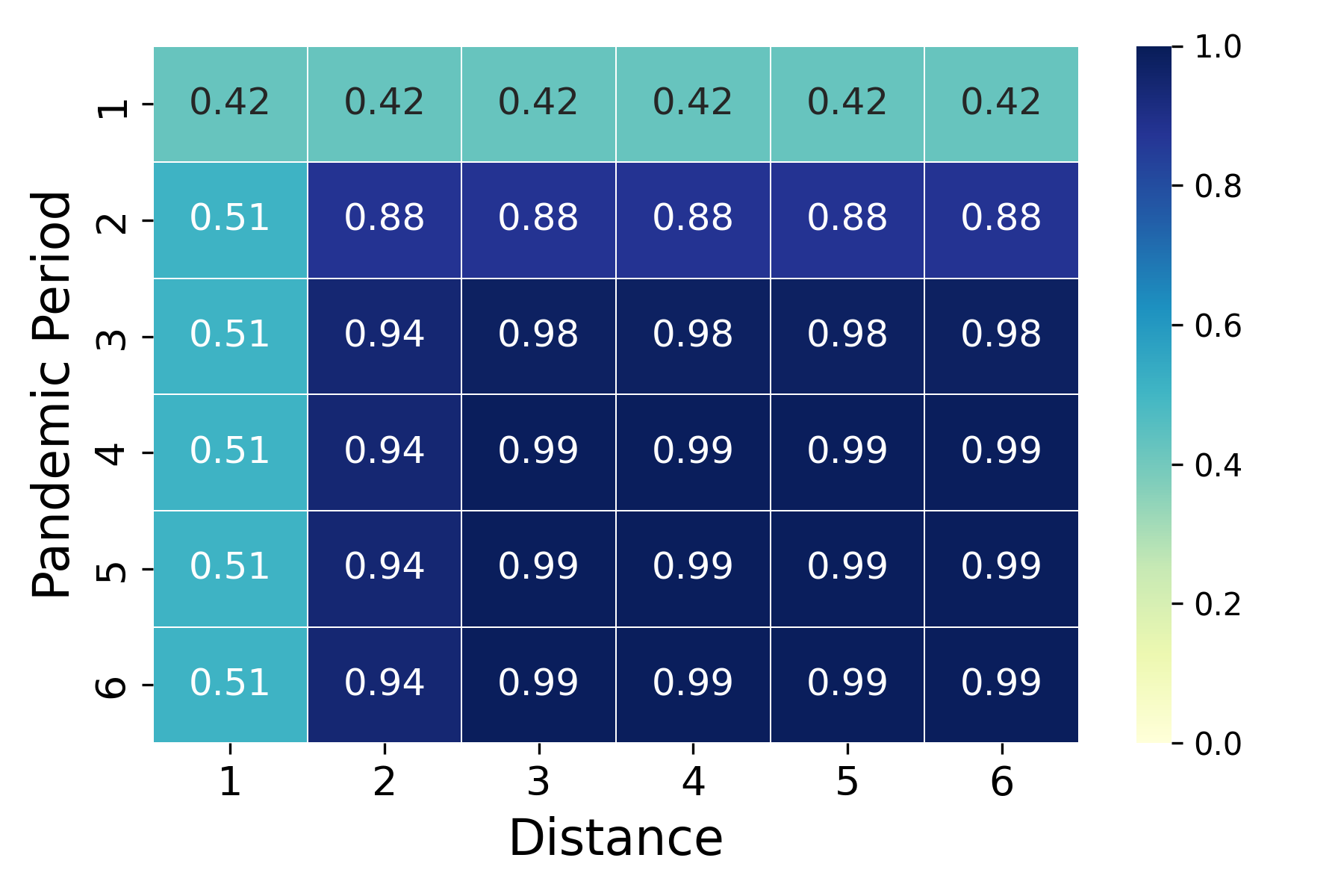}
	\end{minipage}}
	\subfigure[Luxembourg]{
		\begin{minipage}[b]{0.3\linewidth}
			\includegraphics[width=1\linewidth]{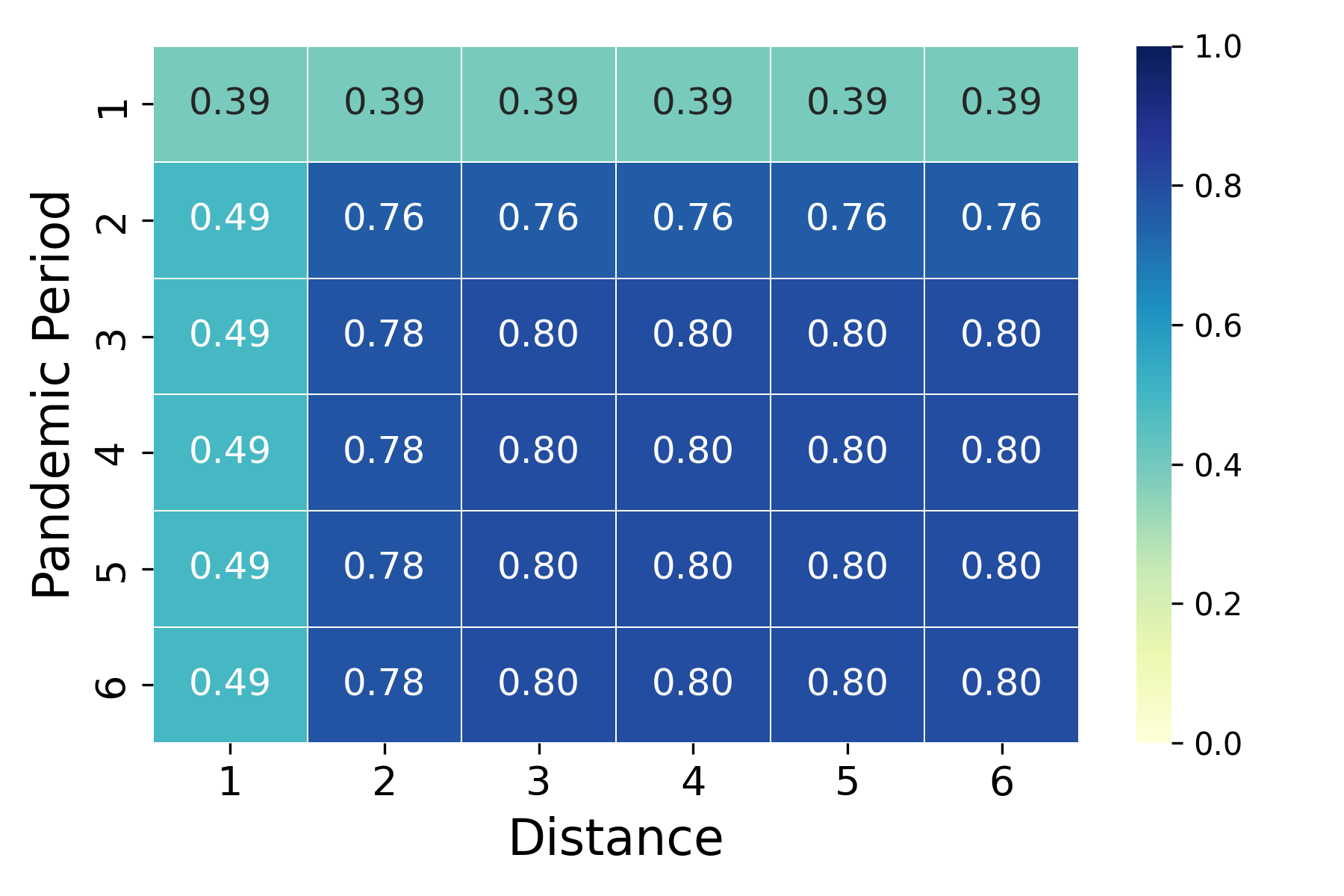}
	\end{minipage}}
		\subfigure[Poland]{
		\begin{minipage}[b]{0.3\linewidth}
			\includegraphics[width=1\linewidth]{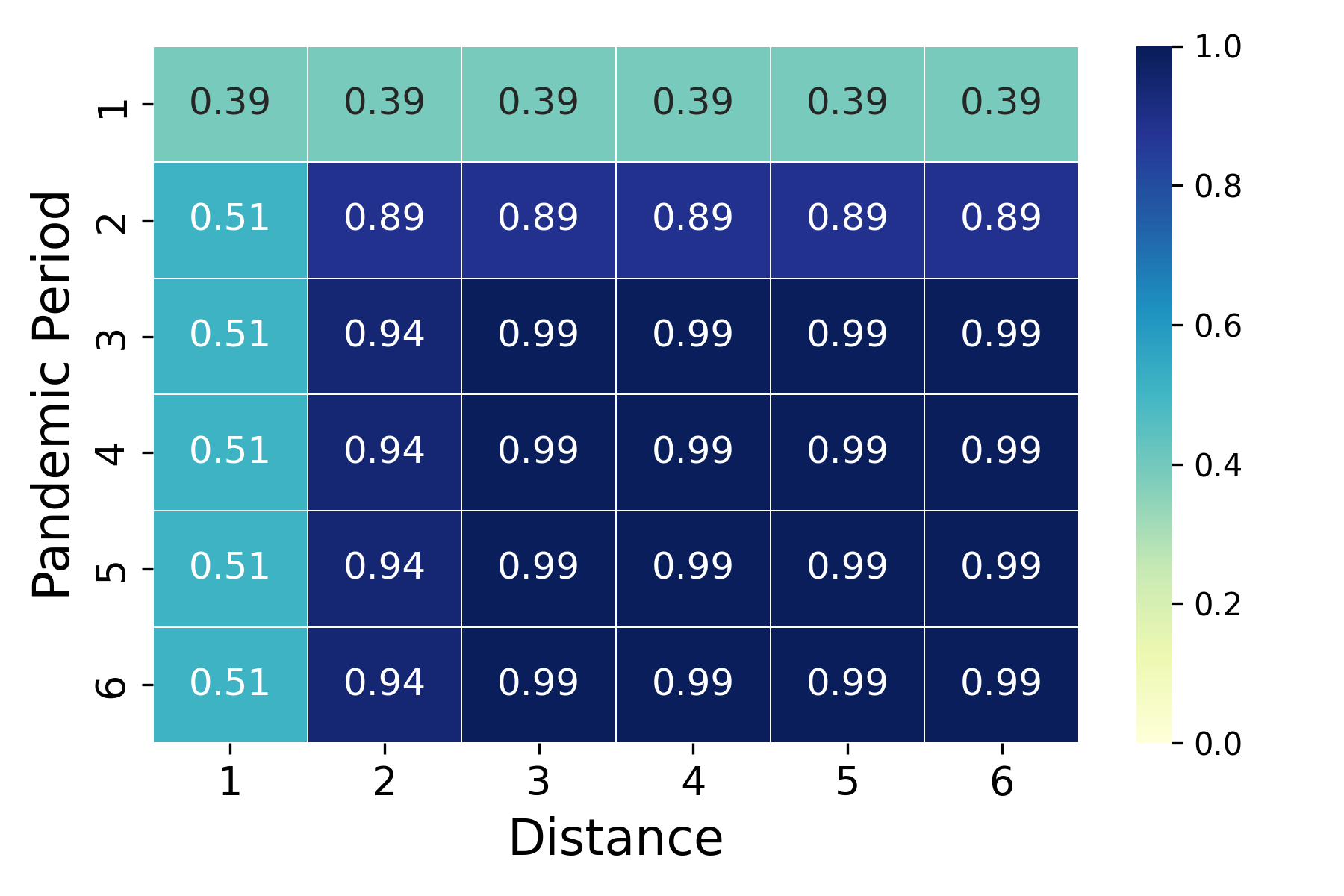}
	\end{minipage}}\\
	\caption{The $PaR$ within the specific distance to the seeded node over the specific time steps in Belgium (See Fig.~(a)), Finland (See Fig.~(b)), Germany (See Fig.~(c)), Italy (See Fig.~(d)), Luxembourg (See Fig.~(e)) and Poland (See Fig.~(f)).}
\label{PaRfig}
\end{figure}

As shown in Fig.~\ref{PaRfig}, for each pandemic period, the $PaR$ values increase with the distance and simulation time steps. Within a specific distance from the seed node, the $PaR$ values increase until all the connected nodes are infected. Overall, the social networks in Belgium, Italy and Poland will, at the end of epidemic simulation, get higher $PaR$ values (close to $100\%$) than in other countries if the epidemic keeps spreading. This can be caused by the higher number of direct and indirect interactions with the popular seed node and its popular neighbours. As shown in Fig.~\ref{BelgiumClus} (c), Fig.~\ref{ItalyClus} (c) and Fig.~\ref{PolandClus} (c), females in Belgium, males in Italy and males Poland all have a positive clustering coefficient, which implies that the seed node and its neighbours with the same preferred sex features cluster densely and leads to extensive epidemic transmission. Moreover, there are less number of fake paths in cases of Belgium, Italy and Poland than that in other countries (See Tab.~\ref{networkinfo}, Fig.~\ref{BelgiumPath}, Fig.~\ref{ItalyPath} and Fig.~\ref{PolandPath}), which leads to a densely connected network and higher number of overall infections.  We also find that the $PaR(2,2)$ values of the social networks in Belgium, Italy, Germany, Luxembourg and Poland are significantly higher than their $PaR(1,1)$ values, which are over $75\%$. This implies the necessity of epidemic control within a distance of 2 -- no later than the second time step. In the case of Finland, the $PaR(3,3)$ values reach the highest infection occurrence and keep lower than $70\%$, which is also significantly higher than $PaR(1,1)$ values but lower than that in other countries, which, considering the increase of infection occurrence, also requires epidemic control no later than the second time step.

\begin{figure}[H] 
	\centering
	\subfigure[Belgium]{
		\begin{minipage}[b]{0.32\linewidth}
			\includegraphics[width=1\linewidth]{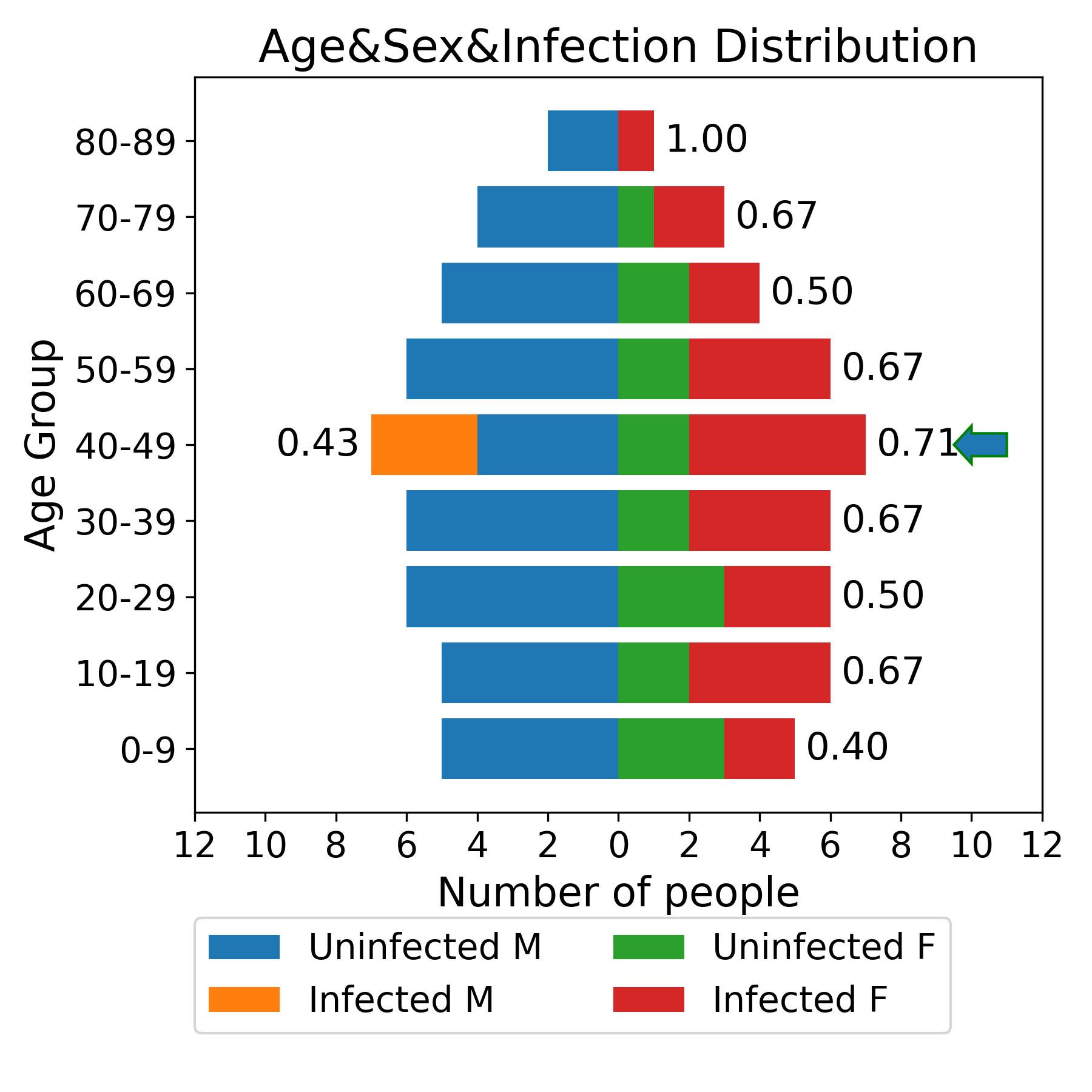}
	\end{minipage}}
	\subfigure[Finland]{
		\begin{minipage}[b]{0.32\linewidth}
			\includegraphics[width=1\linewidth]{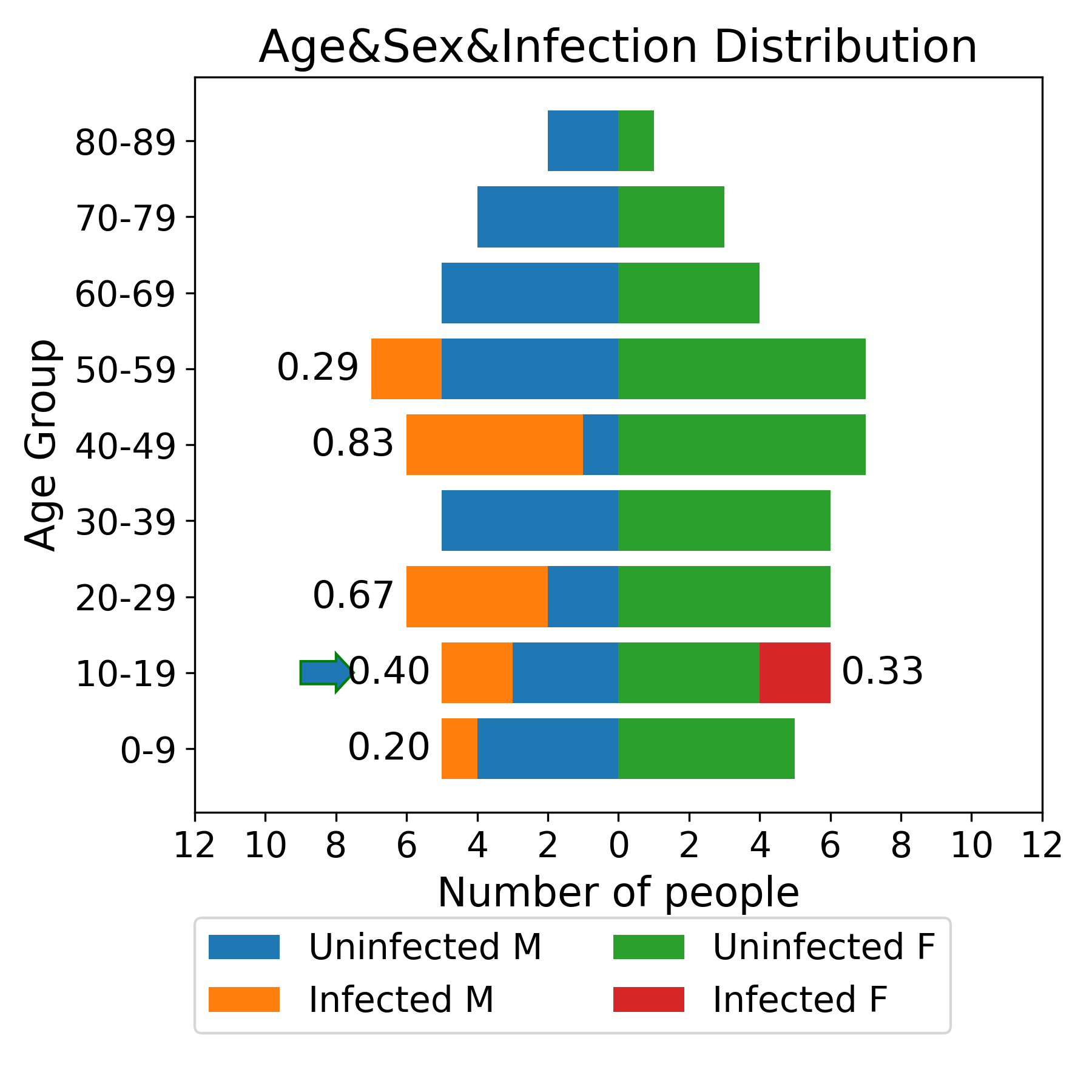}
	\end{minipage}}
	\subfigure[Germany]{
		\begin{minipage}[b]{0.32\linewidth}
			\includegraphics[width=1\linewidth]{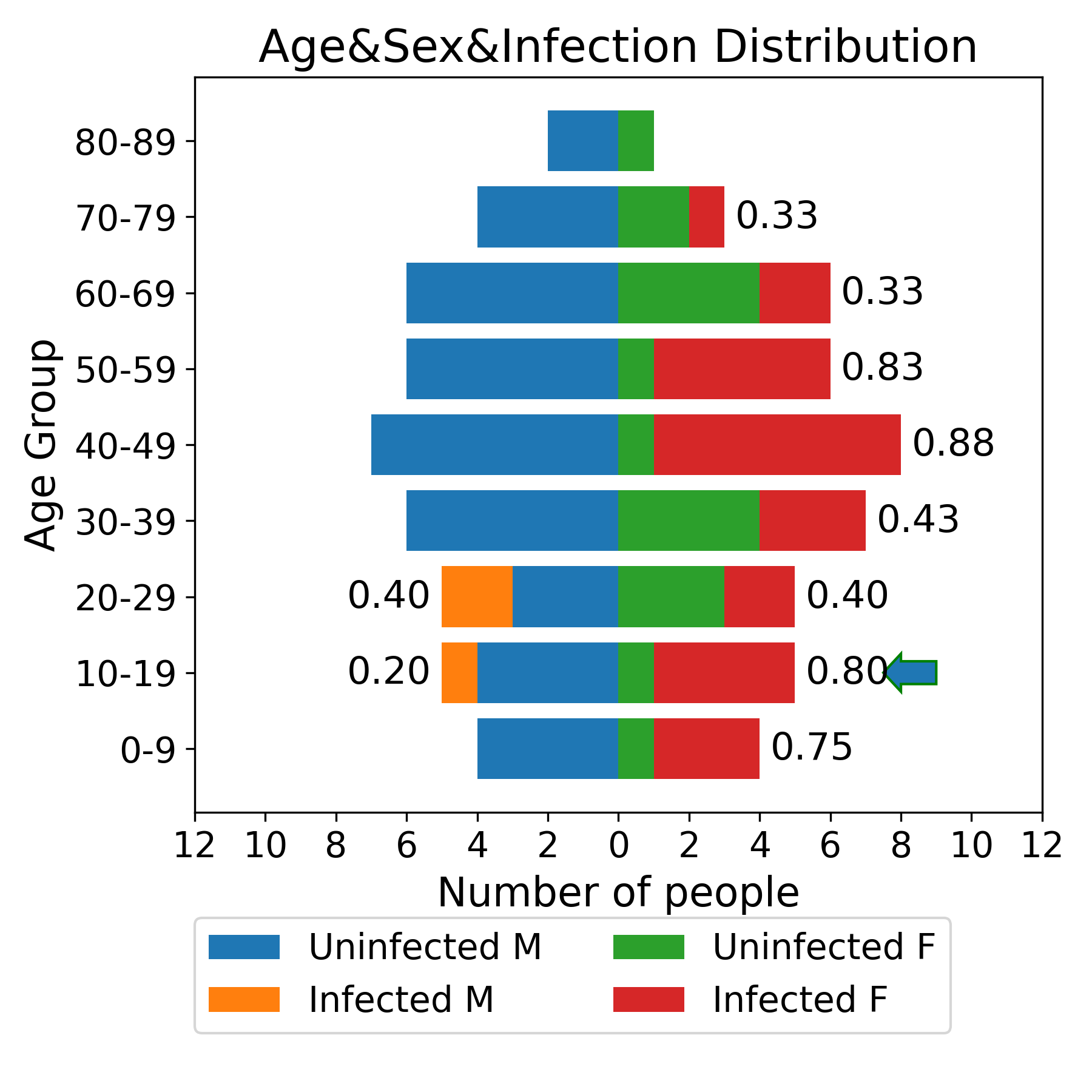}
	\end{minipage}}\\
		\subfigure[Italy]{
		\begin{minipage}[b]{0.32\linewidth}
			\includegraphics[width=1\linewidth]{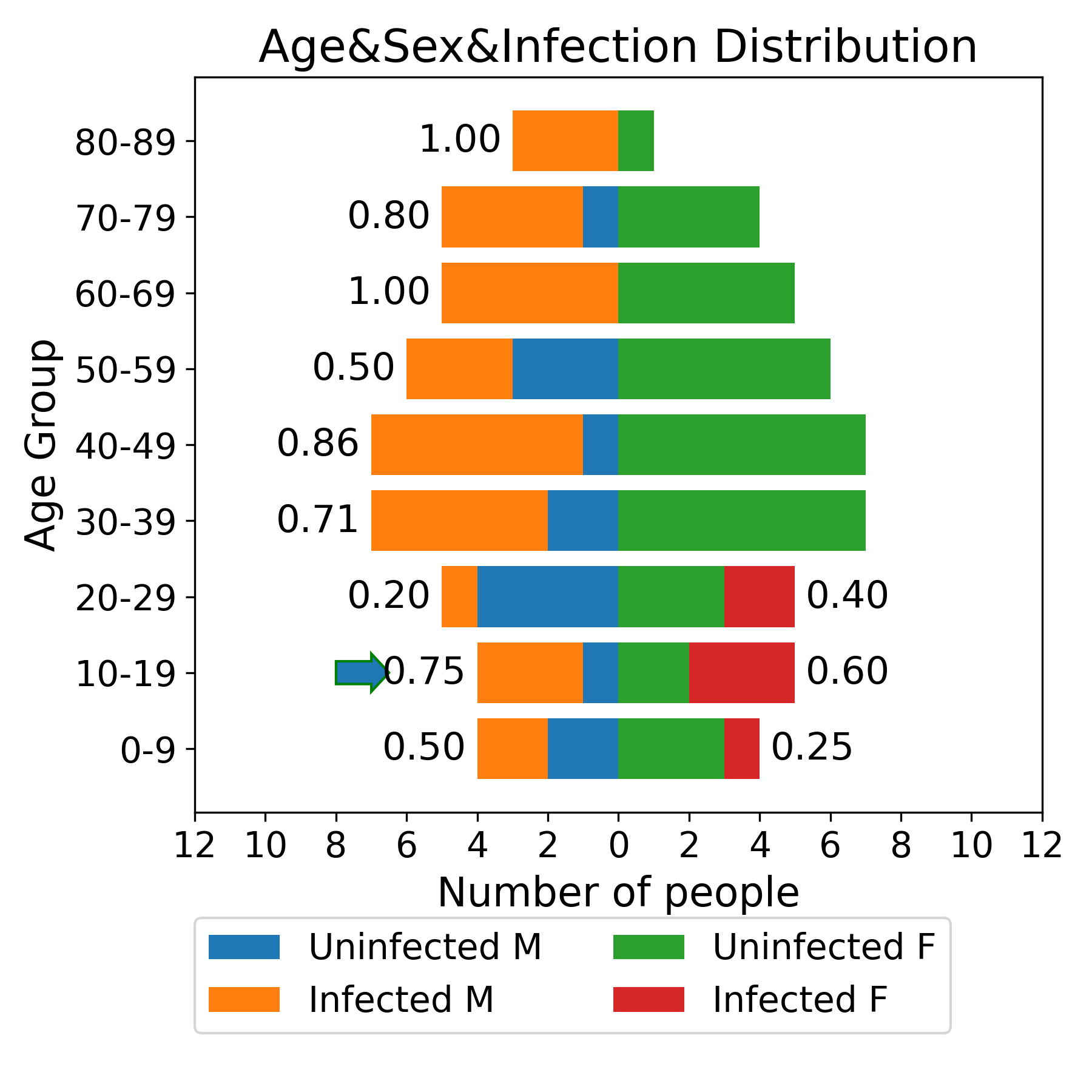}
	\end{minipage}}
	\subfigure[Luxembourg]{
		\begin{minipage}[b]{0.32\linewidth}
			\includegraphics[width=1\linewidth]{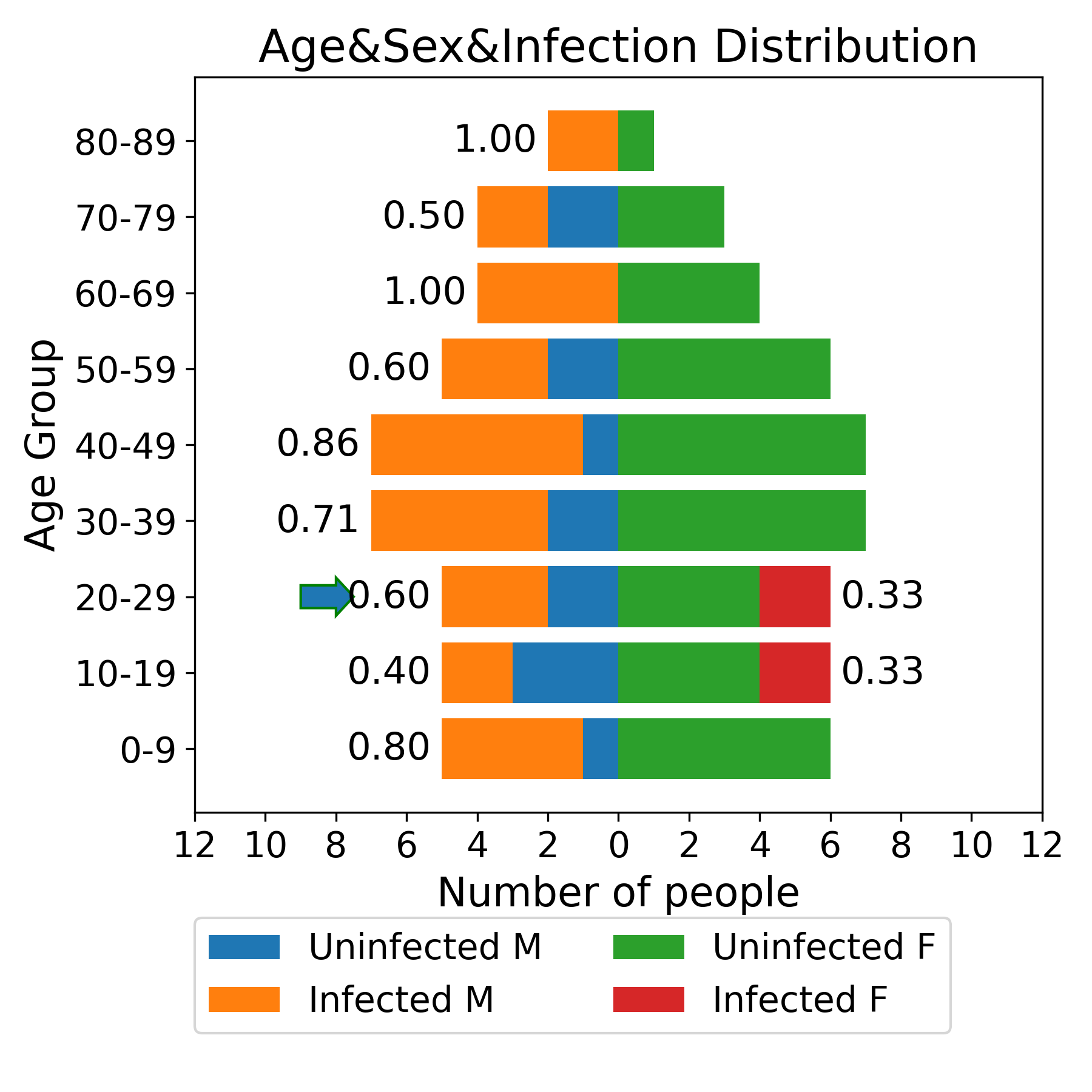}
	\end{minipage}}
		\subfigure[Poland]{
		\begin{minipage}[b]{0.32\linewidth}
			\includegraphics[width=1\linewidth]{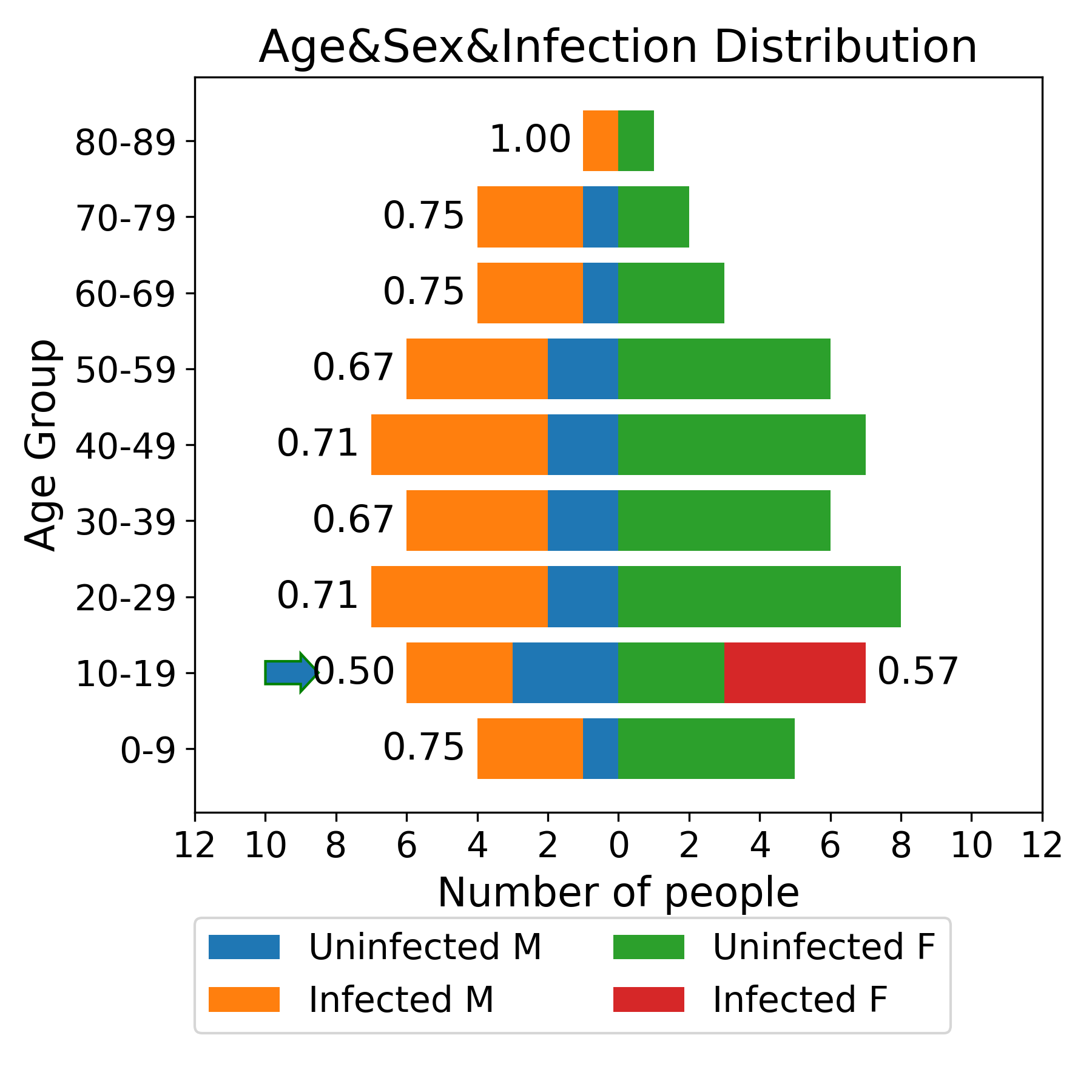}
	\end{minipage}}
	\caption{The simulated age \& sex \& infection-distributions in 2005 of countries including (a) Belgium, (b) Finland, (c) Germany, (d) Italy, (e) Luxembourg and (f) Poland. The arrow points to the specific age \& sex group that includes the seed node for epidemic spread (See Tab.~\ref{startnode}). We also denote each age\&sex group with an infection ratio in the first step of epidemic spread to indicate the infections risk $PaR(1,1)$. }
\label{HNparagesex}
\end{figure}

We calculate the $PaR(1,1)$ of each age \& sex group to better understand the influence of age and sex features on the infection status in the first time step and one hop away from the seed node (See the $PaR(1,1)$ values denoted for each age\&sex bar in Fig.~\ref{HNparagesex}). Given a starting node from female/male group, people with the same sex features and people in a similar age have positive $PaR$ values. For example, given a male starting node whose age ranges between $12$ and $22$ (See Fig.~\ref{HNparagesex} and Tab.~\ref{startnode}), males in Finland, Italy, Luxembourg and Poland tend to have higher $PaR(1,1)$ values than females, and vice versa for Belgium and Germany. The younger age groups and bigger age groups (age groups allocated with more nodes) generally have higher $PaR(1,1)$ values due to their popularity resulting from their age \& sex features and the social contacts developed based on homophily effect (preference for similar age and sex features). Specifically, in Finland, Italy, Luxembourg and Poland, females, in a similar age to the seed node, have positive $PaR(1,1)$ values due to people's preference for social contact with a similar age and young age. These positive $PaR(1,1)$ values indicate the existence of infection risks to the respective age \& sex group. And for the same reason, all females and males in the young age groups in Belgium and Germany also have positive $PaR(1,1)$ values given the female starting node. 
In addition, an uneven distribution of $PaR(1,1)$ values covers the female age groups when the epidemic spreads from a female, and we can also find similar uneven distributions for the $PaR(1,1)$ of males if the epidemic spreads from a male. This results from the different connections developed by the seed node and the randomness of epidemic spread given a transmissibility at $0.80$.

Overall, the heterogeneous features lead to complex social contact patterns and epidemic spreads, which also need to be addressed with heterogeneous and flexible policies. For example, within the first time step and the distance one step away from the epidemic outbreak, we can create different epidemic control policies based on the various $PaR(1,1)$ values for people in each age \& sex group. A higher $PaR(1,1)$ value indicates a higher infection risk under the influence of social network structure, which requires a more extensive quarantine policy in the respective age \& sex group.
More specifically, people with preferred features tend to cluster and thus have higher infection risks within the shortest time and distance to the seed node. For example, females in a similar age or with an age difference around $30$ are popular in Belgium, which given significant number of females age around $[20-60]$, leads to more interactions and the infection risks from multiple exposure to the infection spread. People in Finland have strong preference for young males, which lead to isolated old females and correspondingly, less infection risks to those unconnected people. People with more interactions but higher interaction risks have decisive impact on the infection occurrence. Therefore, epidemic control policies such as suggestions, compulsory vaccination or isolation can be targeted at those popular and risky nodes with preferred features in the first instance.  

\section{Conclusions}
\label{Rep1-2section4}

This study focuses on an emerging research space named Digital Twin-Oriented Complex Networked System (DT-CNS). DT-CNSs aim at faithful representation and modelling of the real networked systems and associated dynamics towards an accurate reflection of reality \citep{wen2022towards}. 

In this study, we propose a DT-CNSs based on Fuzzy Sets. More specifically, we improve the expressive power of the DT-CNSs proposed in the previous works \cite{wen2023dtcns} by introducing heterogeneous feature representation principles, including the crisp and fuzzy value representations. We investigate the impact of various features and feature representation principles and select and validate the best-performing model for a faithful network representation. 

In our empirical analysis, we build DT-CNSs to recreate social contact networks for countries including Belgium, Finland, Germany, Italy, Luxembourg and Poland based on their fuzzy/crisp representation of the real age \& sex distributions and their corresponding preferences for a given age \& sex. We evaluate the outcomes and select the most promising model from the created ones considering the similarity between the target and the recreated social contact matrices. The model selection results suggest that (i) models built on age and sex features outperform the models without these features or with a single age feature, (ii) models built on fuzzy representation principles generally perform better than others and (iii) network models built on sex and fuzzy representation of age ($HN-A_f-S$) are good candidates for the network representation. This suggests that increasing the modelling complexity and using more heterogeneous input information improve the modelling accuracy. In addition, we validate the initialised fuzzy representation principles by introducing a flexible number of fuzzy sets and partial parameter optimisation based on a global sensitivity analysis. The model validation results indicate that (i) increasing flexibility of fuzzy representation principles and (ii) optimisation of partial parameters, even with a small change, improve model performance.

Given the selected network models and fuzzy representation principles, we create realistic social contact networks and investigate their respective disaster resilience when the epidemic spreads from the most popular node. People at risk (PaR), as measured by the proportion of infection to the population within a specific period and distance, changes with time and distance and varies depending on the country. This results from the heterogeneous features and the heterogeneous principles of representation and preference, which require case-dependent and changeable mitigation policies. A higher PaR value indicates a higher infection risk under the influence of social network structure, and requires a more extensive quarantine policy.

To conclude, the structural and dynamic complexity increases when incorporating more features and introducing more flexible representation principles. Such an increase in complexity can potentially improve the faithfulness of network representation by preserving necessary heterogeneous information observed in reality. The feature heterogeneity influences the network structure and induces the epidemic outbreak among specific people. And this requires heterogeneous mitigation policies targeted at heterogeneous populations.

The DT-CNS modelling framework currently focuses on static networks and dyanmic processes with a fixed transmissibility rate. In the future, we aim to develop DT-CNSs with more complex dynamics. Our future research directions on DT-CNS extension generally focuses on three aspects, including (i) preference mutation mechanism that drives network evolution, (ii) transmission mechanisms dependent on node features, and (iii) interrelations between the dynamics of the network and the process. 

\newpage
\appendix

\section{Real age and sex feature distributions}
\label{featapp}
\begin{figure*}[htp] 
	\centering
	\subfigure[Belgium]{
		\begin{minipage}[b]{0.32\linewidth}
			\includegraphics[width=1\linewidth]{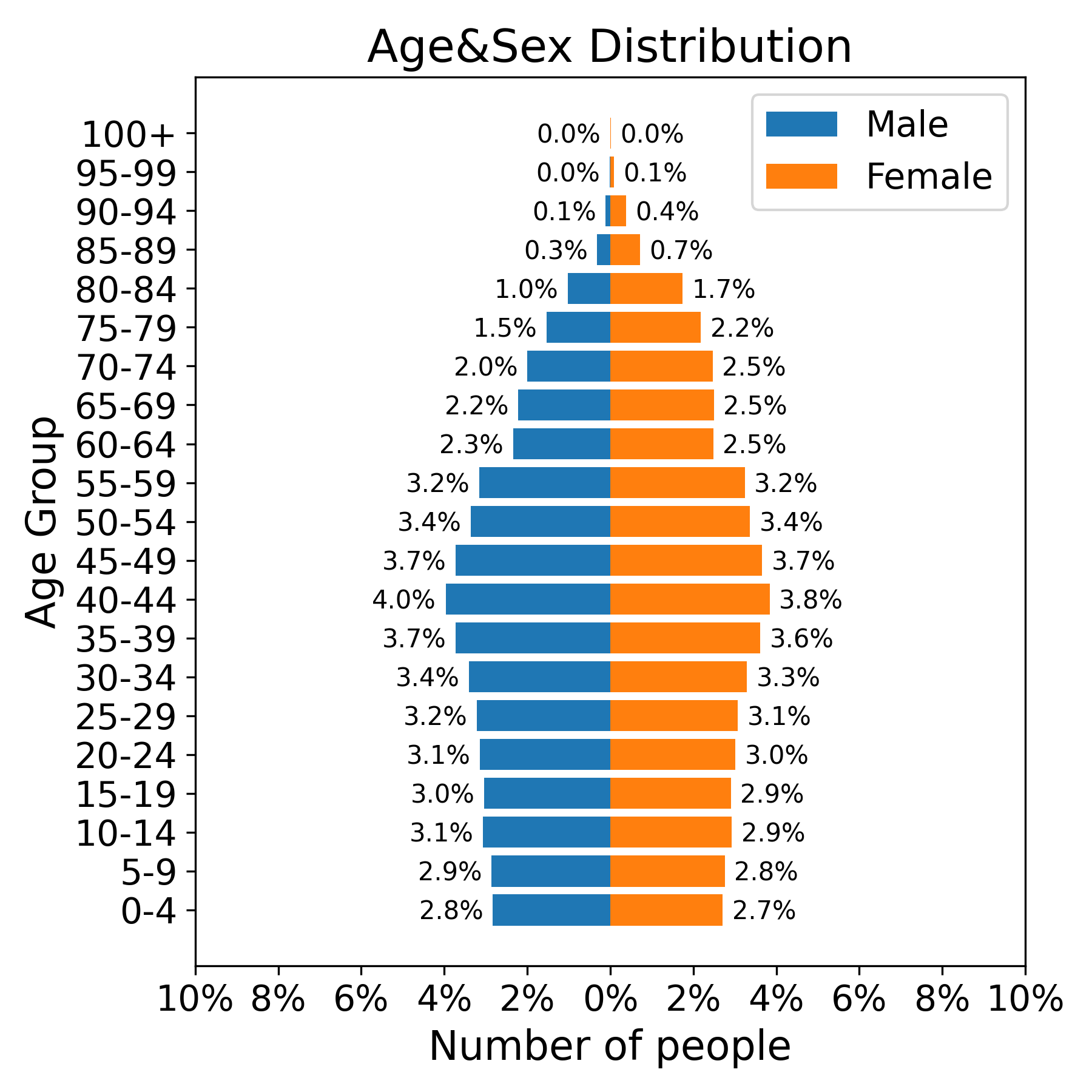}
	\end{minipage}}
	\subfigure[Finland]{
		\begin{minipage}[b]{0.32\linewidth}
			\includegraphics[width=1\linewidth]{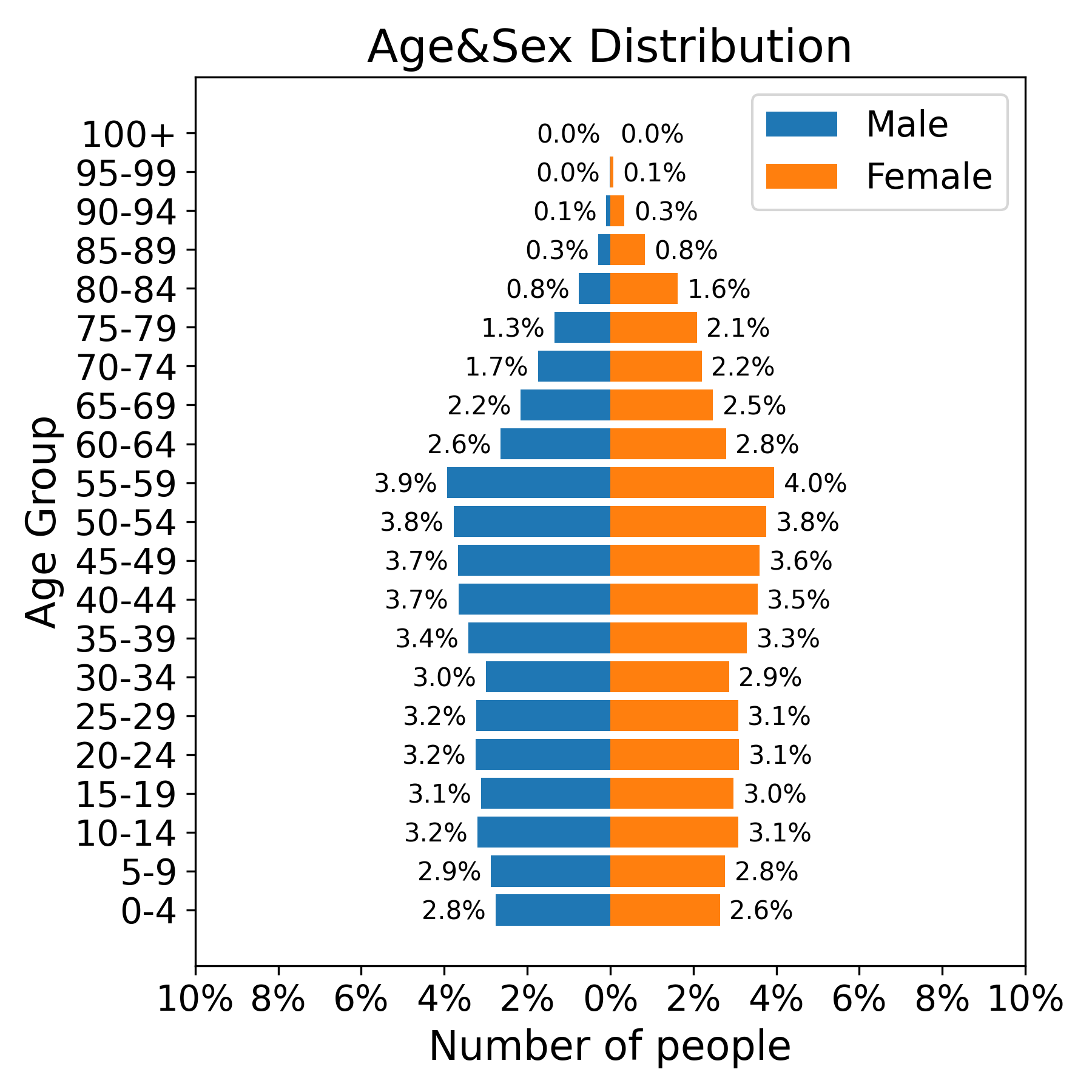}
	\end{minipage}}
	\subfigure[Germany]{
		\begin{minipage}[b]{0.32\linewidth}
			\includegraphics[width=1\linewidth]{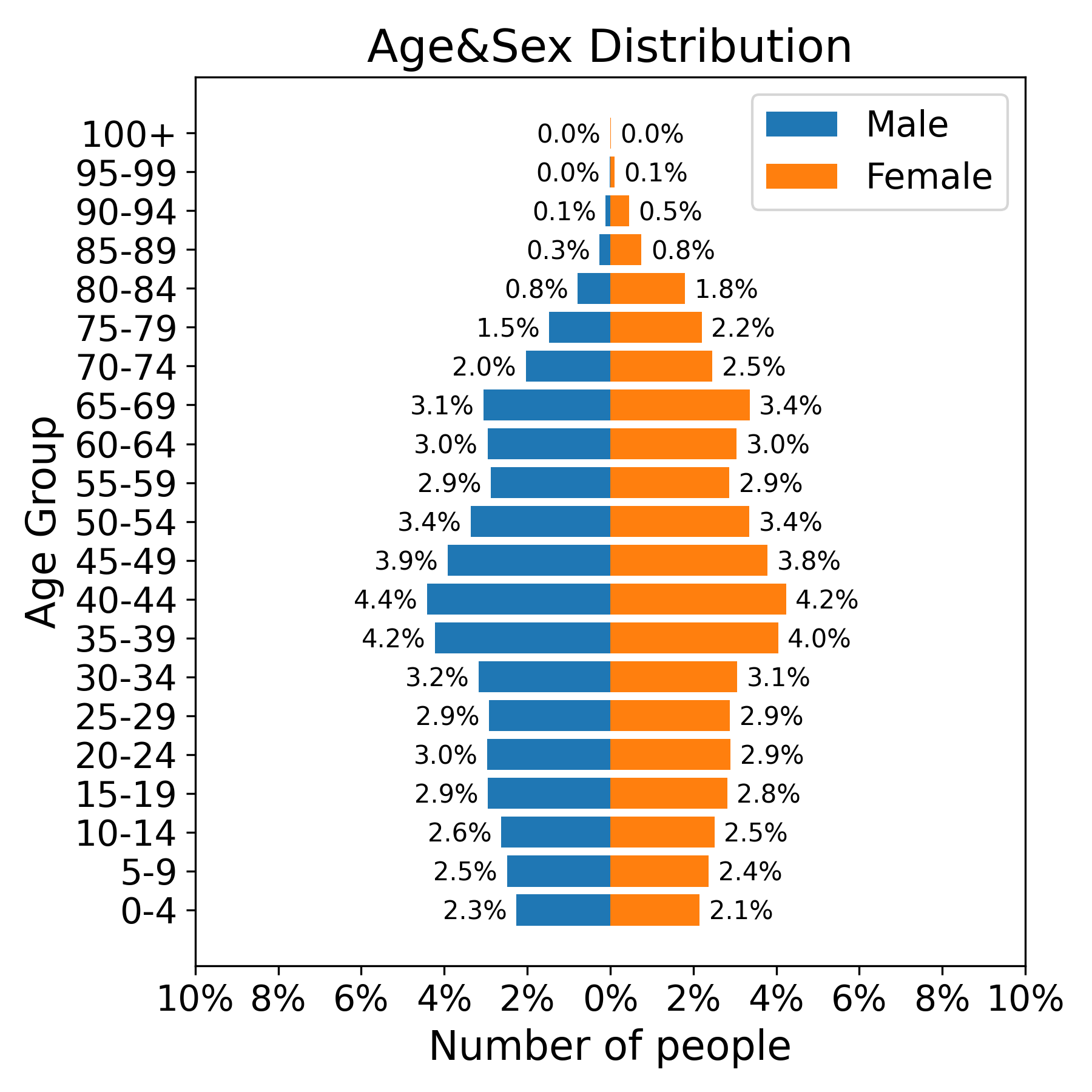}
	\end{minipage}}\\
		\subfigure[Italy]{
		\begin{minipage}[b]{0.32\linewidth}
			\includegraphics[width=1\linewidth]{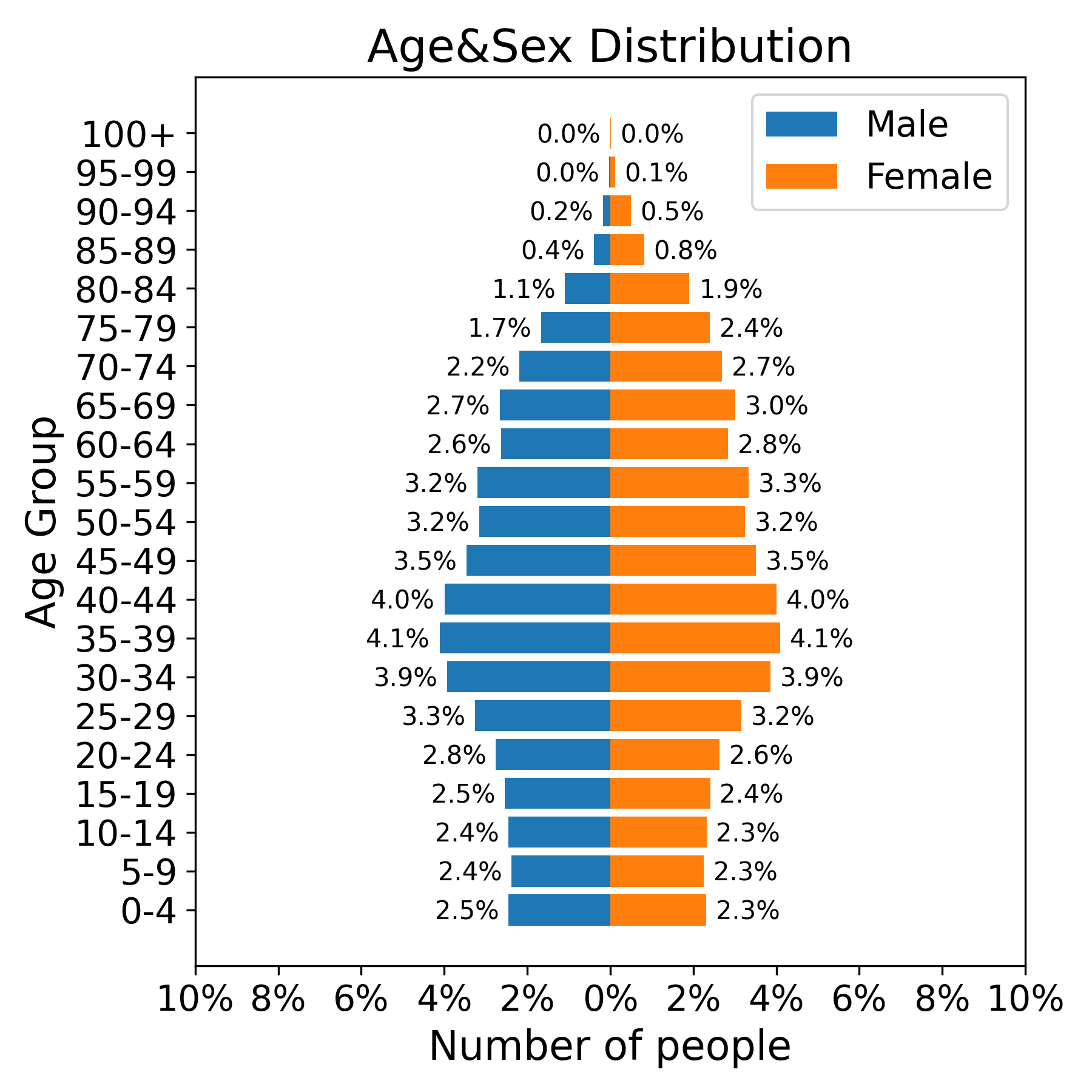}
	\end{minipage}}
	\subfigure[Luxembourg]{
		\begin{minipage}[b]{0.32\linewidth}
			\includegraphics[width=1\linewidth]{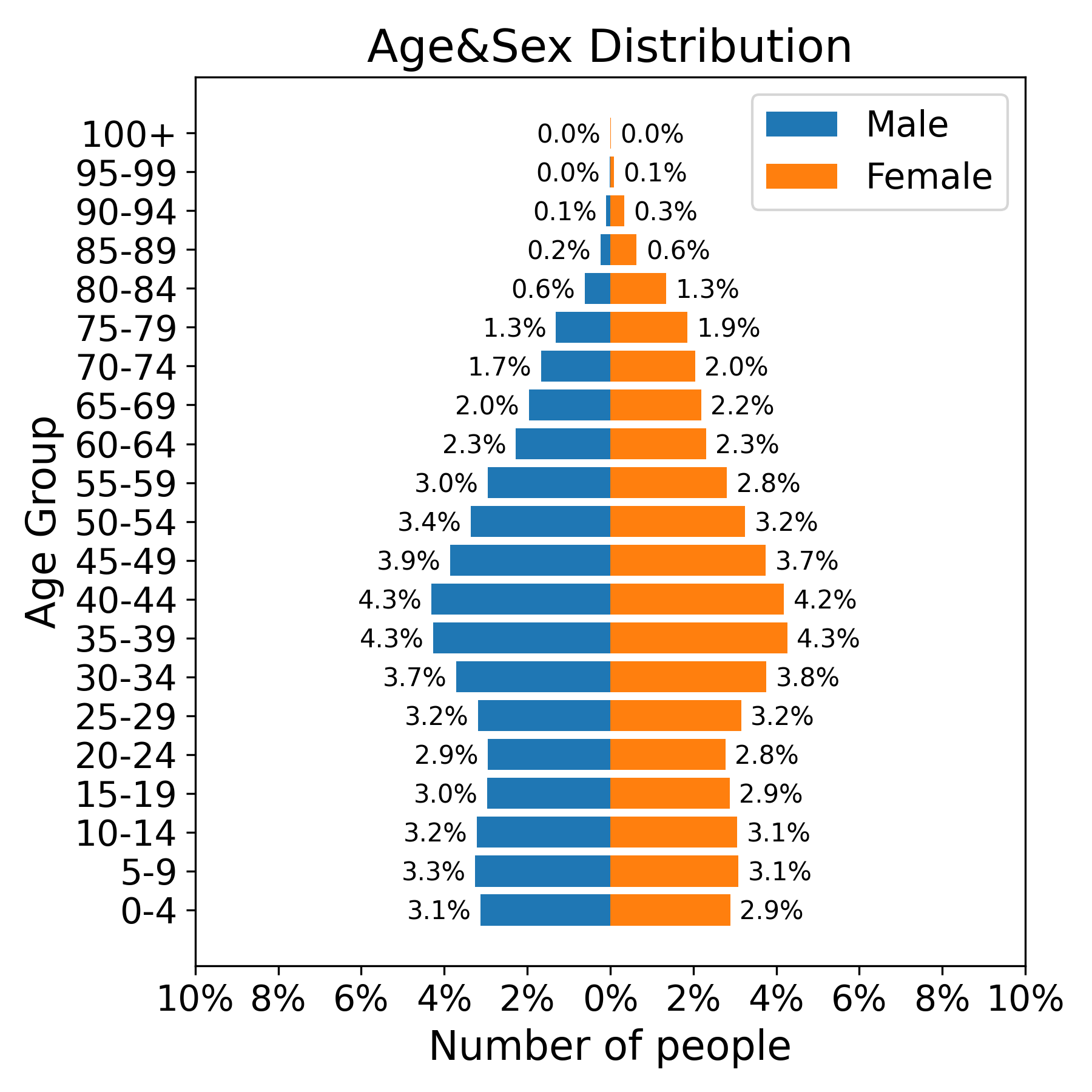}
	\end{minipage}}
		\subfigure[Poland]{
		\begin{minipage}[b]{0.32\linewidth}
			\includegraphics[width=1\linewidth]{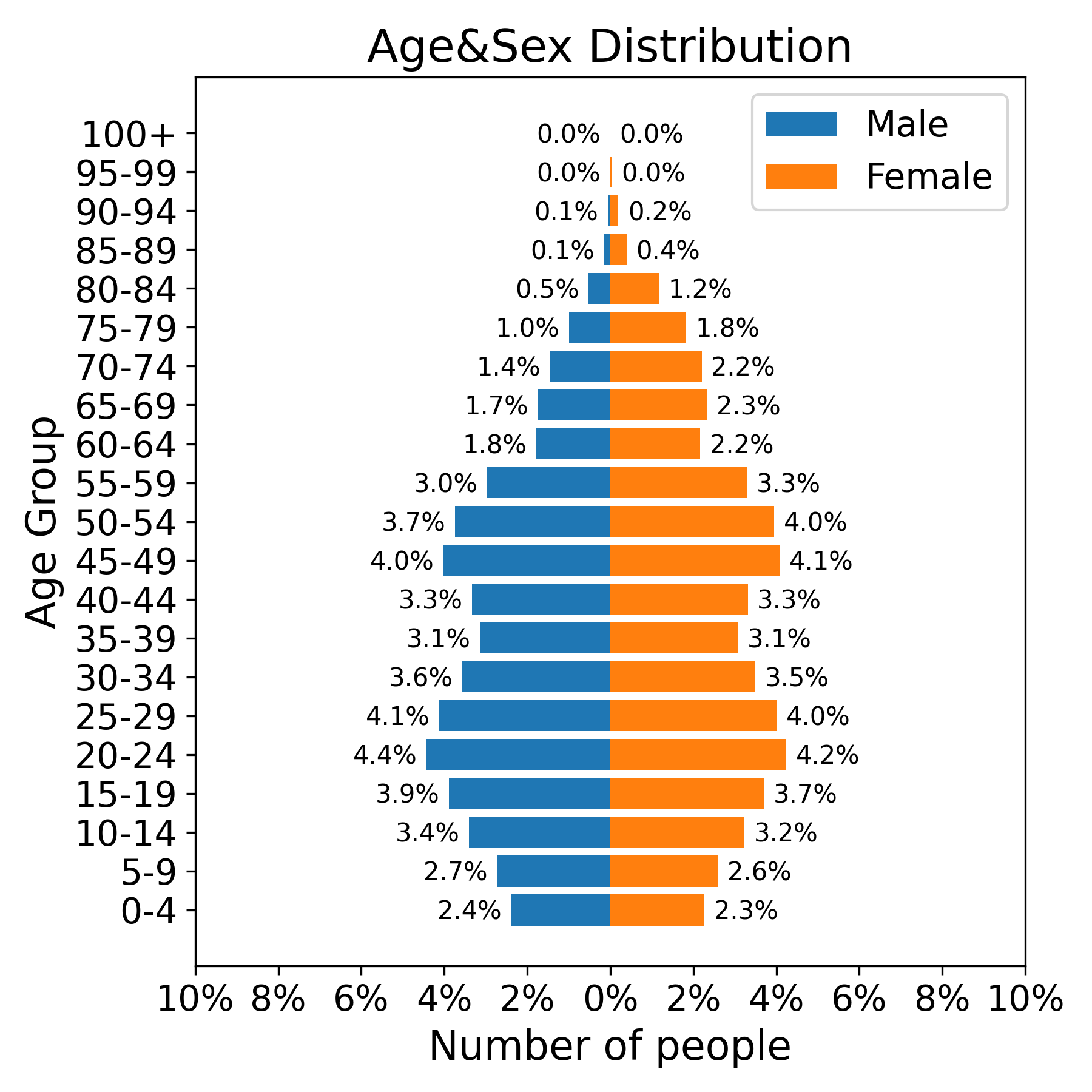}
	\end{minipage}}
	\caption{The age \& sex-structured population profiles in 2005 of countries including (a) Belgium, (b) Finland, (c) Germany, (d) Italy (e) Luxembourg and (f) Poland.}
\label{DNfeat2}
\end{figure*}

\section{Model selection and the recreated social contact matrices}
\label{appMS}

\begin{figure*}[h!]
	\centering
	\subfigure[Target]{
		\begin{minipage}[b]{0.46\linewidth}
			\includegraphics[width=1\linewidth]{BelgiumTarget.png}
	\end{minipage}}
  \hspace{-5mm}
	\subfigure[$RN$]{
		\begin{minipage}[b]{0.46\linewidth}
			\includegraphics[width=1\linewidth]{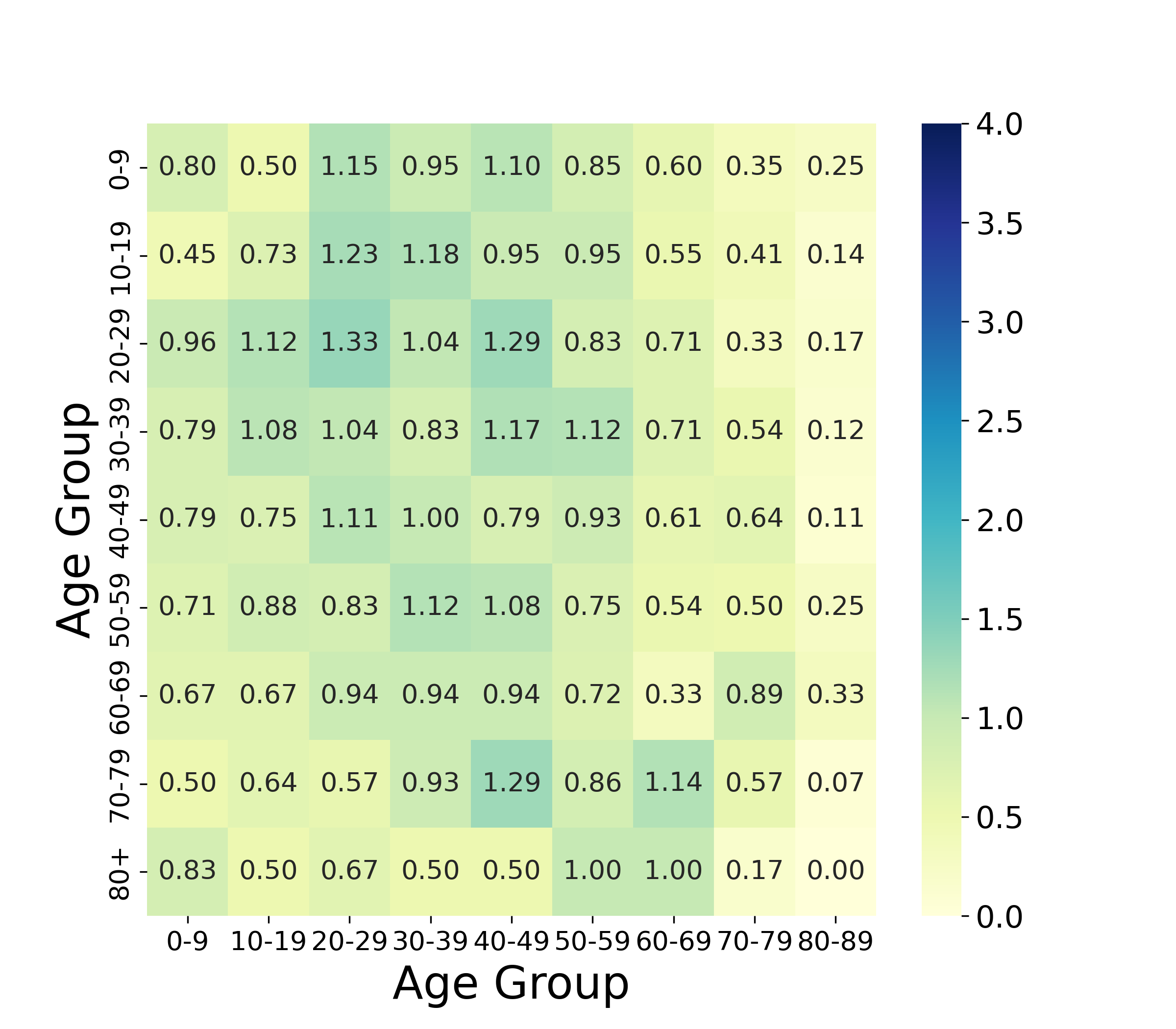}
	\end{minipage}}\\
  \vspace{-3mm}
	\subfigure[$HN-A_c$]{
		\begin{minipage}[b]{0.46\linewidth}
			\includegraphics[width=1\linewidth]{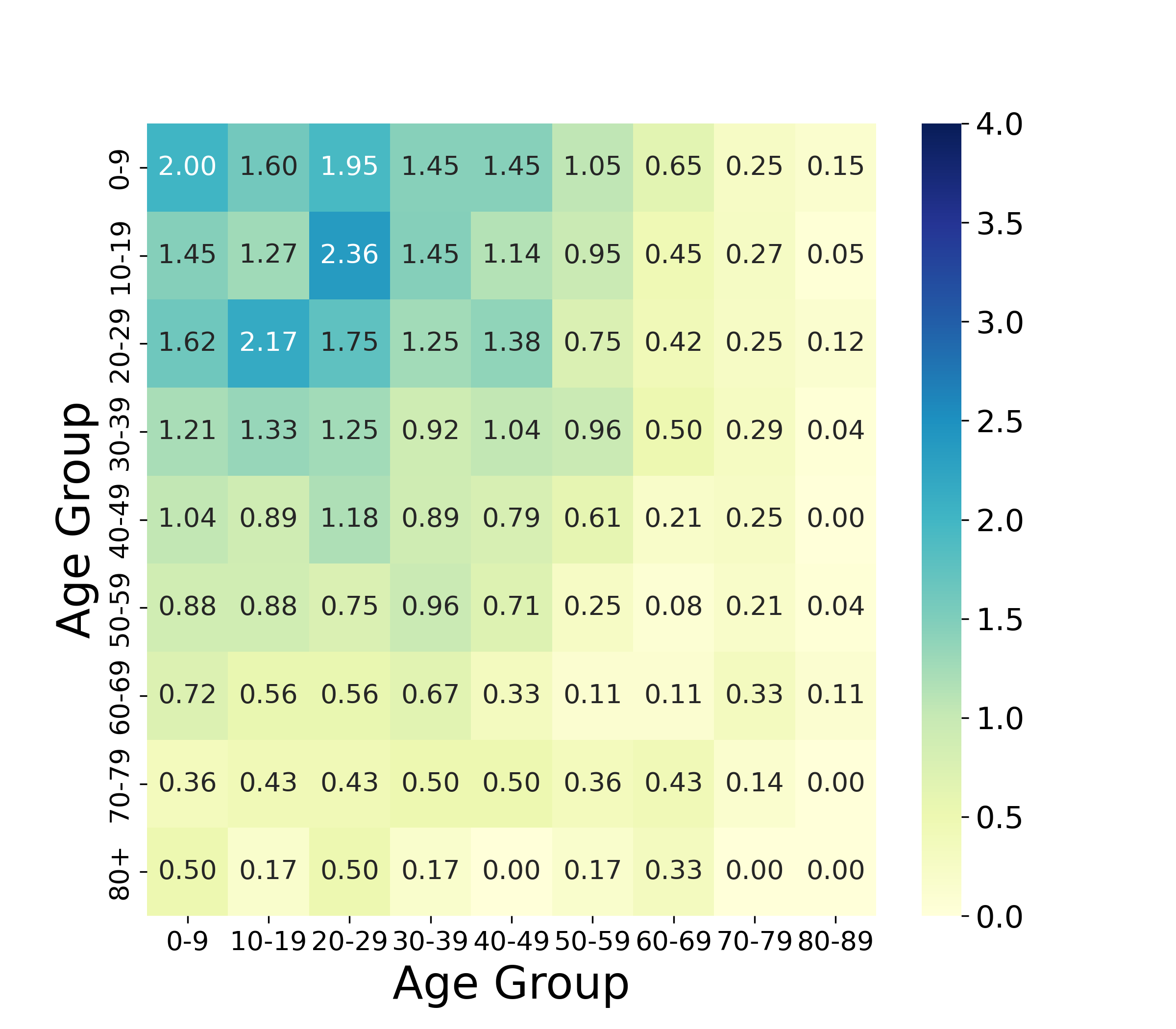}
	\end{minipage}}
  \hspace{-5mm}
		\subfigure[$HN-A_f$]{
		\begin{minipage}[b]{0.46\linewidth}
			\includegraphics[width=1\linewidth]{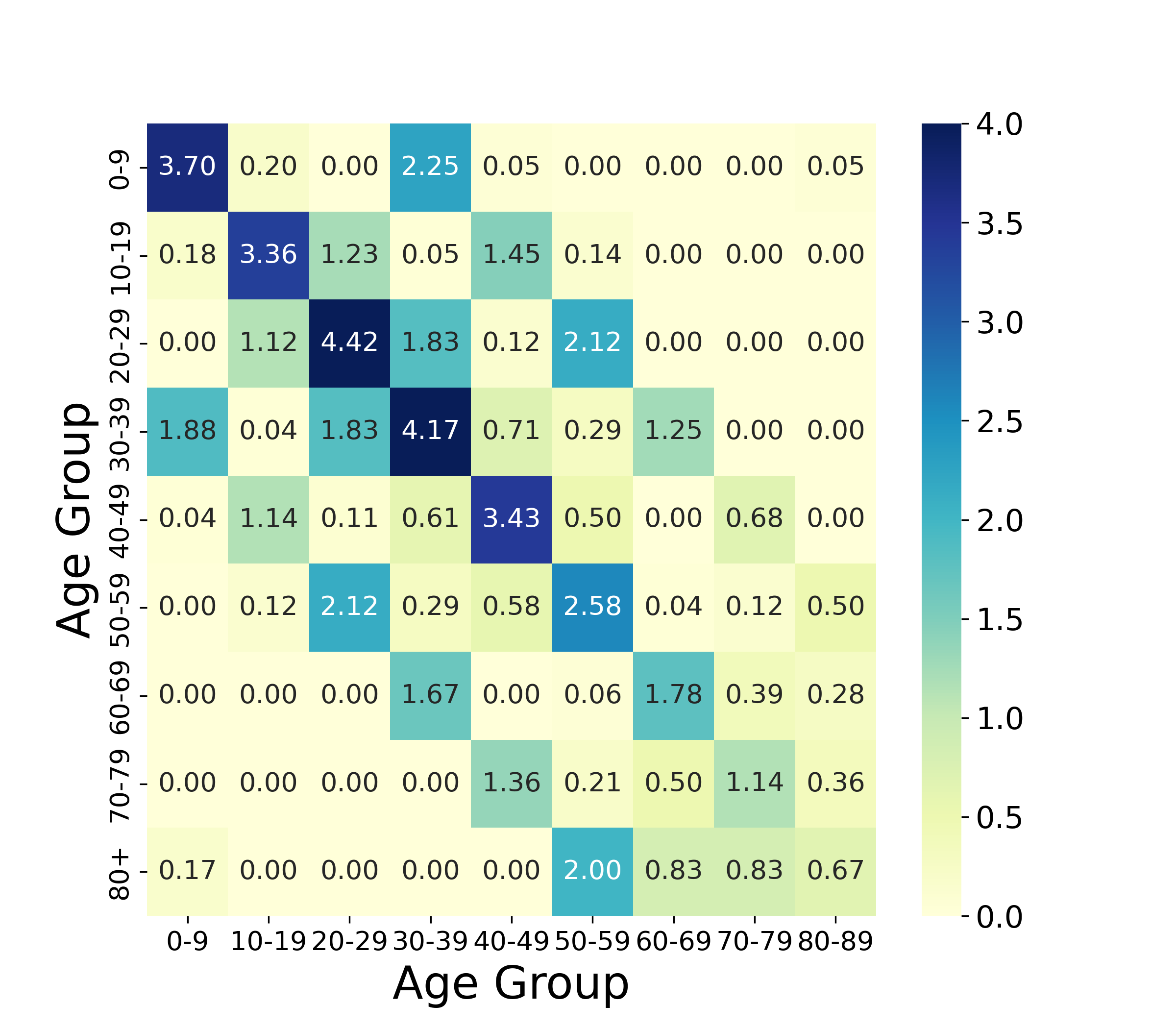}
	\end{minipage}}\\
  \vspace{-3mm}
		\subfigure[$HN-A_c-S$]{
		\begin{minipage}[b]{0.46\linewidth}
	\includegraphics[width=1\linewidth]{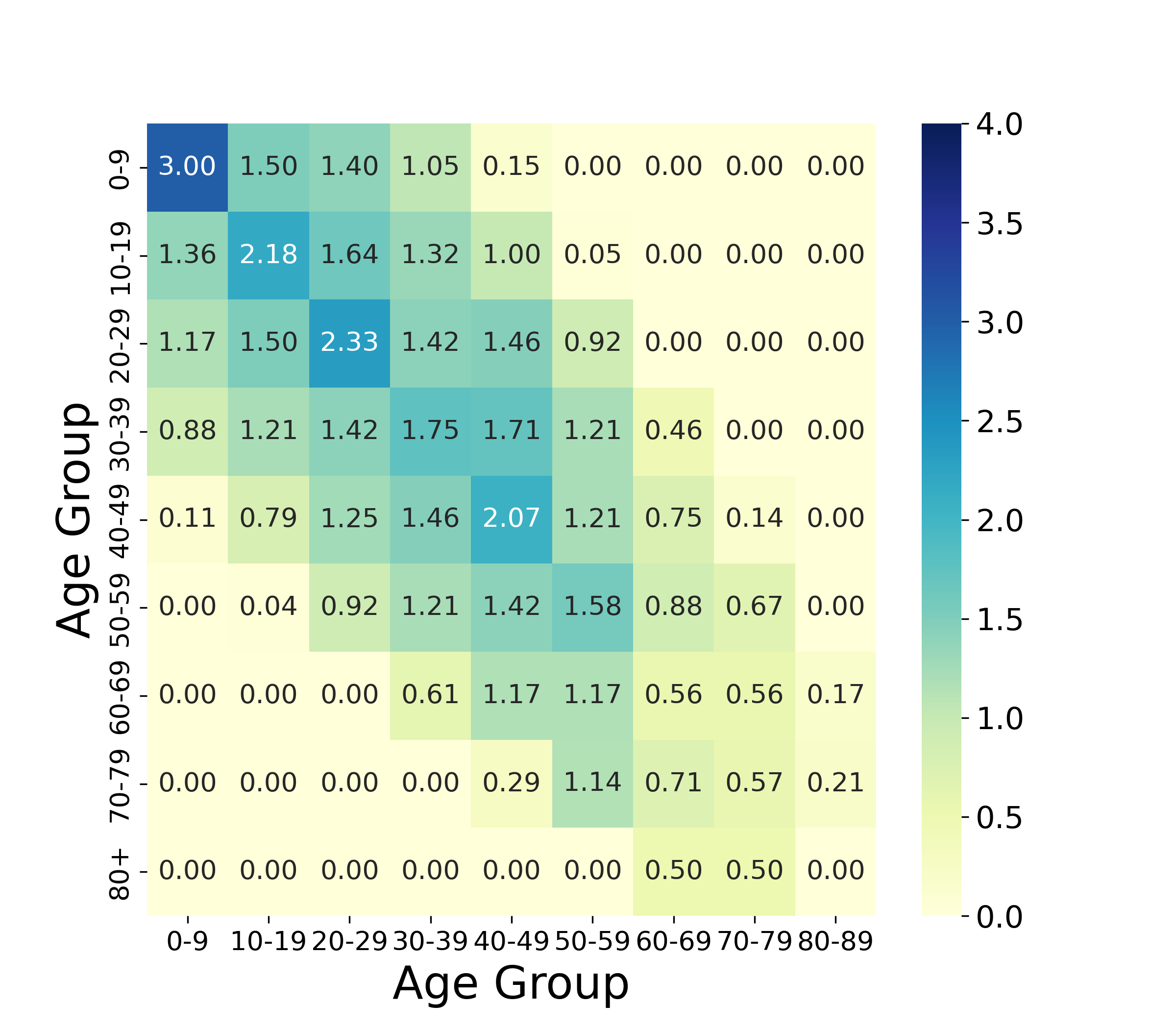}
	\end{minipage}}
  \hspace{-5mm}
		\subfigure[$HN-A_f-S$]{
		\begin{minipage}[b]{0.46\linewidth}
			\includegraphics[width=1\linewidth]{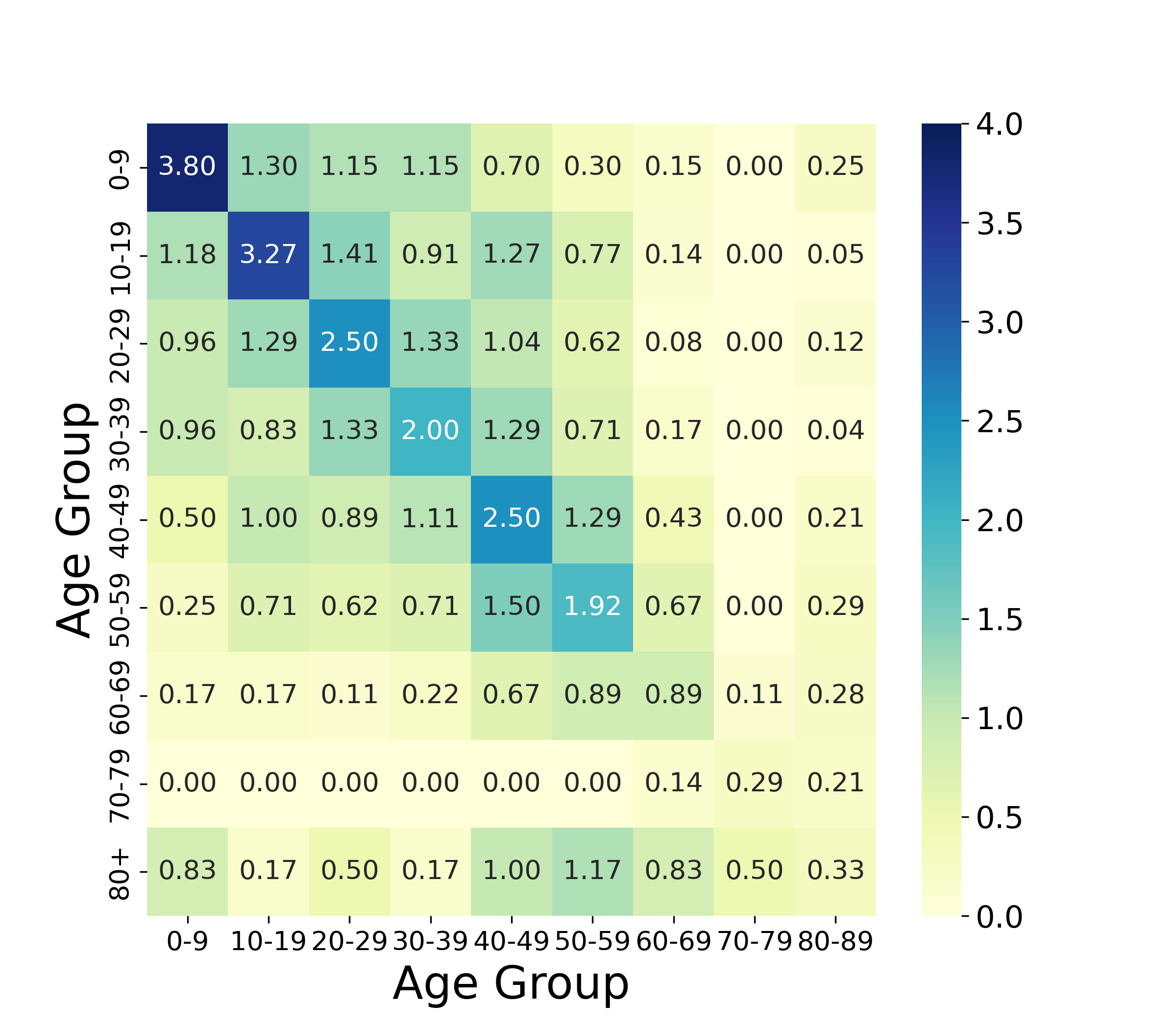}
	\end{minipage}}\\
	\caption{The target social contact matrix ((a) target) and the recreated social contact matrices of Belgium, which are respectively generated by the (b) $RN$, (c) $HN-A_c$, (d) $HN-A_f$, (e) $HN-A_c-S$ and (f) $HN-A_f-S$ models.} 
\label{BelgiumMat}
\end{figure*}

\begin{figure*}[h!]
	\centering
	\subfigure[Target]{
		\begin{minipage}[b]{0.46\linewidth}
			\includegraphics[width=1\linewidth]{GermanyTarget.png}
	\end{minipage}}
  \hspace{-5mm}
	\subfigure[$RN$]{
		\begin{minipage}[b]{0.46\linewidth}
			\includegraphics[width=1\linewidth]{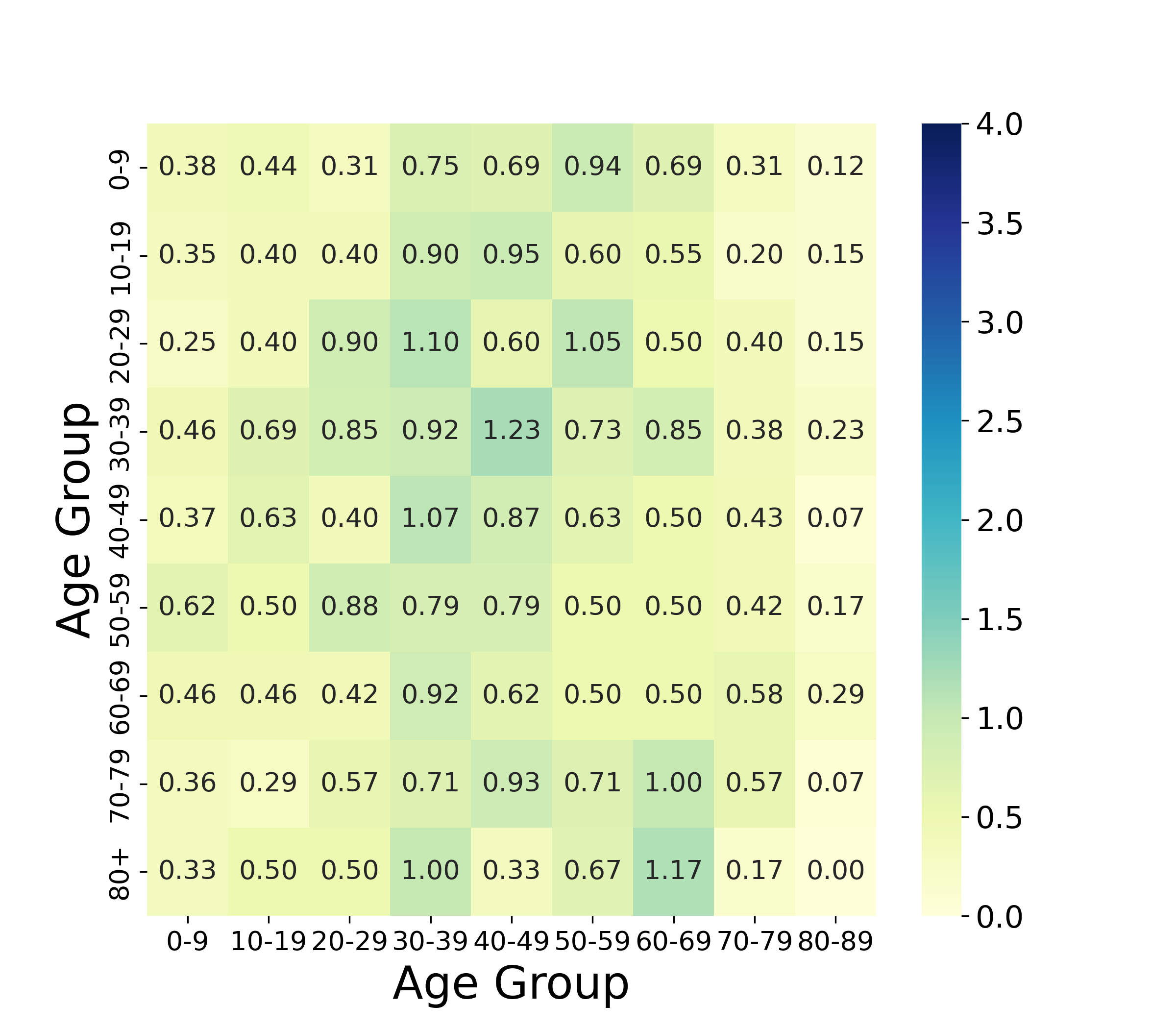}
	\end{minipage}}\\
  \vspace{-3mm}
	\subfigure[$HN-A_c$]{
		\begin{minipage}[b]{0.46\linewidth}
			\includegraphics[width=1\linewidth]{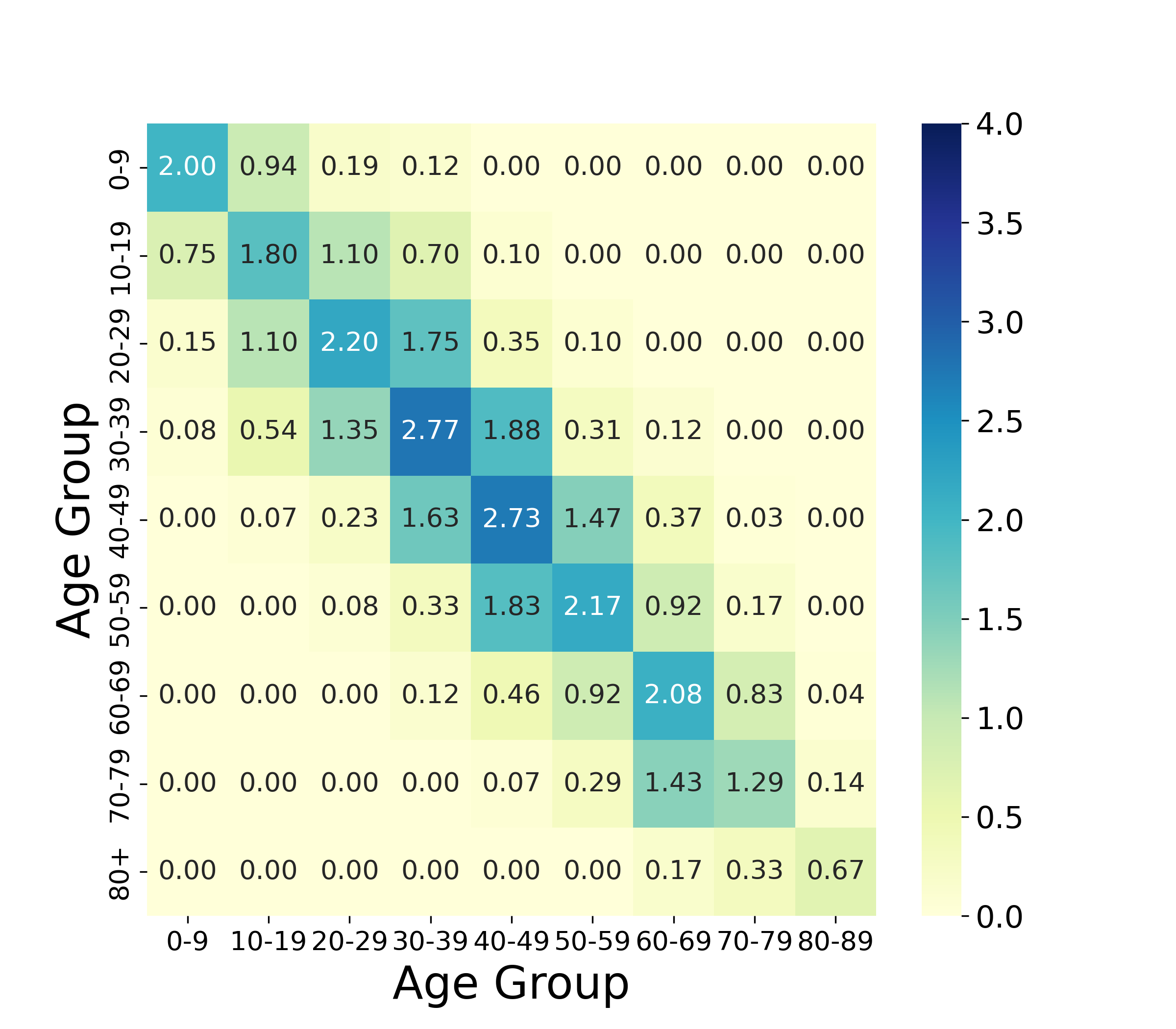}
	\end{minipage}}
  \hspace{-5mm}
		\subfigure[$HN-A_f$]{
		\begin{minipage}[b]{0.46\linewidth}
			\includegraphics[width=1\linewidth]{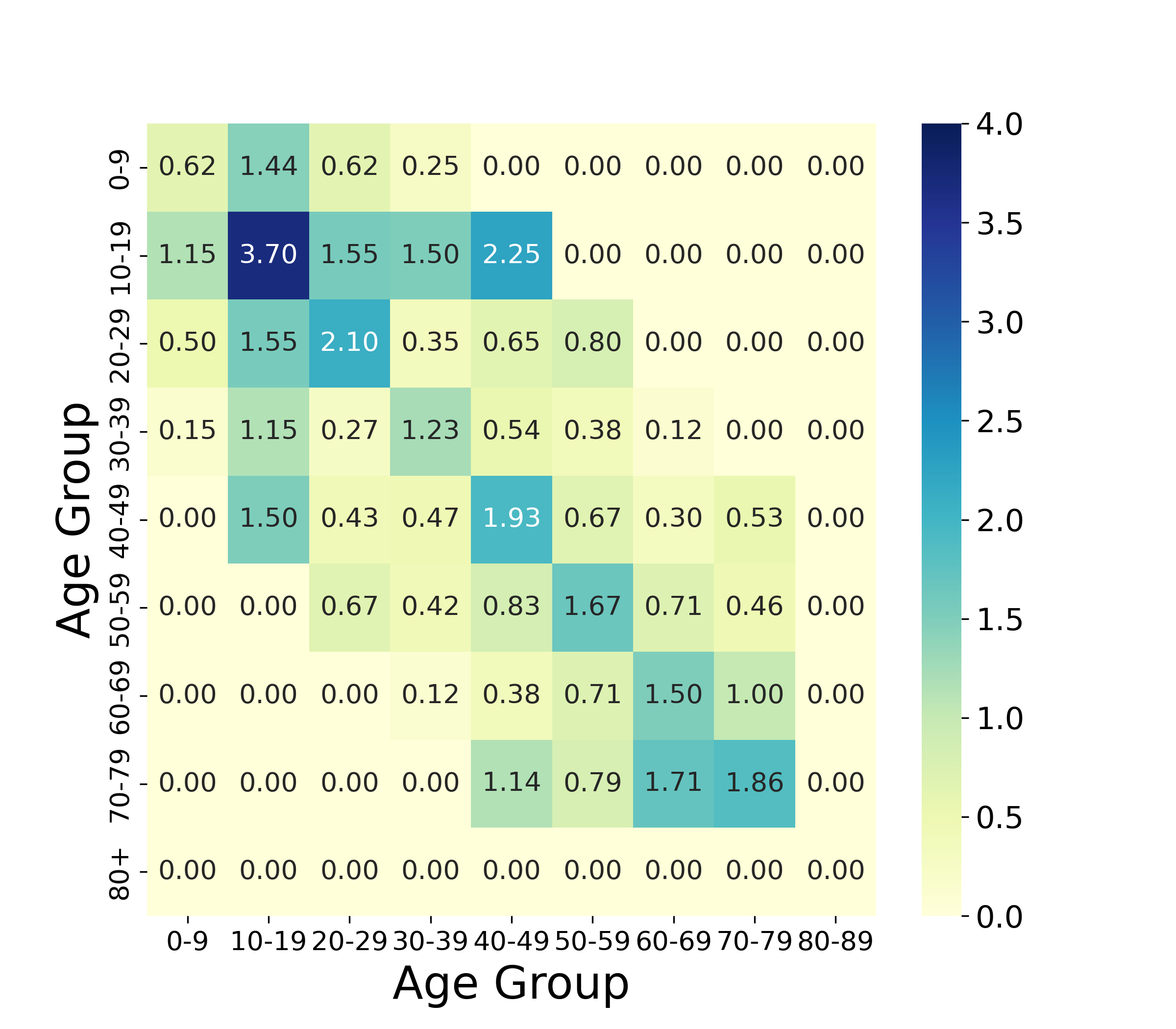}
	\end{minipage}}\\
  \vspace{-3mm}
		\subfigure[$HN-A_c-S$]{
		\begin{minipage}[b]{0.46\linewidth}
	\includegraphics[width=1\linewidth]{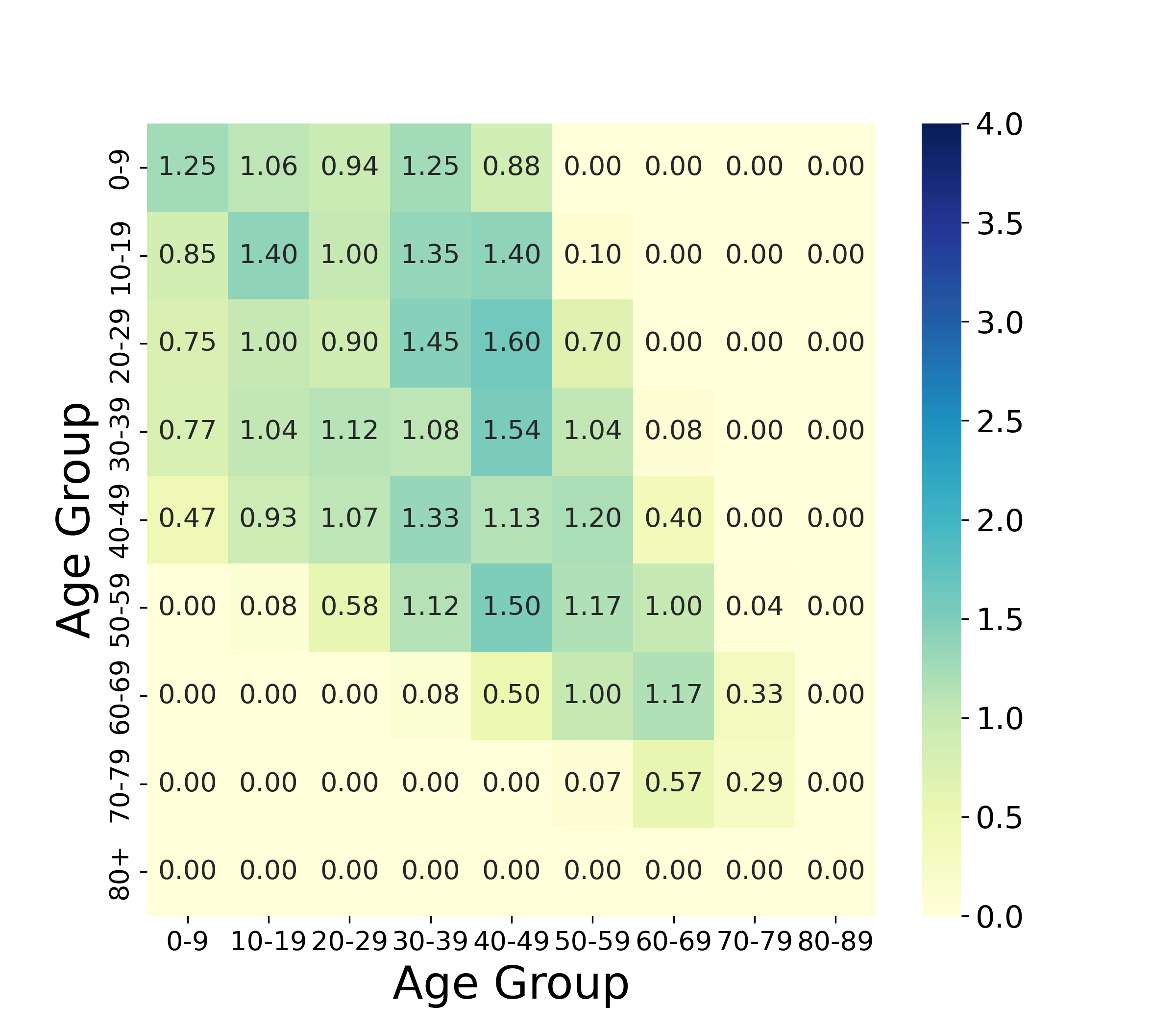}
	\end{minipage}}
  \hspace{-5mm}
		\subfigure[$HN-A_f-S$]{
		\begin{minipage}[b]{0.46\linewidth}
			\includegraphics[width=1\linewidth]{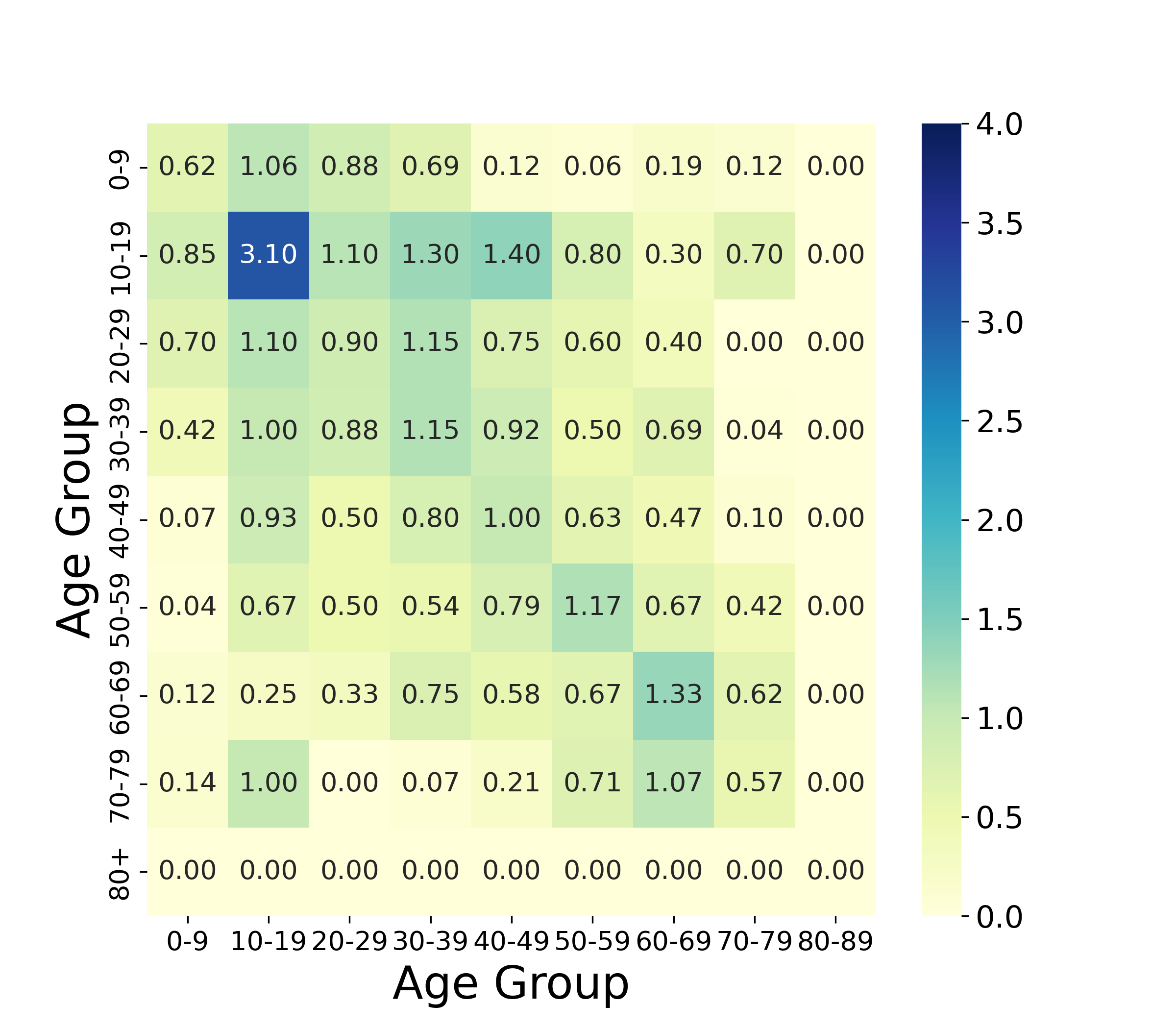}
	\end{minipage}}\\
	\caption{The target social contact matrix ((a) target) and the recreated social contact matrices of Germany, which are respectively generated by the (b) $RN$, (c) $HN-A_c$, (d) $HN-A_f$, (e) $HN-A_c-S$ and (f) $HN-A_f-S$ models.}
\label{GermanyMat}
\end{figure*}

\begin{figure*}[h!]
	\centering
	\subfigure[Target]{
		\begin{minipage}[b]{0.46\linewidth}
			\includegraphics[width=1\linewidth]{ItalyTarget.png}
	\end{minipage}}
  \hspace{-5mm}
	\subfigure[$RN$]{
		\begin{minipage}[b]{0.46\linewidth}
			\includegraphics[width=1\linewidth]{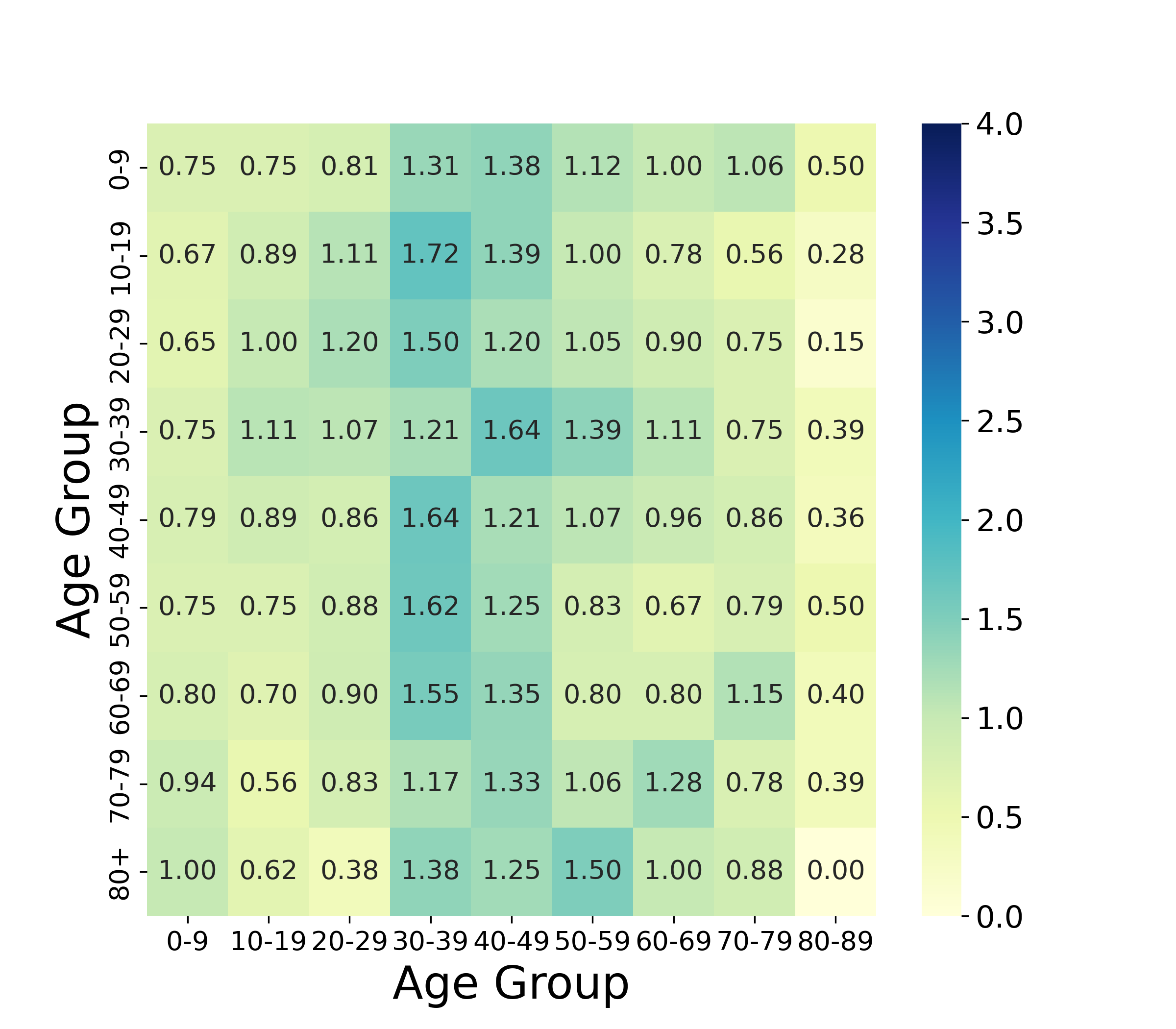}
	\end{minipage}}\\
  \vspace{-3mm}
	\subfigure[$HN-A_c$]{
		\begin{minipage}[b]{0.46\linewidth}
			\includegraphics[width=1\linewidth]{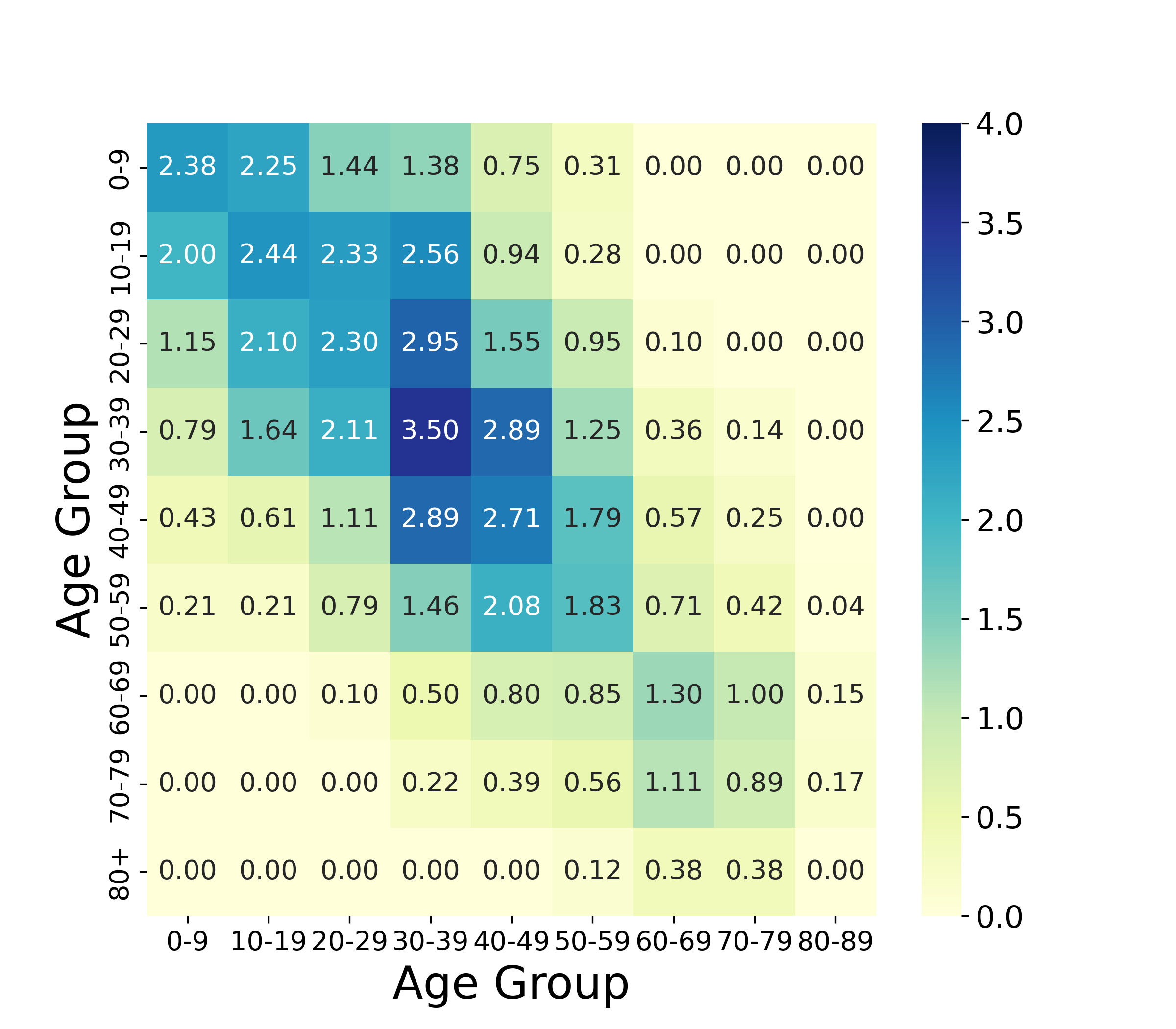}
	\end{minipage}}
  \hspace{-5mm}
		\subfigure[$HN-A_f$]{
		\begin{minipage}[b]{0.46\linewidth}
			\includegraphics[width=1\linewidth]{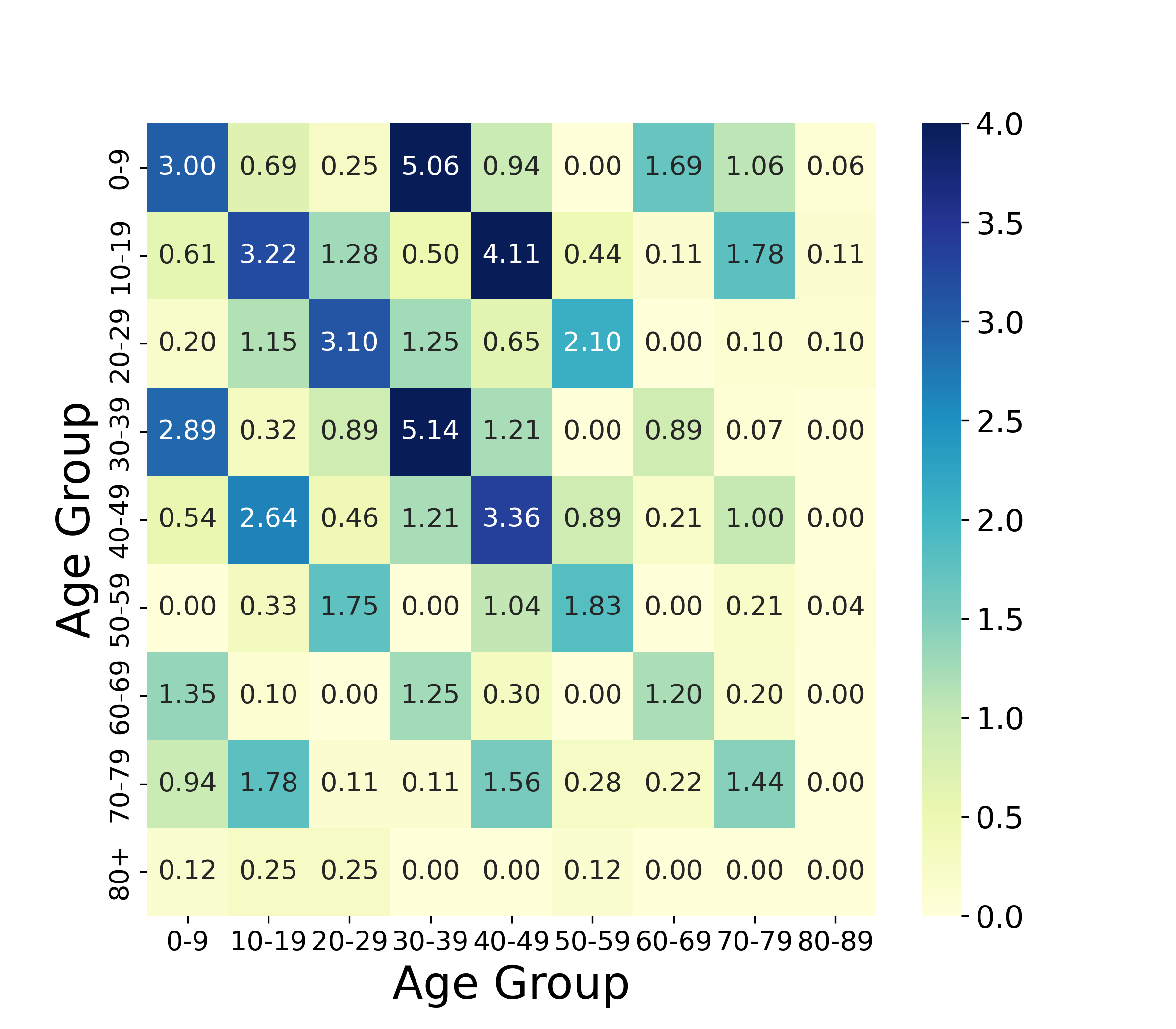}
	\end{minipage}}\\
  \vspace{-3mm}
		\subfigure[$HN-A_c-S$]{
		\begin{minipage}[b]{0.46\linewidth}
	\includegraphics[width=1\linewidth]{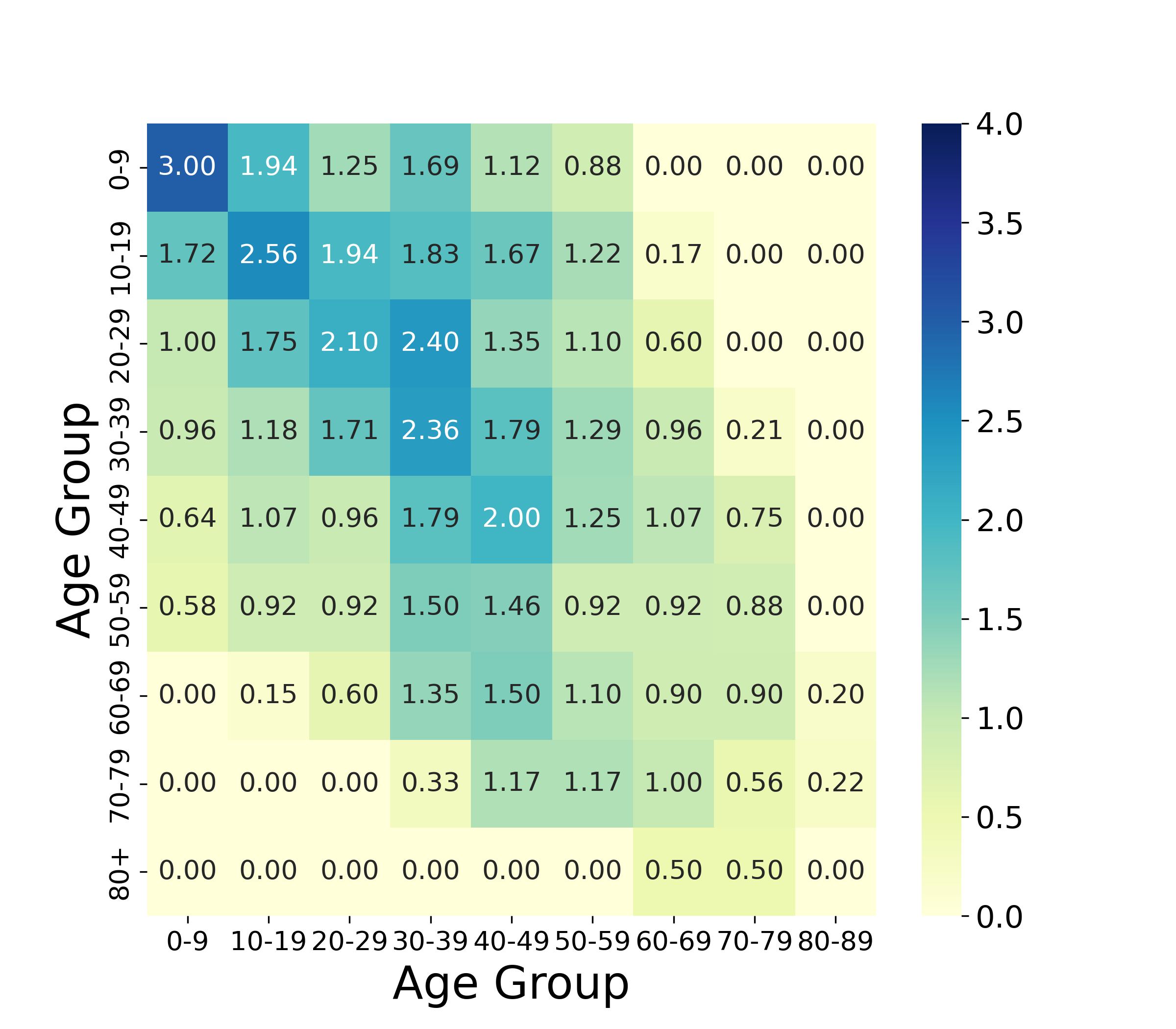}
	\end{minipage}}
  \hspace{-5mm}
		\subfigure[$HN-A_f-S$]{
		\begin{minipage}[b]{0.46\linewidth}
			\includegraphics[width=1\linewidth]{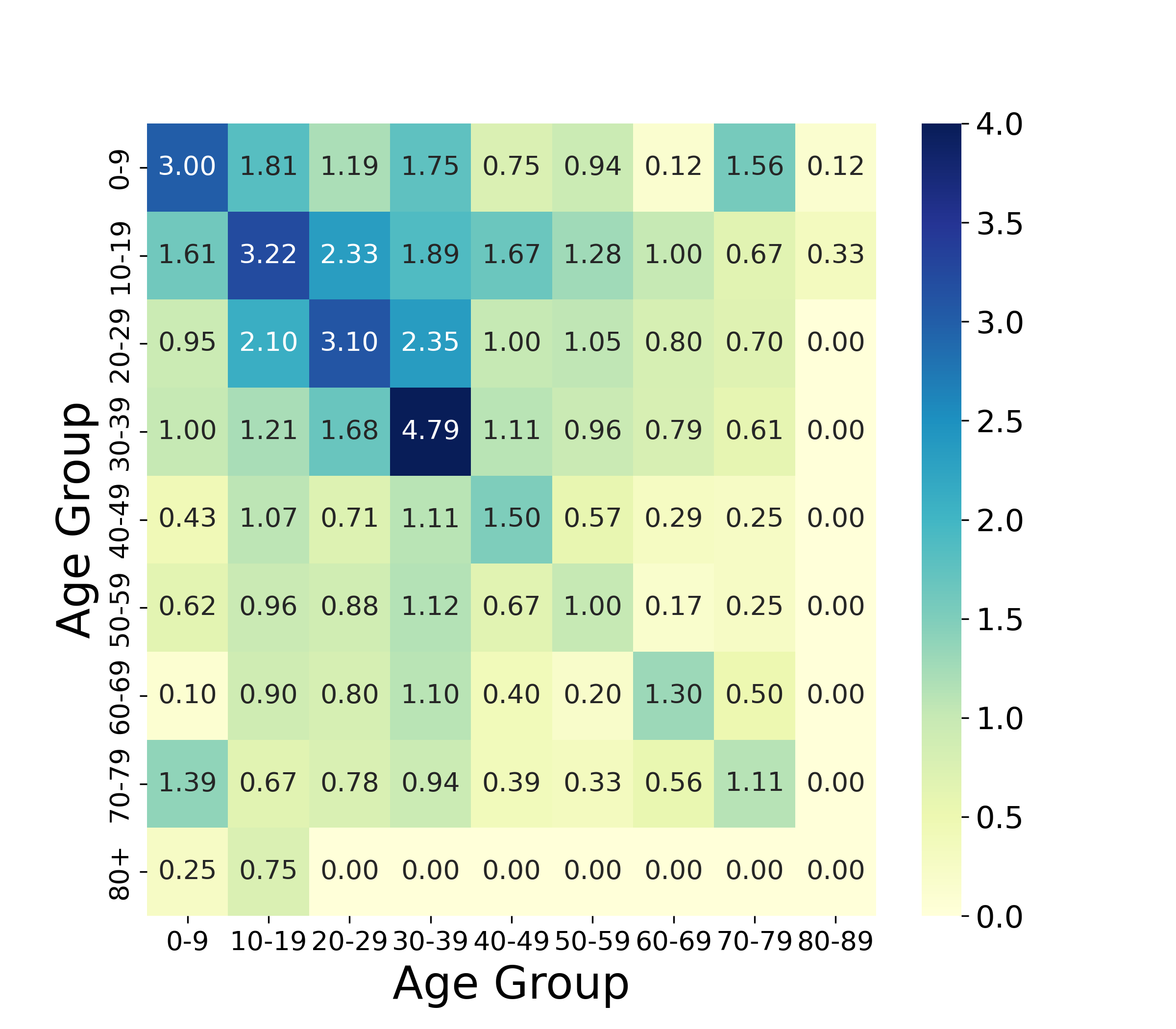}
	\end{minipage}}\\
	\caption{The target social contact matrix ((a) target) and the recreated social contact matrices of Italy, which are respectively generated by the (b) $RN$, (c) $HN-A_c$, (d) $HN-A_f$, (e) $HN-A_c-S$ and (f) $HN-A_f-S$ models.}
\label{ItalyMat}
\end{figure*}

\begin{figure*}[h!]
	\centering
	\subfigure[Target]{
		\begin{minipage}[b]{0.46\linewidth}
			\includegraphics[width=1\linewidth]{LuxembourgTarget.png}
	\end{minipage}}
  \hspace{-5mm}
	\subfigure[$RN$]{
		\begin{minipage}[b]{0.46\linewidth}
			\includegraphics[width=1\linewidth]{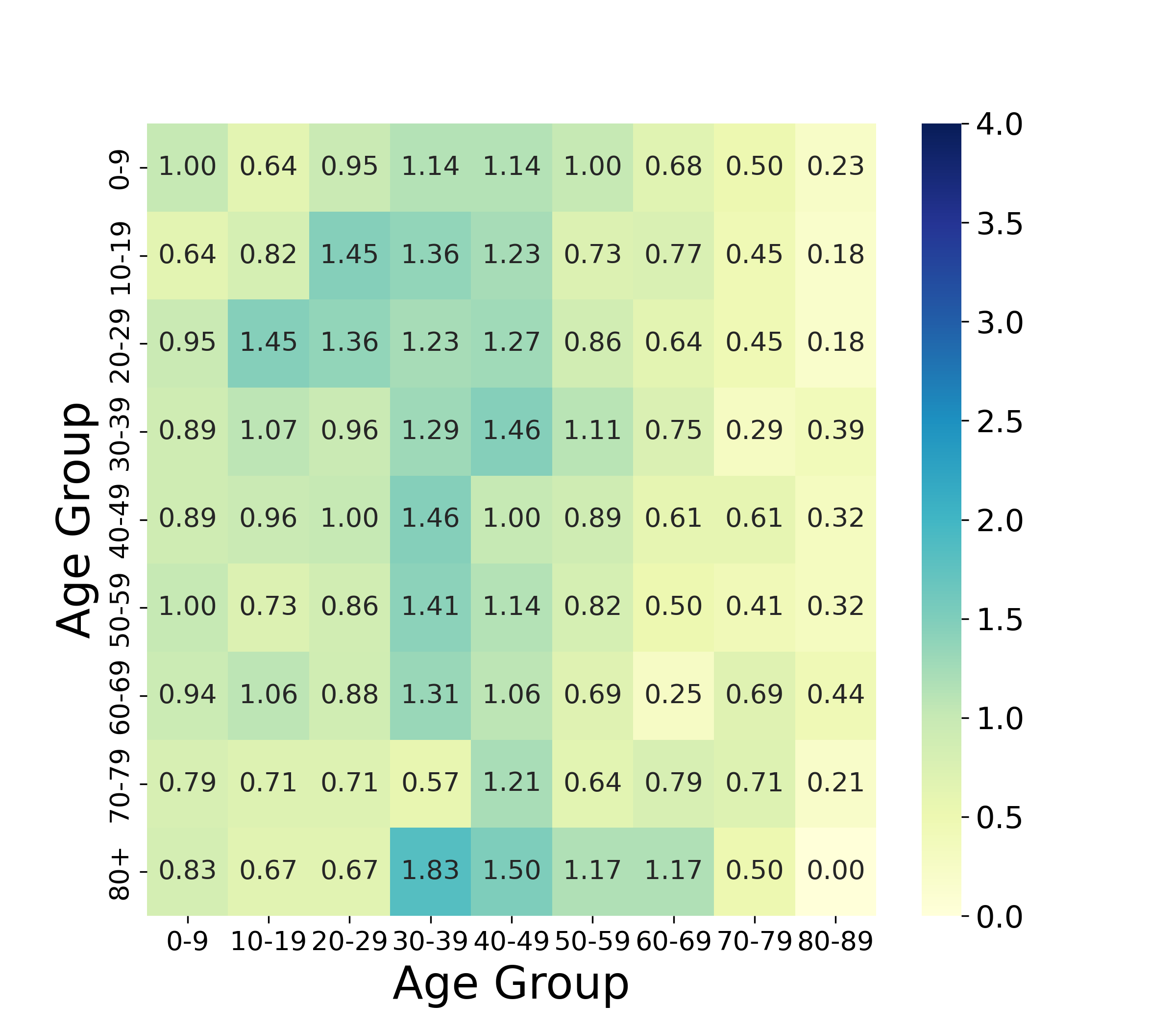}
	\end{minipage}}\\
  \vspace{-3mm}
	\subfigure[$HN-A_c$]{
		\begin{minipage}[b]{0.46\linewidth}
			\includegraphics[width=1\linewidth]{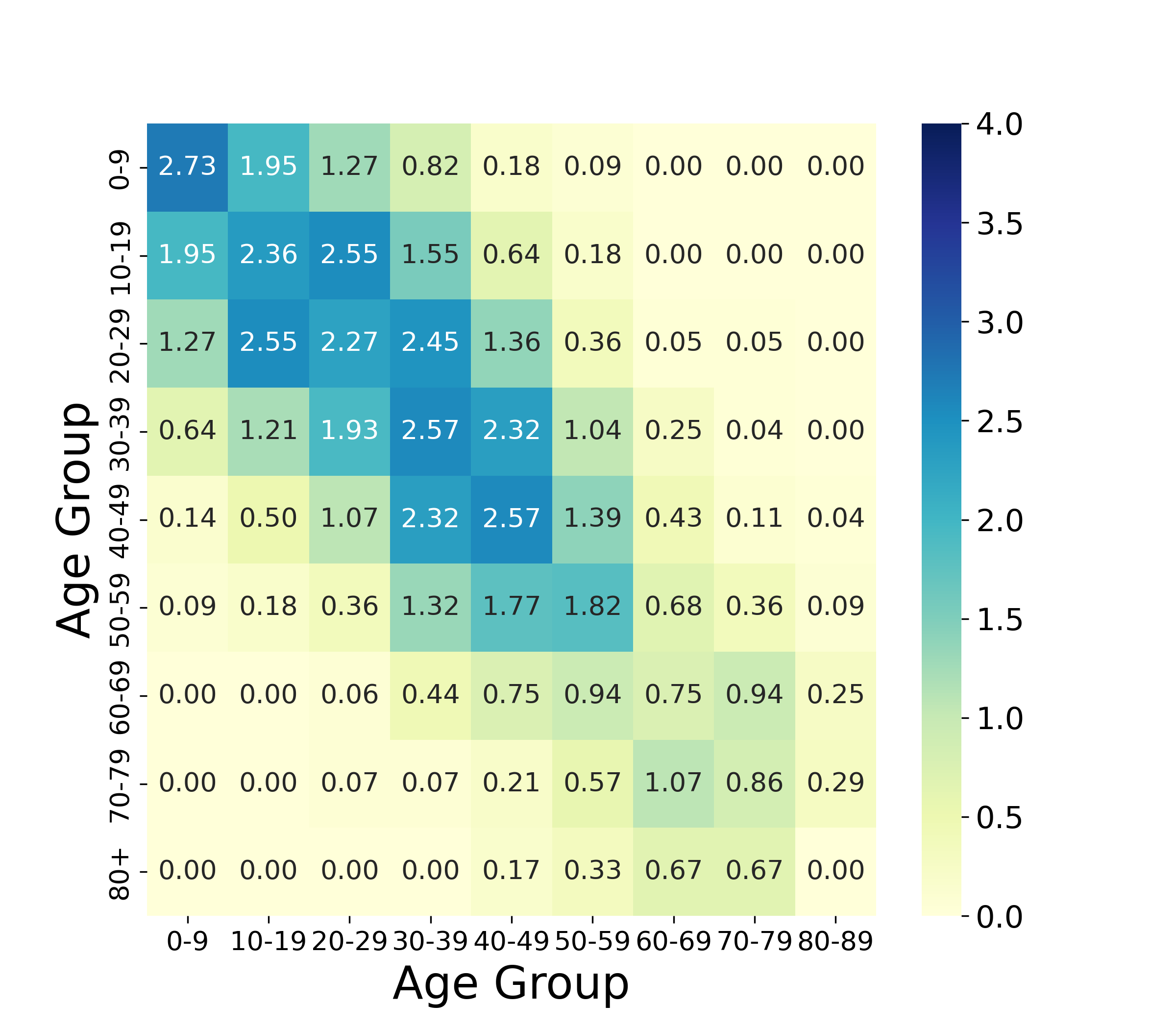}
	\end{minipage}}
  \hspace{-5mm}
		\subfigure[$HN-A_f$]{
		\begin{minipage}[b]{0.46\linewidth}
			\includegraphics[width=1\linewidth]{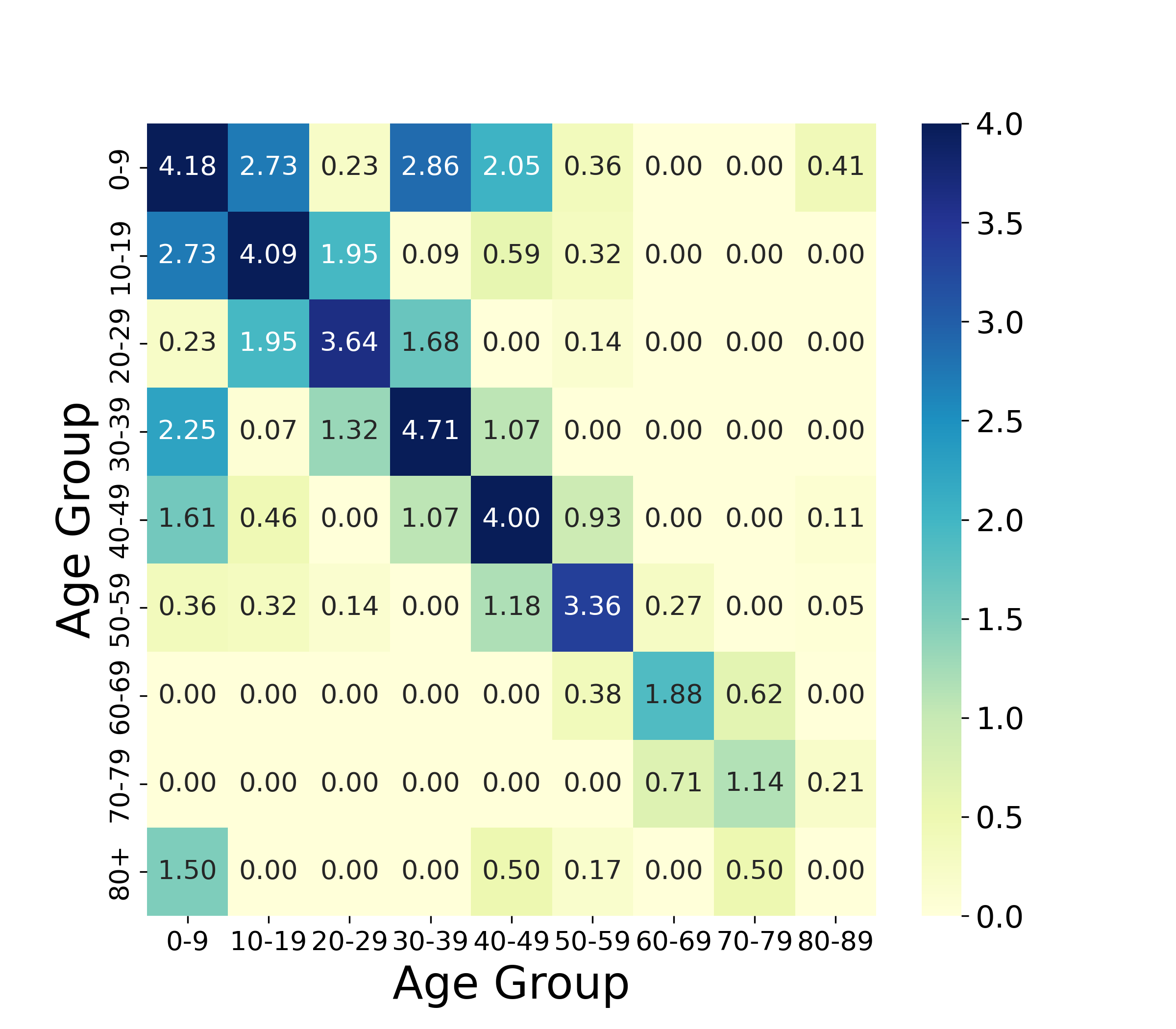}
	\end{minipage}}\\
  \vspace{-3mm}
		\subfigure[$HN-A_c-S$]{
		\begin{minipage}[b]{0.46\linewidth}
	\includegraphics[width=1\linewidth]{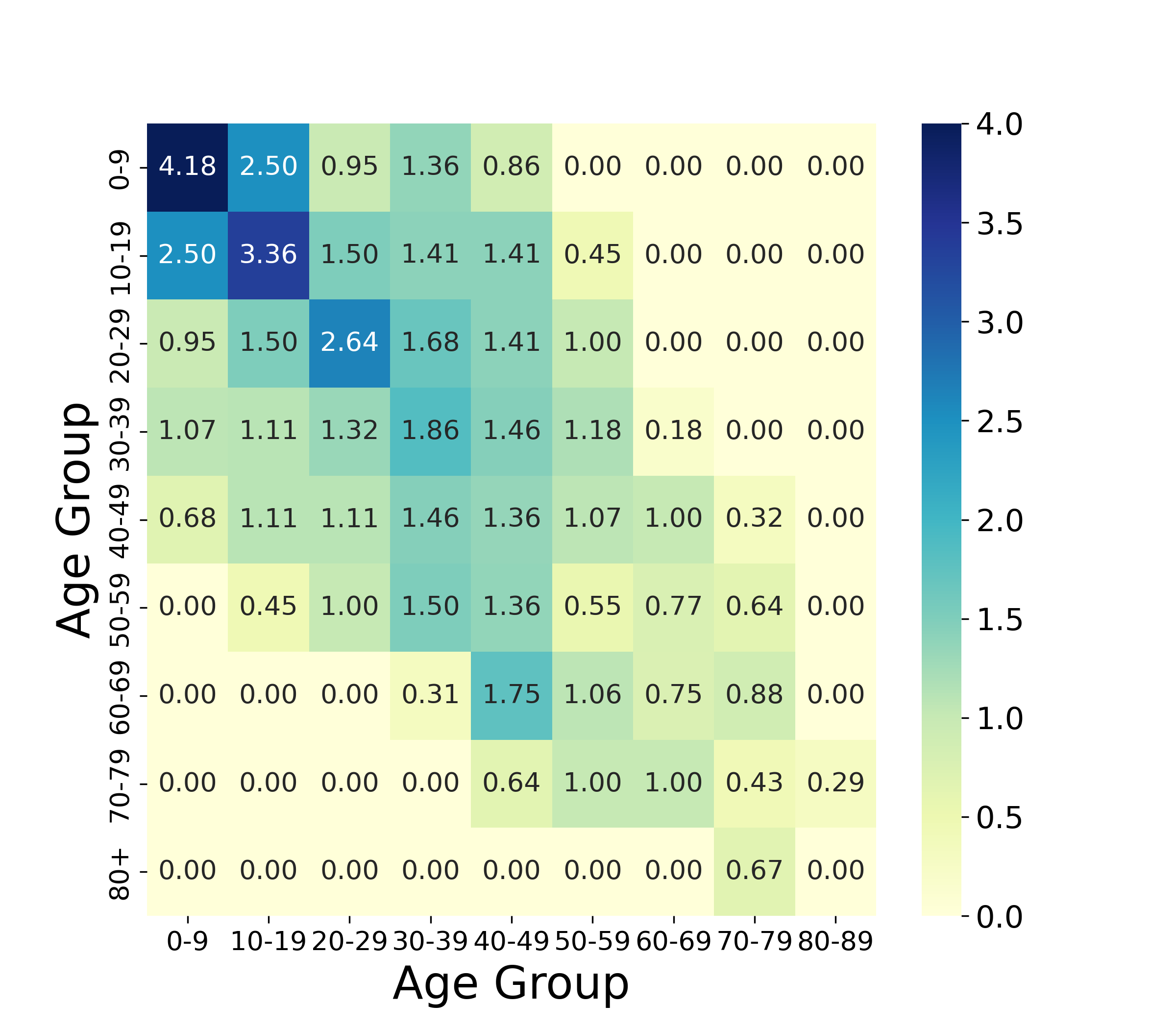}
	\end{minipage}}
  \hspace{-5mm}
		\subfigure[$HN-A_f-S$]{
		\begin{minipage}[b]{0.46\linewidth}
			\includegraphics[width=1\linewidth]{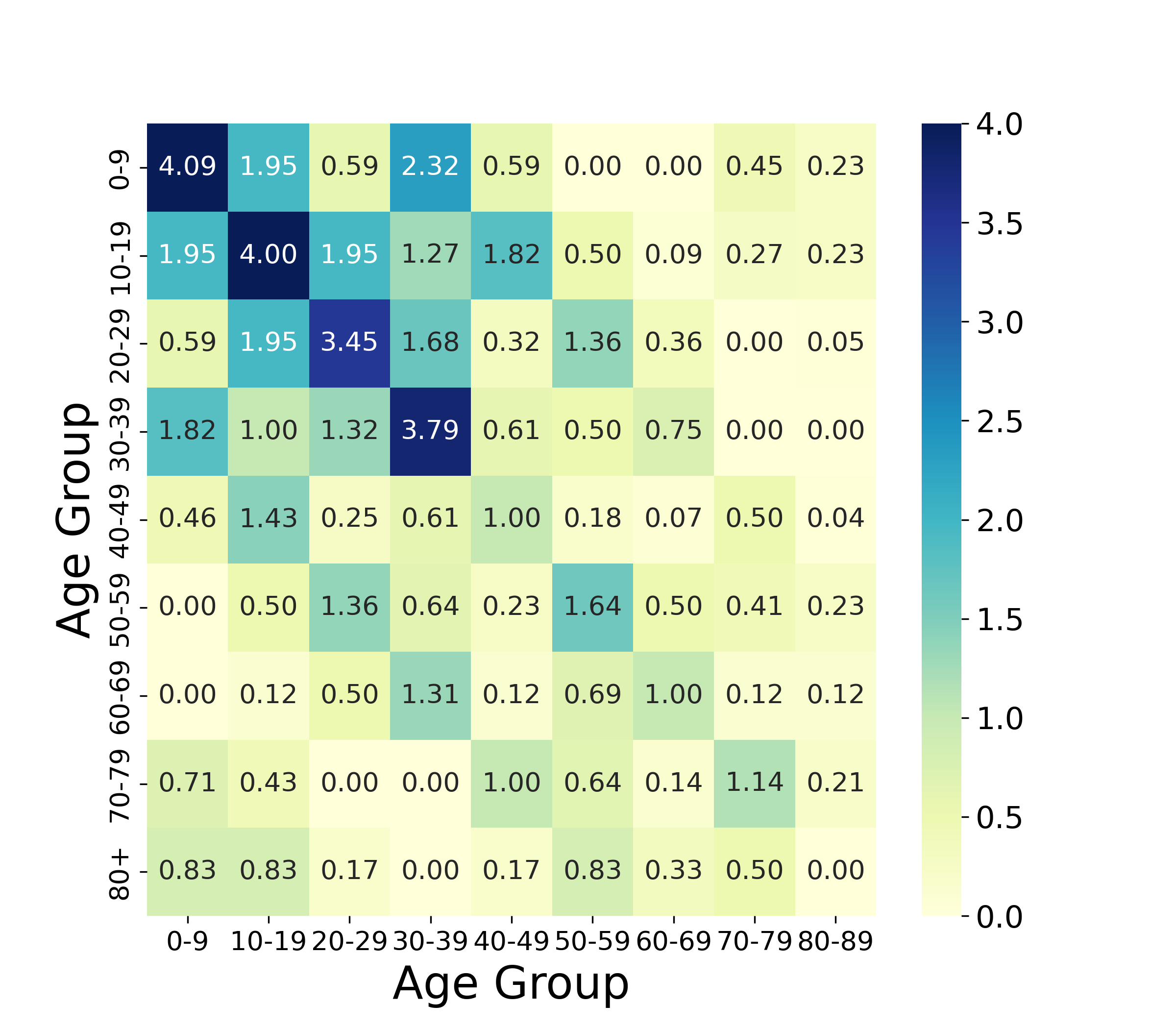}
	\end{minipage}}\\
	\caption{The target social contact matrix ((a) target) and the recreated social contact matrices of Luxembourg, which are respectively generated by the (b) $RN$, (c) $HN-A_c$, (d) $HN-A_f$, (e) $HN-A_c-S$ and (f) $HN-A_f-S$ models.}
\label{LuxembourgMat}
\end{figure*}

\begin{figure*}[h!]
	\centering
	\subfigure[Target]{
		\begin{minipage}[b]{0.46\linewidth}
			\includegraphics[width=1\linewidth]{PolandTarget.png}
	\end{minipage}}
  \hspace{-5mm}
	\subfigure[$RN$]{
		\begin{minipage}[b]{0.46\linewidth}
			\includegraphics[width=1\linewidth]{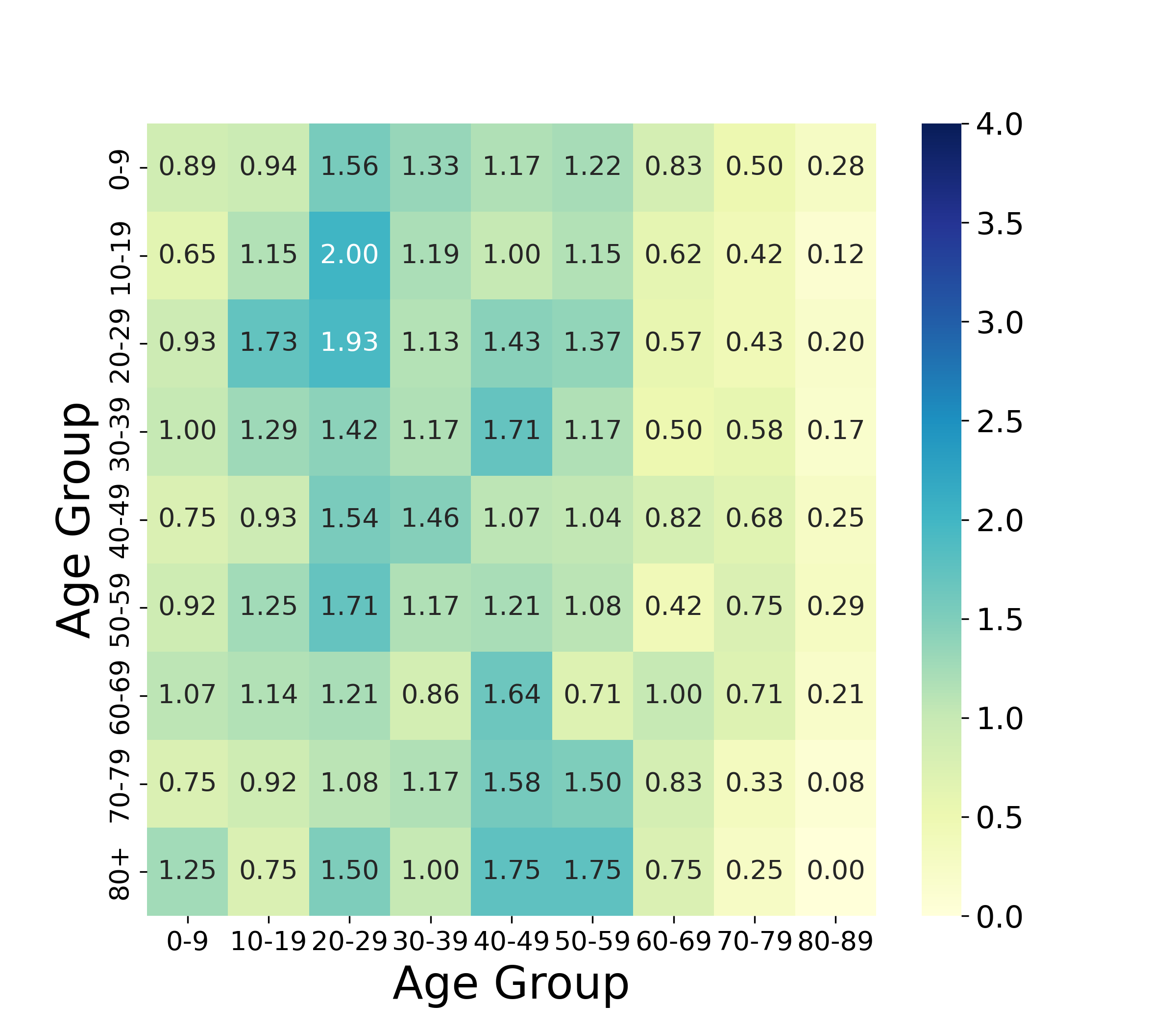}
	\end{minipage}}\\
  \vspace{-3mm}
	\subfigure[$HN-A_c$]{
		\begin{minipage}[b]{0.46\linewidth}
			\includegraphics[width=1\linewidth]{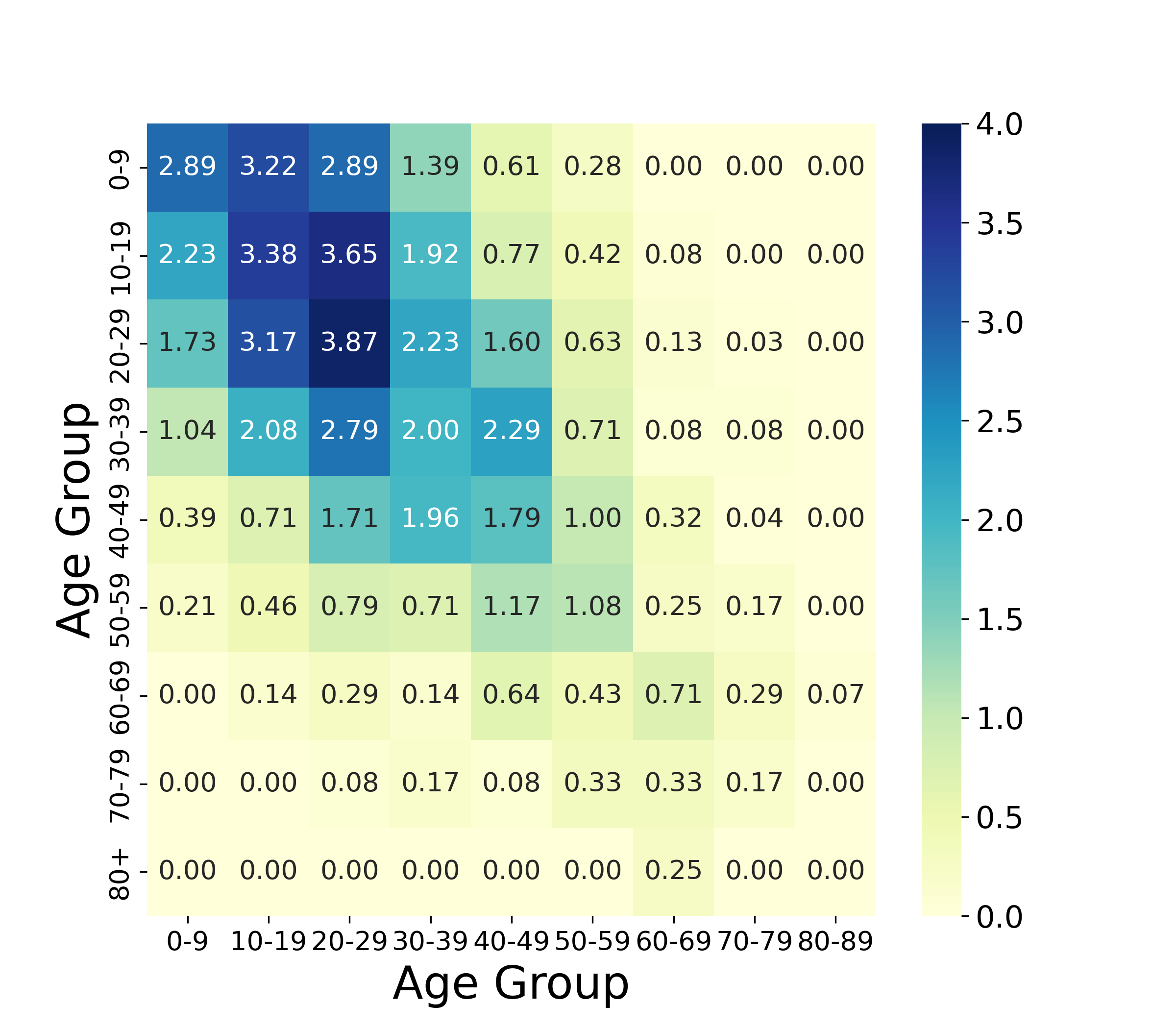}
	\end{minipage}}
  \hspace{-5mm}
		\subfigure[$HN-A_f$]{
		\begin{minipage}[b]{0.46\linewidth}
			\includegraphics[width=1\linewidth]{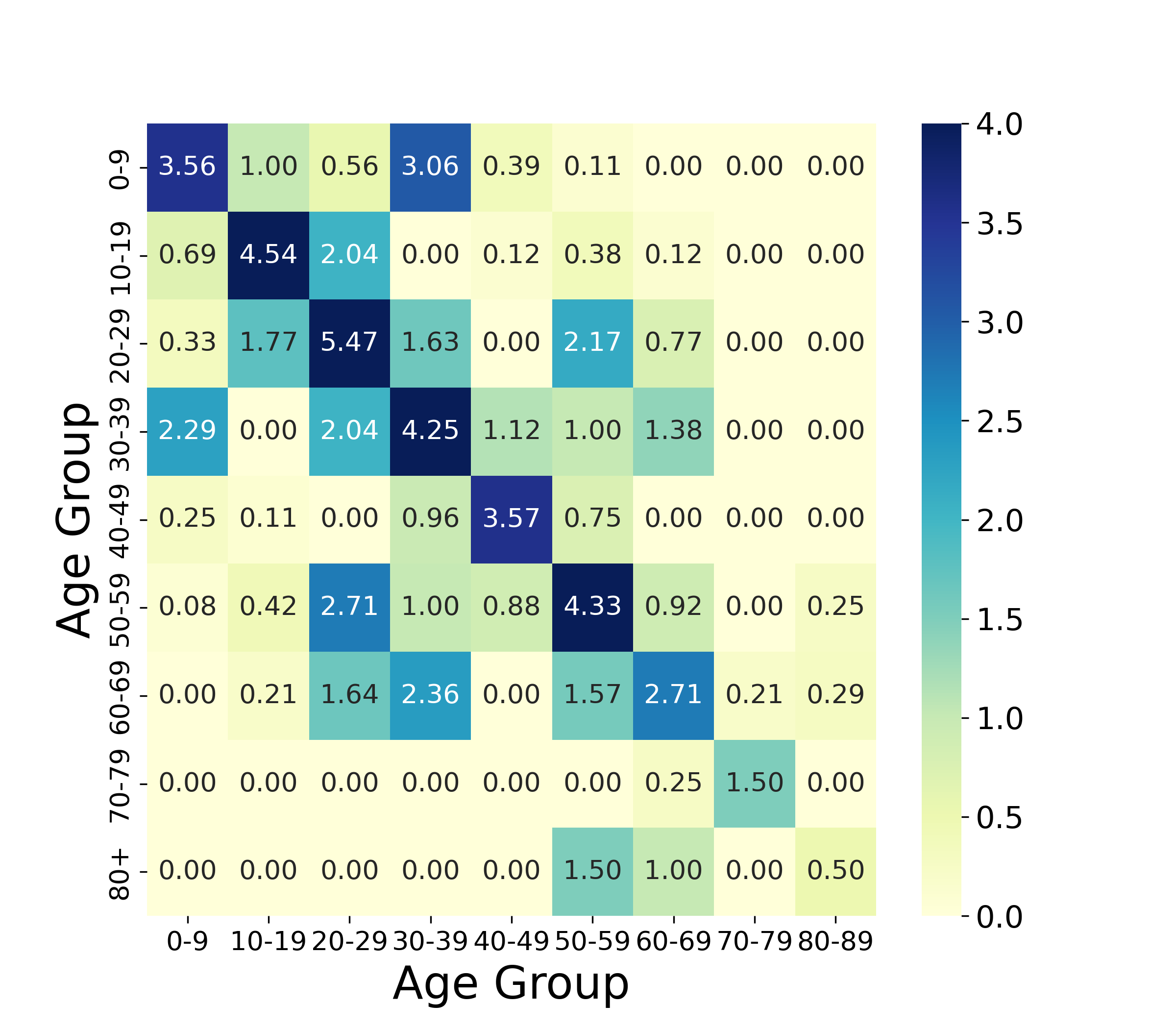}
	\end{minipage}}\\
  \vspace{-3mm}
		\subfigure[$HN-A_c-S$]{
		\begin{minipage}[b]{0.46\linewidth}
	\includegraphics[width=1\linewidth]{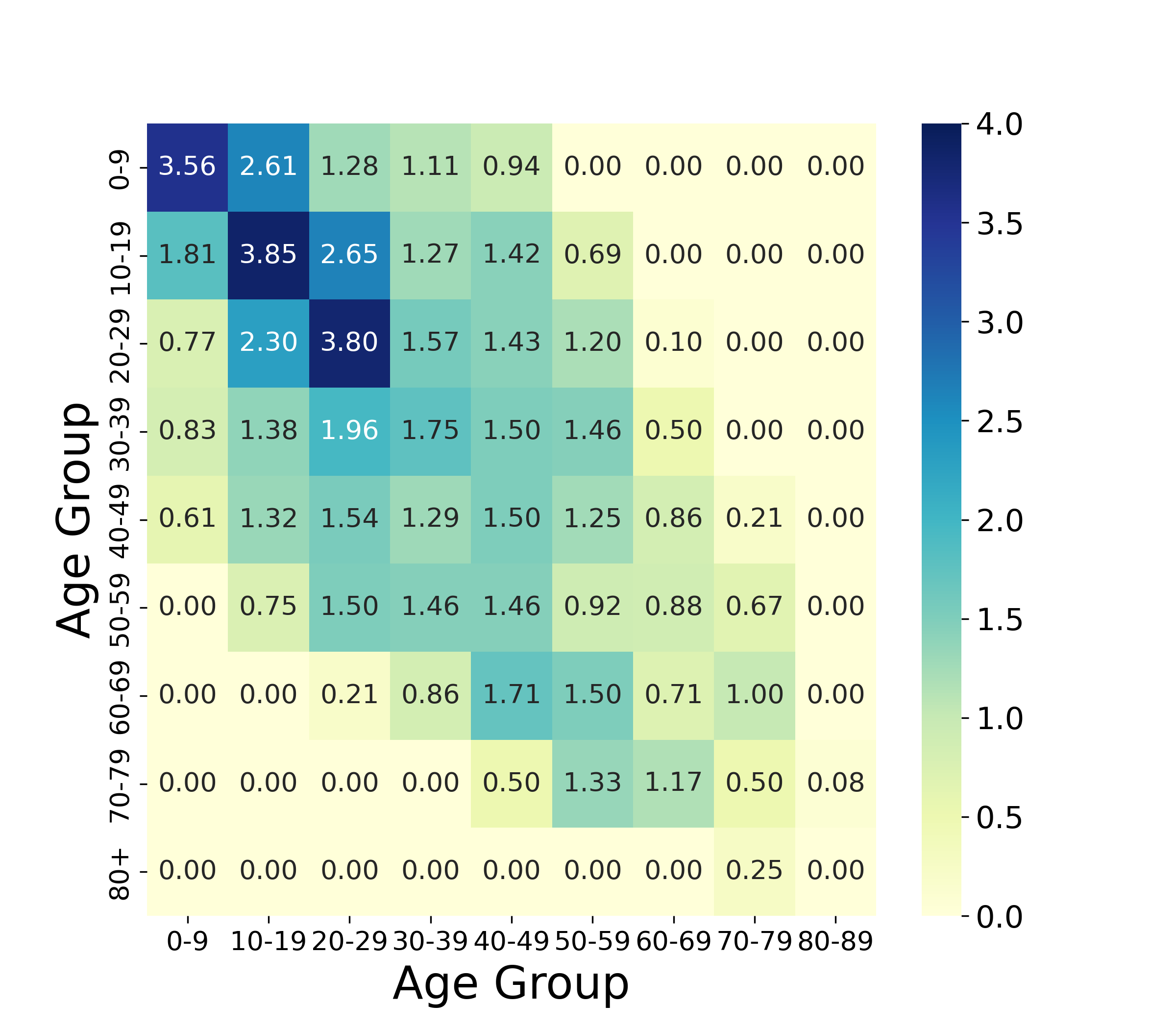}
	\end{minipage}}
  \hspace{-5mm}
		\subfigure[$HN-A_f-S$]{
		\begin{minipage}[b]{0.46\linewidth}
			\includegraphics[width=1\linewidth]{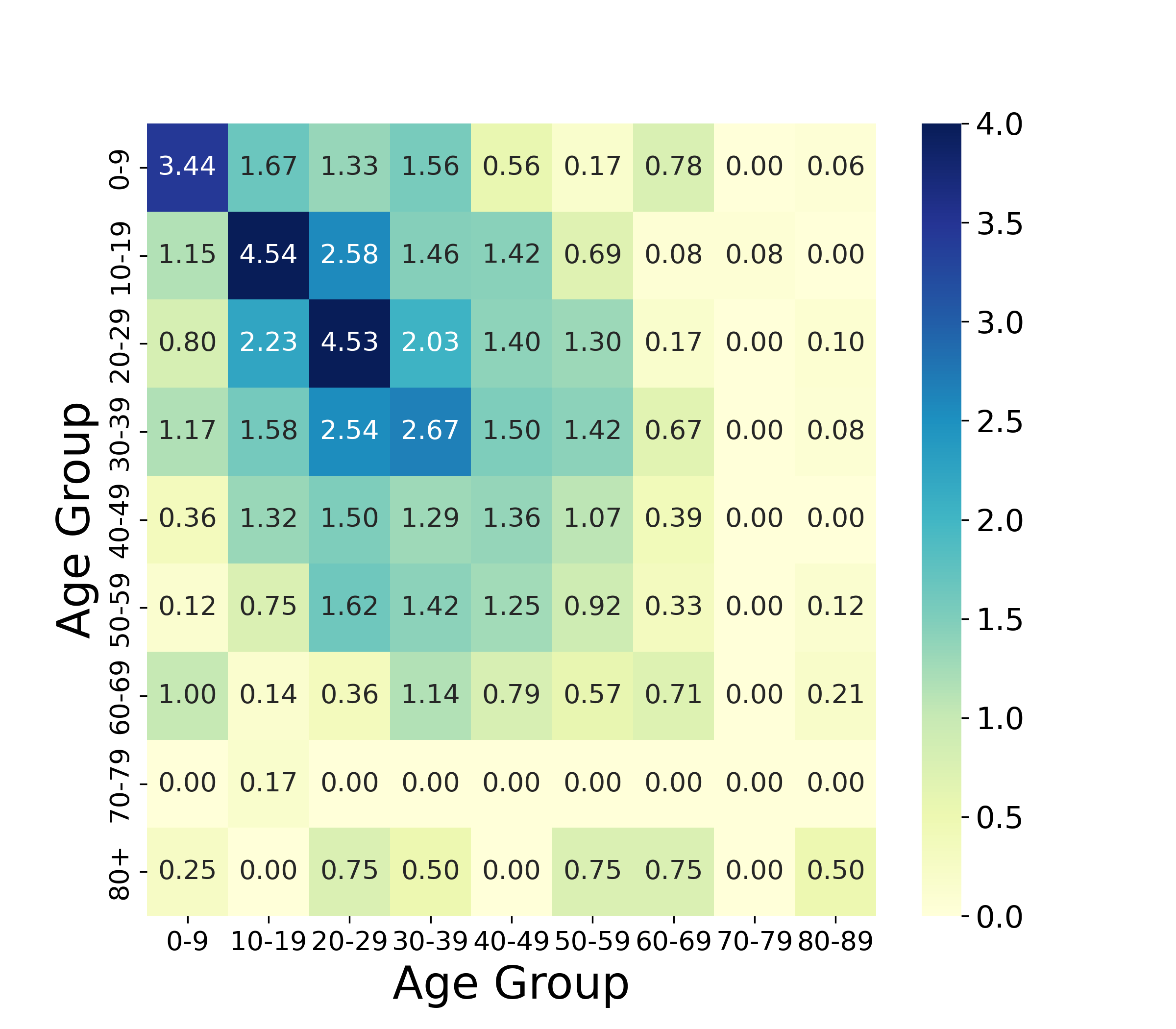}
	\end{minipage}}\\
	\caption{The target social contact matrix ((a) target) and the recreated social contact matrices of Poland, which are respectively generated by the (b) $RN$, (c) $HN-A_c$, (d) $HN-A_f$, (e) $HN-A_c-S$ and (f) $HN-A_f-S$ models.}
\label{PolandMat}
\end{figure*}

\section{Parameter selection based on sensitivity analysis}
\label{paramselect}
\begin{figure*}[h!]
	\centering
  	\subfigure[Parameter rankings based on the log-scale parameter sensitivity]{
		\begin{minipage}[b]{0.94\linewidth}
			\includegraphics[width=1\linewidth]{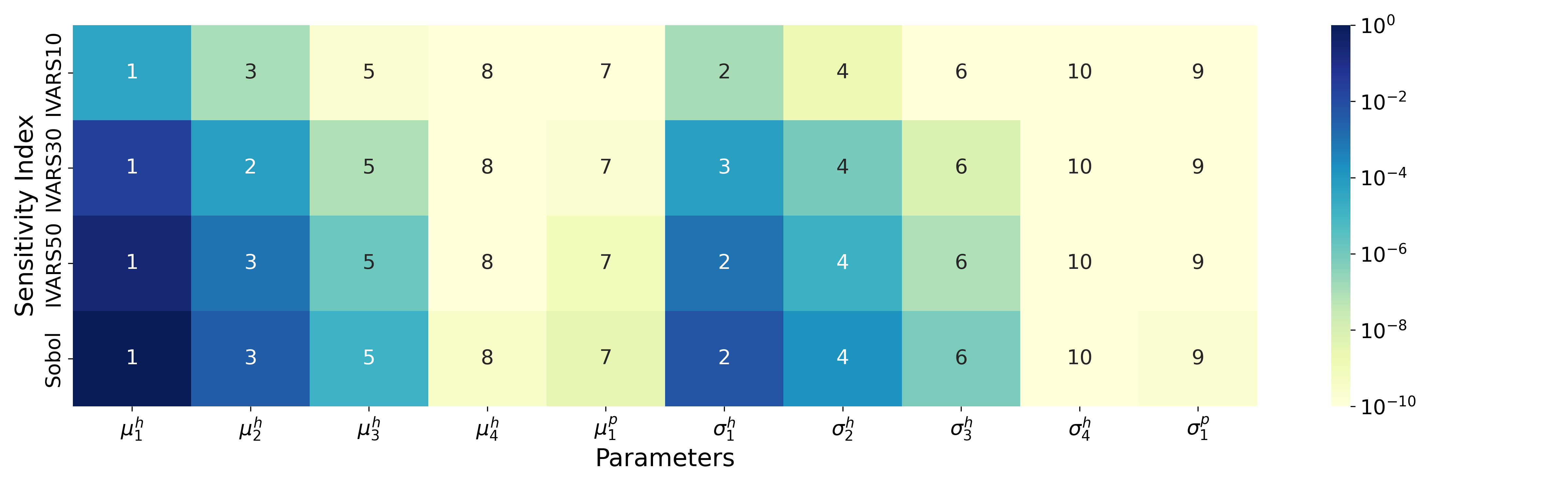}
	\end{minipage}}\\
	\subfigure[Initial representation of age]{
		\begin{minipage}[b]{0.46\linewidth}
			\includegraphics[width=1\linewidth]{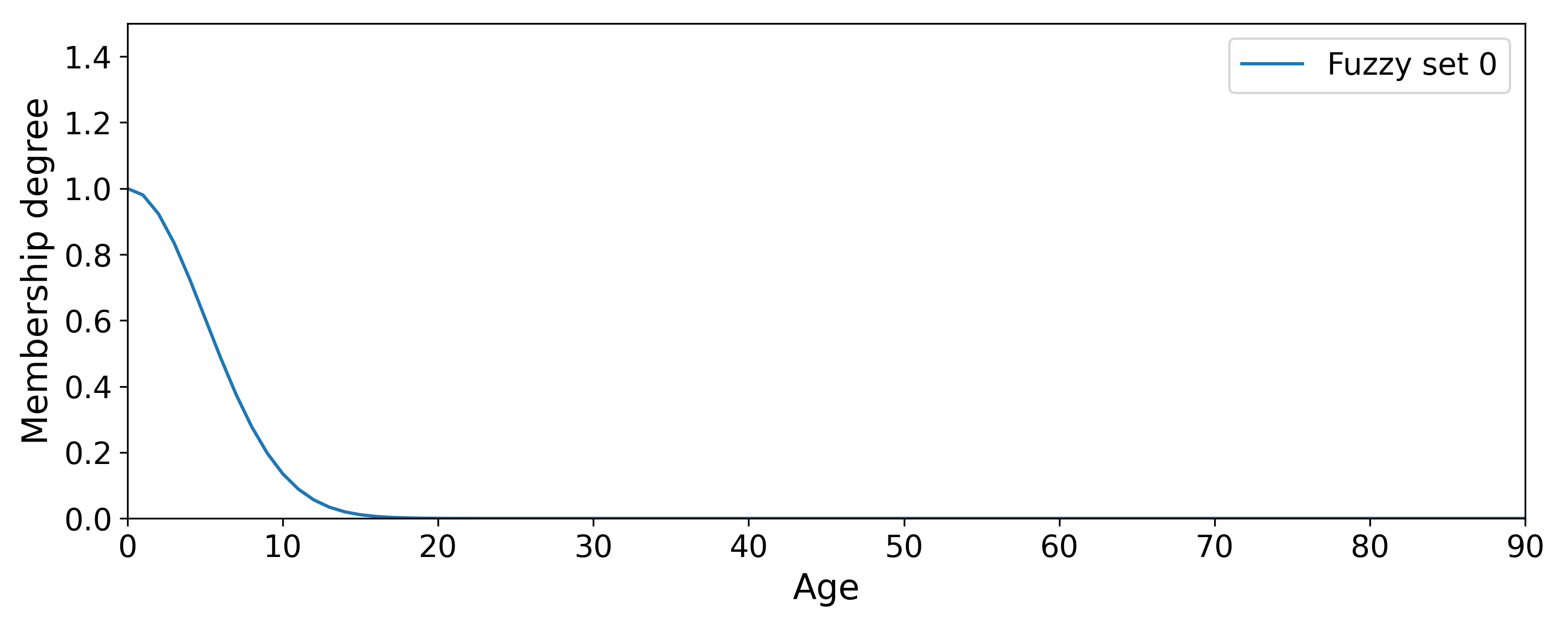}
	\end{minipage}}
	\subfigure[Initial representation of age difference]{
		\begin{minipage}[b]{0.46\linewidth}
			\includegraphics[width=1\linewidth]{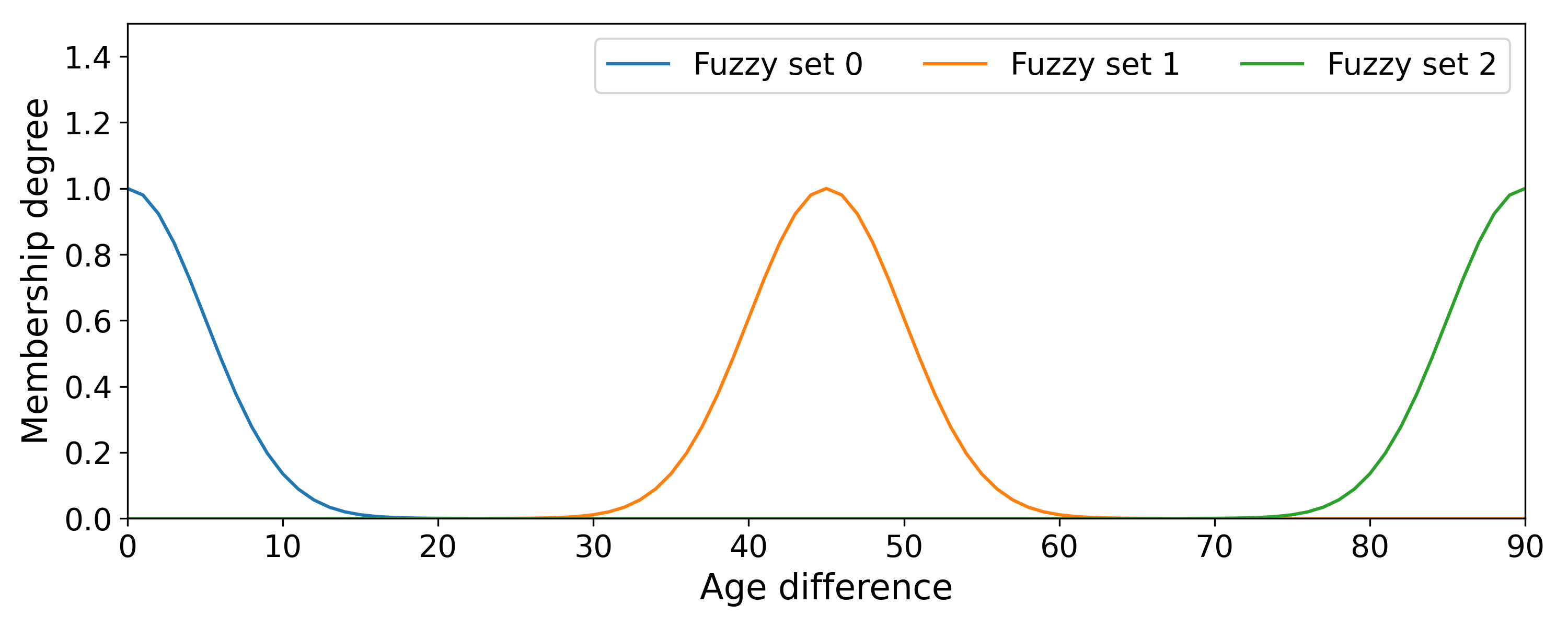}
	\end{minipage}}\\
	\subfigure[Optimised representation of age]{
		\begin{minipage}[b]{0.46\linewidth}
			\includegraphics[width=1\linewidth]{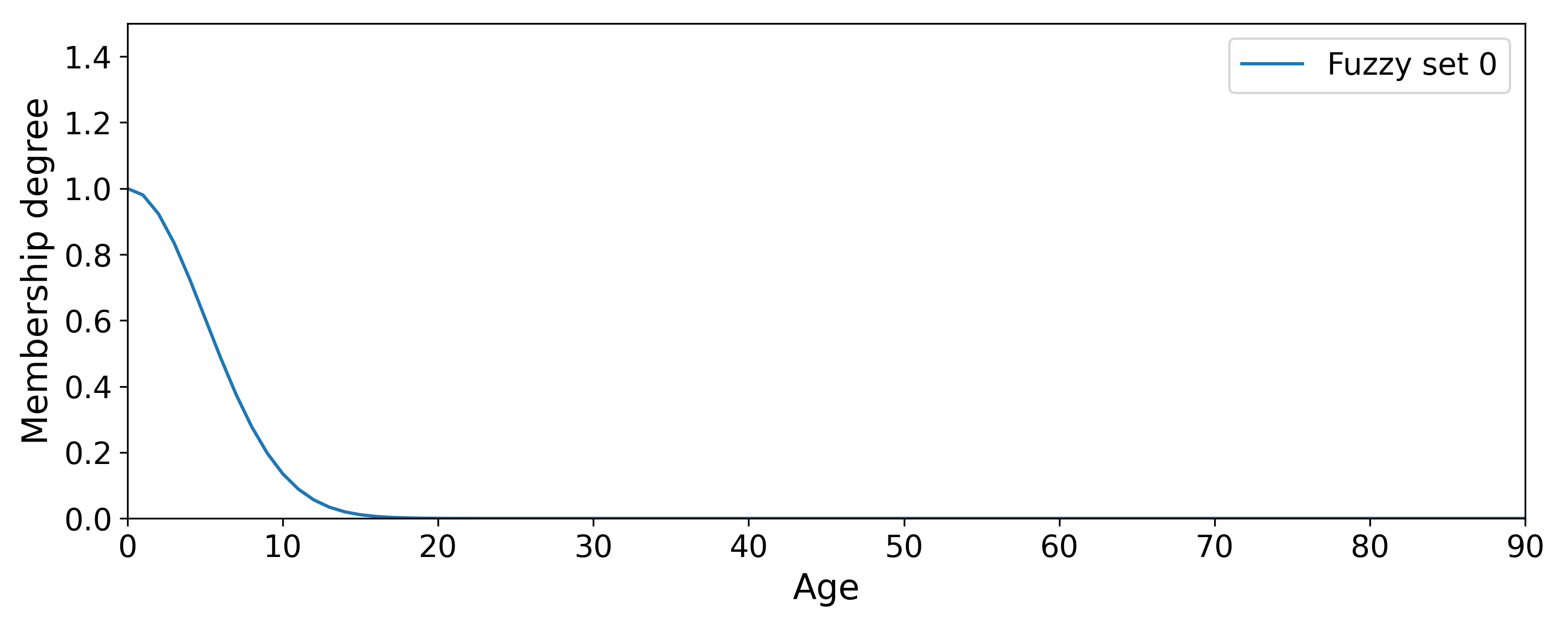}
	\end{minipage}}
	\subfigure[Optimised representation of age difference]{
		\begin{minipage}[b]{0.46\linewidth}
			\includegraphics[width=1\linewidth]{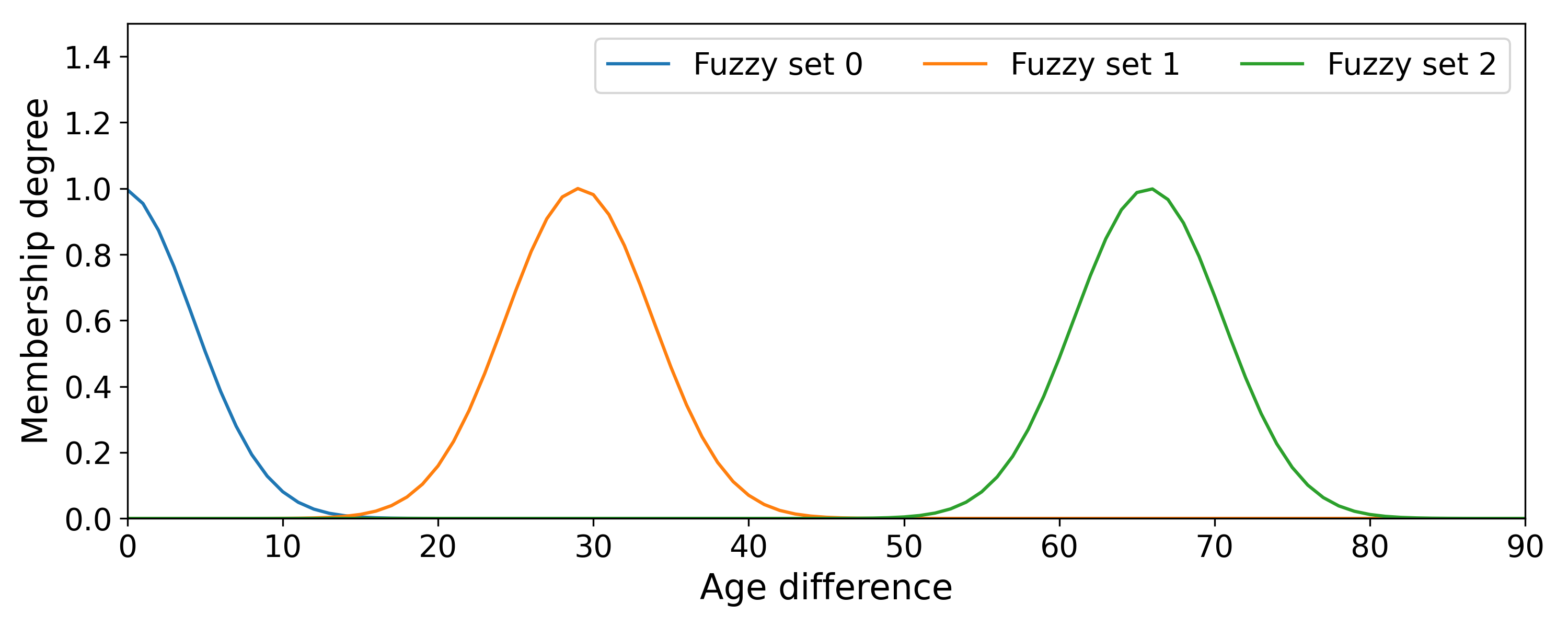}
	\end{minipage}}\\
	\caption{The parameter sensitivity level (in log scale) and the corresponding rankings for parameters of fuzzy sets based on the $IVARS_{10}$, $IVARS_{30}$, $IVARS_{50}$ and $Sobol$(Fig.~(a)), the initialised fuzzy representation of age and age difference (Fig.~(b) and Fig.~(c)) and the optimised fuzzy representation of age and age difference (Fig.~(d) and Fig.~(e)) for Belgium. In Fig.~(a), parameters for the first fuzzy set of age difference ($\mu_1^h$ and $\sigma_1^h$), parameters for the second fuzzy set of age difference ($\mu_2^h$ and $\sigma_2^h$) and parameters of the third fuzzy set of age difference ($\mu_3^h$ and $\sigma_3^h$) rank high considering all types of sensitivity indexes. Therefore, they are optimised in model calibration. The optimised fuzzy representation principles are shown in Fig.~(d) and Fig.~(e).}
\label{fuzzyBselectparam}
\end{figure*}

Fig.~\ref{fuzzyBselectparam} (a) shows the ranks and the values of each parameter's sensitivity metrics (in log scale), including the $IVARS_{10}$, $IVARS_{30}$, $IVARS_{50}$ and $Sobol$ measures for the parameters of the fuzzy sets (one fuzzy set to represent age and six fuzzy sets to represent age difference) for Belgium. The sensitivity metrics enable the same parameter rankings and presents a decreasing impact of the $\mu^h_i$ and $\sigma^h_i$ values as their focus moves from age differences around $0$ to larger age differences. The log value of sensitivity levels decreases as $IVARS_{10}$ transits to $IVARS$ at larger scales, which phenomenon can also be seen in the experiments on other countries. More specifically, $\mu_1^h=0$ and $\sigma_1^h=5$ are the parameters of the Gaussian Membership function for fuzzy set that initialise the age difference around the focus of $0$ within the spread around $[-3\sigma_1^h,3\sigma_1^h]$ (initialised around the range of [-15,15]). They significantly impact the model performance, as characterised with much higher importance than that of other parameters considering all types of sensitivity indices. In addition, $\mu_2^h$ and $\sigma_2^h$ (parameters for fuzzy set that describes the age difference around $30$) and $\mu_3^h$ and $\mu_3^h$ (parameters for fuzzy set that describes the age difference around $60$) also rank high in terms of each sensitivity index. This indicates that tuning the parameters of $\mu_1^h$, $\sigma_1^h$, $\mu_2^h$, $\sigma_2^h$, $\mu_3^h$ and $\sigma_3^h$ can leads to a more significant change of $EU$ distance. Their optimisation can potentially improve the model performance. We optimise the $\mu_1^h$, $\sigma_1^h$, $\mu_2^h$, $\sigma_2^h$, $\mu_3^h$ and $\sigma_3^h$ values based on the initialised parameter set up of the fuzzy sets. As shown in Fig.~\ref{fuzzyBselectparam} (c) and Fig.~\ref{fuzzyBselectparam} (e), the initialised fuzzy set defined with $\mu^h_1$ and $\sigma_1^h$, which describes the uncertain induction of age difference around $0$, is optimised with a less extensive spread from $5.00$ to $4.65$ and the transition of the fuzzy set's location to the left from $0.00$ to $-0.4198$. The negative values of age difference do not make sense but induce a smaller membership values based on the Gaussian Membership functions and adjusts the model performance. More specifically, based on the changes mentioned above, people in Belgium tend to assign smaller membership degrees to any age difference over $0$ for the first fuzzy set.The second fuzzy set of age difference, initialised at around $\mu_2^h=30.00$ with a standard deviation at $\sigma_h^2=5$, centres around $\mu_2^h=29.80$ with a less extensive spread at $\sigma_h^2 = 4.74$ after optimisation. The third fuzzy set of age difference, initialised at around $\mu_3^h=60.00$ with a standard deviation at $\sigma_h^2=5$, centres around $\mu_3^h=65.75$ with a smaller spread at $\sigma_h^2 = 4.79$ after optimisation. The changes mentioned above imply that people in Belgium tend to assign less membership degrees to the age differences at around $0$ and higher membership degrees to the age difference at around $29$ and $65$ (rather than $30$ or $60$). This indicates that people in Belgium pay more attention to their peers or people around their parents'/grandparents' age. They assign larger membership degrees to these differences and consider them when making social contact. These changes finally lead to a decrease in the $EU$ distance from $3.74$ to $3.47$ (see Tab.~\ref{summary}).  

\begin{figure}[h!]
	\centering
  	\subfigure[Parameter rankings based on the log-scale parameter sensitivity]{
		\begin{minipage}[b]{0.94\linewidth}
			\includegraphics[width=1\linewidth]{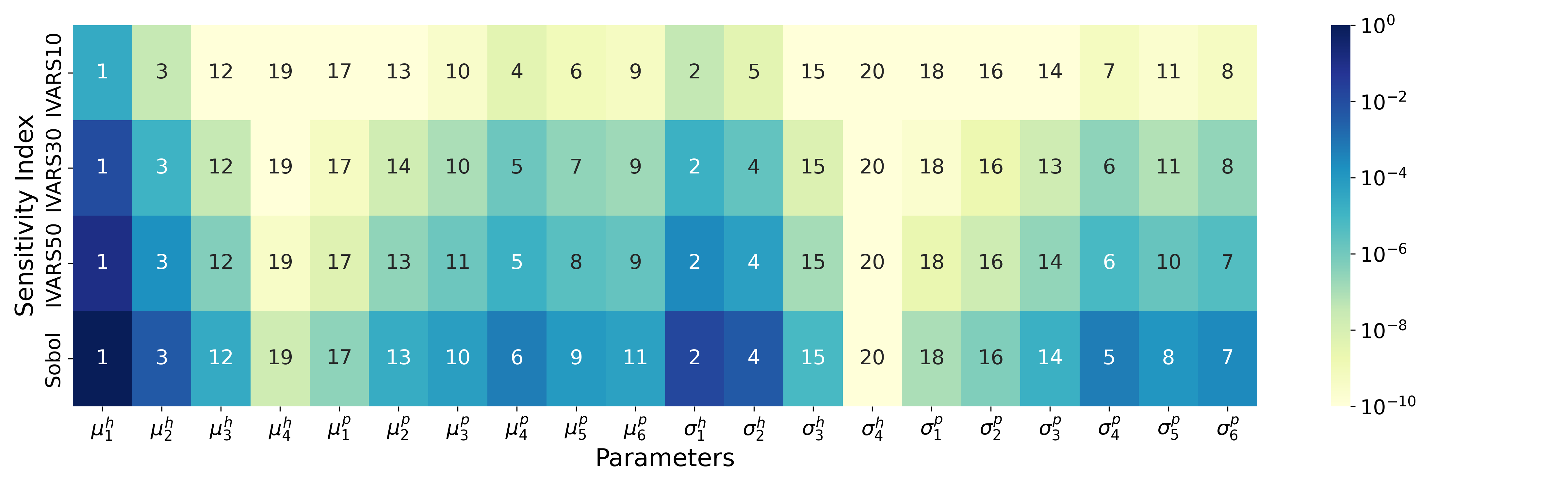}
	\end{minipage}}
	\subfigure[Initial representation of age]{
		\begin{minipage}[b]{0.46\linewidth}
			\includegraphics[width=1\linewidth]{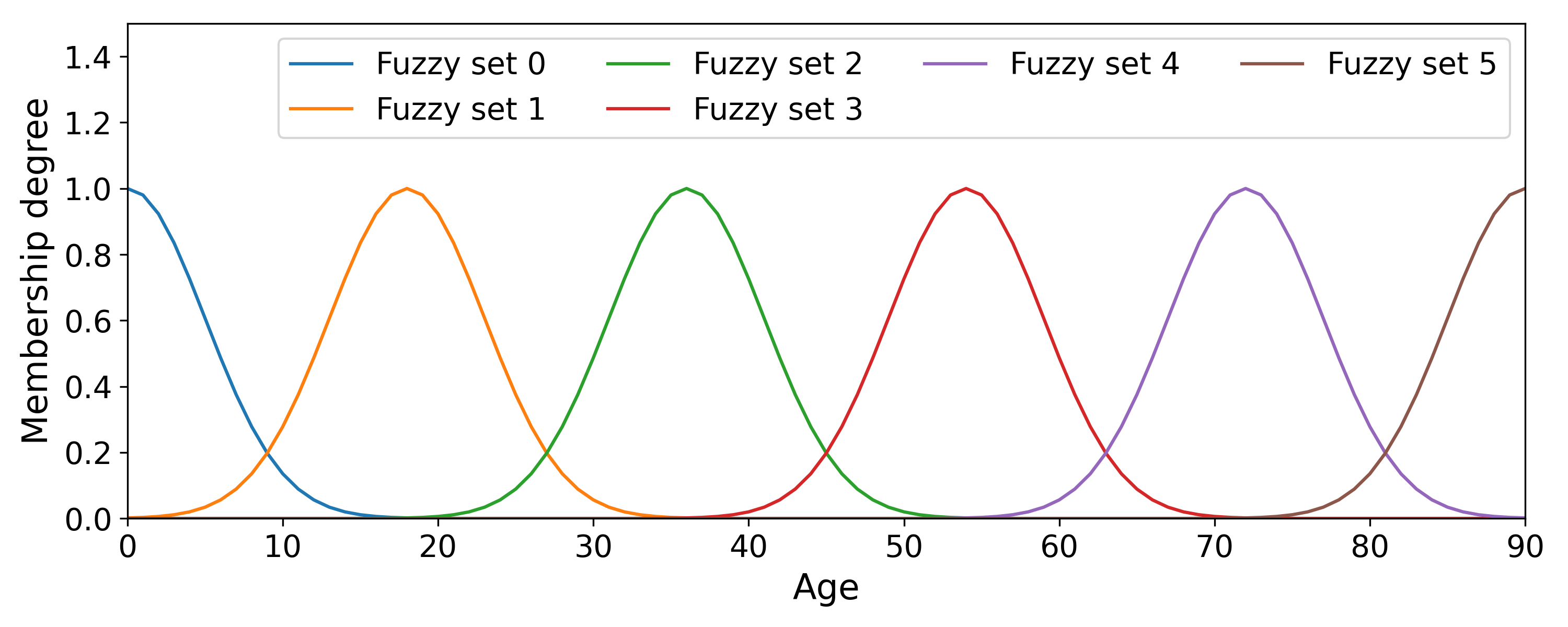}
	\end{minipage}}
	\subfigure[Initial representation of age difference]{
		\begin{minipage}[b]{0.46\linewidth}
			\includegraphics[width=1\linewidth]{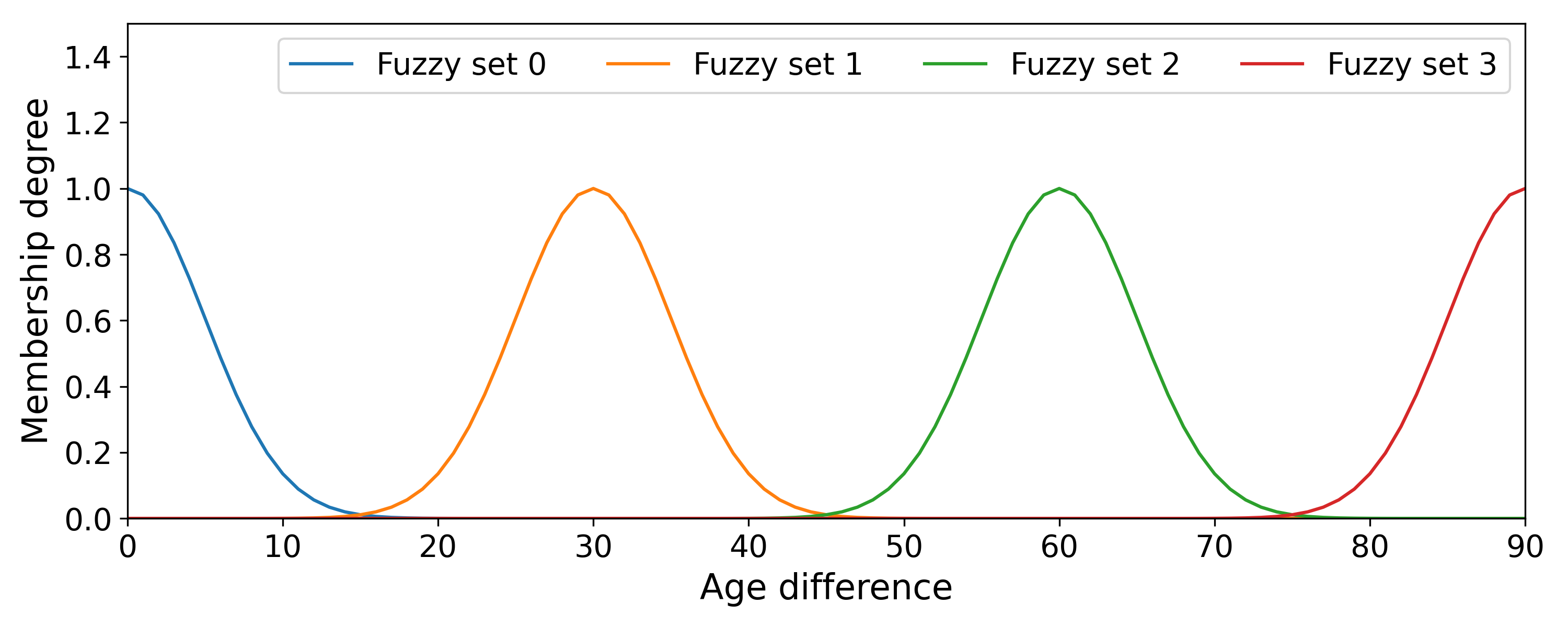}
	\end{minipage}}\\
	\subfigure[Optimised representation of age]{
		\begin{minipage}[b]{0.46\linewidth}
			\includegraphics[width=1\linewidth]{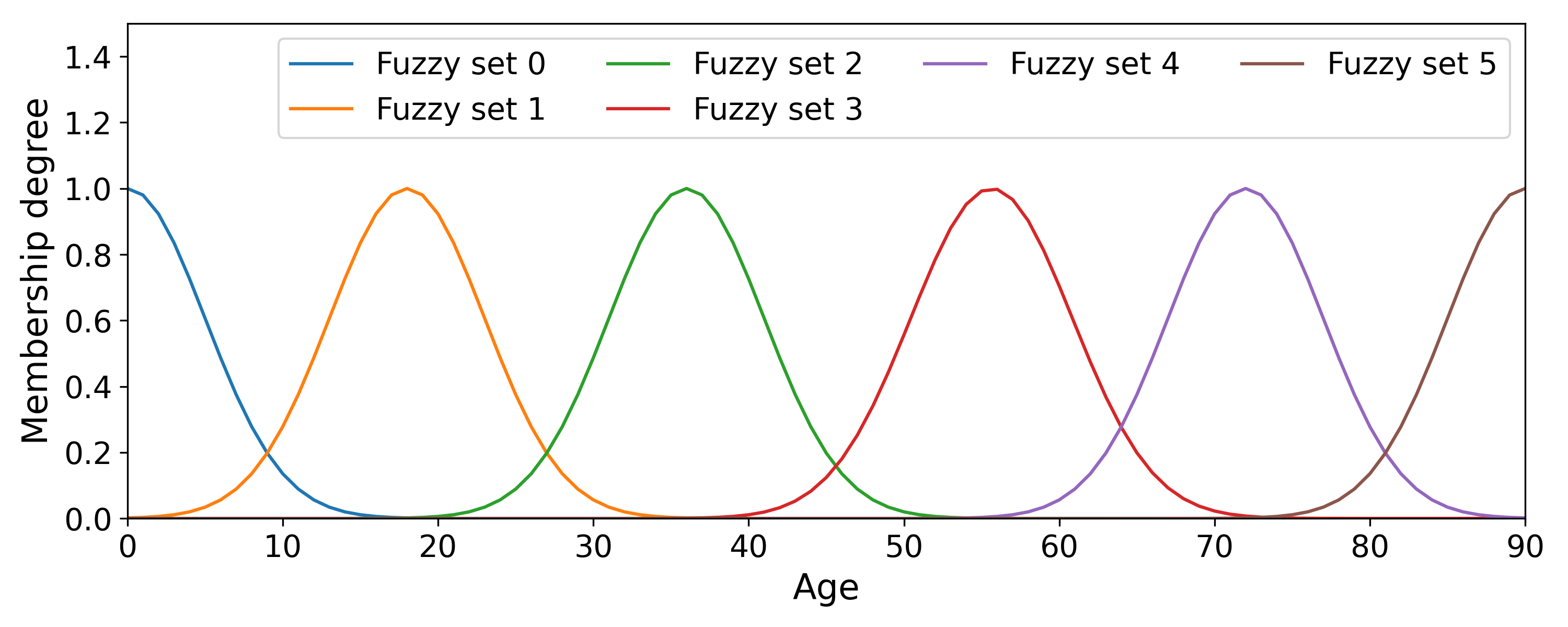}
	\end{minipage}}
	\subfigure[Optimised representation of age difference]{
		\begin{minipage}[b]{0.46\linewidth}
			\includegraphics[width=1\linewidth]{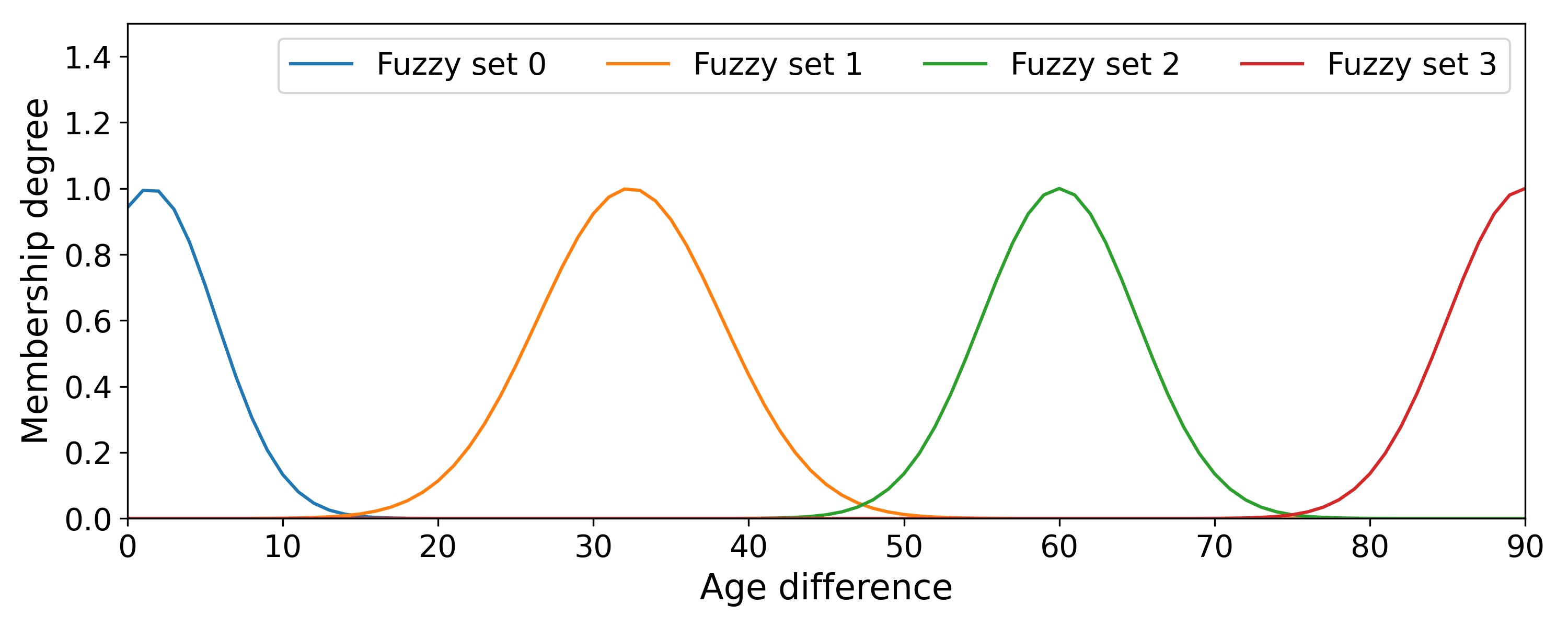}
	\end{minipage}}\\
 	\caption{The parameter sensitivity level (in log scale) and the corresponding rankings for parameters of fuzzy sets based on the $IVARS_{10}$, $IVARS_{30}$, $IVARS_{50}$ and $Sobol$(Fig.~(a)), the initialised fuzzy representation of age and age difference (Fig.~(b) and Fig.~(c)) and the optimised fuzzy representation of age and age difference (Fig.~(d) and Fig.~(e)) for Finland. In Fig.~(a), parameters for the first fuzzy set of age difference ($\mu_1^h$ and $\sigma_1^h$), parameters for the second fuzzy set of age difference ($\mu_2^h$ and $\sigma_2^h$) and parameters of the fourth fuzzy set of age difference ($\mu_4^p$ and $\sigma_4^p$) rank high considering all types of sensitivity indexes. Therefore, they are optimised in model calibration. The optimised fuzzy representation principles are shown in Fig.~(d) and Fig.~(e).}
\label{fuzzyFselectparam}
\end{figure}

Fig.~\ref{fuzzyFselectparam} (a) shows the ranks and the log values of sensitivity metrics for the parameters involved in fuzzy sets (six fuzzy sets to represent age and four fuzzy sets to represent age difference) for Finland. The sensitivity metrics enable the same parameter rankings in the first three places and have similar ranks for the less important parameters. Generally, the impact of the $\mu^h_i$ and $\sigma^h_i$ values decreases as their focus moves from age differences around $0$ to larger age differences. In contrast  the $\mu^p_i$ and $\sigma^p_i$ values fluctuates without a clear trend. $\mu^h_1$, $\sigma^h_1$, $\mu_2^h$ and $\sigma_2^h$, concerned with the age difference around $0$ and $30$, have a more significant impact on the model performance and thus are optimised. These four parameters are followed by $\mu_p^4$ and $\sigma_p^4$, which describe the age values around $54$ and also ranks high considering all the sensitivity metrics. We optimise these parameters based on the initialised parameter set up of the fuzzy sets. As shown in Fig.~\ref{fuzzyFselectparam} (e), the first fuzzy set for age difference, parameterised with $\mu^h_1$ and $\sigma_1^h$, is tuned in the optimisation stage with a less extensive spread at $4.25$ than the initialised spread at $5$ and a transition from the location around $0.00$ to the location around $1.47$. The second fuzzy set for the age difference, parameterised with $\mu^h_2$ and $\sigma^h_2$, centres around $32.36$ rather than $30$ with a more extensive spread at $5.926$ than the initialised one at $5$ after optimisation. The fourth fuzzy set for the age values, parameterised with $\mu^4_p$ and $\sigma^4_p$, transits its focus from $54$ to $55.64$, with a more extensive spread at $5.21$ than the initialised one at $5.00$. The changes mentioned above imply the people's interest in age difference at around $1$ and age difference around $32$ (rather than age difference around $0$ or $30$) and age at around $55$ (rather than $54$) in Finland, different from people in Belgium. These changes lead to a decrease of $EU$ distance from $2.29$ to $2.10$ (see Tab.~\ref{summary}).

\begin{figure*}[h!]
	\centering
 	\subfigure[Parameter rankings based on the log-scale parameter sensitivity]{
		\begin{minipage}[b]{0.94\linewidth}
			\includegraphics[width=1\linewidth]{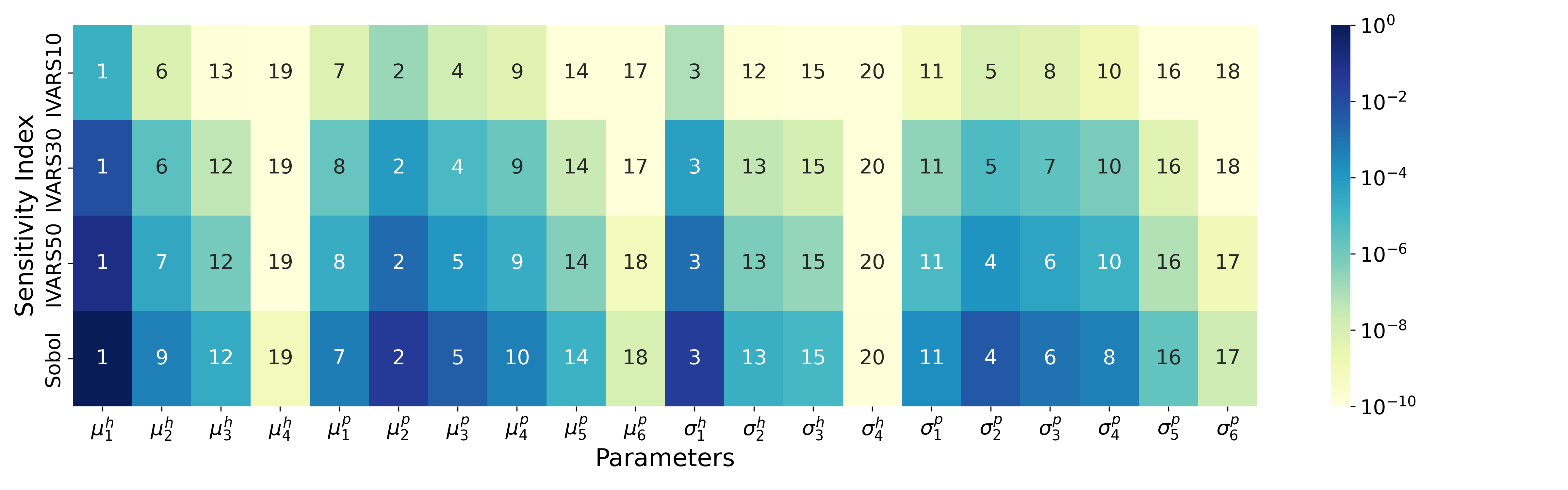}
	\end{minipage}}
	\subfigure[Initial representation of age]{
		\begin{minipage}[b]{0.46\linewidth}
			\includegraphics[width=1\linewidth]{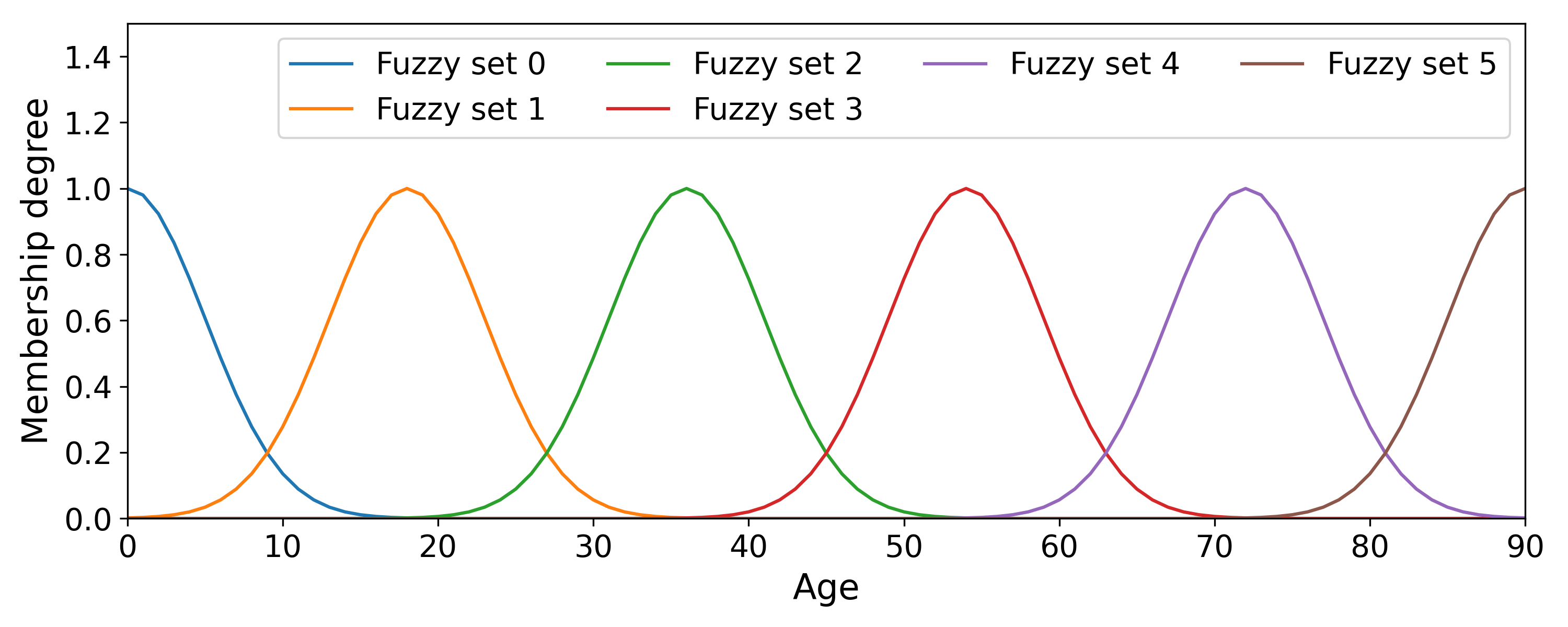}
	\end{minipage}}
	\subfigure[Initial representation of age difference]{
		\begin{minipage}[b]{0.46\linewidth}
			\includegraphics[width=1\linewidth]{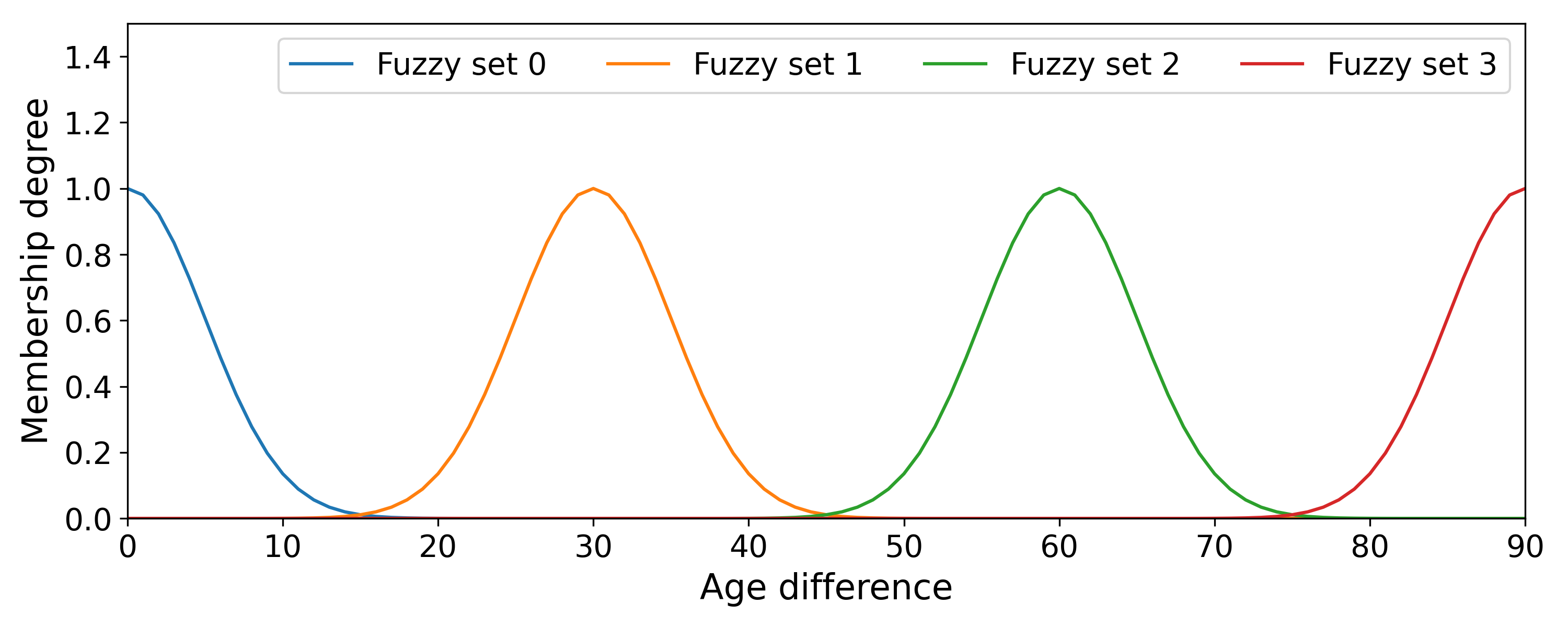}
	\end{minipage}}\\
	\subfigure[Optimised representation of age]{
		\begin{minipage}[b]{0.46\linewidth}
			\includegraphics[width=1\linewidth]{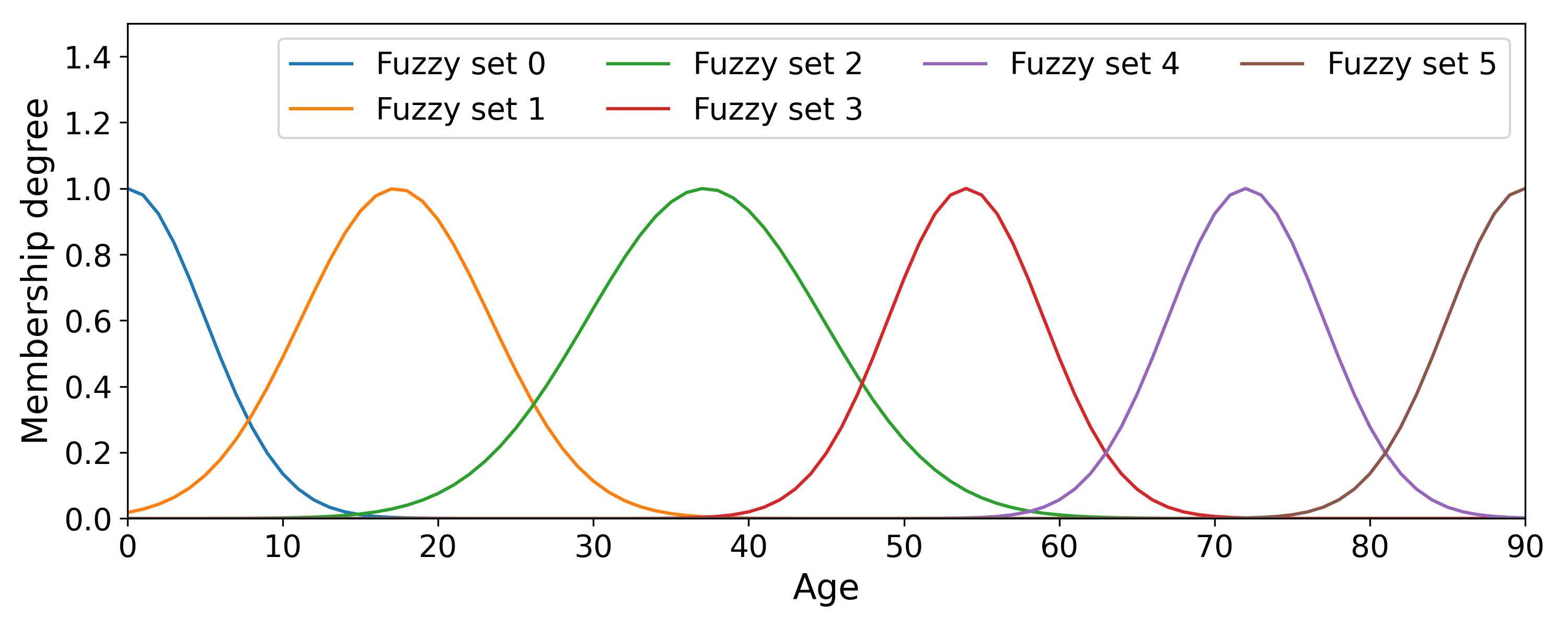}
	\end{minipage}}
	\subfigure[Optimised representation of age difference]{
		\begin{minipage}[b]{0.46\linewidth}
			\includegraphics[width=1\linewidth]{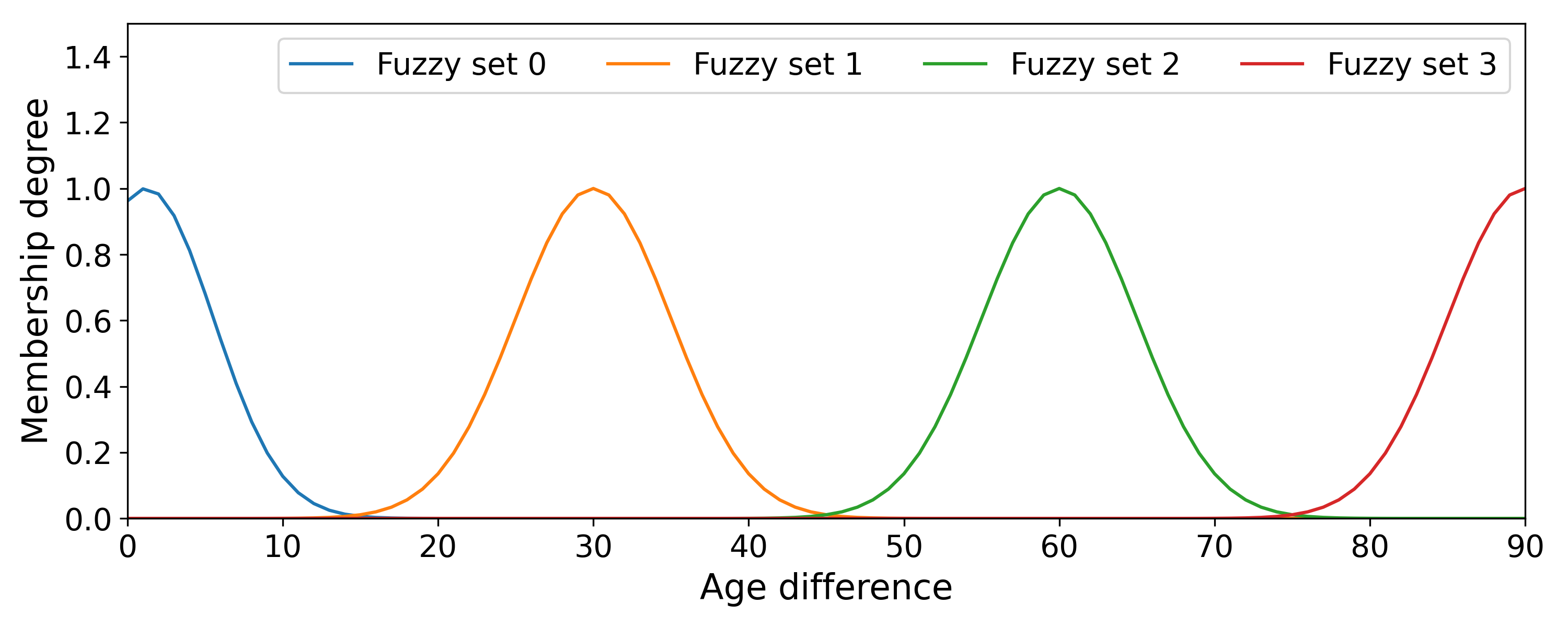}
	\end{minipage}} \\
  	\caption{The parameter sensitivity level (in log scale) and the corresponding rankings for parameters of fuzzy sets based on the $IVARS_{10}$, $IVARS_{30}$, $IVARS_{50}$ and $Sobol$(Fig.~(a)), the initialised fuzzy representation of age and age difference (Fig.~(b) and Fig.~(c)) and the optimised fuzzy representation of age and age difference (Fig.~(d) and Fig.~(e)) for Germany. In Fig.~(a), parameters for the first fuzzy set of age difference ($\mu_1^h$ and $\sigma_1^h$), parameters for the second fuzzy set of age ($\mu_2^p$ and $\sigma_2^p$) and parameters of the third fuzzy set of age ($\mu_3^p$ and $\sigma_3^p$) rank high considering all types of sensitivity indexes. Therefore, they are optimised in model calibration. The optimised fuzzy representation principles are shown in Fig.~(d) and Fig.~(e).}
\label{fuzzyGselectparam}
\end{figure*}

Fig.~\ref{fuzzyGselectparam} (a) shows the ranks and the log values of sensitivity metrics for each parameter involved in the fuzzy sets (six fuzzy sets to represent age and four fuzzy set to represent age difference) for Germany. Similar with the case of Finland, the sensitivity metrics enable the same parameter rankings in the first three places and presents a decreasing impact of the $\mu^h_i$ and $\sigma^h_i$ values as their focus moves from age differences around $0$ to larger age differences. The parameter importance of the $\mu^p_i$ and $\sigma^p_i$ values fluctuates with the transitions of age focuses from young to old. This also indicates that smaller age differences have a larger impact on the model performance in recreating social contact matrices than the larger ones. A more significant impact of the $\mu^h_1$ and $\sigma^h_1$ values on the model performance than the other parameters can be observed. In addition,  parameters for the second fuzzy set of age ($\mu_2^p$ and $\sigma_2^p$) and parameters of the third fuzzy set of age difference ($\mu_3^p$ and $\sigma_3^p$) also rank high considering all types of sensitivity indexes. We optimise these parameters and present the changes of fuzzy sets in Fig.~\ref{fuzzyGselectparam} (c) and Fig.~\ref{fuzzyGselectparam} (e). The first fuzzy set is optimised with a less extensive spread at $4.34$ and a transition from the location around $0.00$ to the location around $1.21$. The second fuzzy set of age has a more extensive spread at $6.09$ and transits the focus from $18.00$ to $17.29$ after optimisation. The third fuzzy set of age has a more extensive spread at $7.56$ and a focus transition from $36.00$ to $37.19$. These changes indicates people's interest in no age difference and the age values around $17$ and $37$ (rather than $18$ or $36$). In addition, the abovementioned changes lead to a decrease of the $EU$ distance from $2.56$ to $2.28$ (see Tab.~\ref{summary}). 

\begin{figure*}[h!]
	\centering
  	\subfigure[Parameter rankings based on the log-scale parameter sensitivity]{
		\begin{minipage}[b]{0.94\linewidth}
			\includegraphics[width=1\linewidth]{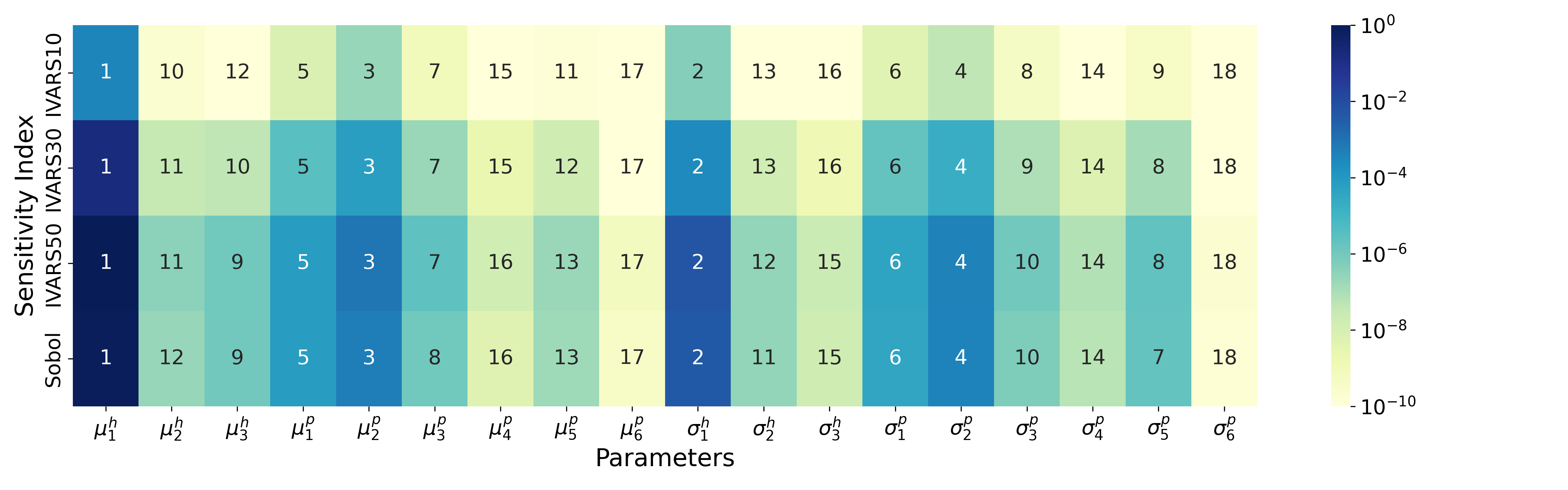}
	\end{minipage}}\\
	\subfigure[Initial representation of age]{
		\begin{minipage}[b]{0.46\linewidth}
			\includegraphics[width=1\linewidth]{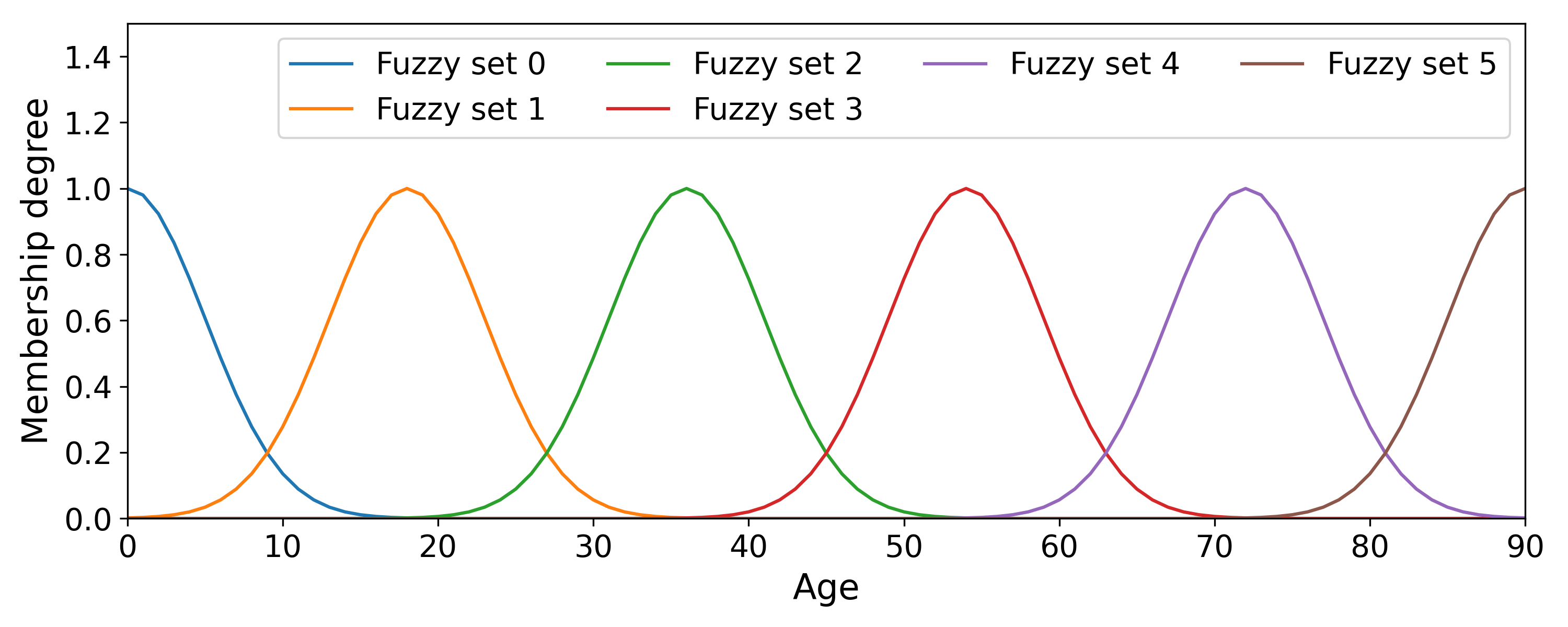}
	\end{minipage}}
	\subfigure[Initial representation of age difference]{
		\begin{minipage}[b]{0.46\linewidth}
			\includegraphics[width=1\linewidth]{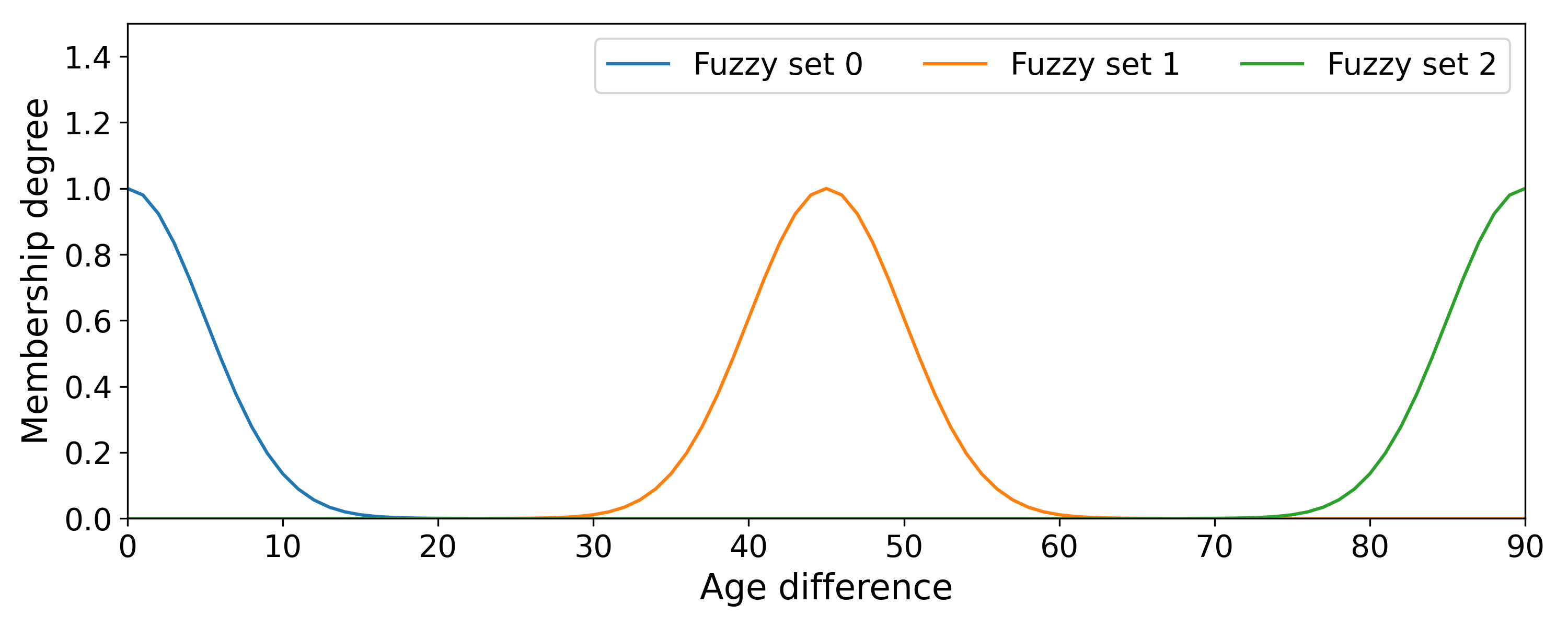}
	\end{minipage}}\\
	\subfigure[Optimised representation of age]{
		\begin{minipage}[b]{0.46\linewidth}
			\includegraphics[width=1\linewidth]{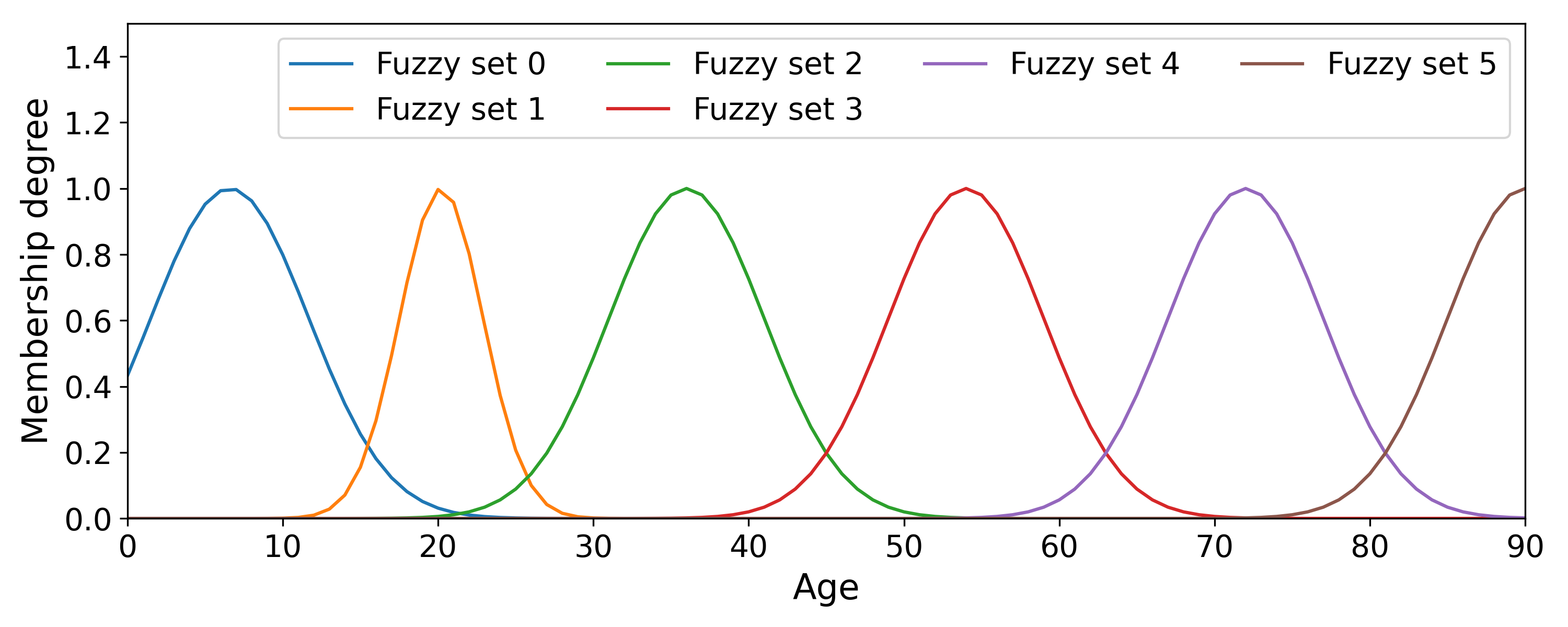}
	\end{minipage}}
	\subfigure[Optimised representation of age difference]{
		\begin{minipage}[b]{0.46\linewidth}
			\includegraphics[width=1\linewidth]{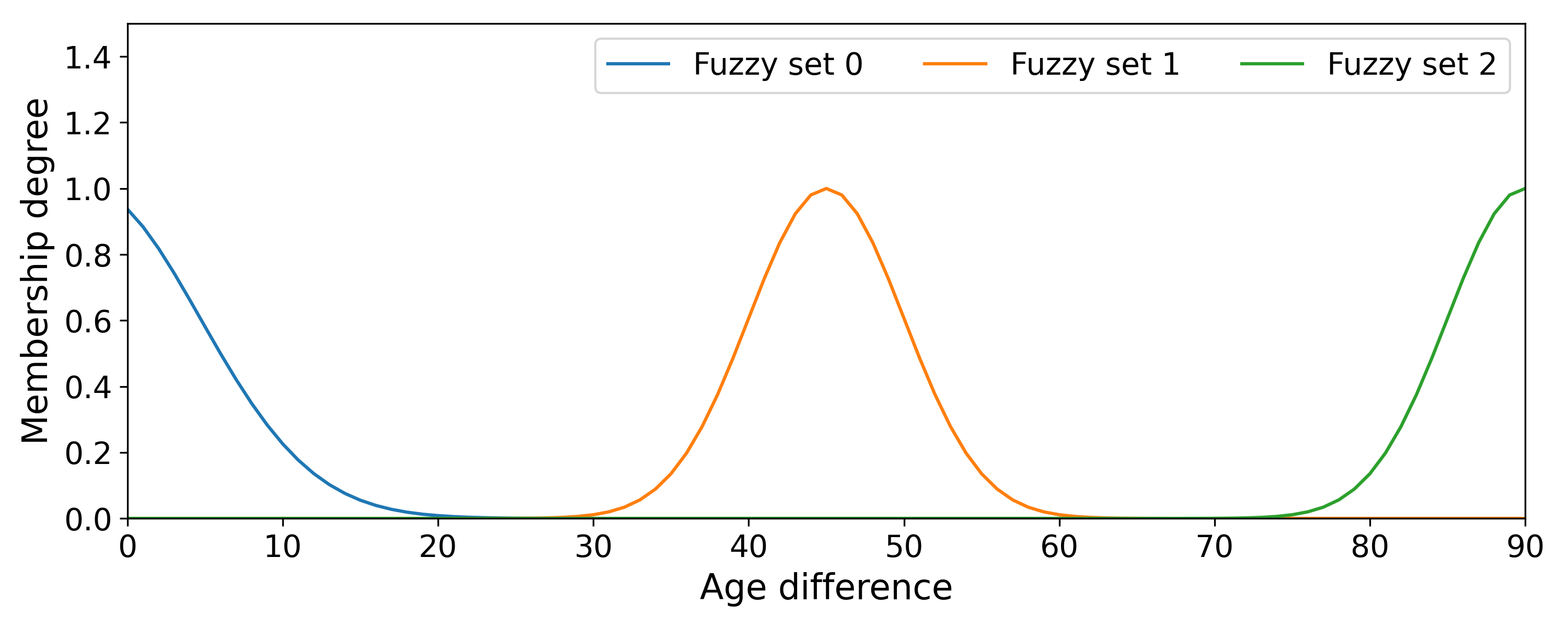}
	\end{minipage}}\\
   	\caption{The parameter sensitivity level (in log scale) and the corresponding rankings for parameters of fuzzy sets based on the $IVARS_{10}$, $IVARS_{30}$, $IVARS_{50}$ and $Sobol$(Fig.~(a)), the initialised fuzzy representation of age and age difference (Fig.~(b) and Fig.~(c)) and the optimised fuzzy representation of age and age difference (Fig.~(d) and Fig.~(e)) for Italy. In Fig.~(a), parameters for the first fuzzy set of age difference ($\mu_1^h$ and $\sigma_1^h$), parameters for the first fuzzy set of age ($\mu_1^p$ and $\sigma_1^p$) and parameters of the second fuzzy set of age ($\mu_2^p$ and $\sigma_2^p$) rank high considering all types of sensitivity indexes. Therefore, they are optimised in model calibration. The optimised fuzzy representation principles are shown in Fig.~(d) and Fig.~(e).}
\label{fuzzyIselectparam}
\end{figure*}

Fig.~\ref{fuzzyIselectparam} (a) shows the ranks and the log values of sensitivity metrics for each parameter involved in the fuzzy sets (eight fuzzy sets to represent age and one fuzzy set to represent age difference) for Italy. The parameters share the same rankings in the first six places considering each sensitivity index. This implies the validity of parameter selection in model calibration as the selected parameters all rank high.  In addition, the parameters of other parameters have small differences and little impact on model calibration and model performance. Similar with the abovementioned countries, the parameter $\mu^h_1$, $\sigma_1^h$ and especially $\mu_1^h$ significantly impact the model performance, indicating the importance of age difference around $0$ (as indicated by parameters for the first fuzzy set of age difference, including $\sigma_1^h$ and $\mu_1^h$). In addition, the age around $18$ and age around $36$ (as indicated by the first fuzzy set of age ($\mu_1^p$ and $\sigma_1^p$) and second fuzzy set of age, related to $\mu^p_2$ and $\sigma^p_2$) also have greater impact on the decision-making of social contact than the other parameters considering all the sensitivity indexes. We optimise the parameter based on the initialised parameter set up of the fuzzy sets. As shown in Fig.~\ref{fuzzyIselectparam} (c) and Fig.~\ref{fuzzyIselectparam} (e), the initialised first fuzzy set of age difference, parameterised with $\mu^h_1$ and $\sigma_1^h$, describes the uncertain induction of age difference around $0$. As shown in Fig.~\ref{fuzzyIselectparam} (b) and Fig.~\ref{fuzzyIselectparam} (d), it is optimised with a larger spread of $7.32$ and a location transition from location around $0$ to location around $-2.63$, indicating the decrease of membership degrees for the first fuzzy set. In addition, the first fuzzy set for age is optimised with a slightly more extensive spread at $5.09$ and a significant focus transition from $0.00$ to $6.59$. The second fuzzy set for age, initialised around $18$ with a spread at $5$, is optimised with the transition of focus to $20.21$ and a less extensive spread at $2.70$. These changes imply the people's interest in age difference around $0$ and the age values around $6$ and $20$ (rather than $0$ or $18$). The abovementioned changes decrease the $EU$ distance from $8.59$ to $8.34$ (see Tab.~\ref{summary}).

\begin{figure*}[h!]
	\centering
  	\subfigure[Parameter rankings based on the log-scale parameter sensitivity]{
		\begin{minipage}[b]{0.94\linewidth}
			\includegraphics[width=1\linewidth]{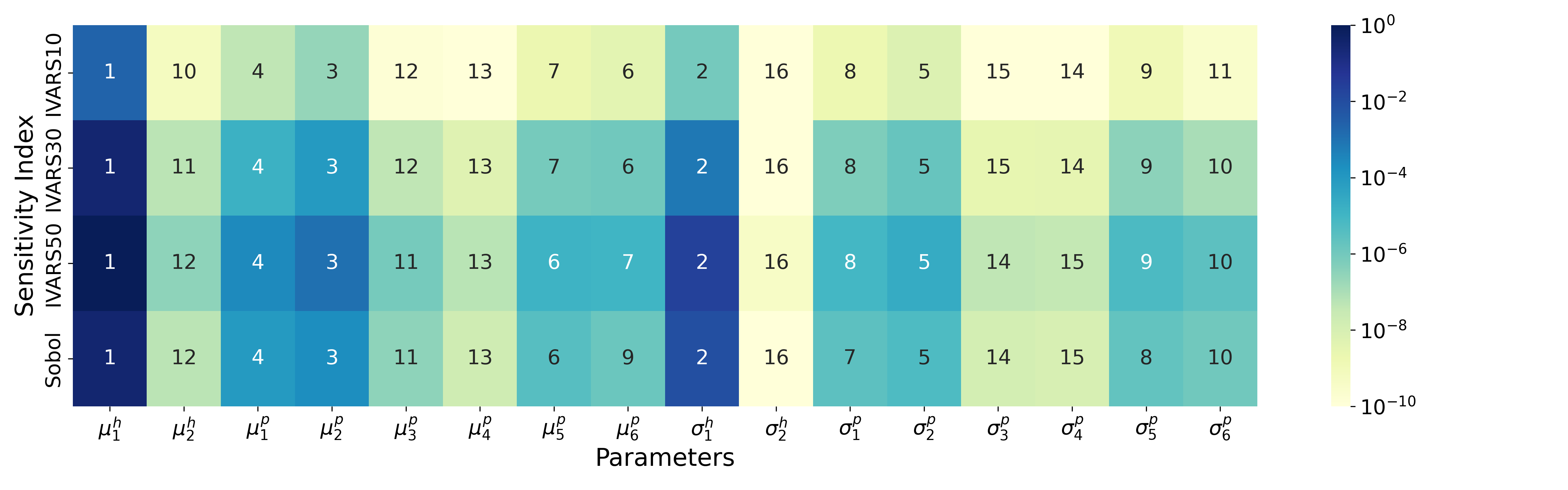}
	\end{minipage}}\\
	\subfigure[Initial representation of age]{
		\begin{minipage}[b]{0.46\linewidth}
			\includegraphics[width=1\linewidth]{ItalyFuzzyP0927.png}
	\end{minipage}}
	\subfigure[Initial representation of age difference]{
		\begin{minipage}[b]{0.46\linewidth}
			\includegraphics[width=1\linewidth]{ItalyFuzzyH0927.png}
	\end{minipage}}\\
	\subfigure[Optimised representation of age]{
		\begin{minipage}[b]{0.46\linewidth}
			\includegraphics[width=1\linewidth]{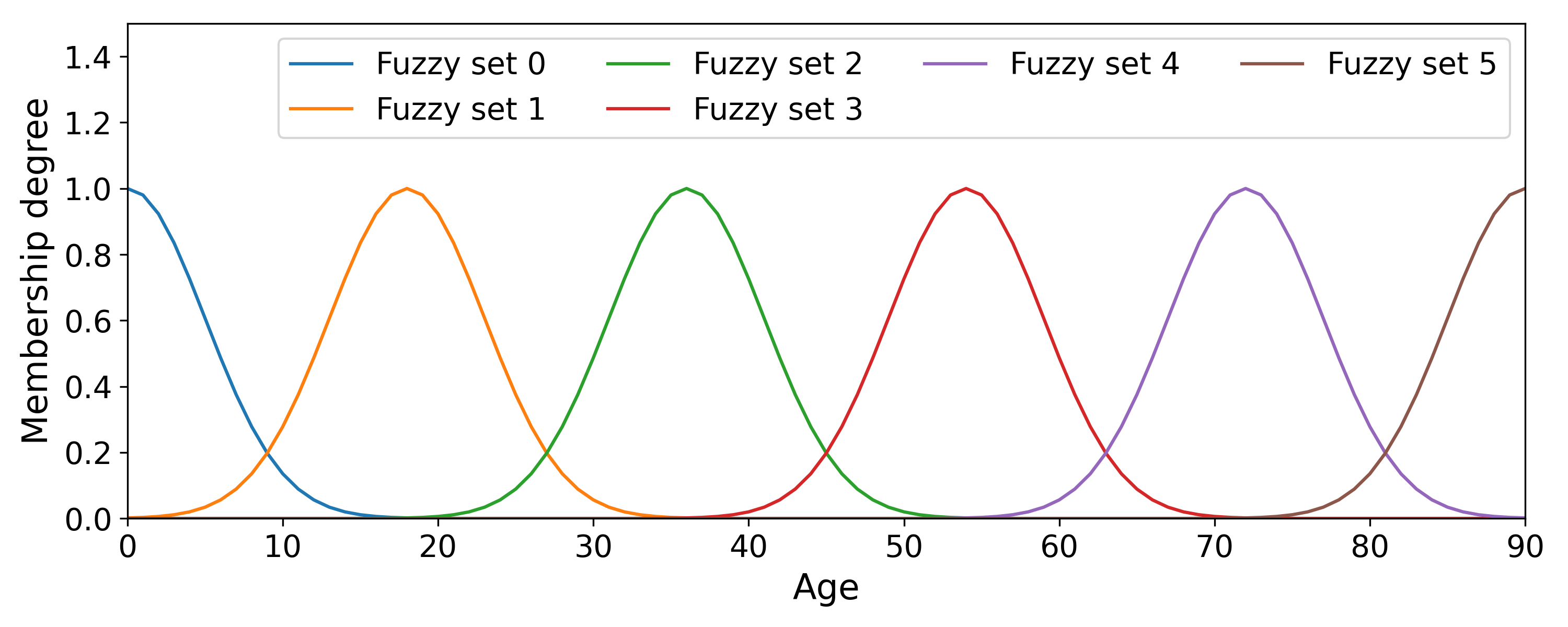}
	\end{minipage}}
	\subfigure[Optimised representation of age difference]{
		\begin{minipage}[b]{0.46\linewidth}
			\includegraphics[width=1\linewidth]{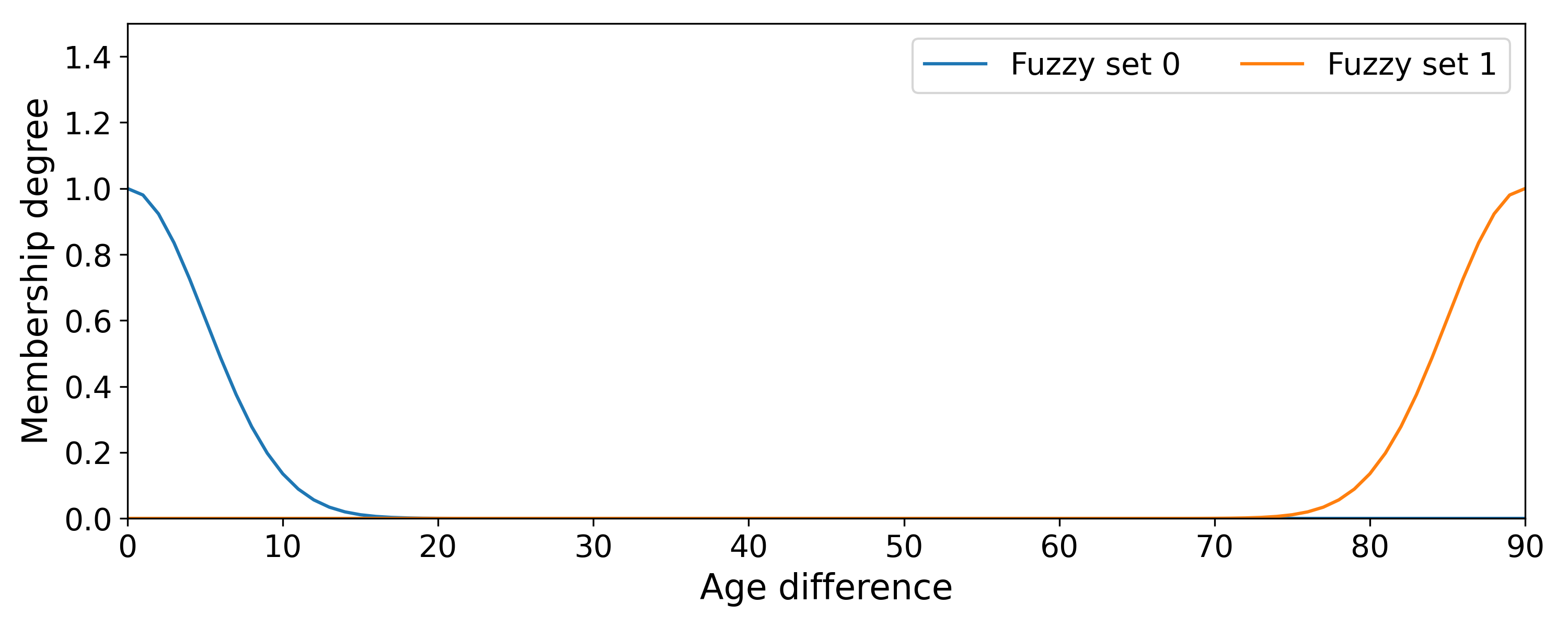}
	\end{minipage}}\\
 \caption{The parameter sensitivity level (in log scale) and the corresponding rankings for parameters of fuzzy sets based on the $IVARS_{10}$, $IVARS_{30}$, $IVARS_{50}$ and $Sobol$(Fig.~(a)), the initialised fuzzy representation of age and age difference (Fig.~(b) and Fig.~(c)) and the optimised fuzzy representation of age and age difference (Fig.~(d) and Fig.~(e)) for Luxembourg. In Fig.~(a), parameters for the first fuzzy set of age difference ($\mu_1^h$ and $\sigma_1^h$), parameter for the first fuzzy set of age ($\mu_1^p$), parameters for the second fuzzy set of age ($\mu_2^p$ and $\sigma_2^p$) and parameter for the fifth fuzzy set of age ($\mu_5^p$) rank high considering all types of sensitivity indexes. Therefore, they are optimised in model calibration. The optimised fuzzy representation principles are shown in Fig.~(d) and Fig.~(e).}
\label{fuzzyLselectparam}
\end{figure*}

Fig.~\ref{fuzzyLselectparam} (a) shows the ranks and the log values of sensitivity metrics for each parameter involved in the fuzzy sets (six fuzzy sets to represent age and two fuzzy set to represent age difference) for Luxembourg. Considering each sensitivity index, we have same parameter ranks in the first five places and similar parameter ranks following these five parameters. The $\mu_1^h$ and $\sigma_1^h$ value, concerned with the representation of zero age difference, significantly impacts the model performance and indicates the importance of age difference around $0$ in the decision-making of social contact. The parameters for the first fuzzy set of age ($\mu_1^p$), the second fuzzy set of age ($\mu_2^h$ and $\sigma_2^h$) and the fifth fuzzy set of age ($\mu_5$) rank afterwards also have a greater impact on model performance than the other parameters. We optimise these parameters based on the initialised parameter set up for the fuzzy set of age difference. As shown in Fig.~\ref{fuzzyLselectparam} (c) and Fig.~\ref{fuzzyLselectparam} (e), the first fuzzy set for age difference is optimised with a more extensive spread of $7.60$ and a focus transition from $0$ to a slightly lower value at $-0.45$, leading to slightly smaller membership values of age differences and indicating less interest from people. Similarly, the first fuzzy set for age is optimised with a focus transition from $0.00$ to a negative value at $-1.23$, which decreases the membership degrees of age values to the first fuzzy set of age after optimisation. The second fuzzy set for age is optimised with a smaller spread at $4.69$ and a significant focus transition from $18$ to $18.15$. The fifth fuzzy set for age, initialised around $72.00$, is optimised with the transition of focus to $73.12$. These changes imply the people's decreased interest in age difference around $0$ and the age values around $0$. They also indicate the people's interest in age around $18$ and $73$ (similar with the unoptimised values around $18$ and $72$). The abovementioned changes decrease the $EU$ distance from $5.07$ to $4.77$ (see Tab.~\ref{summary}), implying the necessity of optimising membership functions for fuzzy representation of features. 

\begin{figure*}[h!]
	\centering
  	\subfigure[Parameter rankings based on the log-scale parameter sensitivity]{
		\begin{minipage}[b]{0.94\linewidth}
			\includegraphics[width=1\linewidth]{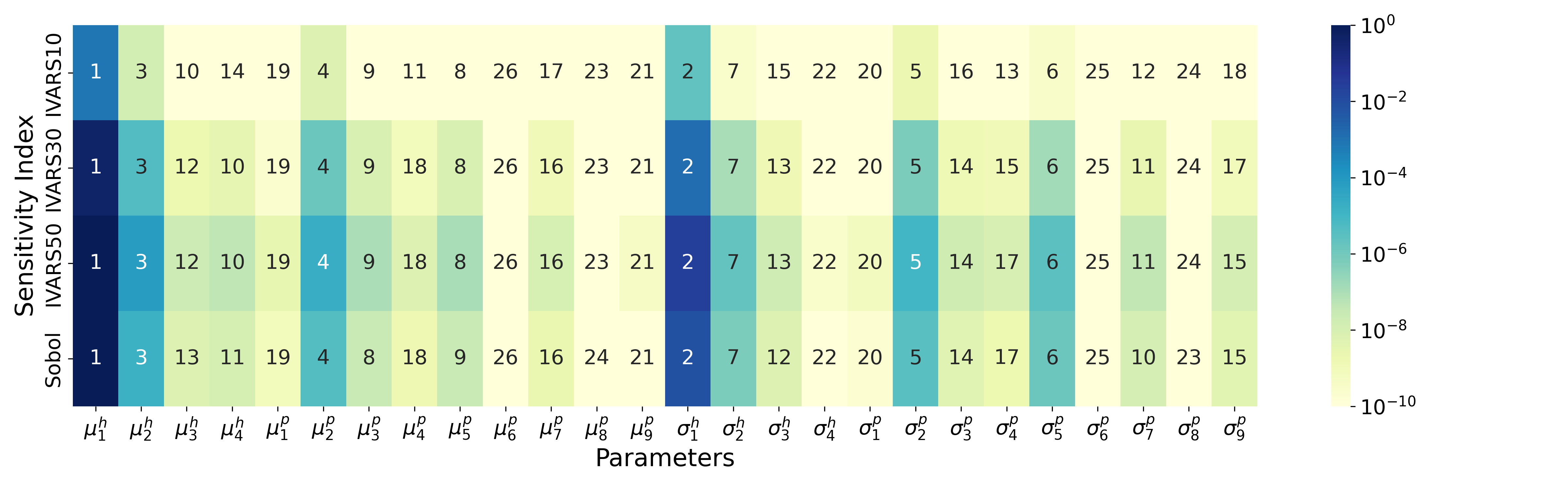}
	\end{minipage}}\\
	\subfigure[Initial representation of age]{
		\begin{minipage}[b]{0.46\linewidth}
			\includegraphics[width=1\linewidth]{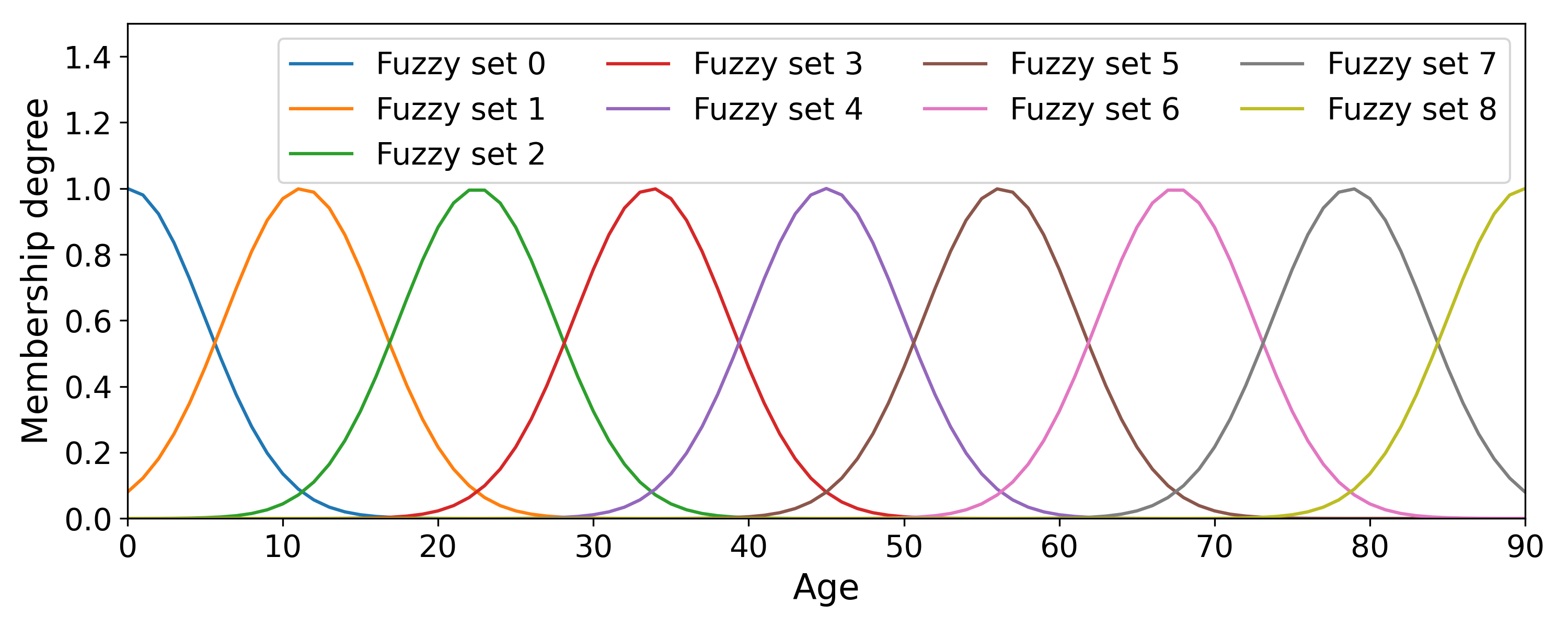}
	\end{minipage}}
	\subfigure[Initial representation of age difference]{
		\begin{minipage}[b]{0.46\linewidth}
			\includegraphics[width=1\linewidth]{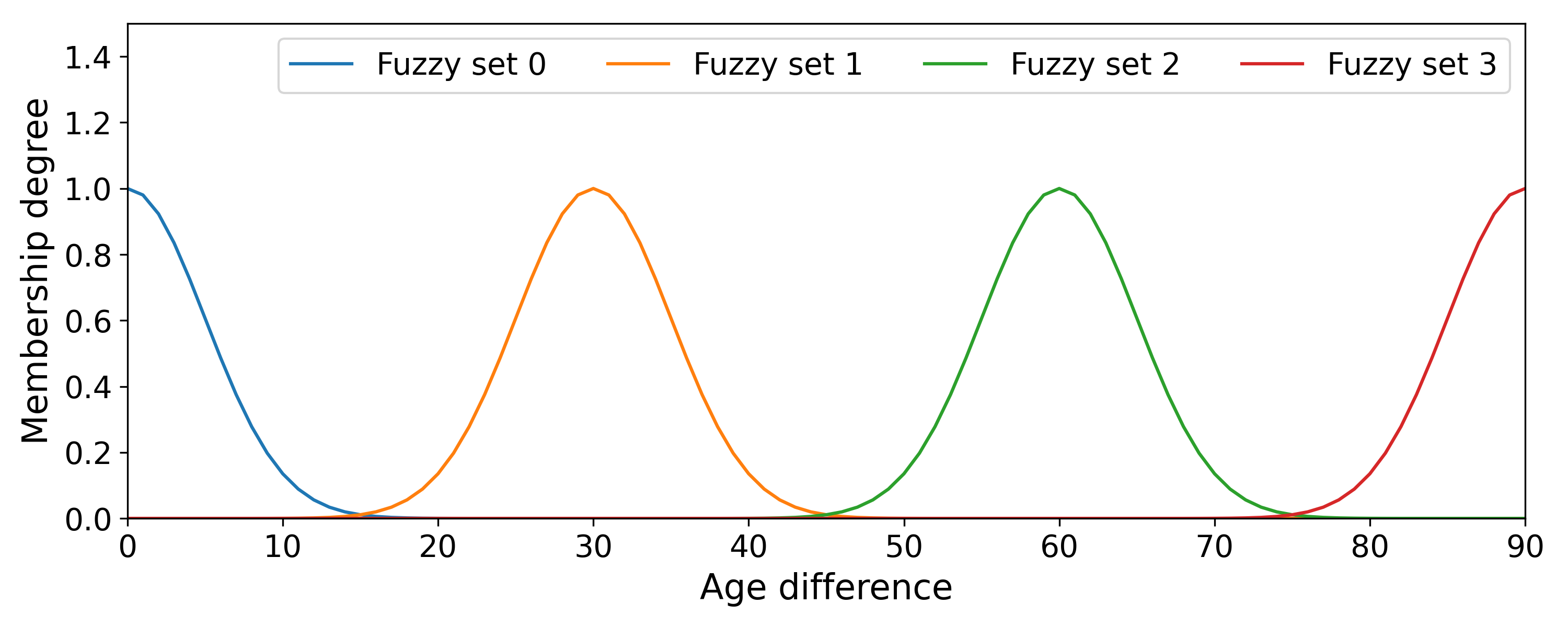}
	\end{minipage}}\\
	\subfigure[Optimised representation of age]{
		\begin{minipage}[b]{0.46\linewidth}
			\includegraphics[width=1\linewidth]{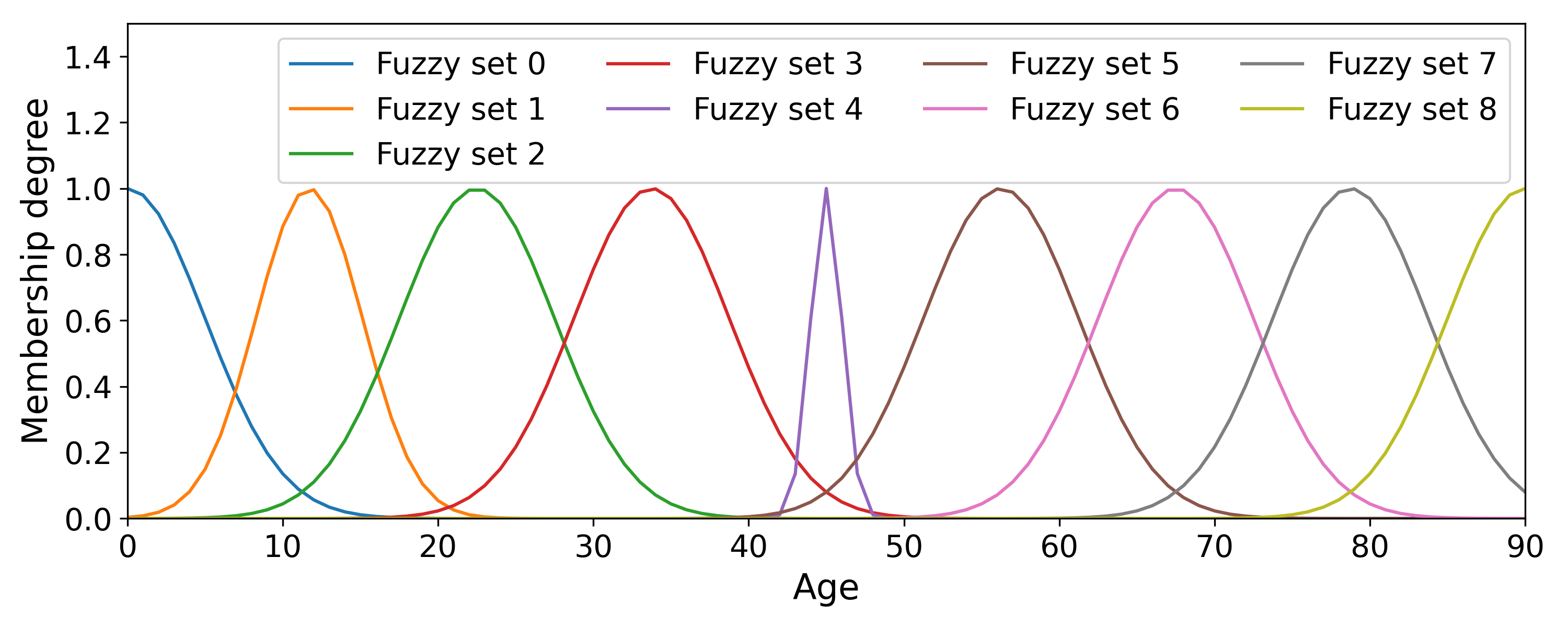}
	\end{minipage}}
	\subfigure[Optimised representation of age difference]{
		\begin{minipage}[b]{0.46\linewidth}
			\includegraphics[width=1\linewidth]{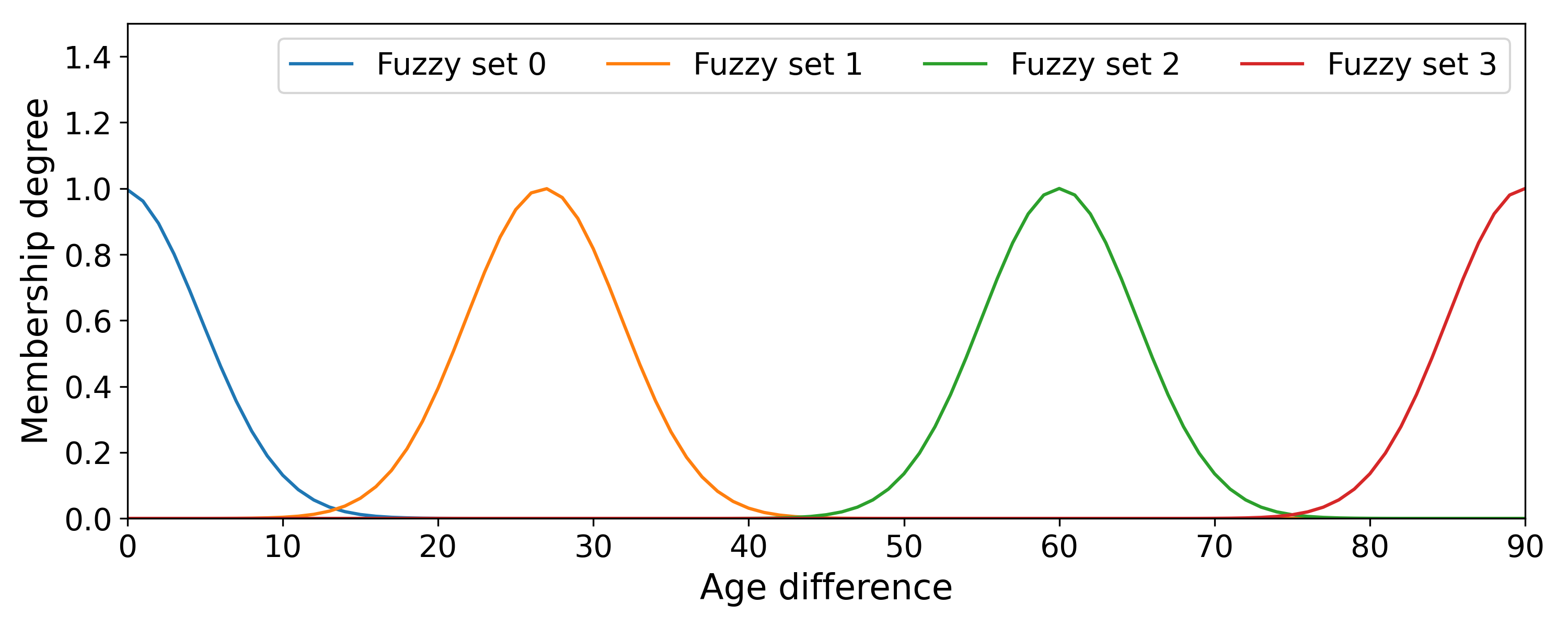}
	\end{minipage}}\\
 \caption{The parameter sensitivity level (in log scale) and the corresponding rankings for parameters of fuzzy sets based on the $IVARS_{10}$, $IVARS_{30}$, $IVARS_{50}$ and $Sobol$(Fig.~(a)), the initialised fuzzy representation of age and age difference (Fig.~(b) and Fig.~(c)) and the optimised fuzzy representation of age and age difference (Fig.~(d) and Fig.~(e)) for Poland. In Fig.~(a), parameter for the first fuzzy set of age difference ($\sigma_1^h$), parameters for the second fuzzy set of age difference ($\mu_2^h$ and $\sigma_2^h$), parameters for the second fuzzy set of age ($\mu_2^p$ and $\sigma_2^p$) and parameter for the fifth fuzzy set of age ($\mu_5^p$) rank high considering all types of sensitivity indexes. Therefore, they are optimised in model calibration. The optimised fuzzy representation principles are shown in Fig.~(d) and Fig.~(e).}
\label{fuzzyPselectparam}
\end{figure*}

Fig.~\ref{fuzzyPselectparam} (a) shows the ranks and the log values of sensitivity metrics for each parameter involved in the fuzzy sets (nine fuzzy sets to represent age and four fuzzy set to represent age difference) in Poland. The $\sigma_1^h$ values, concerned with the representation of age difference around $0$, significantly impact the model performance and indicate their importance in the parameter optimisation. The $\mu_h^2$ and $\sigma_h^2$ for the second fuzzy set of age difference, among the four fuzzy sets of age difference, rank high in terms of parameter importance. This is the same with the cases of Belgium and Finland. In addition, the parameters for the second fuzzy set of age at around $22$ ($\mu_p^2$ and $\sigma_p2$) and the parameter for the fifth fuzzy set of age around $45$ ($\sigma_p^5$) also rank high considering all the sensitivity metrics. We optimise the parameters based on the initialised parameter set up of the fuzzy sets. As shown in Fig.~\ref{fuzzyPselectparam} (c) and Fig.~\ref{fuzzyLselectparam} (e), the fuzzy set for age difference, parameterised with $\mu^h_1$ and $\sigma_1^h$, is further optimised with a similar spread at $5.18$. The second fuzzy set of age difference is optimised with a significant focus transition from $30$ to $26$. This implies people's interest in age difference around $26$ rather than $30$. In addition, the second fuzzy set of age transits its focus from $11.33$ to $11.70$, which is a slight adjustment of people's interest around the age at $11$. The fifth fuzzy set of age is optimised with a significant change in spread, which decreases from $5.00$ to $1.00$, reaching the lower bound of parameter optimisation. This implies that the age closer to $45$ can be assigned with a higher membership degree, indicating the people's interest in the age values that are close to $45$. The abovementioned changes decrease the $EU$ distance from $5.61$ to $5.04$ (see Tab.~\ref{summary}).

\section{Parameter bounds' set-ups in the sensitivity analysis and optimisation process}
\label{app3}

\begin{table}[h]
\centering
\scriptsize
\caption{Parameter bounds' set-ups in the sensitivity analysis and optimisation process}
\label{boundary}
\setlength{\tabcolsep}{3pt}
\renewcommand{\arraystretch}{1.5}
\begin{tabular}{|c|c|c|c|c|c|c|}
\hline
Parameter&Belgium&Finland&Germany&Italy&Luxembourg&Poland\\
\hline
$\mu_1^h$ &[-45,45] & [-9,9]& [-9,9] & [-9,-9] &[-9,-9]& [-5.6250,5.6250] \\
\hline
$\mu_2^h$ & -- & [9,27]& [9,27]&[9,27]&[9,27]& [5.6250,16.8750] \\
\hline
$\mu_3^h$ & --  & [27,45]& [27,45] & [27,45] &[27,45]& [16.8750,28.1250]\\
\hline
$\mu_4^h$ &--  & [45,63]& [45,63] & [45,63] &[45,63]& [20.1250,39.3750]  \\
\hline
$\mu_5^h$ &-- & [63,81]& [63,81] & [63,81] &[63,81]&   [39.3750,50.6250] \\
\hline
$\mu_6^h$ &-- & [81,99]  & [81,99] & [81,99] & [81,99] &  [50.6250,61.8750]\\
\hline
$\mu_7^h$ &-- & --&--&--&--& [61.8750,73.1250]\\
\hline
$\mu_8^h$ &-- & --&--&--&--&[73.1250,84.3750] \\
\hline
$\mu_9^h$ &-- & --&--&--&--& [84.3750,95.6250]\\
\hline
$\mu_1^h$ & [-15,15] &[-15,15] &[-15,15] & [-22.5,22.5]&[-45,45]&[-15,15]\\
\hline
$\mu_2^h$ & [15,45]&[15,45] &[15,45] &[22.5,67.5]& [45,135] & [15,45]\\
\hline
$\mu_3^h$ & [45,75]& [45,75] &[45,75] & [67.5,102.5]& --& [45,75]\\
\hline
$\mu_4^h$ &  [75,105]&[75,105]&[75,105] & -- & --& [75,105]\\
\hline
$\sigma^h_i$& [1,9] & [1,9]& [1,9]& [1,9]& [1,9]& [1,9]\\
\hline
$\sigma^p_i$& [1,9] & [1,9]& [1,9]& [1,9]& [1,9]& [1,9]\\
\hline
\end{tabular}
\end{table}

\section*{Acknowledgements}\label{section-acknowledge}

This work was supported by the Australian Research Council, Dynamics and Control of Complex Social Networks under Grant DP190101087.

\end{document}